\newcommand{\arxivbuild}{1}   % set to 1 by arxiv.tex: appendix included
\documentclass[runningheads]{llncs}
\usepackage[T1]{fontenc}
\usepackage[utf8]{inputenc}
\usepackage{amsmath,amssymb}
\usepackage{graphicx}
\usepackage{booktabs}
\usepackage{multirow}
\usepackage{array,calc,etoolbox}
\usepackage{url}
\usepackage{tikz}
\usepackage[figuresright]{rotating}
\usetikzlibrary{positioning,arrows.meta,fit,calc}
\makeatletter\def\verbatim@font{\scriptsize\ttfamily}\makeatother
\usepackage{xcolor}
\usepackage[hidelinks]{hyperref}
\usepackage{microtype}
\usepackage{orcidlink}

\graphicspath{{figures/}}

\newcommand{\LU}{L_{U}}

\newcommand{\PW}{PW}

\newcommand{\RW}{RW}
\newcommand{\rhoc}{\rho}

\begin{document}

\title{The Power Watershed as a Differentiable Segmentation Layer}
\titlerunning{The Power Watershed as a Differentiable Layer}

\author{Galia Akl\inst{1,2}\orcidlink{0009-0006-1673-4158} \and Laurent Najman\inst{3,4}\orcidlink{0000-0002-6190-0235} \and Hugues Talbot\inst{5}\orcidlink{0000-0002-2179-3498}}
\authorrunning{G. Akl et al.}
\institute{Holy Spirit University of Kaslik, Lebanon\\ \email{galia.r.akl@net.usek.edu.lb}
\and American University of Beirut, Lebanon
\and LIGM, Univ. Gustave Eiffel, CNRS, F-77454 Marne-la-Vall\'ee, France\\ \email{laurent.najman@esiee.fr}
\and Math. Department, Khalifa University, Abu Dhabi, UAE
\and CentraleSup\'elec, Inria, Universit\'e Paris-Saclay, France\\ \email{hugues.talbot@centralesupelec.fr}}

\maketitle

\begin{abstract}
Can a Watershed transform, the classical tool of morphological segmentation, be placed at the end of a deep network and trained through? The Power Watershed (\PW) is the $p\to\infty$ limit of the seeded Random Walker (\RW) with edge weights raised to the power $p$. At every finite $p$ the \RW\ solves a sparse positive-definite linear system and is exactly differentiable with respect to the edge weights and the unaries through one adjoint solve; its multi-label form costs one right-hand side per label, and its limit is the \PW. We study this family as the last layer of a segmentation network, the watershed layer (WS layer). A conditioning study measures how the solve, the gradient and the distance to the limit evolve with $p$, and the implicit gradient is checked against unrolling. On synthetic data with correct edge costs the WS layer transforms a soft prediction into an exact boundary. On nuclei segmentation (PanNuke), a binary WS layer is level with a thresholded U-Net. With the instances as labels, the WS layer turns the U-Net into an instance segmenter whose number of labels is set per image by the markers rather than by the output layout, and the network is trained through that partition: instance $F_1$ rises from 0.656 to 0.795 over a U-Net with connected components, level with the same network followed by a tuned marker watershed. At the training temperature the partition is a watershed of the learned costs, and the finite temperature lets the solve hedge, discounting weak markers and sharing contested pixels. The hedging puts the WS layer 0.015–0.020 $F_1$ ahead of the baseline when trained on a tenth of the data, at the price of having to deploy the finite-p solve rather than its exact limit at inference. Code and the full experimental record are released. 
\end{abstract}

% ---------------------------------------------------------------------------
\section{Introduction}
% ---------------------------------------------------------------------------

Deep segmentation networks such as U-Nets perform well under supervision but have many parameters, which reduces explainability, and produce soft, per-pixel outputs. Boundaries come out rounded, output potentials are rarely binary, and the number of labels is fixed by a one-hot channel layout decided at design time. Watersheds have the opposite profile: results are crisp and directly explainable through labelling and the topographical nature of the input data. As a result, boundaries appear on the strongest edges of a cost map, regions are piecewise constant, and the number of regions is equal to the number of markers, not a property of the algorithm. An interesting combination is thus to add a watershed at the end of a segmenting network and train through it. The obstacle is that a watershed is a combinatorial procedure on a graph, with a gradient that is zero almost everywhere.

This article makes the case that the Power Watershed framework of
Couprie et al.~\cite{couprie2009power,couprie2011power} can be used in this context. It provides its own relaxation, the Random Walker~\cite{grady2006random} at finite $p$, and we evaluate what the resulting watershed layer (WS layer) can be used for. The framework minimises $\sum_e w_e^{\,p}\,|x_u-x_v|^q$ over potentials $x$ fixed on seeds; $q=2$ with finite $p$ is the Random Walker (\RW), and $p\to\infty$ is the Power Watershed (\PW), a maximum spanning forest with a small linear solve on each plateau. Reading $p$ as an inverse temperature of the Gibbs distribution over spanning forests, as the probabilistic watershed of Fita Sanmart\'in et al.~\cite{fitasanmartin2019probabilistic} suggests (the $p\to\infty$ collapse is that of softmax$\to$argmax~\cite{jang2017categorical}), every finite-$p$ member is a convex problem whose solution is exactly differentiable, and the path ends at the \PW, a true watershed rather than a surrogate as in learned watersheds~\cite{wolf2017learned} or post-hoc watershed transforms~\cite{bai2017deep,graham2019hovernet}.

Our contributions are the following. 
(i) A differentiable \RW/\PW\ layer with unaries, written with dense shifted-grid operators so that it is batched, running on the GPU and can be  differentiated in either of two ways: by unrolling a few preconditioned conjugate-gradient steps or implicitly, by one adjoint solve (Sect.~\ref{sec:layer}). Its multi-label form costs one right-hand side per label, and its limit one spanning forest whatever the number of labels, whereas graph cuts are exact only for two labels.
(ii) A conditioning study of solver cost, distance to the limit and gradient support as functions of $p$ (Sect.~\ref{sec:cond}).
(iii) An experimental account on synthetic scenes and PanNuke nuclei~\cite{gamper2019pannuke,gamper2020pannuke} that evaluates where the WS layer helps. With correct edge costs it recovers exact boundaries; as a binary last layer on real images it is level with a thresholded U-Net; with the instances as labels it is level with a tuned marker watershed on the same network, and ahead of it when training data are scarce (Sect.~\ref{sec:exp}).

% ---------------------------------------------------------------------------
\section{Background and related work}
% ---------------------------------------------------------------------------

\paragraph{The $(p,q)$ family.} On a graph $G=(V,E)$ with weights $w_e>0$ and seeds
$M\subset V$ with fixed labels, the \PW\ framework~\cite{couprie2011power} minimises
\begin{equation}
E_{p,q}(x)=\sum_{e=\{u,v\}} w_e^{\,p}\,|x_u-x_v|^q.
\end{equation} The pair $(p{=}1,q{=}1)$ is the graph cut, $(p$ finite$,q{=}2)$ the \RW, and $p\to\infty$ the \PW, whose solution is unique for $q=2$ and is computed by Kruskal's algorithm on the edges in decreasing weight, with a $q$-energy minimisation restricted to each plateau of equal weight~\cite{couprie2011power,najman2017equivalence}. All\`ene et
al.~\cite{allene2007links,allene2010some} showed that the min cut of the powered
weights becomes a maximum spanning forest cut as $p$ grows, and Cousty et
al.~\cite{cousty2009watershed} that this cut is the watershed cut of the graph.
Grady's prior-augmented \RW~\cite{grady2005multilabel} adds unary terms, which we use throughout.

\paragraph{Learning through segmentation algorithms.} Unrolling an iterative
algorithm into a network~\cite{gregor2010learning,monga2021algorithm} and
differentiating an optimisation layer implicitly~\cite{amos2017optnet,bai2019deep}
are well-known techniques; CRF-as-RNN~\cite{zheng2015conditional} unrolls mean-field inference as a
last layer. Closest to us, Cerrone et al.~\cite{cerrone2019learned} learn the edge
weights of a seeded \RW\ end to end from given seeds; we add unaries, the temperature path to the \PW\ with its exact limit, and instances as labels. Wolf et al.~\cite{wolf2017learned} learn
inside a seeded watershed by structured learning, and spanning structures can also be made differentiable by perturbation~\cite{berthet2020perturbed} or stochastic softmax relaxations~\cite{paulus2020stochastic}; ours is the deterministic relaxation that the \PW\ framework already contains. In the morphology literature, Challa et al.~\cite{challa2022triplet} train features with a triplet loss so that a watershed on them classifies hyperspectral pixels, and Chierchia and Perret~\cite{chierchia2019ultrametric} fit ultrametrics, the hierarchical form of the watershed, by gradient descent on a relaxed objective; in both the watershed is differentiated through a surrogate. The deep watershed
transform~\cite{bai2017deep} and HoVer-Net~\cite{graham2019hovernet} predict maps on
which a non-differentiable watershed is run afterwards. Our relaxation is not a surrogate but a member of the same family: the Random Walker is the Power Watershed at finite p, and its limit is the watershed itself.

% ---------------------------------------------------------------------------
\section{The watershed layer}\label{sec:layer}
% ---------------------------------------------------------------------------

\subsection{Energy, temperature and limit}

Let $c_e\ge 0$ be the cost of edge $e$ and, for each pixel $i$, $c^1_i,c^0_i\ge0$
the costs of attaching it to the foreground and background labels. With everything
raised to the inverse temperature $p$,
\begin{equation}
  \begin{aligned}
  E_p(x)&=\sum_{e=\{u,v\}} w_e\,(x_u-x_v)^2+\sum_i a^1_i\,(x_i-1)^2+\sum_i a^0_i\,x_i^2,\\
  w_e&=e^{-p c_e},\qquad a^1_i=e^{-p c^1_i},\qquad a^0_i=e^{-p c^0_i}.
  \end{aligned}
  \label{eq:energy}
\end{equation}
Stationarity yields the symmetric positive definite system
\begin{equation}
  \big(L(w)+\mathrm{diag}(a^0+a^1)\big)\,x=a^1,
  \label{eq:system}
\end{equation}
where $L(w)$ is the weighted graph Laplacian. The unaries are edges to two virtual
label nodes of an augmented graph, so~\eqref{eq:energy} is the seeded $(p,2)$ energy on that graph with the two virtual nodes as seeds. Hence $x^p$ is the probability that a random 2-forest at inverse temperature $p$ connects pixel $i$ to the foreground node\ifnum\arxivbuild=1{} (Appendix~\ref{app:forests})\fi, and $x^\infty=\lim_{p\to\infty}x^p$ is the \PW\ of the augmented graph: a maximum spanning forest flooded from the confident pixels, i.e.\ a marker-controlled watershed whose markers are decided by the unaries. As the last layer of a network, $c^1_i=-\alpha\log s_i$ and $c^0_i=-\alpha\log(1-s_i)$ with $s$ the network's soft foreground probability; $\alpha$ is Grady's $\gamma$ and sets how strongly the unaries compete with the edges.

\paragraph{Many labels.} With $K$ markers (classes or instances), the multi-label
\RW\ solves~\eqref{eq:system} with the shared matrix $L(w)+\mathrm{diag}(\sum_k a^k)$ and one right-hand side $a^k$ per label; summing the equations and using $L\mathbf 1=0$ shows that the $x^k$ sum to one, so the output is a per-pixel distribution over this image's markers. The cost is therefore affine in $K$ and the
problem is exact and convex for every $K$. The limit is one spanning forest irrespective of $K$. This is where the family differs most from graph cuts, exact for two labels but NP-hard beyond, where $\alpha$-expansion~\cite{boykov2001fast} is the usual approximation; in the limit the multi-label problem becomes
tractable~\cite{allene2010some}.

\subsection{WS layer differentiation}

Let $A=L(w)+\mathrm{diag}(a^0+a^1)$ and let $\ell(x)$ be the loss. With
$\lambda=A^{-1}\,\partial\ell/\partial x$ one obtains, for every edge and pixel,
\begin{equation}
  \frac{\partial\ell}{\partial w_e}=-(x_u-x_v)(\lambda_u-\lambda_v),\qquad
  \frac{\partial\ell}{\partial a^1_i}=\lambda_i(1-x_i),\qquad
  \frac{\partial\ell}{\partial a^0_i}=-\lambda_i x_i,
  \label{eq:adjoint}
\end{equation}
and the chain rule through $w_e=e^{-pc_e}$ contributes a factor $-p\,w_e$. This \emph{adjoint solve} therefore yields the gradient with respect to all costs, with memory independent of the solver's depth. We term this the \emph{implicit mode}. Alternatively, $T$ steps of Jacobi-preconditioned conjugate gradients are unrolled and differentiated by automatic differentiation. We term this the \emph{unrolled mode}; this is what we deploy when the truncated solver is the WS layer, and
$T$ then caps how far a boundary can move, since $T$ iterations propagate information at most in a radius of $T$ pixels. The binary experiments below use the unrolled mode with $T=20$ unless stated; the instance experiments use the implicit mode, solving to a relative residual of $10^{-4}$ (at most 100 iterations) in training and $10^{-5}$ at evaluation, so the radius argument does not restrict them. The two modes do not compute the same thing: at $p\ge20$, twenty unrolled steps give the exact gradient of the truncated solver, far from that of the \RW\ (cosine 0.67--0.75 on the edge costs), while the implicit mode matches it to $10^{-3}$ with 30--90 iterations and a constant 0.09\,GB per batch of 16. On an A100 the multi-marker solve costs $41+20K$\,ms per batch of 16 $256^2$ images with $K$ markers each, and the exact limit 29--51\,ms per patch on one CPU core for $K$ up to 128 \ifnum\arxivbuild=1 (Appendix~\ref{app:efficiency})\else~\cite[\S F]{akl2026extended}\fi. On a regular grid all operators are dense shifted differences, so the WS layer is batched and can run on a GPU without sparse matrices; the markers of the multi-label form go in the batch dimension. \ifnum\arxivbuild=1{} The architectures of the trained models are drawn in Appendix~\ref{app:arch}.\fi The exact limit is evaluated on the CPU by the plateau algorithm of~\cite{couprie2011power} with per-label solves on ambiguous plateaus; our implementation agrees with the authors' reference code to single precision~\cite[\S B]{akl2026extended}.

\paragraph{Two difficulties.}
The limit $x^\infty$, seen as a function of the weight vector $w$, is locally constant: it changes only when the order of two edge weights flips, which happens on the finitely many hyperplanes where two weights are equal. Away from those ties $\partial x^\infty/\partial w$ exists and is zero, and on them it is undefined. As $p$ grows, $x^p$ inherits this behaviour: its gradient is well defined for essentially every weight configuration, but it concentrates on fewer and fewer edges and carries less and less information about which direction would improve the loss. At the same time the solve degrades: $\lambda_{\max}(A)$ is at least the largest diagonal entry, hence at least $w_{\max}$, while $\lambda_{\min}(A)$ is at most the weighted degree plus the unary of any pixel (the Rayleigh quotient of its indicator); if some pixel has its four edges at the highest cost, the condition number $\kappa(A)=\lambda_{\max}(A)/\lambda_{\min}(A)$ satisfies $\kappa(A)\gtrsim e^{p\,\Delta c}/4$, with $\Delta c$ the spread of the edge costs. What matters is therefore $p\,\Delta c$, not $p$ alone. With continuous learned costs ties are rare, and the limit reduces to a maximum spanning forest: the plateau solves that distinguish the \PW\ matter only for quantised costs. Section~\ref{sec:cond} measures
both these effects.

% ---------------------------------------------------------------------------
\section{Experiments}\label{sec:exp}
% ---------------------------------------------------------------------------

All code, raw results and the extended version of this article~\cite{akl2026extended} are publicly available\footnote{\url{https://github.com/hugues-talbot/differentiable-power-watershed}}. The metrics we use are: intersection over union (IoU), boundary F-score at a pixel tolerance (BF$_t$), the fraction of pixels with potential in $(0.1,0.9)$ (soft pixels), and, for instance segmentation, F$_1$ of the matches at IoU~0.5, panoptic quality (PQ)~\cite{kirillov2019panoptic}, aggregated Jaccard index (AJI) and the relative object-count error. On PanNuke we also report a label-free edge-alignment score, the fraction of predicted boundary pixels within one pixel of a strong colour edge, since the dataset's semi-automatic contours bound what a boundary score can show. PQ is the product of the segmentation quality SQ, the mean IoU of the matched objects, and the detection quality DQ, an object-level F$_1$; it penalises merged and split objects alike.

\subsection{The cost of trying to reach the limit}\label{sec:cond}
For this experiment, we use three simple images: synthetic shapes with two noise levels $\sigma\in\{0.05,0.1\}$ and the \texttt{camera} image from \texttt{scikit-image}. We use hard-seeded \RW\ on $512\times 512$ 8-bit images with $c_e=((I_u-I_v)/255)^2$. The various seeds are: ground truth eroded by 5~px (edge refinement regime) or 30~px, or only a few scribbles. The reference $x^p$ is a direct sparse LU solve in double precision with diagonal rescaling and iterative refinement; records whose solve is unreliable are excluded. The limit $x^\infty$ is computed by the combinatorial PW on the integer costs of the 8-bit image, so that ties, and hence plateaus, are exact.

\begin{table}[t]
\centering\footnotesize\setlength{\tabcolsep}{5pt}
\caption{Refinement regime ($512^2$, seeds eroded by 5~px). Condition number of
$\LU$, the Laplacian restricted to the unseeded pixels (ARPACK; values above $10^{16}$ exceed double precision and are indicative only), iterations of plain, unpreconditioned conjugate gradients in single precision to a relative residual of
$10^{-3}$, fraction of free pixels whose label differs from the limit, and the number
of edges carrying 90\,\% of the gradient of a squared loss to the
reference labels.}
\label{tab:cond}
\begin{tabular}{llrrrr}
\toprule
image & $p$ & $\kappa(\LU)$ & CG iters & labels $\ne x^\infty$ (\%) & edges, 90\,\% grad.\\
\midrule
clean, $\sigma{=}0.05$ & 10 & $1.8{\cdot}10^{2}$ & 39 & 0.7 & 14276\\
 & 100 & $4.7{\cdot}10^{2}$ & 53 & 0.0 & 5945\\
 & 200 & $5.7{\cdot}10^{3}$ & 63 & 0.0 & 844\\
 & 1000 & $2.3{\cdot}10^{17}$ & 173 & 0.0 & 7\\
noisy, $\sigma{=}0.1$ & 10 & $2.0{\cdot}10^{2}$ & 42 & 3.1 & 14645\\
 & 100 & $1.5{\cdot}10^{10}$ & 89 & 2.2 & 8351\\
 & 500 & $1.0{\cdot}10^{48}$ & 240 & 1.2 & 681\\
\texttt{camera} & 10 & $5.5{\cdot}10^{2}$ & 62 & 9.8 & 23602\\
 & 100 & $1.7{\cdot}10^{7}$ & 104 & 5.5 & 13044\\
 & 500 & $4.1{\cdot}10^{33}$ & 154 & 4.1 & 5503\\
\bottomrule
\end{tabular}
\end{table}

Table~\ref{tab:cond} summarizes the results. On the clean image the labels of the limit are reached at $p\approx100$ while $\kappa$ is still low and CG needs only about 50 iterations. This is a comfortable training window, inside which the gradient support has already more than halved and beyond which it collapses (7 edges at $p=1000$); support is only a proxy for trainability. Since $c_e\in[0,1]$ here, $p$ is also $p\,\Delta c$; the learned costs of Sects.~\ref{sec:binary}--\ref{sec:instances} are trained at $p\,\Delta c\lesssim 50$ (costs in $[0,1]$ at $p\le50$, or a learned scale of 1.85 at $p=20$), inside this window. On noisy and natural images $x^p$ approaches the limit slowly (a few per cent of the free pixels still differ at $p=500$) while $\kappa$ goes far beyond what any solver can resolve.
The iteration counts grow far more slowly than $\sqrt\kappa$, because the ill-conditioning sits in a few near-isolated pixels that barely affect the residual. The double-precision factorisation itself fails beyond $p\approx500$; an exact elimination confirms the table and shows the gap closing slowly beyond (\texttt{camera}: 1.2\,\% at $p=5{\cdot}10^3$), as 8-bit squared costs differ by as little as $1.5{\cdot}10^{-5}$. The last per cent of agreement are out of practical reach of a finite-$p$ solve; they come from the combinatorial step. In the refinement regime 20 unrolled CG steps disagree with the converged solve on 0.04\,\% of the labels (2.9\,\% on the noisy image), independently of the image size; in the global regime (30-px erosion, scribbles) no truncation works.

\subsection{A tunable, spatially coupled softmax}\label{sec:softmax}

\begin{figure}[ht]
\centering
\includegraphics[width=\textwidth]{\ifnum\arxivbuild=1 softmax_examples\else softmax_examples_2rows\fi}
\caption{Simulated soft predictions (ground truth warped by up to 3~px, blurred over 4~px, soft logits) through the WS layer, $\alpha=0.1$, $256^2$, noise 0.05; image edge on the true boundary (upper row) or 4~px away (lower row). Columns: image, soft prediction, $x^p$ for $p=5,20,100,500$, the exact limit, ground truth. The thin protrusion is recovered; with the edge in the wrong place the layer snaps to it.\ifnum\arxivbuild=1{} The two lower rows repeat the scenes with noise 0.1.\fi}
\label{fig:softmax}
\end{figure}

Fig.~\ref{fig:softmax} shows the WS layer on simulated network outputs. With the image edge on the true boundary and $\alpha=0.1$, the boundary score BF$_2$ of the thresholded soft prediction, initially 0.958, becomes 1.000 at $p=200$ and in the limit on the clean scene, and 0.968 at $p=200$ (0.870 in the limit) on the noisy one. The output goes from 22\,\% soft pixels to under 1\,\%: a softmax whose sharpness is spatially coupled through the edges. The scale $\alpha$ does matter: at $\alpha=1$ the unaries of confident pixels compete with the noise edges and everything merges before any label attaches, so both large $p$ and the limit collapse. And when the true boundary is 4~px from the image edge, $p\le10$ is harmless but a larger $p$ snaps the boundary to the wrong edge (BF$_2$ 0.002 in the limit). The WS layer is only as good as its edge costs, which argues for learning them.

\subsection{Real images: a binary last layer}\label{sec:binary}
A small U-Net (32--64--128--256 channels) is trained for 30 epochs with Adam on PanNuke~\cite{gamper2019pannuke,gamper2020pannuke} (nucleus versus background, $256^2$, fold~1 for training, all 2523 patches of fold~2 for testing). Variants: the plain U-Net with binary cross-entropy (BCE), and a U-Net whose logits feed the WS layer as unaries, with edge costs from the image gradient or from a second head of the same network (two more output channels of its last $1{\times}1$ convolution, through a sigmoid, averaged onto the edges). The WS layer is implemented as 20 unrolled PCG steps with $p$ annealed geometrically from 5 to 50 and the loss BCE on the WS layer output plus 0.5 BCE on the logits. The output of the WS layer is evaluated as trained, converged at the final $p$, and as the exact limit.

\begin{table}[t]
\centering\footnotesize\setlength{\tabcolsep}{4pt}
\caption{PanNuke fold 2, binary nucleus segmentation. ``Thresholded'' is the
network's sigmoid output cut at 0.5, with instances as connected components; a threshold sweep moves its IoU by at most 0.003. 95\,\% bootstrap half-widths over the 2523 patches are 0.005--0.010 for IoU, BF$_2$, edge alignment, F$_1$ and PQ, and 0.01--0.05 for the count error.}
\label{tab:binary}
\resizebox{\textwidth}{!}{%
\begin{tabular}{lrrrrrr}
\toprule
output & IoU & BF$_2$ & edge al. & F$_1$@0.5 & PQ & count err.\\
\midrule
U-Net, thresholded & 0.719 & 0.769 & 0.577 & 0.656 & 0.531 & 0.275\\
U-Net + WS layer, image-gradient costs & 0.683 & 0.714 & 0.559 & 0.516 & 0.400 & 0.513\\
\quad its exact limit & 0.174 & 0.251 & 0.804 & 0.209 & 0.163 & 0.676\\
U-Net + WS layer, learned costs & 0.709 & 0.752 & 0.579 & 0.604 & 0.482 & 0.337\\
\quad its exact limit & 0.660 & 0.714 & 0.583 & 0.643 & 0.512 & 0.253\\
\quad trained in the implicit mode (mean of two seeds) & 0.717 & 0.772 & 0.599 & 0.647 & 0.521 & 0.253\\
U-Net + WS layer, learned costs with edge target, converged & \textbf{0.721} & 0.773 & 0.569 & \textbf{0.724} & \textbf{0.588} & \textbf{0.214}\\
U-Net + contour head, marker watershed & 0.718 & \textbf{0.774} & 0.568 & 0.715 & 0.583 & 0.344\\
\bottomrule
\end{tabular}}
\end{table}

Table~\ref{tab:binary} presents the outcome. A binary WS layer with learned costs is 0.01 IoU and 0.02 BF$_2$ below a thresholded U-Net in the unrolled mode and level with it in the implicit mode; the same holds on Oxford-IIIT Pet~\cite{parkhi2012cats}, with objects a hundred pixels across~\cite[\S J]{akl2026extended}. Image-gradient costs are useless on photographs and on cell images alike, as their limit floods whole regions. The exact limit of the learned costs is however deployable at a cost of 0.05 IoU, with a better object-count error than the U-Net. Inside the trained model the WS layer does what it should (its own logits fragment into 350 components per patch where the truth averages 24; its output has 24), but a standalone U-Net trained with BCE for 30 epochs learns a similar sharpening by itself. 
Boundary- and seam-weighted losses, a gap target between touching nuclei, lower annealing ranges and an edge target on the cost head~\cite[\S L, O]{akl2026extended}{} move these metrics by a few thousandths, with one exception: the edge target, a cross-entropy of the cost maps against the instance boundaries, brings the converged WS layer level with the thresholded U-Net on IoU (paired difference $+0.002$, 95\,\% bootstrap interval $[-0.000,+0.005]$) and gives the best raw instance scores of the table. Training in the implicit mode, with the true \RW\ gradient (Sect.~\ref{sec:layer}), matters more: on both seeds every paired interval against the unrolled mode is clear of zero (F$_1$ $+0.035$ and $+0.055$), and the layer is level with the U-Net on IoU and BF$_2$, 0.007--0.012 behind on F$_1$ and PQ; the exact limit of these costs is worse (IoU 0.60\ifnum\arxivbuild=1; Appendix~\ref{app:efficiency}\fi).

Finally, a U-Net given the same edge supervision through a contour channel, with a marker watershed run on that map afterwards, is level with the edge-target WS layer at the pixel level (last two rows); on raw instance scores the layer is slightly ahead, and after an 8-px minimum-area filter the watershed is. 

\subsection{Real images with labeled instances}\label{sec:instances}
The binary WS layer essentially discards the property that makes the \PW{} distinctive. We now let each image carry its own label set, one marker per nucleus plus background, and let the WS layer compute the partition among them (Sect.~\ref{sec:layer}): the U-Net predicts a foreground logit, the edge costs and a marker map; the loss is a per-pixel cross-entropy over this image's markers against the instance mask, with $p$ annealed from 5 to 20. Each marker region $k$ attaches its pixels to label $k$ with $a^k=s^p$ and the background label attaches every pixel with $(1-s)^p$. The loss is differentiable in the edge costs and the attachments, not in the choice of markers: extraction is a discrete step inside the network, run in training as at test time, and the marker map learns from its own regression loss only. Training markers are first the eroded ground-truth instances. Test markers come from a distance-to-boundary head that regresses, for every nucleus pixel, its normalised distance to the nucleus boundary as in Naylor et al.~\cite{naylor2019distance}. From the distance map, the markers are its regional maxima grown into small basins by a watershed of the map; the baseline uses the same procedure, both tuned on fold~3 by PQ (72 settings for the WS layer, 135 for the baseline including its relief; the $p$ schedule is not tuned; \cite[\S D]{akl2026extended}).

\begin{table}[t]
\centering\footnotesize\setlength{\tabcolsep}{4pt}
\caption{PanNuke fold 2, instances. The WS layer with the instances as labels,
against the U-Net with connected components and against the fair baseline: the same
U-Net with the same distance head and a marker watershed run on it afterwards, in the
manner of HoVer-Net~\cite{graham2019hovernet}. Best
in bold, second best underlined; the oracle rows are not ranked. Over
three seeds and three fold pairs the WS layer trained from its own markers spans F$_1$
0.787--0.795 and PQ 0.643--0.649; the baseline is level to 0.008 F$_1$ and 0.005--0.015 PQ ahead, and the two are level on AJI and count error.}
\label{tab:instances}
\resizebox{\textwidth}{!}{%
\begin{tabular}{lrrrrr}
\toprule
output & IoU & F$_1$@0.5 & PQ & AJI & count err.\\
\midrule
U-Net, thresholded, components & \underline{0.719} & 0.656 & 0.531 & 0.494 & 0.275\\
U-Net + distance head, thresholded, components & \textbf{0.725} & 0.664 & 0.539 & 0.501 & 0.313\\
Multi-marker WS layer, distance-head markers & 0.707 & 0.791 & 0.641 & 0.653 & 0.239\\
\quad trained from its own predicted markers & 0.715 & \textbf{0.795} & \underline{0.649} & \textbf{0.668} & \textbf{0.182}\\
\quad the same with ground-truth markers (oracle) & 0.775 & 0.872 & 0.723 & 0.771 & 0.142\\
U-Net + distance head, marker watershed & \underline{0.719} & \underline{0.794} & \textbf{0.655} & \underline{0.658} & \underline{0.190}\\
\quad the same with ground-truth markers (oracle) & 0.774 & 0.867 & 0.708 & 0.765 & 0.066\\
\bottomrule
\end{tabular}}
\end{table}

\begin{figure}[t]
\centering
\includegraphics[width=\textwidth]{\ifnum\arxivbuild=1 pannuke_instances\else pannuke_instances_2rows\fi}
\caption{PanNuke, 128-px crops around touching nuclei, the WS layer trained from its own markers. Columns:
image, ground-truth instances, predicted marker map, the multi-marker WS layer's
partition from the predicted markers, the exact limit from the same markers, and the
layer with the ground-truth markers (oracle). Label maps in random colours with
black contours.}
\label{fig:instances}
\end{figure}

Table~\ref{tab:instances} is where the watershed's strengths show. With the instances as labels the WS layer raises instance F$_1$ from 0.656 to 0.795 and PQ from 0.531 to 0.649 over the U-Net with connected components. The last step of that gain comes from training the solve on the model's own predicted markers, with unmarked nuclei as background in the loss (they still count as misses in every metric). Trained this way the solve \emph{hedges}: through the soft unaries it discounts weak markers and shares contested pixels, so with ground-truth markers it drops 11\,\% of them and its oracle (0.872) falls to the baseline's (0.867). The fair baseline, the same network and distance head followed by a tuned marker watershed, reaches 0.794 and 0.655. Paired, the WS layer is level on F$_1$ ($+0.001$, $[-0.001,+0.004]$), behind on PQ ($-0.005$, $[-0.008,-0.003]$) and ahead on AJI ($+0.009$, $[+0.007,+0.012]$); over seeds and fold pairs the two stay within 0.01. They are close for a precise reason: given the WS layer's own markers and output foreground, a marker watershed on its learned contour map reproduces its partition exactly~\cite[\S D]{akl2026extended}; the two differ through the foreground and the markers only. Nor does the draw depend on the marker head: with a CenterNet-style centre-and-radius detector instead, the WS layer stays at F$_1$ 0.795 and the watershed baseline drops to 0.774~\cite[\S W]{akl2026extended}. With a tenth of the training data (266 images, same number of steps) the WS layer leads by 0.015--0.020 F$_1$ and 0.011--0.015 PQ over three replicates, every paired interval clear of zero: the baseline's tuning leaves a quarter of the nuclei unmarked, while the hedging solve keeps a permissive threshold and absorbs the spurious markers. From 25\,\% on the baseline is ahead again. The residual PQ gap is the price of the hedging: detections are equal (DQ 0.795 against 0.794) and the hedged boundaries are softer (SQ 0.816 against 0.824). Making the marker selection itself differentiable did not help (F$_1$ 0.732 with strengths from the marker head, 0.006--0.013 behind with isolated scorers~\cite[\S S, U]{akl2026extended}). Finally, the exact limit of this model reaches only F$_1$ 0.724 and PQ 0.585. Solved exactly at increasing $p$, the same system traces the path: F$_1$ peaks at the training $p=20$ and falls monotonically to the limit, which the solve reaches by $p\approx500$, while SQ stays at 0.81: a higher $p$ removes the hedging, discounted markers become objects again, and the count error rises from 0.18 to 0.27\ifnum\arxivbuild=1{} (Appendix~\ref{app:psweep})\fi. The model trained from true markers is the mirror image, within 0.006 F$_1$ of its limit but behind on its own markers (Table~\ref{tab:instances}): a deployable limit and tolerance to imperfect markers pulled in opposite directions in every schedule we tried.

% ---------------------------------------------------------------------------
\section{Conclusion}
% ---------------------------------------------------------------------------

The Random Walker at finite $p$ gives the Power Watershed a differentiable form that is exact rather than a surrogate: every $p$ yields a convex problem differentiated through one adjoint solve, the multi-label problem is exact at a cost affine in the number of labels and as $p\to\infty$ the solution converges to that of the combinatorial algorithm. A conditioning study shows what this path costs: the solve and the gradient both degrade as $p$ grows, so training happens at moderate $p$. On learned costs the exact solution tends smoothly to the limit, but the costs are tuned to the training temperature: the instance model peaks there and loses 0.07 F$_1$ at the limit, the binary models lose 0.05 IoU while their instance scores improve, so the deployed solution needs to be the finite-$p$ solve. As a binary last layer, the WS layer is level with a thresholded U-Net. With the instances as labels its partition is, at the training temperature, a watershed of its learned costs between markers, and it matches the same network followed by a tuned marker watershed as baseline; every gain it showed was matched once the baseline received the same supervision. What the end-to-end training adds is hedging, which improves tolerance to imperfect markers and improves performance when annotations are scarce, at the price of some boundary precision and of no longer being able to use the limit. Marker extraction remains a discrete step: the partition loss rewards a marker for the pixels it takes, not for being a nucleus, and three attempts to learn the selection did not help. In future work, we will study the influence of weak or scribble supervision and markers learned without ground-truth instances, where a post-hoc watershed cannot compete, together with training regimes that reconcile the hedging with the limit, and 3D applications. We release the code and the complete experimental record, negative results included.

\subsubsection*{Acknowledgements.} Funding was provided by our respective institutions. In particular, the work of L. Najman was partially supported by Khalifa University of Science and Technology under Award No. KU-INT-FSU-2026-004-8471000027. Computing cluster access was provided by the M\'esocentre Paris-Saclay. Anthropic's Claude Code assistant was used for running the experiments, early drafting and verifying the correctness of the article.

%The work of L. Najman was partially supported by Khalifa University of Science and Technology under Award No. KU-INT-FSU-2026-004-8471000027.

% Bibliography inlined from main.bbl (BibTeX output of refs.bib with splncs04.bst): no BibTeX run needed

\ifnum\arxivbuild=1
\clearpage
\appendix
% Appendix: included only in the extended (HAL / arXiv) build.  Assembled from the
% long-form report `unrolling/finishing-layer-experiments.md` of the research repository.

\section{Questions a reader may ask}\label{app:faq}

The main text had to be short. This section answers, in order, the questions that
the authors and their reviewers asked while writing it, with pointers to where the
evidence is.

\paragraph{What is the WS layer, in one sentence?}
A seeded Random Walker at a finite inverse temperature $p$, with unaries, whose edge
costs and unaries are produced by the network, trained through, and whose $p\to\infty$
limit is the Power Watershed of the same costs. At the temperature it is trained at,
its partition between markers is a marker watershed of its learned contour map
(Appendix~\ref{app:pipelines}, Table~\ref{tab:swap}: identical to three digits on 2523
patches). The finite temperature adds one thing a watershed does not have: the solve
can hedge, i.e.\ discount a weak marker through the soft unaries and share contested
pixels between neighbours.

\paragraph{A post-hoc marker watershed also turns a U-Net into an instance segmenter with
any number of labels. What is new?}
Where the partition happens. In HoVer-Net-style pipelines the watershed runs after
training, on maps the network was trained to predict; the loss never sees the
partition. In the WS layer the partition over a per-image label set is a layer of the
network, and the cross-entropy over that label set trains the edge costs, the
foreground and the marker attachments through the solve (Sect.~\ref{sec:instances}).
Two properties of the $(p,q)$ family make this cheap: the Random Walker's cost is
affine in the number of labels (one shared matrix, one right-hand side per label), and
the limit is one spanning forest whatever that number (Sect.~\ref{sec:layer}), where
graph cuts are exact for two labels only. The number of labels is set per image by
the markers, not by the output layout. The markers themselves remain a discrete step;
see below.

\paragraph{The baseline's watershed is ``tuned''. Does that mean a second training?}
No: a grid search over a few scalars on 300 patches of a held-out fold (fold~3, used
neither for training nor for testing), scored by PQ, five minutes in all. The
baseline has four choices (marker threshold, minimum marker area, prominence of the
distance maxima, and the relief: the contour map, the negative distance map or their
sum; 135 settings). The WS layer's marker extraction is chosen the same way with the
first three (72 settings); its $p$ schedule is fixed. So the honest contrast is not
``end to end versus tuned'': both pipelines tune how markers are read off the distance
map; only the baseline also tunes the assignment step, which the WS layer learns
(Appendix~\ref{app:pipelines}). The tuning stage is where the baseline is brittle: at
a tenth of the training data its grid search picks a high marker threshold and leaves
a quarter of the nuclei unmarked, while the WS layer keeps a permissive threshold at
every data size because its trained solve absorbs the spurious markers
(Appendix~\ref{app:dataeff}).

\paragraph{``The same network'': is the baseline's U-Net trained the same way?}
Same architecture, same three output heads, same targets and training recipe, without
the solve in the loop. The baseline U-Net predicts a foreground logit (cross-entropy
against the binary nucleus mask), a two-channel contour map (the edge target: a
cross-entropy against the grid edges where the instance label changes, seams between
touching nuclei weighted up) and a distance map (the normalised distance of each
nucleus pixel to its boundary, squared error, as in HoVer-Net's distance variant),
trained on fold~1 for 30 epochs with the same optimiser, batch size and augmentation.
The WS-layer model predicts the same four channels with the same three losses, the
contour map becoming the edge costs through a sigmoid and a learned scale, plus the
partition cross-entropy over the markers, which reaches the foreground logit and the
costs through the solve; its foreground loss is an auxiliary term at half weight. So
the heads differ in one respect after training: the baseline's foreground head is
trained only to be a mask and is the best of the study (IoU 0.725 thresholded), while
the layer's foreground head degenerates to speckle, since the solve needs it only as a
soft prior, and the layer's output foreground stays 0.01 IoU below. The distance heads
come out equivalent. Both pipelines then read markers off their distance maps by the
same rule, and the baseline floods a watershed from them inside its thresholded
foreground (Appendix~\ref{app:pipelines}).

\paragraph{Why does the WS layer only match the tuned watershed at full data?}
Because, given the same markers and the same foreground, the two compute the same
partition (Table~\ref{tab:swap}), and the remaining errors are in marker detection,
which both share: with ground-truth markers both reach $F_1$ about 0.87
(Table~\ref{tab:instances}). Every advantage the WS layer showed during the study was
matched once the baseline received the same supervision: the edge target
(Sect.~\ref{sec:binary}, last rows of Table~\ref{tab:binary}) and the distance head
(Sect.~\ref{sec:instances}). Changing the marker head does not change this: with a
CenterNet-style centre-and-radius detector the WS layer stays at $F_1$ 0.795 and the
baseline drops to 0.774 (record section ``A detection head for the markers'').

\paragraph{Why did the oracle with true markers fall from 0.96 to 0.87?}
Because of the hedging. Trained from ground-truth markers, the costs could trust every
marker and the marker attachment overrode the foreground (oracle 0.959). Trained from
the model's own markers, the solve learns to veto a marker where the foreground is
weak and to share contested pixels; with ground-truth markers it then drops 11\,\% of
them, and its oracle (0.872) falls to the watershed's (0.867). The 0.17 $F_1$ between
the earlier layer and its oracle was therefore a property of training on perfect
markers, not headroom (record section ``Training the solve from the model's own markers''). The same hedging explains the layer's PQ
deficit (softer boundaries at equal detection, SQ 0.816 against 0.824) and its loss at
the exact limit, which removes the hedging.

\paragraph{Why can the marker selection not be learned through the layer?}
Because the partition loss reaches a marker only through the pixels it claims. A
spurious marker where the background is clearly stronger claims no pixels at any
strength, so its strength receives no gradient (measured at $10^{-11}$) and is never
pushed down; a true marker whose region spills over the nucleus rim loses rim pixels,
so its strength is pushed down. The strengths therefore learn a rim correction, not a
ranking: the learned strength ranks background candidates above true nuclei (AUC
0.11--0.13). Three designs were tried, strengths from the marker head's own score (which
also degraded the head as a detector, $F_1$ 0.732), a fifth output channel, and a
detached scorer on candidate features, and none beat the same recipe with all
strengths fixed at one (record sections ``Marker selection in the gradient path'' and ``Selection without touching the marker head''). Detection confidences from a
detection head used as fixed strengths do not help either (record section ``A detection head for the markers''). Marker
detection is a different question from pixel partition, and the gradient through the
partition does not answer it.

\paragraph{Is the exact limit the deployed object?}
Not in these experiments. On the instance model, solved exactly at increasing $p$
with costs and markers fixed, accuracy peaks at the training $p=20$ and falls
monotonically to the limit (0.795 to 0.724 $F_1$), which the solve reaches by
$p\approx500$ (Appendix~\ref{app:psweep}). What a higher $p$ removes is the hedging:
discounted markers become objects again. The model trained from true markers is the
mirror image: within 0.006 $F_1$ of its limit, but worse on its own markers. A
deployable limit and tolerance to imperfect markers pulled in opposite directions in
every schedule tried. For the binary models the picture differs: pixel IoU is flat up
to the training $p$ and then loses 0.05 at the limit, while instance $F_1$ keeps rising
to $p\approx200$, because a high $p$ removes the faint blobs that thresholding a soft
output leaves (Table~\ref{tab:psweep} and the record section ``Accuracy against $p$'').

\paragraph{Is there a training regime that lets the combinatorial watershed replace the
WS layer at inference?}
Partly, and the record says what it needs. Two trained models already have a
deployable limit: the binary layer with the edge target trained at $p$ from 5 to 50
(limit 0.016 IoU below the converged solve, no loss on BF$_2$;
Sect.~\ref{sec:binary}) and the multi-marker layer trained from ground-truth markers
at $p$ from 5 to 20 (limit 0.006 $F_1$ below the solve on all of fold~2;
Table~\ref{tab:instances} and the record section ``The exact limit on all of fold
2''). What they share, and what the models with a lost limit lack: costs trained to
cut between every pair of neighbouring markers with the background as a soft seed,
which is what a maximum spanning forest needs (the binary models trained at low $p$
lost their limit entirely); sharp contours from the edge target, which cut the limit's
price from 0.046 to 0.016 IoU; and no hedging, since training from the model's own
markers or in the implicit mode makes the costs rely on finite-$p$ softness that the
limit removes. Those last two are exactly our best finite-$p$ models, so the trade-off
stated in the main text is the obstacle: the tolerance to imperfect markers lives in
the softness, and the limit has none. A regime for the combinatorial watershed must
move that tolerance out of the solve and into the marker rule. Four candidates, none
run, cheapest first: (i) re-tune the marker thresholds for the limit's output on the
tuning fold, since the limit rows reuse thresholds tuned for the solve, which are
permissive because the solve absorbs spurious markers and the limit cannot (a grid
search at 50\,ms per patch, no training; we expect it to recover most of the
0.07 $F_1$, leaving a watershed on learned costs with its own tuned markers, i.e.\
about the baseline's score); (ii) straight-through training, the exact limit in the
forward pass and the Random Walker adjoint at moderate $p$ in the backward pass, so
that the loss sees the deployed partition, hedging cannot be learned, and the gradient
is the relaxation's, which Table~\ref{tab:swap} suggests points the right way;
(iii) a high-$p$ phase at the end of the schedule, limited by the solver (PCG stops
converging beyond $p\approx50$ on these costs, the exact elimination of
Appendix~\ref{app:psweep} takes about 2\,s per patch on one core, enough for a few
fine-tuning epochs); (iv) a margin loss on the ordering of the edge weights, which is
all the limit depends on, i.e.\ the learned-watershed route of Wolf et al.\ with our
path as a warm start. None of these would make the limit better than the finite-$p$
solve on PanNuke, since the solve's advantage is the hedging itself; they would make
``train at finite $p$, deploy at the limit'' a single consistent procedure.

\paragraph{What is the ``diagonal rescaling'' of the reference solve, and is it an
estimate of the largest eigenvalue?}
No; the two uses of the diagonal are different. The reference solve of
Sect.~\ref{sec:cond} factorises the Jacobi-scaled matrix $SAS$ with
$S=\mathrm{diag}(A)^{-1/2}$, so that every diagonal entry of the scaled matrix is one,
solves, maps back with $S$, and then runs two steps of iterative refinement on the
original system. It is needed because at large $p$ the diagonal entries of $A$ are
sums of weights $e^{-pc}$ that span many orders of magnitude from one pixel to the
next, and a sparse LU of the raw matrix loses its pivots to underflow or cancellation;
scaling each row and column by the inverse square root of its diagonal equalises the
magnitudes, and it is the near-optimal diagonal scaling for the condition number (van
der Sluis). The largest diagonal entry appears instead in the bound of
Sect.~\ref{sec:layer}: for a symmetric positive definite matrix, $\lambda_{\max}$ is at
least every diagonal entry (the Rayleigh quotient of a unit vector), which with the
upper bound on $\lambda_{\min}$ from a pixel whose four edges carry the highest cost
gives $\kappa(A)\gtrsim e^{p\,\Delta c}/4$. So the diagonal bounds the condition number
in one place and improves it in the other.

\paragraph{Why does the binary WS layer not beat a thresholded U-Net?}
Because a U-Net trained with cross-entropy for 30 epochs learns a similar sharpening
by itself. Inside the trained model the layer does what it should (its own logits are
speckle, 350 components per patch, which the layer cleans up to the true 24), but the
two-stage system is not better than the one-stage one, on Pet or on PanNuke, under a
plain cross-entropy or under boundary-weighted, seam-weighted or Dice-band losses
(the record's Pet and PanNuke binary sections). Two things did matter: an edge target on the cost head
(Sect.~\ref{sec:binary}) and training in the implicit mode, i.e.\ with the true Random
Walker gradient, which lifts the binary layer to the U-Net's pixel scores
(Appendix~\ref{app:efficiency}). Twenty unrolled conjugate-gradient steps give the
exact gradient of the truncated solver, not of the Random Walker, at $p\ge20$.

\paragraph{So what does training through the watershed buy?}
Three things the experiments support: the assignment step is learned rather than
tuned, so the pipeline has no post-hoc choice of relief or threshold on the
partition; the trained solve tolerates imperfect markers, which shows as the lead
when annotations are scarce and as the best count error of the study; and the whole
model is one differentiable module with a cost affine in the number of labels, so
losses defined on the partition (and, in future work, weak or scribble supervision)
can reach the costs. What it does not buy on PanNuke is accuracy beyond a well-tuned
post-hoc watershed with the same supervision, and the paper says so.

\section{Derivations}\label{app:deriv}

\subsection{Random walks and random forests}\label{app:forests}

Grady reads the seeded Random Walker as follows: $x_i$ is the probability that a
random walk started at pixel $i$, which crosses edge $e$ with probability
proportional to $w_e$, reaches a foreground seed before a background seed.
Sect.~\ref{sec:layer} reads the same $x_i$ as the probability that a random 2-forest
connects $i$ to the foreground node. The two readings are the same harmonic function,
written in two ways; the forest reading is the probabilistic watershed of Fita
Sanmart\'in et al.~\cite{fitasanmartin2019probabilistic}, restated here on the
augmented graph, where the unaries are edges to the label nodes.

Let $G'$ be the augmented graph: the pixel graph, two label nodes $F$ and $B$, and for
every pixel $i$ the edges $(i,F)$ of weight $a^1_i$ and $(i,B)$ of weight $a^0_i$. Let
$L'$ be its Laplacian and $L'_U$ the submatrix on the pixels, i.e.\ $L'$ with the rows
and columns of $F$ and $B$ removed, so that $L'_U=L(w)+\mathrm{diag}(a^0+a^1)$ and
system~\eqref{eq:system} reads $L'_U\,x=a^1$.

\emph{Random walk.} Let $h_i$ be the probability that the walk from $i$ reaches $F$
before $B$. Then $h_F=1$, $h_B=0$, and at every pixel the first step gives
$h_i=\sum_{j\sim i} w_{ij}h_j/\sum_{j\sim i} w_{ij}$, the sum running over the pixel
neighbours and the two label nodes. This says $(L'h)_i=0$ at every pixel with the
boundary values on $F$ and $B$, i.e.\ $L'_U\,h=a^1$. The system is non-singular (it is
diagonally dominant with a strictly dominant row wherever a unary is positive), so
$h=x$.

\emph{Random 2-forest.} By Cramer's rule, $x_i=\det L'_U[i\leftarrow a^1]/\det L'_U$,
where $[i\leftarrow a^1]$ replaces column $i$ by the right-hand side. The all-minors
matrix-tree theorem~\cite{chaiken1982combinatorial} evaluates both determinants.
$\det L'_U$ is the minor of $L'$ with the vertices $F$ and $B$ deleted: it equals the sum,
over the spanning forests of $G'$ with exactly two trees, one containing $F$ and the
other $B$, of the product of the edge weights of the forest. The numerator is, up to a
sign that cancels in the ratio, the minor of $L'$ with rows $\{F,B\}$ and columns
$\{i,B\}$ deleted, which the same theorem evaluates as the sum over those 2-forests in
which $i$ lies in the tree of $F$. Hence
\begin{equation}
x_i=\frac{\sum_{\mathcal F:\,i\sim F} \prod_{e\in\mathcal F} w_e}
         {\sum_{\mathcal F}\prod_{e\in\mathcal F} w_e},
\end{equation}
the probability that $i$ is attached to $F$ when a 2-forest is drawn with probability
proportional to its weight. With $w_e=e^{-pc_e}$ and $a=e^{-pc}$ the weight of a forest
is $e^{-p\sum_{e\in\mathcal F}c_e}$: the distribution is the Gibbs distribution over
2-forests with energy the total cost of the forest, at inverse temperature $p$. As
$p\to\infty$ it concentrates on the forests of minimum total cost, i.e.\ on the
maximum-weight spanning forests separating $F$ from $B$, which is the Power Watershed;
when several forests tie, the measure is uniform over them, and that uniform average
is what the plateau solve of the \PW\ algorithm computes.

With $K$ labels the same theorem, applied with the $K+1$ label nodes as the deleted
set, gives $x^k_i$ as the probability that $i$ lies in the tree of label $k$ when a
$(K{+}1)$-forest, each tree containing exactly one label node, is drawn with probability
proportional to its weight. Since every pixel lies in exactly one tree, the $x^k_i$
sum to one, which is the combinatorial form of Appendix~\ref{app:sumone}.

\subsection{The adjoint gradient}

Take $E(x,w)=\sum_{e=\{u,v\}} w_e (x_u-x_v)^2+\sum_i a^1_i(x_i-1)^2+\sum_i a^0_i x_i^2$.
Stationarity gives $A x=a^1$ with $A=L(w)+\mathrm{diag}(a^0+a^1)$. Differentiating with
respect to one parameter $\theta$, $A\,\partial x/\partial\theta
= \partial a^1/\partial\theta-(\partial A/\partial\theta)\,x$, so for a loss $\ell(x)$
and $\lambda=A^{-1}\partial\ell/\partial x$ (using the symmetry of $A$),
$\partial\ell/\partial\theta=\lambda^\top(\partial a^1/\partial\theta-(\partial
A/\partial\theta)x)$. For $\theta=w_e$, $\partial A/\partial w_e$ is the elementary edge
Laplacian, whose product with $x$ has entries $x_u-x_v$ at $u$ and $x_v-x_u$ at $v$,
giving $\partial\ell/\partial w_e=-(x_u-x_v)(\lambda_u-\lambda_v)$. For $\theta=a^1_i$,
$\partial a^1/\partial a^1_i=e_i$ and $(\partial A/\partial a^1_i)x=x_i e_i$, giving
$\lambda_i(1-x_i)$; for $\theta=a^0_i$ only the second term survives, giving
$-\lambda_i x_i$. The chain rule through $w_e=e^{-pc_e}$ contributes $-p w_e$, whose
growth in $p$ is the algebraic counterpart of the conditioning blow-up.

\subsection{Multi-label solutions sum to one}\label{app:sumone}

With $K$ labels, $A_K=L(w)+\mathrm{diag}(\sum_k a^k)$ and $A_K x^k=a^k$. Summing over
$k$, $A_K\sum_k x^k=\sum_k a^k=A_K\mathbf 1$ since $L\mathbf 1=0$; $A_K$ is positive
definite, hence $\sum_k x^k=\mathbf 1$.

\subsection{The exact limit with $K$ labels}

The plateau algorithm of Couprie et al.\ processes the edges by decreasing weight and
merges the nodes of a plateau unless the plateau connects nodes with two or more
distinct known labels, in which case the $q=2$ energy is minimised on the plateau with
the known nodes as boundary. For two labels the boundary values are scalars; for $K$
labels we solve one system per label with one-hot boundary values and assign the
argmax, which is the limit of the multi-label \RW\ on the plateau. Singleton plateaus
(continuous costs) are processed by a scalar union-find; plateaus of at most 64 edges
by a pure-Python union-find; larger plateaus by the vectorised path.

\subsection{Fixed-point reweighting}\label{sec:reweight}

The following variant of the WS layer is studied in the experimental record below
(sections on reweighting) but is not used in the instance pipeline of the main text.

Instead of raising the weights to a growing $p$, one can keep a benign base
$p_0$ and iterate, multiplying each weight by a factor of the current solution,
\begin{equation}
  w_e\leftarrow w^0_e\,\rhoc(x_u-x_v),\quad
  a^1_i\leftarrow a^{1,0}_i\,\rhoc(x_i-1),\quad
  a^0_i\leftarrow a^{0,0}_i\,\rhoc(x_i),\qquad
  \rhoc(d)=\frac{\varepsilon^2}{\varepsilon^2+d^2},
  \label{eq:reweight}
\end{equation}
and re-solving~\eqref{eq:system} until the pair (weights, solution) stops changing.
This is the half-quadratic / majorise-minimise iteration of a nonconvex
piecewise-constant energy with the Geman--McClure
potential~\cite{geman1992constrained,nikolova2005analysis}; the unaries are
reweighted too because they are edges of the augmented graph. The weights stay in
$(0,1]$ and the smallest factor is set by $\varepsilon$, not by $p$ times the cost
spread, and the tiny weights appear only on the edges across which the solution
jumps. The fixed point is not the \PW: it is a local minimiser that depends on the
starting point, which for a finishing layer is the network's own prediction. Inner
solves are implicit and the outer loop is differentiated by automatic differentiation
(checked against finite differences).

\subsection{Validation against the reference implementation}

Our NumPy implementation of the plateau algorithm was checked against the authors'
reference code (PW\_1.0.1, \url{https://sourceforge.net/projects/powerwatershed}) on
the same 4-connected graphs, integer weights and seeds: on 80 random coarsely quantised
images with two to five labels and on thirteen $512^2$ cases (synthetic, camera,
astronaut; two- and multi-label) the potentials agree to single precision (maximum
difference $6\cdot10^{-8}$, the reference stores them as float) and the labels agree on
every pixel once the same rule is applied to potentials exactly equal to 0.5. The
reference realises the plateaus of the merged graph through a geodesic reconstruction
of the weights from the seeds; without it, its result depends on the order in which
equal weights are sorted. The harness and the records are in the code repository
(\path{experiments/pwref/}, \path{results/pwref/}).

\section{Architectures}\label{app:arch}

% Architecture diagrams (TikZ).  Included from appendix.tex.
% Requires: \usepackage{tikz} \usetikzlibrary{positioning,arrows.meta,fit,calc}
\tikzset{
  arch/.style={font=\footnotesize, node distance=4mm and 5mm, >={Stealth[length=2mm]}},
  blk/.style={draw, rounded corners=2pt, minimum height=7mm, inner sep=3pt, align=center, fill=gray!8},
  net/.style={blk, fill=blue!8},
  ws/.style={blk, fill=orange!15},
  post/.style={blk, fill=red!8, dashed},
  dat/.style={align=center, inner sep=2pt},
  lossn/.style={draw, rounded corners=2pt, inner sep=2pt, align=center, font=\scriptsize, fill=green!8},
  lbl/.style={font=\scriptsize, align=center},
  arr/.style={->},
  grad/.style={->, dashed, gray},
}

% ---------------------------------------------------------------- (a) plain U-Net
\newcommand{\archPlain}{%
\begin{tikzpicture}[arch]
  \node[dat] (img) {image\\$3{\times}H{\times}W$};
  \node[net, right=of img] (unet) {U-Net\\32--64--128--256};
  \node[dat, right=of unet] (logit) {logit\\$1{\times}H{\times}W$};
  \node[blk, right=of logit] (sig) {sigmoid};
  \node[blk, right=of sig] (thr) {threshold\\at 0.5};
  \node[dat, right=of thr] (mask) {mask};
  \node[post, right=of mask] (cc) {connected\\components};
  \node[dat, right=of cc] (inst) {instances};
  \node[lossn, below=of sig] (loss) {BCE loss};
  \draw[arr] (img) -- (unet); \draw[arr] (unet) -- (logit); \draw[arr] (logit) -- (sig);
  \draw[arr] (sig) -- (thr); \draw[arr] (thr) -- (mask); \draw[arr] (mask) -- (cc); \draw[arr] (cc) -- (inst);
  \draw[arr] (sig) -- (loss);
\end{tikzpicture}}

% ---------------------------------------------------------------- (b) binary WS layer
\newcommand{\archBinary}{%
\begin{tikzpicture}[arch, x=1cm, y=1cm]
  \node[dat] (img) at (0,0) {image};
  \node[net] (unet) at (2.0,0) {U-Net\\last $1{\times}1$ conv:\\3 channels};
  \node[dat] (logit) at (5.0,1.0) {ch.\ 0: logit $s$};
  \node[dat] (emap) at (5.0,-1.0) {ch.\ 1--2: edge maps,\\sigmoid, averaged\\onto h/v edges};
  \node[blk] (una) at (8.6,1.0) {unaries\\$a^1=s^{p\alpha}$,\ $a^0=(1-s)^{p\alpha}$};
  \node[blk] (wts) at (8.6,-1.0) {weights $w_e=e^{-p\,c_e}$};
  \node[ws] (solve) at (13.2,0) {WS layer\\solve $(L(w)+\mathrm{diag}(a^0{+}a^1))\,x=a^1$\\unrolled: $K$ PCG steps\\implicit: to tolerance $+$ adjoint};
  \node[dat] (x) at (17.0,0) {potentials\\$x^p\in[0,1]$};
  \node[lossn] (loss) at (17.0,-1.5) {BCE on $x^p$\\$+\,0.5$ BCE on $s$};
  \node[post] (lim) at (13.2,-2.4) {deployment: exact PW on the same weights\\(spanning forest, $p=\infty$)};
  \node[lbl] at (13.2,1.35) {$p$ annealed $5\to50$ during training};
  \draw[arr] (img) -- (unet);
  \draw[arr] (unet.east) -- (logit.west); \draw[arr] (unet.east) -- (emap.west);
  \draw[arr] (logit) -- (una); \draw[arr] (emap) -- (wts);
  \draw[arr] (una.east) -- (solve.west |- una.east); \draw[arr] (wts.east) -- (solve.west |- wts.east);
  \draw[arr] (solve) -- (x); \draw[arr] (x) -- (loss);
  \draw[arr, gray] (wts.south) |- (lim.west);
  \draw[grad] (loss.south) -- ++(0,-1.8) -| (unet.south) node[lbl, pos=0.3, below=1mm] {gradients to the edge maps, the logit and $\alpha$};
\end{tikzpicture}}

% ---------------------------------------------------------------- (c) multi-marker WS layer
\newcommand{\archMulti}{%
\begin{tikzpicture}[arch, x=1cm, y=1cm]
  \node[dat] (img) at (0,0) {image};
  \node[net] (unet) at (2.0,0) {U-Net\\last $1{\times}1$ conv:\\4 channels};
  \node[dat] (logit) at (5.4,0.8) {ch.\ 0: foreground logit\\$\to$ background unary};
  \node[dat] (emap) at (5.4,-0.3) {ch.\ 1--2: edge costs\\$\to$ weights $w_e=e^{-p c_e}$};
  \node[dat] (mk) at (5.4,-1.5) {ch.\ 3: marker map\\(BCE head or distance head)};
  \node[post] (peaks) at (10.6,-1.6) {shared marker extraction:\\$h$-maxima of $d$ grown\\into basins $\to$ markers $1,\dots,K$};
  \node[ws] (solve) at (15.8,0) {multi-marker WS layer\\one right-hand side per marker\\$(L(w)+\mathrm{diag}\sum_k a^k)\,x^k=a^k$,\ $\sum_k x^k=1$};
  \node[dat] (part) at (20.2,0) {partition\\$\arg\max_k x^k$};
  \node[lossn] (loss) at (20.2,-1.6) {cross-entropy over\\this image's markers\\vs.\ the instance mask};
  \node[lbl] at (15.8,1.25) {$p$ annealed $5\to20$};
  \node[lbl] at (10.6,-2.75) {training markers: the true instances eroded,\\then the model's own markers (record, sec.\ 11.5)};
  \draw[arr] (img) -- (unet);
  \draw[arr] (unet.east) -- (logit.west); \draw[arr] (unet.east) -- (emap.west); \draw[arr] (unet.east) -- (mk.west);
  \draw[arr] (mk) -- (peaks);
  \draw[arr] (logit.east) -- (solve.west |- logit.east);
  \draw[arr] (emap.east) -- (solve.west |- emap.east);
  \draw[arr] (peaks.east) -- (solve.west |- peaks.east);
  \draw[arr] (solve) -- (part); \draw[arr] (part) -- (loss);
\end{tikzpicture}}

% ---------------------------------------------------------------- (d) post-processing baseline
\newcommand{\archBaseline}{%
\begin{tikzpicture}[arch, x=1cm, y=1cm]
  \node[dat] (img) at (0,0) {image};
  \node[net] (unet) at (2.2,0) {U-Net\\foreground logit\\$+$ distance head\\($+$ contour head)};
  \node[dat] (fg) at (5.6,0.9) {foreground $s$};
  \node[dat] (dist) at (5.6,-0.9) {distance map $d$,\\contour map $c$};
  \node[blk] (thr) at (9.4,1.0) {threshold at 0.5};
  \node[post] (pk) at (9.4,-1.0) {shared marker extraction:\\$h$-maxima of $d$ grown into basins};
  \node[post] (wsh) at (14.2,0) {marker watershed\\on the relief $c$ (or $-d$)\\inside the mask};
  \node[dat] (inst) at (17.4,0) {instances};
  \node[lossn] (loss) at (2.2,-2.0) {BCE on $s$, losses on $d$ and $c$};
  \node[lbl] at (14.2,-1.45) {outside the training loop;\\threshold and spacing tuned on fold 3};
  \draw[arr] (img) -- (unet);
  \draw[arr] (unet.east) -- (fg.west); \draw[arr] (unet.east) -- (dist.west);
  \draw[arr] (fg) -- (thr); \draw[arr] (dist) -- (pk);
  \draw[arr] (thr.east) -- (wsh.west |- thr.east); \draw[arr] (pk.east) -- (wsh.west |- pk.east);
  \draw[arr] (wsh) -- (inst);
  \draw[arr] (unet.south) -- (loss.north);
\end{tikzpicture}}

Figures~\ref{fig:arch} and~\ref{fig:arch2} draw the four pipelines compared in the main text. In every
case the convolutional part is the same small U-Net (32--64--128--256 channels); the
variants differ only in the number of output channels of its last $1{\times}1$
convolution and in what consumes them. Blue: the network; orange: the WS layer, which
has no trainable parameters of its own except, in later runs, one scale on the edge
costs; red, dashed: operations outside the training loop; green: the losses.

\begin{sidewaysfigure}
\centering
\resizebox{0.8\textwidth}{!}{\archPlain}\\[1mm]
{\footnotesize (a) Plain U-Net: one logit, sigmoid, threshold; instances by connected components.}\\[5mm]
\resizebox{\textwidth}{!}{\archBinary}\\[1mm]
{\footnotesize (b) Binary WS layer: a three-channel head gives the unaries and the edge costs; the layer solves the RW system at the annealed temperature $p$; the exact PW is the deployment option.}
\caption{The pipelines of the binary experiments (Sect.~\ref{sec:binary}).}
\label{fig:arch}
\end{sidewaysfigure}

\begin{sidewaysfigure}
\centering
\resizebox{\textwidth}{!}{\archMulti}\\[1mm]
{\footnotesize (c) Multi-marker WS layer: a fourth channel predicts the marker map; its peaks are this image's labels, one right-hand side each.}\\[5mm]
\resizebox{\textwidth}{!}{\archBaseline}\\[1mm]
{\footnotesize (d) Post-processing baseline: the same U-Net with a distance (and contour) head, thresholded, then a marker watershed run outside the training loop.}
\caption{The pipelines of the instance experiments (Sect.~\ref{sec:instances}).}
\label{fig:arch2}
\end{sidewaysfigure}

\section{The instance pipelines in detail}\label{app:pipelines}

This section describes, as precisely as we can, what the WS layer and its strongest
baseline do in the instance experiments of Sect.~\ref{sec:instances}, which parts they
share, and where they differ. Figure~\ref{fig:pipelines} follows both on the same test
patches.

\paragraph{What is shared.} Both models are the same U-Net (32--64--128--256 channels)
with a four-channel last layer, trained with the same recipe: fold~1, 30 epochs, batch~16,
the eight flips and rotations as augmentation, Adam with learning rate $10^{-3}$ and cosine
decay, one seed. The four channels and their direct supervision are identical:
\begin{itemize}
\item channel 0, the foreground logit, whose sigmoid $s$ is trained with a binary
cross-entropy against the nucleus mask;
\item channels 1--2, the contour (or edge-cost) head, trained against the map of instance
boundaries derived from the instance masks (the edge target of Sect.~\ref{sec:binary}),
seams between touching nuclei weighted five times;
\item channel 3, the distance head of Naylor et al.~\cite{naylor2019distance}: for every
nucleus pixel the Chebyshev distance to the nucleus boundary, capped at 12~px and divided by
the nucleus's maximum, so that every nucleus peaks at~1; sigmoid output, squared error.
\end{itemize}
Test markers are extracted from the predicted distance map $d$ by the same procedure for
both models:
\begin{enumerate}
\item seeds are the regional maxima of $d$ of prominence at least $h$ (the $h$-maxima), kept
where $d>t$, one seed per plateau;
\item each seed is grown into a basin by a watershed of $-d$ inside $\{d>t\}$, so that two
touching nuclei whose maxima are separated by a valley deeper than $h$ get two markers even
when the valley stays above $t$;
\item basins smaller than $a$ pixels are dropped.
\end{enumerate}
The result is a label map with one connected marker region per detected nucleus. The three
parameters are chosen separately for each model by the panoptic quality of its final output
on 300 evenly spaced patches of fold~3, which is used neither for training nor for testing.
Both procedures are therefore identical; the markers are not, because each model predicts
its own $d$ and has its own tuned parameters (Table~\ref{tab:pipelines}).

\paragraph{The baseline's final step.} A marker watershed (4-connected) of a relief $r$ from
the markers, inside the foreground mask $\{s>0.5\}$; foreground components that contain no
marker are left unlabelled. The relief is chosen on fold~3 among the contour map $c$ (mean
of the two contour channels), $-d$, and $c-d$; the choice hardly matters (PQ 0.654--0.655
on fold~2 for the three). This is the post-processing of HoVer-Net~\cite{graham2019hovernet}
applied to Naylor et al.'s distance map; it receives no gradient. Keeping the unmarked
foreground components as extra instances lowers F$_1$ to 0.780 and PQ to 0.637 and raises
AJI to 0.666, so they are dropped.

\paragraph{The WS layer's final step.} The multi-marker Random Walker of
Sect.~\ref{sec:layer} at the final temperature $p=20$, solved to convergence: every marker
region $k$ attaches its pixels to label $k$ with the unary $a^k=s^{p}$, the background
label attaches every pixel with $(1-s)^{p}$ (the costs $-\log s$ and $-\log(1-s)$ clipped at 8), the edge weights are $e^{-p\,c_e}$ with $c_e$
the edge costs of channels 1--2 (times a learned scale), and every pixel takes the label of
its largest probability $x^k$. During training the same solve is differentiated, the loss
being the per-pixel cross-entropy of the label probabilities against the instance map (seam
pixels weighted six times), added to the three direct losses above; $p$ is annealed from 5
to 20. The training markers are the ground-truth nuclei eroded by 2--3~px for the first ten
epochs, then a mix, and the model's own extracted markers from epoch 15; nuclei that receive
no marker are given to the background label in the loss, as they would be at test time.
The extraction is thus inside the training loop but outside the gradient: the solve's
gradient reaches the edge-cost channels through $w$ and the foreground channel through the
unaries $a^k=s^p$ and $a^0=(1-s)^p$, never the distance channel, which learns only from its
own squared error. A variant that puts the selection in the gradient path (every regional
maximum above $t=0.1$ a candidate, with a strength $\sigma_k$ computed from the mean
distance logit over the candidate by a sigmoid with learned offset and temperature, entering
as $a^k=(\sigma_k s)^p$, duplicates and background candidates pushed towards zero strength
by the cross-entropy) lost 0.063 F$_1$ and 0.048 PQ against this recipe on fold~2: the
strengths barely separated true from spurious candidates, and the gradient reaching the
distance channel doubled the share of unmarked nuclei (26\,\% against 12.7\,\%; extended report, Sect.~14).

\begin{table}[h]
\centering\footnotesize\setlength{\tabcolsep}{4pt}
\caption{Marker extraction chosen on fold~3, and the markers it produces on fold~2.
``One / none / several'': the fraction of ground-truth nuclei covered by exactly one, no,
or several markers; ``background'': markers per patch that touch no nucleus.}
\label{tab:pipelines}
\resizebox{\textwidth}{!}{%
\begin{tabular}{lll}
\toprule
 & U-Net + marker watershed & U-Net + WS layer\\
\midrule
grid searched ($t$ / $a$ / $h$) & 0.1--0.5 / 4, 9, 16 / 0.1--0.3 & 0.2--0.7 / 1, 4, 9, 16 / 0.05--0.2\\
chosen $t$, $a$, $h$ & 0.5, 4, 0.2 & 0.2, 4, 0.2\\
final step & watershed of $c-d$ in $\{s>0.5\}$ & multi-marker RW, $p=20$\\
markers: one / none / several & 76.6 / 21.8 / 1.6\,\% & 85.5 / 12.7 / 1.8\,\%\\
background markers per patch & 1.6 & 4.3\\
\bottomrule
\end{tabular}}
\end{table}

The baseline's tuning picks a high threshold: it misses a fifth of the nuclei but keeps
spurious markers rare, because its watershed turns every marker inside the foreground into
an object. The WS layer's tuning picks a low one: it detects more nuclei and lets its solve,
trained on such markers, absorb the spurious ones. An extended grid ($t$ from 0.4 to 0.7)
gives the baseline the same choice; the WS layer's $t=0.2$ is the lowest value of its grid,
and for the earlier distance-head model of the record an extended grid down to $t=0.1$
still chose $0.2$.

\paragraph{Separating the markers from the final step.} To compare the two final steps
with everything else held fixed, and in particular without any HoVer-Net-style
post-processing on the WS layer's side, the two models were crossed
(Table~\ref{tab:swap}): each model's markers through each final step, all on fold~2.

\begin{table}[h]
\centering\footnotesize\setlength{\tabcolsep}{4pt}
\caption{Markers $\times$ final step on fold~2 (2523 patches). Rows 1 and 4 are the two
methods of Table~\ref{tab:instances}; row 2 replaces the WS layer's solve by a plain marker
watershed inside the WS layer's own output foreground; row 3 uses the WS layer's markers in
the baseline's watershed and foreground.}
\label{tab:swap}
\resizebox{\textwidth}{!}{%
\begin{tabular}{llrrrrrr}
\toprule
markers & final step & IoU & F$_1$@0.5 & PQ & SQ & AJI & count err.\\
\midrule
WS layer's & WS layer's solve & 0.715 & 0.795 & 0.649 & 0.816 & 0.668 & 0.182\\
WS layer's & watershed of the WS layer's contour map, in its foreground & 0.715 & 0.795 & 0.649 & 0.816 & 0.668 & 0.182\\
WS layer's & baseline's watershed of $c-d$ in $\{s>0.5\}$ & 0.726 & 0.780 & 0.638 & 0.818 & 0.674 & 0.203\\
baseline's & baseline's watershed of $c-d$ in $\{s>0.5\}$ & 0.719 & 0.794 & 0.655 & 0.824 & 0.658 & 0.190\\
baseline's & WS layer's solve & 0.620 & 0.705 & 0.576 & 0.817 & 0.597 & 0.291\\
\bottomrule
\end{tabular}}
\end{table}

Three things follow. First, given the WS layer's own markers and its own output foreground,
a plain marker watershed reproduces the WS layer to three decimals on every metric (row 2).
This is largely built in: the foreground, hence the IoU and the number of objects, is held
fixed, so the two can only differ in how they split touching nuclei, and at $p=20$ they
split them the same way. At test time, then, the WS layer differs from the baseline through
its foreground (which pixels the background label wins) and its markers, not through the
split itself. Second, the WS layer's markers in the baseline's watershed and foreground
(row 3) lose 0.015 F$_1$ and 0.011 PQ against the WS layer and gain 0.006 AJI: the baseline's
watershed turns the WS layer's extra, spurious markers into objects, which the WS layer's
solve partly absorbs. Third, the baseline's markers in the WS layer's solve (row 5) do
poorly (F$_1$ 0.705), because the solve was trained with the WS layer's own denser markers
and gives every nucleus without a marker to the background; the two halves of each method
are tuned to each other.

\begin{figure}[h]
\centering
\includegraphics[width=\textwidth]{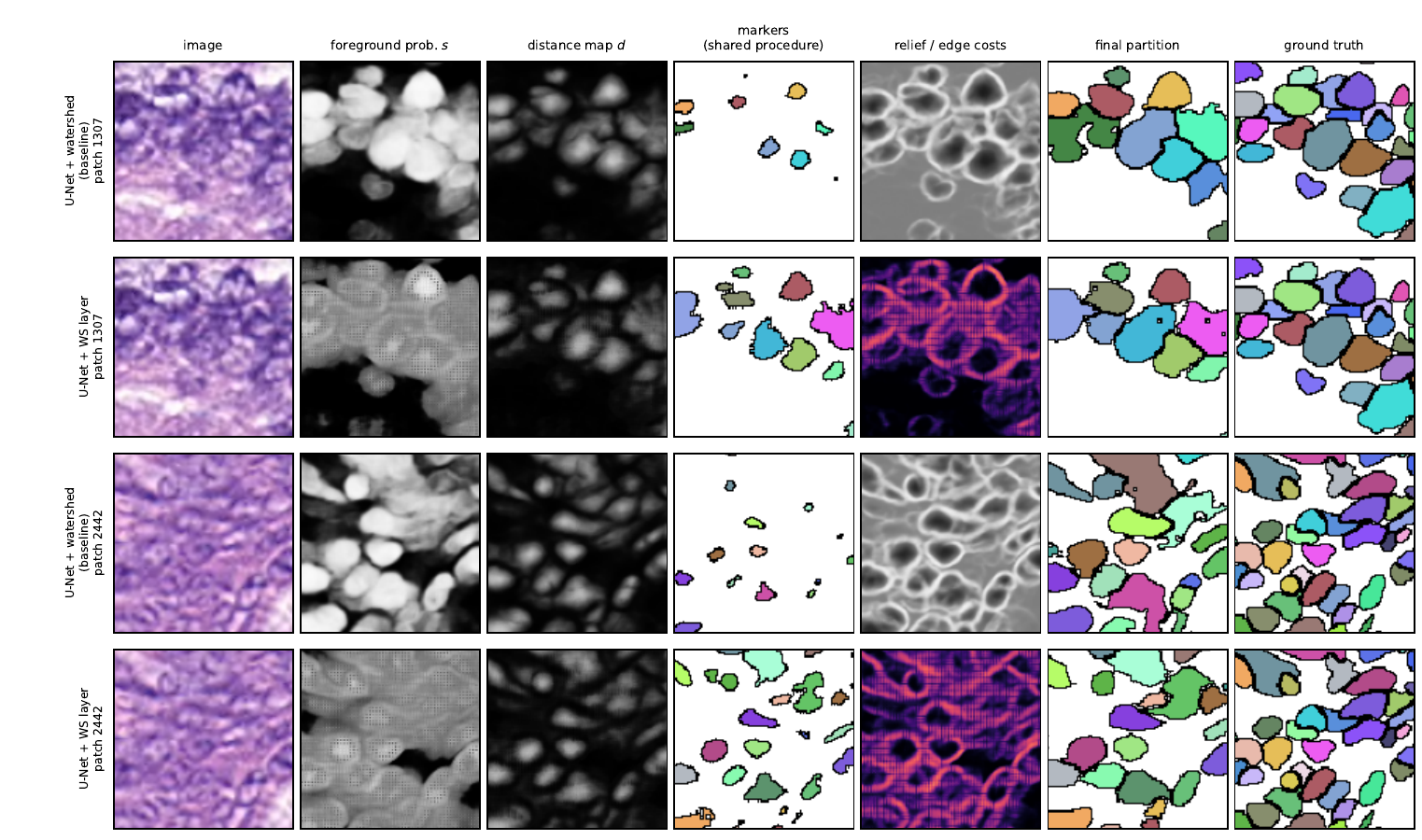}
\caption{The two instance pipelines step by step on two fold-2 patches (112-px crops around
touching nuclei). For each patch, the baseline (upper row) and the WS layer (lower row): the
image, the foreground probability $s$, the distance map $d$, the markers extracted from $d$
by the shared procedure (each model with its own tuned parameters), the baseline's relief
$c-d$ and the WS layer's learned edge costs, the final partition, and the ground truth.}
\label{fig:pipelines}
\end{figure}

\section{Data efficiency}\label{app:dataeff}

The multi-marker WS layer (trained from its own predicted markers, unmarked nuclei sent
to the background) and the post-processing baseline (U-Net with a distance head and a
marker watershed, tuned on fold~3) were trained on nested random subsets of fold~1 of
10, 25 and 50\,\% (subset seed~0), with the number of epochs scaled so that every run
makes the same number of gradient steps as the 30-epoch run on the full fold, and
evaluated on all 2523 patches of fold~2 (Fig.~\ref{fig:dataeff}). The last column gives
the paired difference, layer minus baseline, with 95\,\% bootstrap intervals.

\begin{table}[h]
\centering\footnotesize\setlength{\tabcolsep}{4pt}
\caption{Training-set size against F$_1$@0.5 / PQ / AJI on fold~2.}
\label{tab:dataeff}
\resizebox{\textwidth}{!}{%
\begin{tabular}{lllll}
\toprule
training set & WS layer & U-Net + dist.\ head + watershed & U-Net + components & layer $-$ baseline, F$_1$ / PQ\\
\midrule
10\,\% (266) & 0.765 / 0.620 / 0.635 & 0.745 / 0.605 / 0.620 & 0.622 / 0.500 & $+0.020$ [$+0.017$, $+0.023$] / $+0.015$ [$+0.013$, $+0.017$]\\
25\,\% (664) & 0.785 / 0.640 / 0.659 & 0.789 / 0.646 / 0.662 & 0.653 / 0.530 & $-0.004$ [$-0.007$, $-0.001$] / $-0.006$ [$-0.008$, $-0.004$]\\
50\,\% (1328) & 0.787 / 0.641 / 0.663 & 0.794 / 0.653 / 0.666 & 0.659 / 0.536 & $-0.007$ [$-0.009$, $-0.004$] / $-0.012$ [$-0.014$, $-0.010$]\\
100\,\% (2656) & 0.795 / 0.649 / 0.668 & 0.794 / 0.655 / 0.658 & 0.664 / 0.539 & $+0.001$ [$-0.002$, $+0.004$] / $-0.005$ [$-0.007$, $-0.003$]\\
\bottomrule
\end{tabular}}
\end{table}

\begin{figure}[h]
\centering
\includegraphics[width=\textwidth]{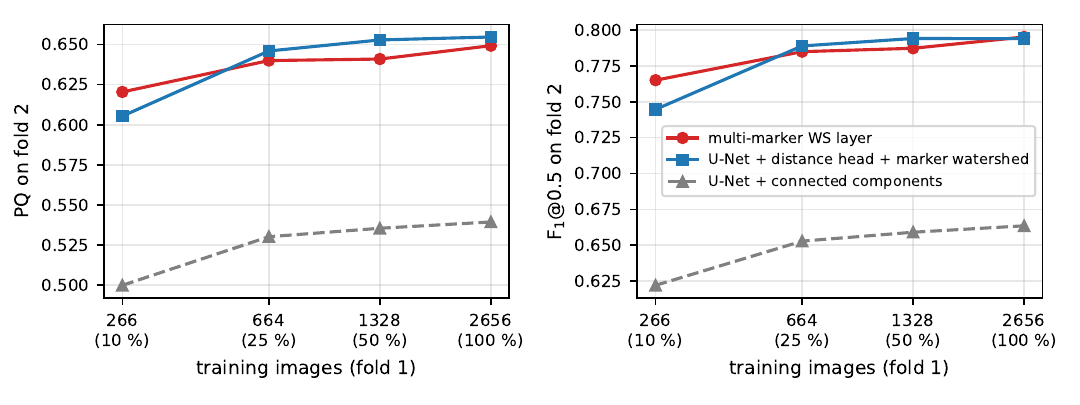}
\caption{PQ and F$_1$@0.5 on fold~2 against the number of training images from fold~1.}
\label{fig:dataeff}
\end{figure}

The WS layer degrades more gracefully only at 10\,\%: it loses 0.030 F$_1$ and 0.029 PQ
with respect to the full fold where the baseline loses 0.049 and 0.049, and it leads on
every metric; the margin is 2.5 times the seed spread of the layer and was reproduced on
a second GPU to 0.001. The gain is in detection, not in boundaries (equal segmentation
quality): the baseline's tuning on fold~3 raises its marker threshold and leaves
24.5\,\% of the nuclei without a marker, whereas the layer keeps a low threshold and its
solve, trained on its own markers, absorbs the spurious ones. From 25\,\% on, the
baseline is ahead again by 0.004--0.007 F$_1$ and 0.006--0.012 PQ, as with the full
fold.

The 10\,\% point was replicated for both methods with a second seed on the same subset
and with a second random draw of 266 images (Table~\ref{tab:dataeff10}). The margin is
stable, $0.017\pm0.003$ F$_1$ and $0.013\pm0.002$ PQ, with the WS layer at F$_1$ 0.765 in
every run, and the mechanism is the same in each replicate: the baseline leaves about
24\,\% of the nuclei unmarked, the WS layer 14--16\,\%, at equal segmentation quality. The
larger fractions were run once. One caveat remains: the baseline at 10\,\% over-fits
mildly under the fixed number of steps, and early stopping could recover part of the
margin.

\begin{table}[h]
\centering\footnotesize\setlength{\tabcolsep}{4pt}
\caption{The 10\,\% point replicated: F$_1$@0.5 / PQ on fold~2 and the paired difference,
WS layer minus baseline, with 95\,\% bootstrap intervals.}
\label{tab:dataeff10}
\resizebox{\textwidth}{!}{%
\begin{tabular}{llll}
\toprule
run & WS layer & baseline & layer $-$ baseline, F$_1$ / PQ\\
\midrule
seed 0, draw 0 & 0.765 / 0.620 & 0.745 / 0.605 & $+0.020$ [$+0.017$, $+0.023$] / $+0.015$ [$+0.013$, $+0.017$]\\
seed 1, draw 0 & 0.765 / 0.620 & 0.749 / 0.609 & $+0.015$ [$+0.012$, $+0.018$] / $+0.011$ [$+0.009$, $+0.014$]\\
seed 0, draw 1 & 0.765 / 0.624 & 0.749 / 0.612 & $+0.017$ [$+0.014$, $+0.020$] / $+0.012$ [$+0.010$, $+0.014$]\\
\bottomrule
\end{tabular}}
\end{table}

\section{Cost and accuracy of the two differentiation modes}\label{app:efficiency}

This section supports the efficiency statements of Sect.~\ref{sec:layer}. All timings are
medians of seven runs after two warm-up runs on one NVIDIA A100 (40\,GB, PyTorch 2.11,
single precision), on $256^2$ fold-2 patches; memory is the peak GPU allocation above
the inputs. $T$ is the number of unrolled steps and $K$ the number of markers.

\paragraph{Gradient accuracy.} We compare, on eight fold-2 patches and the binary WS
layer with learned costs (Sect.~\ref{sec:binary}), the gradient of the binary
cross-entropy with respect to the edge costs and to the unaries against a reference:
automatic differentiation through a double-precision PCG run to a relative residual of
$10^{-12}$. Central finite differences on the converged solve agree with that reference
to a median relative error below $10^{-8}$. Figure~\ref{fig:effgrad} gives the median
relative error. The unrolled mode computes the exact gradient of the \emph{truncated}
solver, as its double-precision curve shows, but that is not the gradient of the \RW. At
$p=5$, $T=20$ steps are already close (cosine 0.993); at $p=20$ and $p=50$ the same 20
steps leave a relative error of 0.9 (cosine 0.75 and 0.67), and $T=50$ and $T=200$
respectively are needed to fall below 5\,\%. The implicit mode at the training
tolerance ($10^{-4}$, at most 100 iterations) stays within $1.6\cdot10^{-3}$ at every
$p$, with 30, 62 and 92 forward iterations at $p=5$, 20 and 50 and as many adjoint ones.

\begin{figure}[h]
\centering
\includegraphics[width=\textwidth]{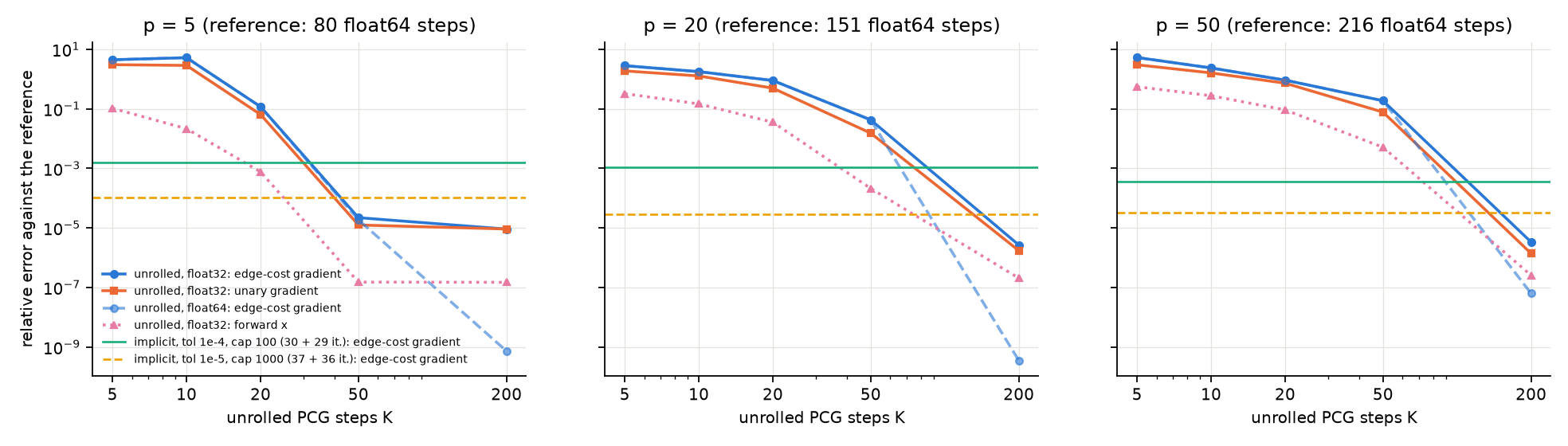}
\caption{Relative error of the gradient against a double-precision reference, as a
function of the number of unrolled steps $T$, at $p=5$, 20 and 50. Horizontal lines: the
implicit mode at two tolerances.}
\label{fig:effgrad}
\end{figure}

\paragraph{Time and memory, binary layer.} Unrolling stores every step for the backward
pass: for a batch of 16 its memory grows by 24\,MB per step (0.55\,GB at $T=20$,
4.9\,GB at $T=200$) whatever $p$, while the implicit mode needs a constant 0.09\,GB
(Fig.~\ref{fig:effcost}). In time, forward plus backward, the implicit mode at $p=20$
costs as much as about 45 unrolled steps, at $p=50$ as much as about 65.

\begin{figure}[h]
\centering
\includegraphics[width=\textwidth]{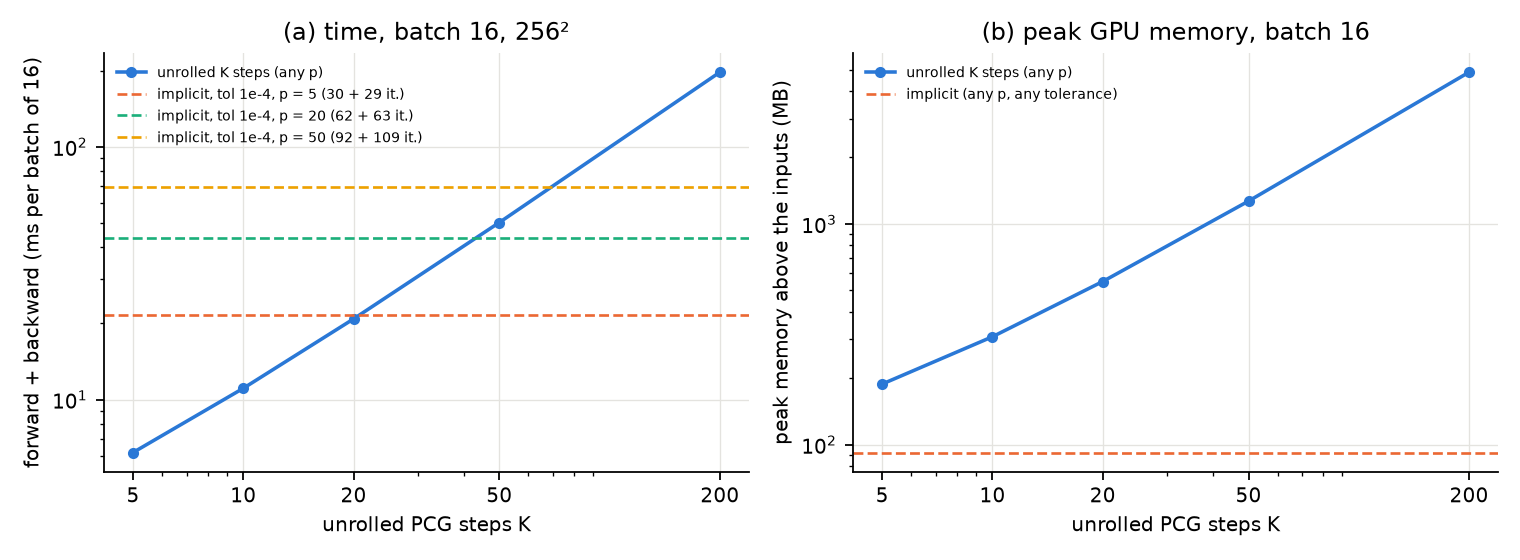}
\caption{Binary WS layer, batch of 16: forward plus backward time and peak memory of $T$
unrolled steps against the implicit mode.}
\label{fig:effcost}
\end{figure}

\paragraph{The multi-marker layer.} Figure~\ref{fig:effk} times the instance layer at
$p=20$ on 16 fold-2 patches with $K$ predicted markers kept (27 on average when all are
kept). In the training setting (tolerance $10^{-4}$, at most 100 iterations; the cap is
reached on these patches) the forward pass of a batch of 16 costs $41+20.3K$\,ms for
$K\ge8$, affine to within 0.5\,\%, and the backward pass the same; at batch size 1 the
time is flat up to $K=16$, since the GPU is then bound by kernel launches. Memory is
4.8\,MB per marker and image, 9.8\,GB for a batch of 16 at $K=128$. The converged
setting used at evaluation ($10^{-5}$) needs 190--470 iterations and is two to four
times slower. The exact $K$-label limit on one CPU core (Numba) costs 29--51\,ms per
patch for $K$ from 1 to 128, nearly independent of $K$ because it is one spanning forest
with per-label solves only on plateaus; this is 2 to 7 times faster than the converged
GPU solve at batch size 1. The U-Net itself costs 0.9\,ms per image at batch 16.

\begin{figure}[h]
\centering
\includegraphics[width=\textwidth]{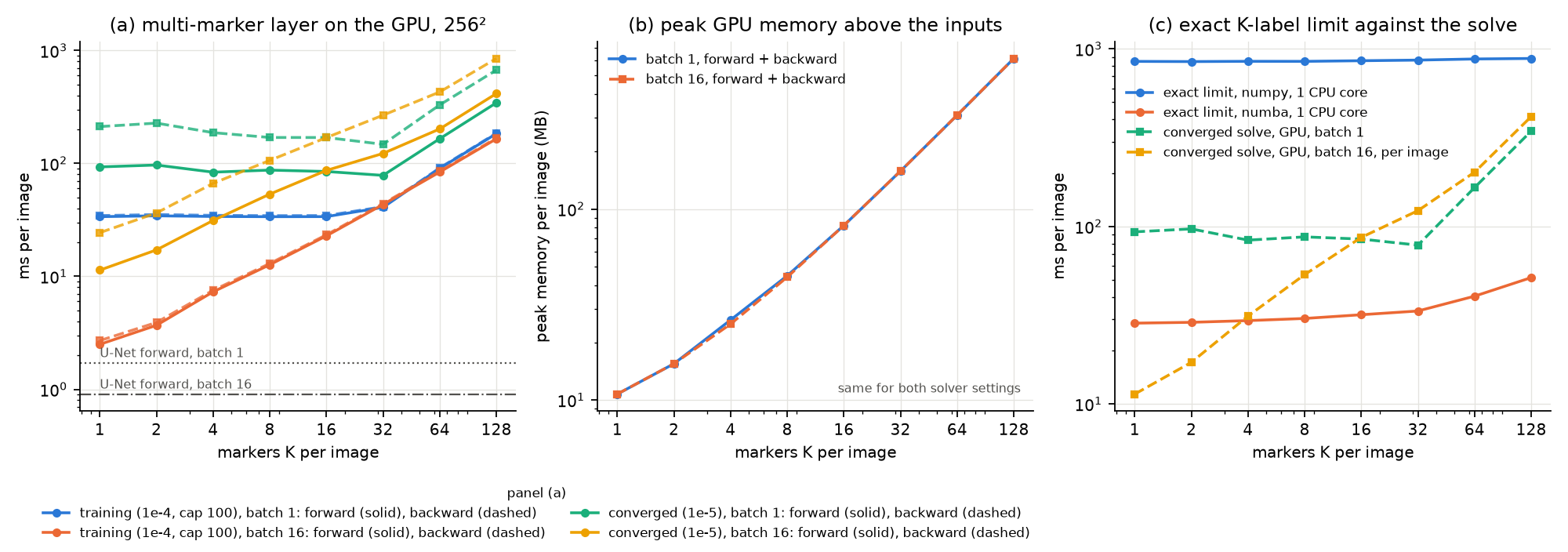}
\caption{Multi-marker WS layer against the number of markers $K$ per image: (a) time
per image, training and converged settings, batch 1 and 16; (b) peak GPU memory per
image; (c) the exact limit on one CPU core against the converged GPU solve.}
\label{fig:effk}
\end{figure}

\paragraph{Training in the two modes.} The binary WS layer with learned costs was
retrained with the recipe of Sect.~\ref{sec:binary} ($p$ from 5 to 50, 30 epochs) in
each mode and with two seeds (Table~\ref{tab:effmodes}). The implicit mode raises every
metric of the layer; the paired differences against the unrolled run of the same seed
are $+0.008$/$+0.009$ IoU, $+0.018$/$+0.024$ BF$_2$, $+0.035$/$+0.055$ F$_1$ and
$+0.034$/$+0.050$ PQ, all with 95\,\% bootstrap intervals clear of zero. The layer is
then level with the thresholded U-Net on IoU and BF$_2$ and 0.007--0.012 behind on
F$_1$ and PQ. Its exact limit, however, is worse than that of the unrolled runs: the
costs learned with the true \RW\ gradient at the end of the schedule are not closer to a
watershed. An implicit epoch takes twice as long as an unrolled one (26\,s against
13\,s), as the number of iterations grows with $p$ along the schedule (about 20 at the
start, 210 at the end; 3--4\,\% of the forward solves reach the cap of 1000).

\begin{table}[h]
\centering\footnotesize\setlength{\tabcolsep}{4pt}
\caption{Binary WS layer with learned costs trained in the unrolled mode ($T=20$) or in
the implicit mode, fold~1 to fold~2. The layer rows are on all 2523 patches, the
exact-limit rows on 307 of them.}
\label{tab:effmodes}
\begin{tabular}{llrrrrrr}
\toprule
run & output & IoU & BF$_2$ & edge al. & F$_1$@0.5 & PQ & count err.\\
\midrule
U-Net & thresholded & 0.719 & 0.769 & 0.577 & 0.656 & 0.531 & 0.275\\
unrolled, seed 0 & layer & 0.708 & 0.751 & 0.580 & 0.609 & 0.486 & 0.330\\
 & limit & 0.668 & 0.719 & 0.575 & 0.630 & 0.501 & 0.246\\
unrolled, seed 1 & layer & 0.709 & 0.750 & 0.577 & 0.594 & 0.473 & 0.341\\
 & limit & 0.620 & 0.676 & 0.574 & 0.575 & 0.456 & 0.298\\
implicit, seed 0 & layer & 0.716 & 0.770 & 0.601 & 0.644 & 0.519 & 0.256\\
 & limit & 0.602 & 0.677 & 0.650 & 0.595 & 0.486 & 0.417\\
implicit, seed 1 & layer & 0.719 & 0.773 & 0.596 & 0.649 & 0.523 & 0.249\\
 & limit & 0.604 & 0.684 & 0.643 & 0.611 & 0.498 & 0.394\\
\bottomrule
\end{tabular}
\end{table}

\section{From the training temperature to the exact limit}\label{app:psweep}

The multi-marker WS layer trained from its own predicted markers (Table~\ref{tab:instances})
was evaluated on all 2523 patches of fold~2 with its costs, unaries and test markers fixed,
solving its system exactly at increasing $p$ (Table~\ref{tab:psweep}). The Jacobi-PCG used
in training is accurate only near the training $p$: on 256 patches it matches the exact
solve to 0.015\,\% of the pixels at $p\le20$, differs on 0.5\,\% at $p=35$, and from
$p=100$ reaches its iteration cap in single and double precision with 5--12\,\% of the
pixels wrong. Sparse LU also loses accuracy for $p\ge100$ by cancellation in the pivots.
The sweep therefore uses an elimination free of subtractions (the GTH variant of Gaussian
elimination, in the log domain), about 2\,s per patch at any $p$; its outputs sum to one
within $10^{-13}$, and it agrees with sparse LU on 32 patches at $p\le100$ and with the
training solver at $p\le20$.

Accuracy rises up to the training $p$ and then falls monotonically towards the limit,
which the curve reaches by $p\approx500$; most of the 0.07 F$_1$ and 0.06 PQ gap has opened
by $p=100$. The quality of the matched nuclei (SQ) stays between 0.809 and 0.816 from
$p=20$ on, so the loss is in detection. The fraction of pixels labelled unlike the limit
decreases steadily; on three patches it is zero at $p=3000$, direct evidence that the
finite-$p$ solve tends to our exact limit on the same costs.

The binary models behave differently (Fig.~\ref{fig:psweep}b). For the WS layer with learned
costs of Table~\ref{tab:binary}, trained up to $p=50$, IoU is flat up to the training $p$
(0.707--0.710) and then falls to the limit (0.660), while the instance scores of its
connected components keep rising past the training $p$, to F$_1$ 0.653 and PQ 0.521 at
$p=200$ against 0.602 and 0.480 at $p=50$, and stay within 0.01 F$_1$ of the limit
(0.643): a higher $p$ removes the faint blobs that thresholding a soft output leaves. The
same holds for a second binary model trained with an edge target on its costs (the run
\texttt{edge\_bce} of the record; not the edge-target row of Table~\ref{tab:binary}): F$_1$
0.434 at the training $p$, 0.529 at $p=200$, 0.519 at the limit. Agreement with the limit
converges more slowly than for the instance model (0.69\,\% and 0.16\,\% of the pixels still
differ at $p=500$), since sigmoid edge costs lie in $[0,1]$ and $p\,\Delta c$ stays small.
The layer's PCG is accurate for the learned-cost model up to $p=200$ but not at $p=500$
(0.44\,\% of the pixels wrong), and for the edge-target model it is wrong on 0.8--3\,\% of
the pixels from $p=100$ on; the rows of Table~\ref{tab:binary} are at the training $p$, where
the exact solve reproduces them.

\begin{figure}[h]
\centering
\includegraphics[width=\textwidth]{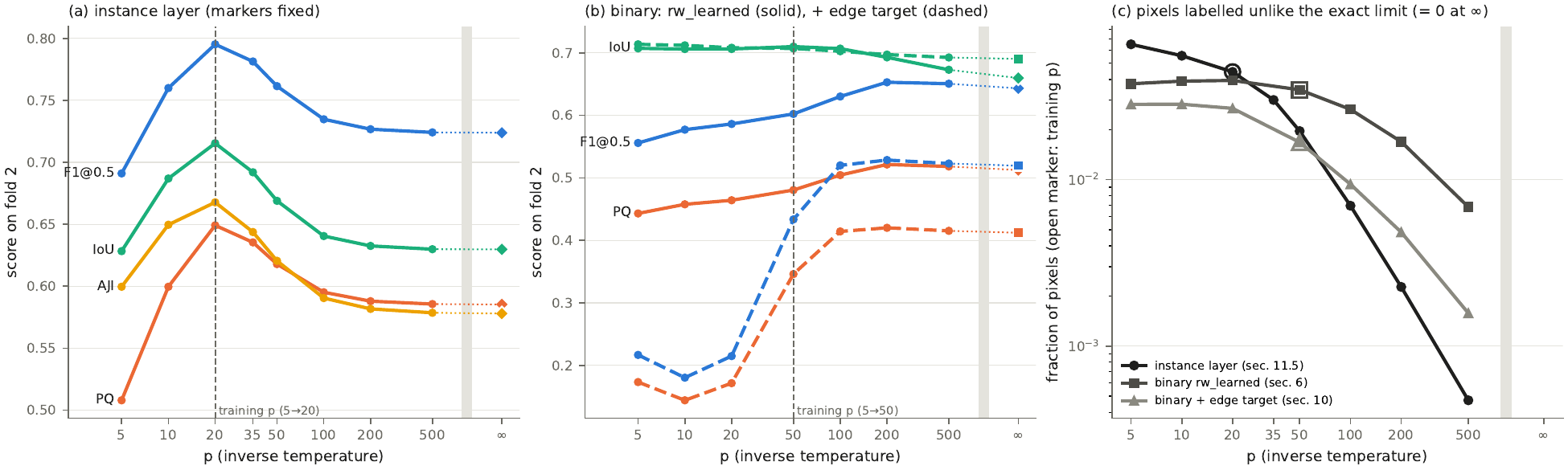}
\caption{Accuracy on fold~2 against $p$, solved exactly with costs and markers fixed:
(a) the instance WS layer, (b) two binary WS layers (learned costs, solid; with an edge
target, dashed); the diamonds at $\infty$ are the exact \PW\ limit. (c) Fraction of pixels
labelled unlike the limit; open markers at the training $p$.}
\label{fig:psweep}
\end{figure}

\begin{table}[h]
\centering\footnotesize\setlength{\tabcolsep}{4pt}
\caption{Instance WS layer on fold~2 solved exactly at increasing $p$, costs and markers
fixed. The last two columns give the pixels labelled unlike the limit, over all pixels and
over the pixels that are foreground in either output.}
\label{tab:psweep}
\resizebox{\textwidth}{!}{%
\begin{tabular}{lrrrrrrrrr}
\toprule
$p$ & IoU & BF$_2$ & F$_1$@0.5 & PQ & SQ & AJI & count err. & px $\ne x^\infty$ & fg px $\ne x^\infty$\\
\midrule
5 & 0.628 & 0.630 & 0.691 & 0.508 & 0.735 & 0.600 & 0.231 & 6.51\,\% & 39.0\,\%\\
10 & 0.687 & 0.730 & 0.760 & 0.600 & 0.789 & 0.650 & 0.210 & 5.56\,\% & 32.5\,\%\\
20 (training) & 0.715 & 0.775 & 0.795 & 0.649 & 0.816 & 0.668 & 0.182 & 4.45\,\% & 25.3\,\%\\
35 & 0.692 & 0.752 & 0.781 & 0.635 & 0.813 & 0.644 & 0.190 & 3.01\,\% & 17.9\,\%\\
50 & 0.669 & 0.731 & 0.761 & 0.618 & 0.812 & 0.621 & 0.216 & 1.96\,\% & 12.2\,\%\\
100 & 0.641 & 0.706 & 0.735 & 0.595 & 0.810 & 0.590 & 0.258 & 0.70\,\% & 4.6\,\%\\
200 & 0.633 & 0.699 & 0.727 & 0.588 & 0.809 & 0.582 & 0.270 & 0.23\,\% & 1.5\,\%\\
500 & 0.630 & 0.697 & 0.724 & 0.586 & 0.809 & 0.579 & 0.274 & 0.047\,\% & 0.32\,\%\\
$\infty$ (exact limit) & 0.630 & 0.697 & 0.724 & 0.585 & 0.809 & 0.578 & 0.274 & 0 & 0\\
\bottomrule
\end{tabular}}
\end{table}

\section*{Experimental record}
The remaining sections reproduce the long-form experimental report of the research
repository (\texttt{unrolling/finishing-layer-experiments.md}), lightly edited: every
run, every table and the negative results, in the order in which they were done.
Section numbers of the main text are not those of the report.
\section{Conditioning study (hard seeds, 512$^2$)}

Cases: synthetic shapes with noise $\sigma$ $\in$ \{0.05, 0.1\}, 8-bit, costs c\_e = ((I\_u $-$ I\_v)/255)$^2$; seeds = ground truth eroded by r = 5 px (refinement regime) or r = 30 px, or a few scribbles (global regime); the \texttt{camera} image with Otsu labels eroded likewise. Reference x\^{}p by a diagonally scaled sparse LU in float64 (flagged unreliable when its residual exceeds 1e-8, which happens for the noisy cases beyond p $\approx$ 500); limit x\^{}$\infty$ by the exact MSF + plateau algorithm on the integer costs. Figures: \path{results/conditioning/figures512/*.png}, table \texttt{summary.md}.

\textbf{Where the limit is reached, and what it costs} (refinement regime, r = 5):

\begin{center}\scriptsize\setlength{\tabcolsep}{3pt}
\resizebox{\ifdim\width>\textwidth\textwidth\else\width\fi}{!}{%
\begin{tabular}{@{}llllll@{}}
\toprule\noalign{}
case & p & $\kappa$(L\_U) & CG fp32 iters to 1e-3 & labels $\ne$ x\^{}$\infty$ & edges carrying 90 \% of the gradient \\
\midrule\noalign{}

clean, $\sigma$ = 0.05 & 10 & 1.8e2 & 39 & 0.7 \% & 14 276 \\
& 100 & 4.7e2 & 53 & 0 & 5 945 \\
& 200 & 5.7e3 & 63 & 0 & 844 \\
& 1000 & 2.3e17 & 173 & 0 & 7 \\
noisy, $\sigma$ = 0.1 & 10 & 2.0e2 & 42 & 3.1 \% & 14 645 \\
& 100 & 1.5e10 & 89 & 2.2 \% & 8 351 \\
& 500 & 1.0e48 & 240 & 1.2 \% & 681 \\
camera & 10 & 5.5e2 & 62 & 9.8 \% & 23 602 \\
& 100 & 1.7e7 & 104 & 5.5 \% & 13 044 \\
& 500 & 4.1e33 & 154 & 4.1 \% & 5 503 \\
\bottomrule
\end{tabular}}
\end{center}

\begin{itemize}

\item
  On the clean image the labels of the limit are reached at p $\approx$ 100 while $\kappa$ is still a few hundred and CG to a loose tolerance needs about 50 iterations: a comfortable training window. The gradient support has already shrunk by half there and is gone by p $\approx$ 500 (the zero-temperature collapse of the note).
\item
  On noisy or natural images x\^{}p approaches the limit only slowly (a few per cent of the free pixels still differ at p = 500) while $\kappa$ explodes far beyond anything CG can resolve; the float64 LU itself fails beyond p $\approx$ 500--1000. The central tension of the note is real there: the last few per cent of agreement with the limit are not reachable by a finite-p solve. They have to come from deploying the combinatorial limit (section 3 measures what that costs).
\item
  The CG iteration count to a loose tolerance grows much more slowly than $\surd\kappa$ because the ill-conditioning sits in near-isolated pixels that barely affect the residual; Jacobi preconditioning helps by one to two orders of magnitude in $\kappa$.
\item
  Natural images are the hard case. On \texttt{astronaut} (Otsu labels eroded by 5 px) x\^{}p still disagrees with the limit on 15 \% of the free pixels at p = 10, 11 \% at p = 100 and 6 \% at p = 500, where the float64 reference already fails; the textures give $\kappa$ $\approx$ 5e9 at p = 100. The PW limit of a raw-gradient graph on a photograph is simply far from any finite-p solution, which is the same message as the Pets results below: on photographs the edge costs have to be learned.
\item
  \textbf{Resolution does not change the cost in the refinement regime.} At 1024$^2$ with the same 5-px erosion, CG needs 39 / 44 / 54 iterations to 1e-3 at p = 10 / 100 / 200, against 39 / 53 / 63 at 512$^2$: the work is set by the width of the band, not by the image. In the global regime (r = 30) it grows with the image (160--205 at 1024$^2$ against 245--260 at 512$^2$ for a band that is relatively narrower). Truncation at 1024$^2$, clean, p = 100: 20 CG steps disagree with the converged solve on 0.01 \% of the labels, 50 Jacobi steps on 0.8 \%. Figures: \path{results/conditioning/figures1024/}.
\end{itemize}

\textbf{Truncation} (label disagreement with the converged x\^{}p after K steps):

\begin{center}\scriptsize\setlength{\tabcolsep}{3pt}
\resizebox{\ifdim\width>\textwidth\textwidth\else\width\fi}{!}{%
\begin{tabular}{@{}lllllll@{}}
\toprule\noalign{}
case, p = 100 & CG K=10 & CG K=20 & CG K=50 & Jacobi K=10 & Jacobi K=50 & Jacobi K=200 \\
\midrule\noalign{}

clean, r = 5 & 1.2 \% & 0.04 \% & 0 & 24 \% & 1.3 \% & 0.01 \% \\
noisy, r = 5 & 18 \% & 2.9 \% & 0.5 \% & 25 \% & 13 \% & 0.6 \% \\
camera, r = 5 & 20 \% & 11 \% & 3 \% & 52 \% & 18 \% & 7 \% \\
clean, r = 30 & 25 \% & 23 \% & 16 \% & 26 \% & 25 \% & 23 \% \\
\bottomrule
\end{tabular}}
\end{center}

\begin{itemize}

\item
  In the refinement regime (band of 10 px) 20 unrolled CG steps suffice on clean images, plain Jacobi needs ten times more. In the global regime (r = 30, scribbles) no truncation works: the note\textquotesingle s "match the application to the method\textquotesingle s weakness" is confirmed quantitatively.
\end{itemize}

\textbf{Cost transform} (\path{figures_transforms/}, four 512$^2$ cases, squared cost vs \textbar $\Delta$I\textbar{} vs dense rank of the distinct values; all three share the same limit). The transform only moves the p scale. Measured at equal agreement with the limit, it does not buy a cheaper solve: on the clean refinement case the squared cost reaches the limit\textquotesingle s labels at p = 100 for 53 CG iterations, \textbar $\Delta$I\textbar{} at p = 50 for 115, the rank at p = 50 for 259. On the natural image the rank and \textbar $\Delta$I\textbar{} get closer to the limit than the squared cost ever does (1--2 \% of the free pixels at p = 100--200 against 4 \% at p = 500) for about twice the iterations, but with the float64 reference already failing there. What the transforms do change is where the gradient dies: with the rank, 90 \% of the gradient mass sits on 20 edges by p = 50, at the very p where the limit is first reached, while with the squared cost 6 000 edges still carry it at the corresponding point. So for annealing, the squared cost, which spends the dynamic range on the strong edges, is the better path: the limit is reached while the gradient is still alive. The rank normalisation keeps its other role, fixing the output distribution of a learned cost head so that p means the same thing on every image.

\section{The WS layer as a tunable softmax (no training, 256$^2$)}

Simulated U-Net output: ground truth elastically displaced by up to 3 px, blurred ($\sigma$ = 4 px), soft logits with smooth noise; four scenes $\times$ noise $\sigma$ $\in$ \{0.05, 0.1\} $\times$ \{image edge on the true boundary, image edge shifted by 4 px\}. Edge costs from the image, unaries $\alpha$($-$log s, $-$log(1 $-$ s)), $\alpha$ $\in$ \{0.03, 0.1, 0.3, 1\}. Converged solves (implicit layer) for p $\in$ {[}1, 500{]}, the exact limit, and truncated layers. Figures: \path{results/softmax/figures/}.

\begin{center}\scriptsize\setlength{\tabcolsep}{3pt}
\resizebox{\ifdim\width>\textwidth\textwidth\else\width\fi}{!}{%
\begin{tabular}{@{}lllllllll@{}}
\toprule\noalign{}
noise & image edge & $\alpha$ & soft prediction BF$_2$ & best p & BF$_2$ at best p & soft pixels at best p & limit BF$_2$ & limit IoU \\
\midrule\noalign{}

0.05 & aligned & 0.1 & 0.958 & 200 & \textbf{1.000} & 0.3 \% & \textbf{1.000} & 1.000 \\
0.05 & aligned & 1 & 0.958 & 50 & 0.954 & 21 \% & 0.000 & 0.08 \\
0.1 & aligned & 0.1 & 0.958 & 200 & 0.968 & 2.4 \% & 0.870 & 0.966 \\
0.05 & shifted 4 px & 0.1 & 0.963 & 1 & 0.959 & 100 \% & 0.002 & 0.865 \\
0.1 & shifted 4 px & 0.03 & 0.955 & 2 & 0.955 & 100 \% & 0.688 & 0.920 \\
\bottomrule
\end{tabular}}
\end{center}

(BF$_2$ = boundary F-score at 2 px tolerance, computed against the ground truth.)

\begin{itemize}

\item
  \textbf{Crisper and more correct.} With $\alpha$ $\le$ 0.1 and the image edge where the true boundary is, the WS layer turns a 0.96 boundary score into 1.00 at p = 200 and in the limit, with the thin protrusion that the soft prediction rounds off fully recovered (\texttt{examples.png}), and the output goes from mostly soft to binary. This is the "better, tunable softmax": p is the temperature, and sharpening is spatially coupled through the edges rather than per pixel.
\item
  \textbf{The unary scale is a real hyperparameter.} With $\alpha$ = 1 the unaries of confident pixels compete with the noise edges, nothing attaches to a label before the whole image has merged, and both large p and the limit collapse to one label. In the trained setting the network sets this scale itself through the logit magnitude.
\item
  \textbf{The failure mode is as predicted.} When the true boundary is 4 px away from the image edge, p $\le$ 10 is harmless but larger p snaps the boundary to the wrong edge and the limit is poor (BF$_2$ 0.002 at $\alpha$ = 0.1). The WS layer is only as good as its edge costs, which argues for learning them.
\item
  \textbf{Truncation} (\texttt{truncation.png}, clean, aligned, $\alpha$ = 0.1): 10 unrolled PCG steps at p = 50 agree with the converged solve on 99.9 \% of the labels and reach the full boundary score; Jacobi needs about 50 steps for the same.
\end{itemize}

\section{U-Net with the WS layer as its last layer, trained end to end (Oxford-IIIT Pet)}

Binary pet / background from the trimap (border band counted as pet), 128$^2$, small U-Net (32-64-128-256), Adam, 30 epochs, batch 16, horizontal flips. Variants:

\begin{itemize}

\item
  \texttt{unet}: BCE on the logits.
\item
  \path{rw_img}: logits $\to$ unary costs; edge costs = squared colour differences; the WS layer = 20 unrolled Jacobi-PCG steps in float32, p annealed geometrically from 5 to 50; BCE on the WS layer output plus 0.5 BCE on the logits.
\item
  \path{rw_learned}: as \path{rw_img}, but the edge costs come from a second head of the same U-Net (sigmoid, averaged onto the edges).
\end{itemize}

Test split (3669 images): IoU, boundary F-scores at 1 and 2 px, accuracy within 3 px of the true boundary, fraction of pixels in (0.1, 0.9). Every "thresholded at 0.5" row is the practitioner\textquotesingle s pipeline, the network\textquotesingle s sigmoid output cut at 0.5 (and, where instance metrics appear, its connected components); it is the baseline every layer row is to be read against. Section 9 adds the tuned-threshold version of it. For the RW variants the output is evaluated as trained (20 steps), converged at p = 50, and as the exact PW limit on 300 images (deployment at p = $\infty$).

\begin{center}\scriptsize\setlength{\tabcolsep}{3pt}
\resizebox{\ifdim\width>\textwidth\textwidth\else\width\fi}{!}{%
\begin{tabular}{@{}lllllll@{}}
\toprule\noalign{}
variant & output & IoU & BF$_1$ & BF$_2$ & band acc. & soft pixels \\
\midrule\noalign{}

unet & sigmoid, thresholded at 0.5 & 0.853 & 0.586 & 0.724 & 0.778 & 13.4 \% \\
rw\_img & sigmoid, thresholded at 0.5 & 0.838 & 0.500 & 0.633 & 0.768 & 12.6 \% \\
rw\_img & WS layer, 20 steps & 0.840 & 0.508 & 0.681 & 0.742 & 18.0 \% \\
rw\_img & converged & 0.842 & 0.508 & 0.681 & 0.749 & 18.9 \% \\
rw\_img & PW limit (300) & 0.405 & 0.134 & 0.260 & 0.510 & 0.2 \% \\
rw\_learned & sigmoid, thresholded at 0.5 & 0.840 & 0.498 & 0.638 & 0.765 & 15.2 \% \\
rw\_learned & WS layer, 20 steps & 0.852 & 0.584 & 0.722 & 0.771 & 14.2 \% \\
rw\_learned & converged & 0.853 & 0.586 & 0.724 & 0.775 & 14.9 \% \\
rw\_learned & PW limit (300) & 0.822 & 0.561 & 0.695 & 0.749 & 0 \\
\bottomrule
\end{tabular}}
\end{center}

Figures: \path{results/pets/figures/} (training curves, example panels with the learned edge costs).

\begin{itemize}
\item
  \textbf{Honest outcome at 128$^2$: no gain over the plain U-Net.} The learned-cost variant matches the U-Net on every metric (0.853 vs 0.853 IoU, 0.724 vs 0.724 BF$_2$); the image-gradient variant is worse, and its limit is useless (0.41 IoU): raw colour edges on photographs are not where pet boundaries are. The U-Net trained with BCE for 30 epochs is already nearly binary (13 \% soft pixels), so there was little softness to remove at this resolution.
\item
  \textbf{What the WS layer does do.} Inside the trained model the WS layer adds about 0.085 BF$_2$ over its own logits (0.638 $\to$ 0.722): the network learned to leave the boundary decision to the WS layer and the learned cost head is a clean contour detector (\path{examples_rw_learned.png}). The 20-step layer is within 0.002 of the converged solve, so K = 20 is enough here.
\item
  \textbf{Deployment at p = $\infty$ works, with a price.} The exact combinatorial PW on the learned costs, which nothing was ever trained with directly, loses 0.03 IoU and 0.03 BF$_2$ against the converged finite-p solve, and is perfectly binary. That is the train-at-finite-p / deploy-at-the-limit scheme of the note, demonstrated.
\item
  \textbf{The 256$^2$ repeat says the same.} Same recipe at 256 px (\path{results/pets256/}):

  {\def\LTcaptype{none} % do not increment counter
  \begin{center}\scriptsize\setlength{\tabcolsep}{3pt}
\resizebox{\ifdim\width>\textwidth\textwidth\else\width\fi}{!}{%
\begin{tabular}{@{}lllllll@{}}
  \toprule\noalign{}
  variant & output & IoU & BF$_1$ & BF$_2$ & band acc. & soft pixels \\
  \midrule\noalign{}

  unet & sigmoid, thresholded at 0.5 & 0.842 & 0.364 & 0.503 & 0.685 & 18.7 \% \\
  rw\_learned & sigmoid, thresholded at 0.5 & 0.817 & 0.211 & 0.310 & 0.670 & 22.2 \% \\
  rw\_learned & WS layer, 20 steps & 0.844 & 0.359 & 0.498 & 0.678 & 17.8 \% \\
  rw\_learned & converged & 0.845 & 0.369 & 0.506 & 0.688 & 19.5 \% \\
  rw\_learned & PW limit (200) & 0.795 & 0.296 & 0.412 & 0.637 & 0 \\
\bottomrule
\end{tabular}}
\end{center}

  }

  Parity with the plain U-Net on every metric, a 0.19 BF$_2$ gain of the WS layer over its own logits, and a 0.05 IoU price for the exact limit. Higher resolution did not open a gap; the boundary scores of both methods simply drop because the 1--2 px tolerance is now tighter relative to the trimap\textquotesingle s own boundary band.
\end{itemize}

\section{U-Net with the WS layer as its last layer on real nuclei (PanNuke)}

Code: \path{experiments/convert_pannuke.py}, \path{experiments/exp_unet_seg.py} (dataset-agnostic version of \path{exp_unet_pets.py}, reusing its model and loss), \path{experiments/seg_metrics.py}, \path{experiments/plot_pannuke.py}. Raw results, per-image metrics and figures: \path{results/pannuke/}.

\textbf{Data.} PanNuke (Gamper et al. 2019/2020), 256$^2$ H\&E patches at 40$\times$, 19 tissue types, semi-automatic instance labels. Train on fold 1 (2656 patches), test on all of fold 2 (2523 patches; Breast 749, Colon 468, HeadNeck 169, Adrenal 148, ...). Binary target = union of the five nucleus channels (17 \% of the pixels; 23.7 nuclei per patch on average; 0.09 \% of the pixels sit on an interface between two touching nuclei). Native 256 px, flips and 90$^\circ$ rotations, otherwise the Pets recipe: small U-Net (32-64-128-256), Adam 1e-3 with cosine decay, 30 epochs, batch 16, p annealed 5 $\to$ 50, K = 20 unrolled Jacobi-PCG steps, BCE on the WS layer output + 0.5 BCE on the logits. Training takes 11 s per epoch; evaluating fold 2 takes a few minutes, dominated by the CPU-side metrics.

\textbf{Metrics.} Pixel level as before (IoU, boundary F at 1, 2 and 3 px, accuracy in a 3-px band around the true boundary, fraction of pixels in (0.1, 0.9)). Instance level, on the 4-connected components of the thresholded output (4-adjacency is the graph the WS layer works on) against the instance masks: object-count error \textbar n\_pred $-$ n\_gt\textbar{} / n\_gt, F1 of the matches at IoU $\ge$ 0.5, panoptic quality PQ and AJI, all aggregated over the whole test set. Because the target is a binary mask, touching nuclei merge for every method; the row "ground truth / binary mask" (components of the true binary mask against the instances) is the ceiling any binary method can reach on these metrics. Label-free edge alignment: fraction of predicted boundary pixels within 1 px of a strong image edge (top 10 \% of the colour-gradient magnitude after a $\sigma$ = 1 px Gaussian). The same score for the ground-truth boundary calibrates it. The exact PW limit (p = $\infty$) is computed on 307 patches: 300 evenly spaced through fold 2 plus the 8 patches with the most touching-nuclei pixels, which also serve as the example panels.

\textbf{A note on the limit\textquotesingle s cost.} With float32 costs quantised to 2$^{-20}$, a 256$^2$ patch has about 200 000 plateaus of which 40 000 have 2--7 tied edges; the general plateau path of \path{power_watershed} (sparse graph + connected components per plateau) made the exact limit cost 50--60 s per patch. A pure-Python route for plateaus of $\le$ 64 edges, with the same semantics (checked bit-identical), brings it to 2 s per patch.

\textbf{Results on all of fold 2} (2523 patches; the PW limit on 307). Instance metrics on 4-connected components of the thresholded output; "count err." = mean \textbar n\_pred $-$ n\_gt\textbar{} / n\_gt.

\begin{center}\scriptsize\setlength{\tabcolsep}{3pt}
\resizebox{\ifdim\width>\textwidth\textwidth\else\width\fi}{!}{%
\begin{tabular}{@{}lllllllllllll@{}}
\toprule\noalign{}
variant & output & IoU & BF$_1$ & BF$_2$ & BF$_3$ & band acc. & soft px & edge align. & \href{mailto:F1@0.5}{\nolinkurl{F1@0.5}} & PQ & AJI & count err. \\
\midrule\noalign{}

unet & sigmoid, thresholded at 0.5 & \textbf{0.719} & \textbf{0.652} & \textbf{0.769} & \textbf{0.825} & \textbf{0.807} & 15.4 \% & 0.577 & \textbf{0.656} & \textbf{0.531} & \textbf{0.494} & 0.275 \\
rw\_img & sigmoid, thresholded at 0.5 & 0.669 & 0.547 & 0.672 & 0.737 & 0.806 & 13.2 \% & 0.481 & 0.303 & 0.240 & 0.529 & 5.01 \\
rw\_img & WS layer, 20 steps & 0.683 & 0.571 & 0.714 & 0.794 & 0.775 & 19.9 \% & 0.559 & 0.516 & 0.400 & 0.396 & 0.513 \\
rw\_img & converged & 0.686 & 0.570 & 0.713 & 0.794 & 0.779 & 20.3 \% & 0.548 & 0.531 & 0.411 & 0.336 & 0.441 \\
rw\_img & PW limit (307) & 0.181 & 0.217 & 0.254 & 0.273 & 0.540 & 0 & 0.796 & 0.194 & 0.150 & 0.109 & 0.675 \\
rw\_learned & sigmoid, thresholded at 0.5 & 0.674 & 0.560 & 0.684 & 0.749 & 0.800 & 17.7 \% & 0.481 & 0.083 & 0.067 & 0.533 & 25.4 \\
rw\_learned & WS layer, 20 steps & 0.709 & 0.628 & 0.752 & 0.813 & 0.799 & 16.1 \% & 0.579 & 0.604 & 0.482 & 0.456 & 0.337 \\
rw\_learned & converged & 0.710 & 0.630 & 0.751 & 0.812 & 0.805 & 16.2 \% & 0.572 & 0.602 & 0.480 & 0.366 & 0.322 \\
rw\_learned & PW limit (307) & 0.661 & 0.591 & 0.711 & 0.768 & 0.771 & 0 & 0.577 & 0.618 & 0.490 & 0.425 & \textbf{0.246} \\
ground truth & binary mask & 1 & 1 & 1 & 1 & 1 & 0 & 0.559 & 0.913 & 0.883 & 0.701 & 0.088 \\
\bottomrule
\end{tabular}}
\end{center}

(The AJI of the raw logits is inflated by their hundreds of tiny components, which AJI only counts in the denominator by area; F1, PQ and the count error are the honest instance numbers.)

\textbf{Per tissue type} (the three most common in fold 2; \path{figures/summary.md} has the full tables, \path{rw_learned.json} all 15 types with $\ge$ 50 patches):

\begin{center}\scriptsize\setlength{\tabcolsep}{3pt}
\resizebox{\ifdim\width>\textwidth\textwidth\else\width\fi}{!}{%
\begin{tabular}{@{}llllllll@{}}
\toprule\noalign{}
tissue (n) & variant / output & IoU & BF$_2$ & edge align. & \href{mailto:F1@0.5}{\nolinkurl{F1@0.5}} & PQ & count err. \\
\midrule\noalign{}

Breast (749) & unet / logits & 0.722 & 0.756 & 0.585 & 0.690 & 0.547 & 0.198 \\
& rw\_learned / layer & 0.713 & 0.738 & 0.589 & 0.635 & 0.498 & 0.313 \\
& rw\_learned / PW limit (89) & 0.667 & 0.703 & 0.586 & 0.684 & 0.535 & 0.174 \\
& ground truth / binary & 1 & 1 & 0.553 & 0.933 & 0.907 & 0.072 \\
Colon (468) & unet / logits & 0.698 & 0.738 & 0.558 & 0.547 & 0.431 & 0.274 \\
& rw\_learned / layer & 0.681 & 0.711 & 0.565 & 0.495 & 0.384 & 0.319 \\
& rw\_learned / PW limit (62) & 0.610 & 0.658 & 0.534 & 0.463 & 0.353 & 0.283 \\
& ground truth / binary & 1 & 1 & 0.539 & 0.836 & 0.796 & 0.123 \\
HeadNeck (169) & unet / logits & 0.672 & 0.744 & 0.648 & 0.640 & 0.519 & 0.424 \\
& rw\_learned / layer & \textbf{0.693} & \textbf{0.758} & 0.635 & 0.593 & 0.475 & 0.309 \\
& ground truth / binary & 1 & 1 & 0.629 & 0.927 & 0.897 & 0.061 \\
\bottomrule
\end{tabular}}
\end{center}

Paired per-patch differences (2523 patches, bootstrap 95 \% intervals), \path{rw_learned} layer minus \texttt{unet}: IoU $-$0.009 {[}$-$0.012, $-$0.007{]}, BF$_2$ $-$0.017 {[}$-$0.020, $-$0.014{]}, band accuracy $-$0.008; the WS layer is better than the U-Net on 25 \% of the patches. The gap grows with crowding: $-$0.004 IoU on patches with fewer than 10 nuclei, $-$0.016 on patches with more than 50. Within the trained \path{rw_learned} model, layer minus its own logits: +0.035 IoU, +0.068 BF$_2$, better on 81 \% / 86 \% of the patches. Limit minus converged on the same 307 patches: $-$0.046 IoU {[}$-$0.066, $-$0.027{]}, $-$0.034 BF$_2$.

Figures: \path{results/pannuke/figures/training.png}, \texttt{examples\_\textless{}variant\textgreater{}.png} (full patches) and \texttt{examples\_\textless{}variant\textgreater{}\_crop.png} (64-px crops around the densest cluster of touching nuclei), for the 8 fold-2 patches with the most touching-nuclei pixels; all eight are Colon, where crowding is worst.

\begin{itemize}

\item
  \textbf{Honest outcome: on real nuclei the WS layer does not help, and costs a little.} The learned-cost variant is 0.01 IoU and 0.017 BF$_2$ below the plain U-Net, consistently across tolerances (1, 2, 3 px), tissues (HeadNeck is the one exception, +0.02 IoU and +0.014 BF$_2$ on 169 patches), and at the instance level (F1 0.60 vs 0.66, PQ 0.48 vs 0.53). The difference is small but not noise. The image-gradient variant is clearly worse, as on Pets.
\item
  \textbf{Why: the WS layer is a smoother here, and nuclei are small.} With p $\le$ 50 the WS layer diffuses the unaries across every edge whose cost is not large, so the output is softer than the U-Net\textquotesingle s (16 \% soft pixels against 15 \%, and visibly blurrier blobs in \path{examples_rw_learned_crop.png}), and it bridges the 1--2 px gaps between touching nuclei that the U-Net\textquotesingle s logits keep. On Pets the object was 100 px across and this cost nothing; on nuclei about 25 px across it costs the boundary metrics directly and the instance count indirectly (the IoU gap grows with the number of nuclei per patch). The learned edge head is nevertheless a clean contour detector that does respond to the interfaces between touching nuclei (last column of the crop panels), but its contours are 2--3 px wide and the costs along them are not large enough at p = 50 to stop the diffusion. Lower p than Pets was never tried, nor a sharper cost parametrisation (e.g. the costs through an exponential rather than a sigmoid).
\item
  \textbf{The division of labour is extreme.} As on Pets, the network leaves the decision to the WS layer: its own logits are a checkerboard of speckle (348 components per patch against 24 true nuclei, F1 0.08) that the WS layer cleans up into 23.8 components with F1 0.60. The WS layer adds 0.035 IoU and 0.068 BF$_2$ over its own logits, but the two-stage system does not beat the one-stage one.
\item
  \textbf{Deployment at p = $\infty$ costs 0.05 IoU at the pixel level but is the best instance segmenter of the RW outputs.} The exact PW limit on the learned costs loses 0.046 IoU and 0.034 BF$_2$ against the converged finite-p solve, and the panels show why: in crowded regions it floods whole clusters into one component. Yet on the same 307 patches its instance metrics are better than the WS layer\textquotesingle s (F1 0.618 vs 0.581, PQ 0.49 vs 0.46, count error 0.25 vs 0.34), because it is binary and removes the faint spurious blobs that survive thresholding the soft output; on Breast it even reaches the U-Net\textquotesingle s PQ (0.535 vs 0.547) with a lower count error. The train-at-finite-p / deploy-at-the-limit scheme therefore works on real data in the sense that the limit is usable (unlike with image-gradient costs, 0.18 IoU), but it is not free.
\item
  \textbf{The label-noise caveat, measured.} The ground-truth boundaries lie within 1 px of a strong colour edge only 55.9 \% of the time; every method\textquotesingle s output does so at least as often (U-Net 57.7 \%, layer 57.9 \%, limit 57.7 \%). Either the labels are indeed a pixel or two off the visible edge, or nucleus boundaries on H\&E are often not strong colour edges at all (cytoplasm and nucleus of similar hue). In both cases boundary F-scores at 1 px against these labels cannot separate a method that follows the image from one that follows the label noise, which is part of the reason the BF$_1$ gap ($-$0.023) is larger than the BF$_3$ gap ($-$0.012). The \path{rw_img} limit shows the other side: it follows image edges 80 \% of the time and is useless. A dataset with pixel-accurate manual boundaries (CoNSeP, MoNuSeg) is needed for the pixel-level boundary claim; the instance metrics above do not depend on the exact boundary and already tell the same story.
\item
  \textbf{What the WS layer would need to help here.} Edge costs that are sharp and large where nuclei touch (a cost head trained with an explicit boundary target, or costs from the instance interfaces that the data provide), a target that leaves the interface pixels as background so that separating nuclei is rewarded, and a lower p, or a p-dependent cost scale, so that the WS layer stops being a smoother before it is a cutter. Section 6.1 tries the target, the lower p and the sharper parametrisation.
\end{itemize}

\subsection{Letting the WS layer cut: gap target, lower p, sharper edge costs}

Code: \texttt{-{}-gap}, \texttt{-{}-edge-param}, \texttt{-{}-edge-cmax}, \texttt{-{}-edge-learn-scale}, \texttt{-{}-tag} in \path{experiments/exp_unet_seg.py} (class \texttt{SegmenterX}), \path{experiments/run_gap.sh}, \path{experiments/compare_gap.py} (table, paired statistics), \path{experiments/seam_stats.py} (does the output cut the seams?), \path{plot_pannuke.py\ -\/-runs}. Raw results, per-image metrics and figures: \path{results/pannuke/gap/} (\texttt{compare.md}, \path{seam_stats.json}, \path{figures/}).

\textbf{What was changed.} Three of the four things section 6 asked for, together. (i) \emph{Gap target}: the binary mask with the 2-px seam between touching nuclei set to background, for training, validation and test, for every variant, so that separating nuclei is rewarded instead of punished (the seam is 0.09 \% of the pixels, so pixel metrics against the gap mask and against the binary mask are comparable; the U-Net of section 6 scores 0.717 against the gap mask instead of 0.719). The instance metrics are unchanged, and the "ground truth" ceiling row becomes the gap mask itself, F1 0.998. (ii) \emph{Lower p}: 5 $\to$ 20 and 2 $\to$ 10 instead of 5 $\to$ 50, so that the unaries s\^{}p stay soft and anchor the solution while a contour can still be a hard cut. (iii) \emph{Sharper edge costs}: 8$\cdot\sigma$(h) instead of $\sigma$(h), with the scale 8 learnable (initialised at the unary clamp c\_max = 8), so that the cost head can make a contour as expensive as a confident unary. Everything else is section 6\textquotesingle s recipe (fold 1 $\to$ fold 2, 30 epochs, K = 20, BCE + 0.5 aux BCE). Five runs: the U-Net under the gap target, and the learned-cost layer under the gap target at 5 $\to$ 50, at 5 $\to$ 20, at 5 $\to$ 20 with sharp edges and at 2 $\to$ 10 with sharp edges.

\textbf{Results on all of fold 2} (2523 patches; the PW limit on the same 307 patches as section 6). The first rows are section 6\textquotesingle s runs. Instance metrics on the raw 4-connected components of the thresholded output.

\begin{center}\scriptsize\setlength{\tabcolsep}{3pt}
\resizebox{\ifdim\width>\textwidth\textwidth\else\width\fi}{!}{%
\begin{tabular}{@{}llllllllllll@{}}
\toprule\noalign{}
target & run & output & IoU & BF$_1$ & BF$_2$ & BF$_3$ & band acc. & soft px & \href{mailto:F1@0.5}{\nolinkurl{F1@0.5}} & PQ & count err. \\
\midrule\noalign{}

binary & unet (sec. 6) & sigmoid, thresholded at 0.5 & \textbf{0.719} & \textbf{0.652} & \textbf{0.769} & \textbf{0.825} & \textbf{0.807} & 15.4 \% & 0.656 & 0.531 & 0.275 \\
binary & rw\_learned, p 5 $\to$ 50 (sec. 6) & WS layer, 20 steps & 0.709 & 0.628 & 0.752 & 0.813 & 0.799 & 16.1 \% & 0.604 & 0.482 & 0.337 \\
binary & rw\_learned, p 5 $\to$ 50 (sec. 6) & PW limit (307) & 0.661 & 0.591 & 0.711 & 0.768 & 0.771 & 0 & 0.618 & 0.490 & 0.246 \\
gap & unet & sigmoid, thresholded at 0.5 & 0.717 & 0.651 & 0.768 & 0.825 & 0.806 & 15.5 \% & 0.667 & 0.541 & 0.296 \\
gap & rw\_learned, p 5 $\to$ 50 & WS layer, 20 steps & 0.705 & 0.624 & 0.747 & 0.809 & 0.797 & 16.4 \% & 0.610 & 0.487 & 0.357 \\
gap & rw\_learned, p 5 $\to$ 50 & PW limit (307) & 0.664 & 0.587 & 0.710 & 0.768 & 0.770 & 0 & 0.621 & 0.490 & 0.246 \\
gap & rw\_learned, p 5 $\to$ 20 & WS layer, 20 steps & 0.707 & 0.625 & 0.748 & 0.810 & 0.797 & 15.8 \% & 0.635 & 0.508 & 0.313 \\
gap & rw\_learned, p 5 $\to$ 20 & PW limit (307) & 0.476 & 0.482 & 0.562 & 0.597 & 0.669 & 0 & 0.510 & 0.417 & 0.510 \\
gap & rw\_learned, p 5 $\to$ 20, 8$\cdot\sigma$ edges & WS layer, 20 steps & 0.706 & 0.624 & 0.747 & 0.810 & 0.795 & 16.1 \% & 0.625 & 0.499 & 0.295 \\
gap & rw\_learned, p 5 $\to$ 20, 8$\cdot\sigma$ edges & PW limit (307) & 0.435 & 0.446 & 0.522 & 0.555 & 0.641 & 0 & 0.474 & 0.390 & 0.567 \\
gap & rw\_learned, p 2 $\to$ 10, 8$\cdot\sigma$ edges & WS layer, 20 steps & 0.712 & 0.639 & 0.759 & 0.819 & 0.796 & 14.5 \% & \textbf{0.678} & \textbf{0.548} & \textbf{0.225} \\
gap & rw\_learned, p 2 $\to$ 10, 8$\cdot\sigma$ edges & converged & 0.716 & 0.639 & 0.760 & 0.820 & 0.801 & 14.8 \% & \textbf{0.680} & \textbf{0.549} & 0.226 \\
gap & rw\_learned, p 2 $\to$ 10, 8$\cdot\sigma$ edges & PW limit (307) & 0.508 & 0.496 & 0.582 & 0.620 & 0.671 & 0 & 0.523 & 0.435 & 0.555 \\
gap & ground truth & gap mask & 1 & 1 & 1 & 1 & 1 & 0 & 0.998 & 0.990 & 0.002 \\
\bottomrule
\end{tabular}}
\end{center}

Paired per-patch differences (2523 patches, bootstrap 95 \% intervals), layer minus the U-Net trained on the same gap target: p 5 $\to$ 50: IoU $-$0.012 {[}$-$0.015, $-$0.009{]}, BF$_2$ $-$0.021; p 5 $\to$ 20: $-$0.010, $-$0.021; p 5 $\to$ 20 sharp: $-$0.011, $-$0.021; \textbf{p 2 $\to$ 10 sharp: IoU $-$0.005 {[}$-$0.007, $-$0.002{]}, BF$_2$ $-$0.009 {[}$-$0.012, $-$0.007{]}}, better than the U-Net on 33 \% of the patches, and +0.007 IoU on the 486 patches with fewer than 10 nuclei ($-$0.010 on the 179 with more than 50). On HeadNeck (169 patches) the p 2 $\to$ 10 layer is above the U-Net on every metric (IoU 0.697 vs 0.670, BF$_2$ 0.765 vs 0.741, F1 0.671 vs 0.649, PQ 0.547 vs 0.528). Limit minus converged on the 307 patches: $-$0.037 IoU at p = 50, $-$0.23 at p = 20, $-$0.20 at p = 10. The learnable edge scale went \emph{down}, from 8 to 3.4 (p 5 $\to$ 20) and 4.1 (p 2 $\to$ 10), within the first six epochs and stayed there (\path{figures/training.png}; the dotted p axis there shows the 5 $\to$ 50 schedule only).

\textbf{Does anything cut the seams?} \path{seam_stats.json}: fraction of the 142 211 seam pixels of fold 2 predicted as background, and fraction of the 8999 seams (components of the touch mask) entirely background. U-Net, binary target: 15.1 \% / 10.6 \%; U-Net, gap target: 19.1 \% / 13.0 \%. Layer, gap target: 16.8 \% / 11.9 \% at p = 50, 18.5 \% at p = 20, 19.9 \% / 13.6 \% at p = 10 with sharp edges. In every run the WS layer output cuts \emph{fewer} seam pixels than its own logits do (25--31 \% for the logits), and the converged solve fewer still (11--18 \%); at best the WS layer matches the U-Net. The crops (\path{figures/examples_rw_p10_sharp_crop.png}, \path{examples_unet_gap_crop.png}) show the same thing: at p 2 $\to$ 10 the network\textquotesingle s own logits are now a recognisable nucleus detector with dark seams (54 components per patch instead of 340 at p = 50), the WS layer blurs those seams half shut, and the exact limit floods whole crops into one component or none.

\begin{itemize}

\item
  \textbf{Honest outcome: the three changes together close half of the pixel-level gap and make the WS layer the best raw instance segmenter, but the WS layer still does not cut, and it still does not beat the U-Net.} The gap target on its own changes little (U-Net: F1 0.656 $\to$ 0.667, seam pixels cut 15 $\to$ 19 \%; the WS layer at p 5 $\to$ 50 is slightly \emph{worse} than under the binary target, $-$0.005 IoU paired). Lower p at the old parametrisation changes nothing at the pixel level. Only the lowest schedule with the sharp parametrisation moves: $-$0.005 IoU and $-$0.009 BF$_2$ instead of $-$0.012 and $-$0.021, and instance F1 0.678, PQ 0.548, count error 0.225 against the gap-target U-Net\textquotesingle s 0.667 / 0.541 / 0.296 at the same threshold and without any post-processing (section 7\textquotesingle s min-area-filtered, boundary-trained U-Net, F1 0.714, remains better). But the seam statistics say the gain is not the mechanism we wanted: the WS layer cuts no more seam pixels than the U-Net and a third fewer than its own logits, the network \emph{lowered} the edge-cost scale it was given rather than raising it, and the inner logits at p = 10 are far closer to a plain U-Net\textquotesingle s output than at p = 50. At p $\approx$ 10 the unaries s\^{}p are strong enough to anchor every moderately confident pixel, so the WS layer degenerates toward a mild, edge-aware smoothing of a U-Net that now does most of the work itself, and the result is a system that behaves like the U-Net plus a small denoiser: fewer spurious components (hence the instance numbers), slightly blurrier boundaries (hence BF$_1$--BF$_3$ still below).
\item
  \textbf{The limit is lost at low p.} Costs learned at p = 10 or 20 mean nothing at p = $\infty$: the exact PW on them scores 0.44--0.51 IoU and about 10 components per patch, because the network relied on unaries whose p-th power no longer saturates and on contours with holes that a maximum-spanning-forest path walks through. The train-at-finite-p / deploy-at-the-limit scheme of section 6 ($-$0.04 IoU at p = 50) and the finite-p gains above therefore pull in opposite directions on this data; one cannot have both with a single schedule. What is left untried, and would be needed before concluding that a cut-capable last layer cannot help on nuclei: supervising the cost head directly with the seam mask (an edge target rather than hoping the segmentation loss finds it), a loss that pays for the separation of instances rather than for pixels near seams, and a dataset with pixel-accurate boundaries where a 1-px seam is a real image feature.
\end{itemize}

\section{Does a boundary- or instance-level loss open a gap? (PanNuke, Pets)}

Code: \texttt{-\/-loss\ \{bce,boundary,instance\}} in \path{experiments/exp_unet_seg.py}, \path{experiments/run_losses.sh}, \path{experiments/compare_losses.py} (tables, paired statistics, \texttt{losses.png}). Raw results, per-image metrics and figures: \texttt{results/pannuke\_losses/\{boundary,instance\}/} and \path{results/pets_losses/boundary/}.

\textbf{Hypothesis.} Sections 3 and 6 trained everything with a per-pixel BCE, and under that loss the plain U-Net learned to sharpen its own output as well as the WS layer does. A loss that pays for boundary placement or for separating neighbouring objects should favour a model with an explicit spatial prior, if the WS layer is to be worth anything. Both losses are applied to the final output of \emph{both} variants (for \path{rw_learned} the same loss on the WS layer output, plus 0.5 times the same loss on the logits, as the aux BCE before), so the comparison stays fair; everything else is the recipe of section 6 (fold 1 $\to$ fold 2, 30 epochs, p annealed 5 $\to$ 50, K = 20).

\begin{itemize}

\item
  \textbf{Boundary loss.} Per-pixel BCE with the pixels within 3 px of the true boundary weighted $\times$5 (band = max-pool minus min-pool of the target with a 7$\times$7 window), plus a differentiable boundary-F surrogate at a 2-px tolerance (Bokhovkin \& Burnaev 2019: soft boundaries as the inner morphological gradient x $-$ minpool$_3$(x), precision and recall through max-pooled tolerance regions, loss = 1 $-$ 2PR/(P+R)), weight 1.
\item
  \textbf{Instance loss (PanNuke only).} The 2-px seam between touching nuclei (pixels with a 4-neighbour in a different instance, 0.09 \% of the pixels) is set to \emph{background} in the target, and the seam plus a 1-px flank is weighted $\times$21 in the BCE. This is the "target that leaves the interface pixels as background" of section 6: separating touching nuclei is now rewarded, where the binary union target punished it. Test metrics are still computed against the original binary mask and the instance masks.
\end{itemize}

\textbf{Results on all of fold 2} (2523 patches; the PW limit on the same 307 patches as section 6). The \texttt{bce} rows are section 6\textquotesingle s runs. Instance metrics on the raw 4-connected components, as before; see the next table for what they become after a minimum-area filter.

\begin{center}\scriptsize\setlength{\tabcolsep}{3pt}
\resizebox{\ifdim\width>\textwidth\textwidth\else\width\fi}{!}{%
\begin{tabular}{@{}lllllllllllll@{}}
\toprule\noalign{}
loss & variant & output & IoU & BF$_1$ & BF$_2$ & BF$_3$ & band acc. & soft px & edge align. & \href{mailto:F1@0.5}{\nolinkurl{F1@0.5}} & PQ & count err. \\
\midrule\noalign{}

bce & unet & sigmoid, thresholded at 0.5 & 0.719 & 0.652 & 0.769 & 0.825 & 0.807 & 15.4 \% & 0.577 & 0.656 & 0.531 & 0.275 \\
bce & rw\_learned & WS layer, 20 steps & 0.709 & 0.628 & 0.752 & 0.813 & 0.799 & 16.1 \% & 0.579 & 0.604 & 0.482 & 0.337 \\
bce & rw\_learned & converged & 0.710 & 0.630 & 0.751 & 0.812 & 0.805 & 16.2 \% & 0.572 & 0.602 & 0.480 & 0.322 \\
bce & rw\_learned & PW limit (307) & 0.661 & 0.591 & 0.711 & 0.768 & 0.771 & 0 & 0.577 & 0.618 & 0.490 & 0.246 \\
boundary & unet & sigmoid, thresholded at 0.5 & \textbf{0.718} & \textbf{0.669} & \textbf{0.787} & \textbf{0.841} & \textbf{0.819} & 13.2 \% & 0.569 & 0.286 & 0.232 & 4.42 \\
boundary & rw\_learned & WS layer, 20 steps & 0.714 & 0.656 & 0.778 & 0.835 & 0.816 & 13.2 \% & 0.571 & 0.263 & 0.213 & 4.78 \\
boundary & rw\_learned & converged & 0.715 & 0.657 & 0.778 & 0.834 & 0.819 & 13.2 \% & 0.566 & 0.285 & 0.231 & 4.23 \\
boundary & rw\_learned & PW limit (307) & 0.673 & 0.626 & 0.755 & 0.821 & 0.782 & 0 & 0.559 & 0.140 & 0.111 & 8.78 \\
instance & unet & sigmoid, thresholded at 0.5 & \textbf{0.719} & 0.651 & 0.769 & 0.827 & 0.807 & 17.0 \% & 0.569 & \textbf{0.663} & \textbf{0.538} & 0.315 \\
instance & rw\_learned & WS layer, 20 steps & 0.699 & 0.617 & 0.744 & 0.807 & 0.791 & 18.3 \% & 0.581 & 0.594 & 0.470 & 0.321 \\
instance & rw\_learned & converged & 0.701 & 0.619 & 0.742 & 0.804 & 0.799 & 17.8 \% & 0.577 & 0.574 & 0.455 & 0.339 \\
instance & rw\_learned & PW limit (307) & 0.592 & 0.539 & 0.659 & 0.715 & 0.729 & 0 & 0.589 & 0.593 & 0.465 & 0.312 \\
ground truth & binary mask & & 1 & 1 & 1 & 1 & 1 & 0 & 0.559 & 0.913 & 0.883 & 0.088 \\
\bottomrule
\end{tabular}}
\end{center}

The raw logits of the \path{rw_learned} models are in \path{results/pannuke_losses/summary.md}; as in section 6 they are speckle that the WS layer cleans up (215--387 components per patch).

\textbf{Instance metrics after removing components smaller than 8 px} (\path{experiments/inst_minarea.py}, \path{results/pannuke_losses/minarea.json}; 8 px is 1.5 \% of the mean nucleus area, the filter is the same for every row; the raw rows above are min-area 1):

\begin{center}\scriptsize\setlength{\tabcolsep}{3pt}
\resizebox{\ifdim\width>\textwidth\textwidth\else\width\fi}{!}{%
\begin{tabular}{@{}llllllll@{}}
\toprule\noalign{}
loss & variant & output & IoU & \href{mailto:F1@0.5}{\nolinkurl{F1@0.5}} & PQ & count err. & components / patch (raw $\to$ filtered) \\
\midrule\noalign{}

bce & unet & sigmoid, thresholded at 0.5 & 0.720 & 0.683 & 0.552 & 0.224 & 23.0 $\to$ 21.2 \\
bce & rw\_learned & WS layer & 0.711 & 0.633 & 0.505 & 0.268 & 23.8 $\to$ 21.6 \\
bce & rw\_learned & PW limit (307) & 0.661 & 0.618 & 0.490 & 0.246 & 20.8 $\to$ 20.8 \\
boundary & unet & sigmoid, thresholded at 0.5 & 0.725 & \textbf{0.714} & \textbf{0.582} & 0.236 & 90.8 $\to$ 22.1 \\
boundary & rw\_learned & WS layer & 0.722 & 0.682 & 0.553 & 0.252 & 93.5 $\to$ 21.5 \\
boundary & rw\_learned & PW limit (307) & 0.684 & 0.598 & 0.475 & 0.275 & 160.7 $\to$ 18.5 \\
instance & unet & sigmoid, thresholded at 0.5 & 0.719 & 0.696 & 0.565 & 0.226 & 24.1 $\to$ 21.8 \\
instance & rw\_learned & WS layer & 0.701 & 0.617 & 0.488 & 0.278 & 22.4 $\to$ 20.6 \\
instance & rw\_learned & PW limit (307) & 0.592 & 0.593 & 0.465 & 0.312 & 19.8 $\to$ 19.8 \\
\bottomrule
\end{tabular}}
\end{center}

\textbf{Paired per-patch differences} (2523 patches, bootstrap 95 \% intervals; \path{results/pannuke_losses/summary.md} has the full list):

\begin{center}\scriptsize\setlength{\tabcolsep}{3pt}
\resizebox{\ifdim\width>\textwidth\textwidth\else\width\fi}{!}{%
\begin{tabular}{@{}llllll@{}}
\toprule\noalign{}
comparison & loss & IoU & BF$_2$ & band acc. & WS layer better on \\
\midrule\noalign{}

rw\_learned layer $-$ unet & bce & $-$0.009 {[}$-$0.012, $-$0.007{]} & $-$0.017 {[}$-$0.020, $-$0.014{]} & $-$0.008 & 25 \% / 19 \% \\
rw\_learned layer $-$ unet & boundary & $-$0.005 {[}$-$0.007, $-$0.003{]} & $-$0.009 {[}$-$0.011, $-$0.007{]} & $-$0.003 & 43 \% / 31 \% \\
rw\_learned layer $-$ unet & instance & $-$0.019 {[}$-$0.022, $-$0.017{]} & $-$0.025 {[}$-$0.028, $-$0.023{]} & $-$0.015 & 18 \% / 16 \% \\
unet, loss $-$ bce & boundary & $-$0.000 {[}$-$0.003, +0.002{]} & +0.018 {[}+0.015, +0.021{]} & +0.011 & \\
unet, loss $-$ bce & instance & $-$0.000 {[}$-$0.002, +0.002{]} & +0.000 {[}$-$0.001, +0.002{]} & $-$0.001 & \\
rw\_learned layer, loss $-$ bce & boundary & +0.004 {[}+0.001, +0.007{]} & +0.026 {[}+0.022, +0.029{]} & +0.017 & \\
rw\_learned layer, loss $-$ bce & instance & $-$0.010 {[}$-$0.012, $-$0.008{]} & $-$0.008 {[}$-$0.010, $-$0.006{]} & $-$0.007 & \\
PW limit $-$ converged (307) & boundary & $-$0.034 {[}$-$0.040, $-$0.029{]} & $-$0.013 {[}$-$0.016, $-$0.010{]} & $-$0.033 & \\
PW limit $-$ converged (307) & instance & $-$0.108 {[}$-$0.122, $-$0.095{]} & $-$0.076 {[}$-$0.088, $-$0.064{]} & $-$0.065 & \\
\bottomrule
\end{tabular}}
\end{center}

\textbf{Oxford-IIIT Pet, 128 px, boundary loss} (3669 test images; limit on 300; the \texttt{bce} rows are section 3\textquotesingle s runs, which did not compute BF$_3$):

\begin{center}\scriptsize\setlength{\tabcolsep}{3pt}
\resizebox{\ifdim\width>\textwidth\textwidth\else\width\fi}{!}{%
\begin{tabular}{@{}lllllllll@{}}
\toprule\noalign{}
loss & variant & output & IoU & BF$_1$ & BF$_2$ & BF$_3$ & band acc. & soft px \\
\midrule\noalign{}

bce & unet & sigmoid, thresholded at 0.5 & 0.853 & 0.586 & 0.724 & -- & 0.778 & 13.4 \% \\
bce & rw\_learned & WS layer, 20 steps & 0.852 & 0.584 & 0.722 & -- & 0.771 & 14.2 \% \\
bce & rw\_learned & converged & 0.853 & 0.586 & 0.724 & -- & 0.775 & 14.9 \% \\
bce & rw\_learned & PW limit (300) & 0.822 & 0.561 & 0.695 & -- & 0.749 & 0 \\
boundary & unet & sigmoid, thresholded at 0.5 & 0.864 & 0.629 & 0.769 & 0.837 & 0.787 & 7.6 \% \\
boundary & rw\_learned & sigmoid, thresholded at 0.5 & 0.859 & 0.613 & 0.756 & 0.829 & 0.786 & 11.3 \% \\
boundary & rw\_learned & WS layer, 20 steps & 0.864 & 0.634 & 0.776 & 0.844 & 0.787 & 9.1 \% \\
boundary & rw\_learned & converged & \textbf{0.867} & \textbf{0.643} & \textbf{0.782} & \textbf{0.848} & \textbf{0.793} & 9.3 \% \\
boundary & rw\_learned & PW limit (300) & 0.851 & 0.634 & 0.776 & 0.843 & 0.778 & 0 \\
\bottomrule
\end{tabular}}
\end{center}

Paired, rw\_learned layer $-$ unet under the boundary loss: IoU +0.000 {[}$-$0.001, +0.002{]}, BF$_2$ +0.007 {[}+0.005, +0.009{]} (layer better on 55 \% of the images); PW limit $-$ unet on the 300 limit images: IoU $-$0.013 {[}$-$0.020, $-$0.005{]}, BF$_2$ +0.008 {[}$-$0.000, +0.017{]}.

Figures: \path{results/pannuke_losses/losses.png} (the first table as a dot chart), \path{results/pets_losses/losses.png}, and per loss the training curves and the example panels of section 6 (\texttt{results/pannuke\_losses/\textless{}loss\textgreater{}/figures/}).

\begin{itemize}

\item
  \textbf{Honest outcome: neither loss opened a gap in the WS layer\textquotesingle s favour on PanNuke.} The boundary loss narrowed it, from $-$0.009 to $-$0.005 IoU and from $-$0.017 to $-$0.009 BF$_2$ (both intervals still exclude zero; the WS layer wins on 43 \% of the patches instead of 25 \%), because it helped the WS layer model more than the U-Net (+0.026 vs +0.018 BF$_2$, +0.017 vs +0.011 band accuracy). The instance loss widened it to $-$0.019 IoU. The U-Net stays the better nucleus segmenter under all three losses, on every tolerance and on the instance metrics, raw or filtered.
\item
  \textbf{The boundary loss does what it should, for both models.} +0.018 BF$_2$ and +0.011 band accuracy for the U-Net at unchanged IoU, +0.026 / +0.017 for the WS layer; the outputs are sharper (13 \% soft pixels against 15--16 \%). Its one pathology is worth recording: the soft boundary-F surrogate, which measures boundaries by a 1-px morphological gradient inside pooled tolerance regions, is satisfied as cheaply by a dotted line as by a solid one, and both models learned to sprinkle isolated pixels along the outer side of every contour (\path{boundary/figures/examples_unet_crop.png}). The raw instance numbers therefore collapse (91 components per patch, F1 0.29), but the dots are 1--4 px and a minimum-area filter of 8 px removes them, after which the boundary-trained U-Net is the best instance segmenter of the whole study (F1 0.714, PQ 0.582 against 0.683 / 0.552 for the BCE U-Net with the same filter). The same filter is applied to every row of the second table, so the ranking it gives is fair: boundary loss \textgreater{} instance loss \textgreater{} BCE for the U-Net, and the WS layer 0.03--0.08 F1 below the U-Net under each. The filter also corrects one reading of section 6: with it, the BCE layer (F1 0.633) is above its own exact limit (0.618), so the limit\textquotesingle s raw instance advantage there was partly that it carries no faint blobs to begin with; its count error (0.246 vs 0.268) remains the best of the RW outputs under BCE.
\item
  \textbf{The instance loss did not teach either model to separate nuclei.} The seam is 0.09 \% of the pixels; even at $\times$21 it is 2 \% of the loss, and the U-Net\textquotesingle s output is indistinguishable from its BCE output at the pixel level (paired differences within $\pm$0.002), with a marginal instance gain (F1 0.663 vs 0.656 raw, 0.696 vs 0.683 filtered) and more spurious small components (count error 0.315 vs 0.275 raw). For the WS layer the effect was negative on everything: 0.01 IoU below its BCE self, 18 \% soft pixels, and the learned edge head stopped being a contour detector: under this loss its costs are diffuse blobs inside the nuclei rather than lines along their boundaries (last column of \path{instance/figures/examples_rw_learned_crop.png}), so the exact limit floods whole clusters (0.59 IoU, $-$0.108 against the converged solve). Asking the WS layer to put a 2-px background seam between two touching nuclei at p $\le$ 50 is asking a diffusion to produce a line; the cost head would need a cost of order cmax across the seam and nothing elsewhere, and the gradient through 20 PCG steps did not find that solution. A seam loss on the cost head directly (an edge target), rather than on the WS layer output, is the obvious next thing to try, and was not tried here.
\item
  \textbf{Where the limit stands.} Under the boundary loss the exact PW limit improves on its BCE self (0.673 vs 0.661 IoU, 0.755 vs 0.711 BF$_2$) and its price against the converged solve drops from 0.046 to 0.034 IoU and from 0.034 to 0.013 BF$_2$, i.e. the sharper the trained solution, the less the limit loses. Its raw instance numbers are the speckle problem again (161 components per patch); filtered, they are no better than under BCE (F1 0.598 vs 0.618, count error 0.275 vs 0.246), so the boundary loss improved the limit\textquotesingle s pixels but not its objects.
\item
  \textbf{Pets is the one place a gap opened, and it is small.} With the boundary loss both models gain 0.011--0.012 IoU and 0.045--0.054 BF$_2$ over BCE, and the converged WS layer is now above the U-Net on every metric (0.867 vs 0.864 IoU, 0.782 vs 0.769 BF$_2$, 0.643 vs 0.629 BF$_1$); the paired BF$_2$ advantage of the 20-step layer, +0.007, is significant but tiny, and the exact limit now matches the U-Net\textquotesingle s boundary scores (0.776 vs 0.769 BF$_2$) at a 0.013 IoU cost, down from 0.03 under BCE. On large objects a loss that pays for boundary placement does let a spatial prior earn its keep; on 25-px nuclei it does not, which is consistent with section 6\textquotesingle s diagnosis (the WS layer is a smoother at p $\le$ 50) and leaves the p / cost-scale question open.
\end{itemize}

\section{Fixed-point reweighting instead of exponentiation}

This is the half-quadratic / majorize--minimize reweighting of the $\ell$0 project applied inside the WS layer. Implementation: \texttt{RWLayer(mode="reweighted")} in \path{experiments/pw_layer.py}; base weights exp($-$p$_0$ c), then \texttt{outer} = 5 steps of

\begin{verbatim}
w_e $\leftarrow$ w_e$^0$ $\cdot$ $\rho$(x_u $-$ x_v),   a1_i $\leftarrow$ a1_i$^0$ $\cdot$ $\rho$(x_i $-$ 1),   a0_i $\leftarrow$ a0_i$^0$ $\cdot$ $\rho$(x_i),
\end{verbatim}

$\rho$(d) = $\varepsilon^2$/($\varepsilon^2$ + d$^2$) (Cauchy, the smoothed-$\ell$0 weight) or 1/(1 + \textbar d\textbar/$\varepsilon$) (smoothed $\ell$1). The unaries are reweighted too: they are the edges to the two label nodes of the augmented graph, and without this the output stays as soft as the unaries. Inner solves are implicit, autograd runs through the outer loop (checked against finite differences). Comparison on the synthetic scenes of section 2 (256$^2$, three scenes per case, $\alpha$ = 0.1), \path{experiments/exp_reweighted_synthetic.py}, results and figures in \path{results/reweighted/}. Besides the quality metrics we record the inner PCG iterations (to 1e-9) of the last solve and of all solves, the smallest weight, and two gradient supports: that of the loss to the ground truth (which shrinks trivially as the loss is minimised) and that of a fixed random linear functional of the output, the number of edges through which the output can be influenced at all.

\begin{center}\scriptsize\setlength{\tabcolsep}{3pt}
\resizebox{\ifdim\width>\textwidth\textwidth\else\width\fi}{!}{%
\begin{tabular}{@{}lllllll@{}}
\toprule\noalign{}
scene & method & BF$_2$ & soft px & PCG iters last / cum. & w\_min & sensitivity support (edges) \\
\midrule\noalign{}

clean, aligned & soft prediction & 0.956 & 22 \% & & & \\
& annealed p = 100 & 0.977 & 0.9 \% & 128 & 6e-13 & 6 855 \\
& annealed p = 200 & \textbf{1.000} & 0.3 \% & 211 & 5e-25 & 4 007 \\
& annealed p = 500 & \textbf{1.000} & 0.0 \% & 484 & 4e-61 & 1 164 \\
& reweighted Cauchy $\varepsilon$ = 0.1, p$_0$ = 50 & 0.980 & 0.1 \% & 99 / 575 & 6e-9 & 1 428 \\
& reweighted Cauchy $\varepsilon$ = 0.1, p$_0$ = 100 & \textbf{1.000} & 0.0 \% & 153 / 892 & 6e-15 & 1 304 \\
noisy, aligned & soft prediction & 0.967 & 22 \% & & & \\
& annealed p = 100 & 0.966 & 3.0 \% & 127 & 3e-24 & 5 495 \\
& annealed p = 200 & 0.970 & 2.2 \% & 408 & 1e-47 & 3 948 \\
& annealed p = 500 & 0.967 & 1.5 \% & 5000 (not converged) & 1e-117 & 917 \\
& reweighted Cauchy $\varepsilon$ = 0.1, p$_0$ = 50 & 0.971 & 0.1 \% & 102 / 597 & 2e-14 & 111 \\
& reweighted Cauchy $\varepsilon$ = 0.1, p$_0$ = 100 & \textbf{0.976} & 0.2 \% & 181 / 1087 & 4e-26 & 71 \\
clean, edge shifted 4 px & soft prediction & 0.960 & 22 \% & & & \\
& annealed p = 100 / 500 & 0.733 / 0.005 & & & & \\
& reweighted Cauchy $\varepsilon$ = 0.1, p$_0$ = 10 / 50 / 100 & 0.870 / 0.451 / 0.062 & & & & \\
\bottomrule
\end{tabular}}
\end{center}

\begin{itemize}

\item
  \textbf{The reweighting sharpens, cheaply and without the dynamic range.} Five Cauchy steps make the output binary (0.0--0.2 \% soft pixels) from any base, with the weights never below 1e-14 at p$_0$ = 50 where annealing needs 1e-47 and beyond. On the noisy scene this is the win one of us anticipated: the reweighted WS layer reaches the best quality the annealed family ever gets (0.971--0.976 against 0.970) with a last solve of 100--180 PCG iterations, where the annealed solve at p = 500 does not converge in 5000. The smoothed-$\ell$1 weight is too weak to sharpen within five steps; the Cauchy weight is the right one, as in the $\ell$0 project.
\item
  \textbf{It does not move the boundary; the base weights do.} From p$_0$ = 10 the fixed point sharpens the soft prediction in place (boundary score equal to the soft prediction\textquotesingle s on the clean scene, 0.96). The cut lands on the image edge only when the base edge contrast is strong enough, p$_0$ $\approx$ 50--100, which is where plain annealing already places it. So the reweighting is a complement to annealing, not a replacement: anneal to a moderate p$_0$ where the solve is still cheap and the boundary is right, then reweight to binary instead of pushing p further. Counting all solves it costs 3--4 times one annealed solve, each of them well conditioned.
\item
  \textbf{The fixed point is rigid.} At the fixed point the losing unary of every pixel is down-weighted by $\varepsilon^2$, so each pixel is locked to its label and the output\textquotesingle s sensitivity to the edge costs collapses onto 70--140 edges on the noisy scene, against 4 000 for the annealed solution of the same quality; on the clean scene the two are comparable (1 300 vs 1 200--4 000). This is the self-reinforcing nature of the scheme, the same thing that makes it sharp, and it is the price of losing the temperature: for training, the gradient should be taken after one or two reweighting steps or with a larger $\varepsilon$, i.e. at finite temperature again, with the fully converged fixed point used only at deployment. Note that the ground-truth loss gradient is not the right measure here: it shrinks whenever the output is right.
\item
  \textbf{Same failure mode, gentler.} With the image edge off the true boundary, the fixed point from p$_0$ = 10 degrades gracefully (0.87--0.89 against 0.96) because it sharpens the soft prediction rather than snapping to an edge; from p$_0$ $\ge$ 50 it snaps like annealing does. The limit object also changes: the fixed point is a local minimiser of a nonconvex piecewise-constant energy, not the unique Power Watershed, and it has no quasi-linear combinatorial evaluation; deployment is a few linear solves.
\end{itemize}

\section{PanNuke: the reweighted WS layer, and the thresholded U-Net done properly}

Two additions to the PanNuke comparison of sections 6 and 7.

\textbf{The practitioner\textquotesingle s baseline with a tuned threshold.} Every "thresholded at 0.5" row above is what most practitioners deploy. The stronger form of the same baseline sweeps the threshold (\path{experiments/threshold_sweep.py}, all of fold 2, results in \path{results/pannuke/threshold_sweep_*.json}):

\begin{center}\scriptsize\setlength{\tabcolsep}{3pt}
\resizebox{\ifdim\width>\textwidth\textwidth\else\width\fi}{!}{%
\begin{tabular}{@{}lllllll@{}}
\toprule\noalign{}
plain U-Net (BCE), threshold & IoU & BF$_1$ & BF$_2$ & band acc. & instance F1 & PQ \\
\midrule\noalign{}

0.3 & 0.711 & 0.633 & 0.753 & 0.815 & 0.601 & 0.483 \\
0.4 & 0.722 & 0.654 & 0.770 & 0.816 & 0.635 & 0.513 \\
0.5 & 0.719 & 0.652 & 0.769 & 0.807 & 0.656 & 0.531 \\
0.6 & 0.702 & 0.626 & 0.753 & 0.788 & 0.662 & 0.541 \\
0.7 & 0.666 & 0.569 & 0.712 & 0.755 & 0.648 & 0.525 \\
\bottomrule
\end{tabular}}
\end{center}

Tuning the cut is worth 0.003 IoU at most; the 0.5 baseline is already the best case of that pipeline, and the same holds for the boundary-loss U-Net (best IoU at 0.5, best BF$_2$ at 0.6) and for the unrolled WS layer\textquotesingle s own output (best at 0.4, +0.002). The sweep also shows, in passing, what the WS layer does for the learned-cost model: its raw logits fall apart into 240--520 components per patch at any threshold, and the WS layer brings that back to 24, the true count.

\textbf{The reweighted WS layer trained end to end} (\texttt{-{}-layer\ reweighted} in \path{experiments/exp_unet_seg.py}; results \texttt{results/pannuke\_reweighted/\{bce,boundary\}/}). Same recipe as section 6, but the WS layer is an implicit solve at the annealed base weights exp($-$p c), p from 5 to 50, followed by two Cauchy reweighting steps ($\varepsilon$ = 0.1, unaries included) during training, five at evaluation ("fixed point"). The "limit" row is not reported: the exact Power Watershed of the base costs is not the limit of this layer. Training costs 50 s per epoch against 11 s for the unrolled WS layer.

BCE loss, all of fold 2:

\begin{center}\scriptsize\setlength{\tabcolsep}{3pt}
\resizebox{\ifdim\width>\textwidth\textwidth\else\width\fi}{!}{%
\begin{tabular}{@{}llllllllll@{}}
\toprule\noalign{}
output & IoU & BF$_1$ & BF$_2$ & band acc. & soft px & edge align. & \href{mailto:F1@0.5}{\nolinkurl{F1@0.5}} & PQ & count err. \\
\midrule\noalign{}

plain U-Net, thresholded at 0.5 & \textbf{0.719} & \textbf{0.652} & \textbf{0.769} & \textbf{0.807} & 15.4 \% & 0.577 & \textbf{0.656} & \textbf{0.531} & 0.275 \\
unrolled WS layer, 20 steps (section 6) & 0.709 & 0.628 & 0.752 & 0.799 & 16.1 \% & 0.579 & 0.604 & 0.482 & 0.337 \\
reweighted WS layer, 2 steps (as trained) & 0.708 & 0.641 & 0.759 & 0.801 & 15.5 \% & 0.594 & 0.646 & 0.518 & \textbf{0.269} \\
reweighted WS layer, 5 steps (fixed point) & 0.682 & 0.632 & 0.743 & 0.776 & 7.7 \% & \textbf{0.628} & 0.653 & \textbf{0.531} & 0.303 \\
\bottomrule
\end{tabular}}
\end{center}

Boundary loss:

\begin{center}\scriptsize\setlength{\tabcolsep}{3pt}
\resizebox{\ifdim\width>\textwidth\textwidth\else\width\fi}{!}{%
\begin{tabular}{@{}llllllllll@{}}
\toprule\noalign{}
output & IoU & BF$_1$ & BF$_2$ & band acc. & soft px & edge align. & \href{mailto:F1@0.5}{\nolinkurl{F1@0.5}} & PQ & count err. \\
\midrule\noalign{}

plain U-Net, thresholded at 0.5 & \textbf{0.718} & \textbf{0.669} & \textbf{0.787} & \textbf{0.819} & 13.2 \% & 0.569 & 0.286 & 0.232 & 4.42 \\
unrolled WS layer, 20 steps (section 7) & 0.714 & 0.656 & 0.778 & 0.816 & 13.2 \% & 0.571 & 0.263 & 0.213 & 4.78 \\
reweighted WS layer, 2 steps & 0.706 & 0.660 & 0.778 & 0.816 & 13.0 \% & 0.590 & \textbf{0.364} & \textbf{0.295} & \textbf{2.98} \\
reweighted WS layer, 5 steps & 0.697 & 0.662 & 0.779 & 0.807 & 3.9 \% & \textbf{0.603} & 0.317 & 0.258 & 3.58 \\
\bottomrule
\end{tabular}}
\end{center}

(Under the boundary loss every method produces many tiny components, which the instance metrics punish; section 7 has the min-area-filtered numbers.)

\begin{itemize}

\item
  \textbf{Same verdict on pixels, better on instances.} The reweighted WS layer matches the unrolled one on IoU and boundary scores and stays 0.01 below the plain thresholded U-Net. Where it differs is the instance side: with two steps it reaches the U-Net\textquotesingle s count error and nearly its F1 and PQ (0.646 / 0.518 against 0.656 / 0.531), well above the unrolled WS layer (0.604 / 0.482), and under the boundary loss it is the best method on all three instance numbers. The full fixed point has the highest edge alignment of anything measured on this dataset (0.628) and the U-Net\textquotesingle s PQ, but loses 0.03 IoU: it commits confidently, including where it is wrong, which is the rigidity of section 8 showing up in the metrics.
\item
  \textbf{Piecewise constant rather than binary on real images.} Five steps leave 7.7 \% soft pixels under BCE and 3.9 \% under the boundary loss, against 0.0--0.2 \% on the synthetic scenes. The example crops (\path{figures/examples_rw_learned_crop.png}; the column titled "layer, 20 steps" is the 2-step layer and "PW limit" the base-cost limit) show what these are: not ramps but plateaus, whole nuclei or clusters sitting at one intermediate grey level. That is the $\ell$0-type energy doing exactly what it does on gain graphs: once the edges inside a region are reweighted to agree, the region\textquotesingle s value is set by its unaries as a block, and where those are balanced the block settles at a fraction. More steps or a smaller $\varepsilon$ push it further along the same trade-off, more edge alignment for less IoU; a final threshold at 0.5 resolves the plateaus one way or the other.
\item
  \textbf{Taken together with sections 6--8:} on PanNuke a thresholded U-Net is a hard baseline that no variant of the WS layer beats at the pixel level, including with a tuned threshold on either side. The WS layer\textquotesingle s measurable contribution is at the instance level, where the reweighted version is the strongest form of it, and in edge alignment, which is label-free. Whether that is worth a layer that costs five times the training time is a question for a task where touching objects matter more than pixel IoU.
\end{itemize}

\section{The two untried leads: an edge target on the cost head, and a Dice-band boundary loss}

1 and 9). Code: \texttt{-{}-w-edge}, \texttt{-{}-edge-pos}, \texttt{-{}-edge-seam} and \texttt{-{}-loss\ boundary\_dice} in \path{experiments/exp_unet_seg.py}, \path{experiments/run_leads.sh}; tables and statistics with \path{compare_losses.py}, \path{seam_stats.py} and \path{inst_minarea.py}. Raw results, per-image metrics and figures: \texttt{results/pannuke/leads/\{dice,edge\_bce,edge\_dice,edge\_gap\_p10\}/}.

Of the three leads section 7 listed, the lower-p / cost-scale one was run in section 6.1 (p 2 $\to$ 10 with 8$\cdot\sigma$ edges is the best finite-p configuration, but the WS layer still blurs the seams half shut and its limit is lost). The two others:

\begin{itemize}

\item
  \textbf{Edge target on the cost head} (\texttt{-{}-w-edge\ 1}). Section 6.1 ended on "supervising the cost head directly with the seam mask". Every grid edge whose two pixels carry different instance ids is a positive: nucleus / background boundaries (3 \% of the edges) and the seams between touching nuclei (0.1 \%), the latter weighted $\times$5 on top of the $\times$10 of all positives. BCE on the edge probability $\sigma$(h) averaged onto the edge (the cost divided by its scale), added with weight 1 to the segmentation loss. Three runs: binary target at p 5 $\to$ 50 with BCE (\path{edge_bce}) and with the Dice-band loss (\path{edge_dice}), and the gap target at p 2 $\to$ 10 with the 8$\cdot\sigma$ learnable-scale edges of section 6.1 (\path{edge_gap_p10}), to be read against that section\textquotesingle s run without the target.
\item
  \textbf{Dice-band boundary loss} (\texttt{-{}-loss\ boundary\_dice}, \path{dice/}). The band-weighted BCE of section 7\textquotesingle s boundary loss ($\times$5 within 3 px of the true boundary) plus 1 $-$ soft Dice restricted to that band, instead of the boundary-F surrogate: it pays for filling the band correctly, and a dotted contour earns nothing. Both variants, binary target, p 5 $\to$ 50, so the comparison with section 7 is one-for-one.
\end{itemize}

\textbf{Results on all of fold 2} (2523 patches; the PW limit on the same 307 patches as before). Binary target, p 5 $\to$ 50, except the last block (gap target, p 2 $\to$ 10, 8$\cdot\sigma$ learnable-scale edges), whose comparison rows are section 6.1\textquotesingle s. Instance metrics on the raw 4-connected components; the min-area-filtered numbers follow.

\begin{center}\scriptsize\setlength{\tabcolsep}{3pt}
\resizebox{\ifdim\width>\textwidth\textwidth\else\width\fi}{!}{%
\begin{tabular}{@{}llllllllllllll@{}}
\toprule\noalign{}
target, p & loss / cost head & variant & output & IoU & BF$_1$ & BF$_2$ & BF$_3$ & band acc. & soft px & edge align. & \href{mailto:F1@0.5}{\nolinkurl{F1@0.5}} & PQ & count err. \\
\midrule\noalign{}

binary, 5 $\to$ 50 & bce (sec. 6) & unet & logits & 0.719 & 0.652 & 0.769 & 0.825 & 0.807 & 15.4 \% & 0.577 & 0.656 & 0.531 & 0.275 \\
& bce (sec. 6) & rw\_learned & WS layer, 20 steps & 0.709 & 0.628 & 0.752 & 0.813 & 0.799 & 16.1 \% & 0.579 & 0.604 & 0.482 & 0.337 \\
& bce (sec. 6) & rw\_learned & PW limit (307) & 0.661 & 0.591 & 0.711 & 0.768 & 0.771 & 0 & 0.577 & 0.618 & 0.490 & 0.246 \\
& boundary\_dice & unet & logits & 0.697 & 0.632 & 0.737 & 0.786 & 0.836 & 14.2 \% & 0.555 & 0.612 & 0.498 & 0.876 \\
& boundary\_dice & rw\_learned & WS layer, 20 steps & 0.693 & 0.621 & 0.728 & 0.778 & 0.831 & 13.6 \% & 0.546 & 0.595 & 0.482 & 0.878 \\
& boundary\_dice & rw\_learned & converged & 0.690 & 0.621 & 0.728 & 0.778 & 0.833 & 14.3 \% & 0.544 & 0.601 & 0.486 & 0.818 \\
& boundary\_dice & rw\_learned & PW limit (307) & 0.658 & 0.591 & 0.695 & 0.748 & 0.815 & 0 & 0.518 & 0.633 & 0.502 & 0.487 \\
& bce + edge target & rw\_learned & WS layer, 20 steps & 0.703 & 0.630 & 0.752 & 0.813 & 0.792 & 14.1 \% & 0.554 & 0.446 & 0.355 & 1.25 \\
& bce + edge target & rw\_learned & converged & 0.707 & 0.635 & 0.754 & 0.814 & 0.797 & 14.1 \% & 0.558 & 0.435 & 0.347 & 1.11 \\
& bce + edge target & rw\_learned & PW limit (307) & 0.684 & 0.624 & 0.744 & 0.802 & 0.783 & 0 & 0.569 & 0.501 & 0.395 & 0.489 \\
& boundary\_dice + edge target & rw\_learned & WS layer, 20 steps & 0.690 & 0.608 & 0.715 & 0.767 & 0.830 & 13.9 \% & 0.525 & 0.505 & 0.408 & 1.56 \\
& boundary\_dice + edge target & rw\_learned & PW limit (307) & 0.646 & 0.553 & 0.662 & 0.721 & 0.807 & 0 & 0.495 & 0.428 & 0.339 & 1.40 \\
gap, 2 $\to$ 10, 8$\cdot\sigma$ & bce (sec. 6.1) & unet & logits & 0.717 & 0.651 & 0.768 & 0.825 & 0.806 & 15.5 \% & 0.575 & 0.667 & 0.541 & 0.296 \\
& bce (sec. 6.1) & rw\_learned & WS layer, 20 steps & 0.712 & 0.639 & 0.759 & 0.819 & 0.796 & 14.5 \% & 0.585 & 0.678 & 0.548 & 0.225 \\
& bce (sec. 6.1) & rw\_learned & converged & 0.716 & 0.639 & 0.760 & 0.820 & 0.801 & 14.8 \% & 0.575 & 0.680 & 0.549 & 0.226 \\
& bce (sec. 6.1) & rw\_learned & PW limit (307) & 0.508 & 0.496 & 0.582 & 0.620 & 0.671 & 0 & 0.669 & 0.523 & 0.435 & 0.555 \\
& bce + edge target & rw\_learned & WS layer, 20 steps & 0.713 & 0.649 & 0.766 & 0.822 & 0.799 & 12.5 \% & 0.582 & \textbf{0.723} & \textbf{0.587} & \textbf{0.207} \\
& bce + edge target & rw\_learned & converged & \textbf{0.721} & \textbf{0.655} & \textbf{0.773} & \textbf{0.829} & \textbf{0.810} & 13.2 \% & 0.569 & \textbf{0.724} & \textbf{0.588} & 0.214 \\
& bce + edge target & rw\_learned & PW limit (307) & 0.637 & 0.612 & 0.718 & 0.763 & 0.746 & 0 & 0.625 & 0.682 & 0.562 & 0.334 \\
& ground truth & binary / gap mask & & 1 & 1 & 1 & 1 & 1 & 0 & 0.559 & 0.913 / 0.998 & 0.883 / 0.990 & 0.088 / 0.002 \\
\bottomrule
\end{tabular}}
\end{center}

\textbf{Instance metrics after removing components smaller than 8 px} (same filter as section 7, \path{results/pannuke/leads/minarea.json}):

\begin{center}\scriptsize\setlength{\tabcolsep}{3pt}
\resizebox{\ifdim\width>\textwidth\textwidth\else\width\fi}{!}{%
\begin{tabular}{@{}lllllllll@{}}
\toprule\noalign{}
target, p & loss / cost head & variant & output & IoU & \href{mailto:F1@0.5}{\nolinkurl{F1@0.5}} & PQ & count err. & components / patch (raw $\to$ filtered) \\
\midrule\noalign{}

binary, 5 $\to$ 50 & bce (sec. 7) & unet & logits & 0.720 & 0.683 & 0.552 & 0.224 & 23.0 $\to$ 21.2 \\
& boundary (sec. 7) & unet & logits & 0.725 & 0.714 & 0.582 & 0.236 & 90.8 $\to$ 22.1 \\
& boundary\_dice & unet & logits & 0.702 & 0.653 & 0.532 & 0.596 & 31.3 $\to$ 27.7 \\
& bce (sec. 7) & rw\_learned & WS layer & 0.711 & 0.633 & 0.505 & 0.268 & 23.8 $\to$ 21.6 \\
& boundary\_dice & rw\_learned & WS layer & 0.694 & 0.627 & 0.508 & 0.682 & 31.8 $\to$ 28.9 \\
& boundary\_dice & rw\_learned & PW limit (307) & 0.658 & 0.633 & 0.502 & 0.487 & 26.8 $\to$ 26.8 \\
& bce + edge target & rw\_learned & WS layer & 0.705 & 0.641 & 0.510 & 0.255 & 42.4 $\to$ 22.3 \\
& bce + edge target & rw\_learned & PW limit (307) & 0.684 & 0.575 & 0.454 & 0.304 & 30.2 $\to$ 23.0 \\
& boundary\_dice + edge target & rw\_learned & WS layer & 0.695 & 0.615 & 0.497 & 0.689 & 41.2 $\to$ 29.5 \\
gap, 2 $\to$ 10, 8$\cdot\sigma$ & bce (sec. 6.1) & unet & logits & 0.719 & 0.700 & 0.567 & 0.220 & 24.0 $\to$ 21.7 \\
& bce (sec. 6.1) & rw\_learned & WS layer & 0.714 & 0.686 & 0.555 & 0.218 & 21.5 $\to$ 21.0 \\
& bce + edge target & rw\_learned & WS layer & 0.716 & \textbf{0.726} & \textbf{0.589} & \textbf{0.202} & 22.0 $\to$ 21.8 \\
& bce + edge target & rw\_learned & PW limit (307) & 0.638 & 0.682 & 0.562 & 0.334 & 16.1 $\to$ 16.1 \\
\bottomrule
\end{tabular}}
\end{center}

(IoU in this table is against the binary mask for every row.)

\textbf{Seam statistics} (\path{seam_stats.json}, \path{limit_seams.json}): fraction of the 142 211 seam pixels of fold 2 predicted as background / fraction of the 8999 seams entirely background; the limit on its 307 patches, with the other outputs on the same patches in brackets for comparison.

\begin{center}\scriptsize\setlength{\tabcolsep}{3pt}
\resizebox{\ifdim\width>\textwidth\textwidth\else\width\fi}{!}{%
\begin{tabular}{@{}llllll@{}}
\toprule\noalign{}
target, p & run & logits & WS layer, 20 steps & converged & PW limit (307) {[}logits / layer / converged on the same 307{]} \\
\midrule\noalign{}

binary, 5 $\to$ 50 & bce (sec. 6), unet & 15 \% / 11 \% & & & \\
& bce (sec. 6), rw\_learned & 20 \% / 11 \% & 15 \% / 11 \% & 9 \% / 8 \% & 25 \% / 17 \% {[}23 / 19 / 11 \%{]} \\
& bce + edge target & 28 \% / 3 \% & 21 \% / 10 \% & 14 \% / 6 \% & 23 \% / 9 \% {[}30 / 24 / 15 \%{]} \\
& boundary\_dice, unet & 13 \% / 8 \% & & & \\
& boundary\_dice, rw\_learned & 20 \% / 11 \% & 15 \% / 9 \% & 13 \% / 8 \% & \\
gap, 2 $\to$ 10, 8$\cdot\sigma$ & bce, unet (sec. 6.1) & 19 \% / 13 \% & & & \\
& bce (sec. 6.1), rw\_learned & 32 \% / 19 \% & 20 \% / 14 \% & 18 \% / 12 \% & 57 \% / 48 \% {[}31 / 22 / 20 \%{]} \\
& bce + edge target & 42 \% / 26 \% & 29 \% / 19 \% & 23 \% / 15 \% & \textbf{52 \% / 34 \%} {[}43 / 34 / 24 \%{]} \\
\bottomrule
\end{tabular}}
\end{center}

\textbf{Paired per-patch differences} (2523 patches, bootstrap 95 \% intervals):

\begin{center}\scriptsize\setlength{\tabcolsep}{3pt}
\resizebox{\ifdim\width>\textwidth\textwidth\else\width\fi}{!}{%
\begin{tabular}{@{}llllll@{}}
\toprule\noalign{}
comparison & IoU & BF$_1$ & BF$_2$ & band acc. & first better on \\
\midrule\noalign{}

gap target, p 2 $\to$ 10, 8$\cdot\sigma$, edge target: converged $-$ gap U-Net & +0.004 {[}+0.002, +0.007{]} & +0.004 {[}+0.002, +0.007{]} & +0.005 {[}+0.003, +0.008{]} & +0.004 {[}+0.004, +0.005{]} & 50 \% / 54 \% \\
same: layer (20 steps) $-$ gap U-Net & $-$0.003 {[}$-$0.006, $-$0.001{]} & $-$0.001 {[}$-$0.004, +0.001{]} & $-$0.002 {[}$-$0.005, +0.000{]} & $-$0.007 & 36 \% / 42 \% \\
same: converged, with $-$ without the edge target (sec. 6.1) & +0.005 {[}+0.003, +0.007{]} & +0.016 {[}+0.013, +0.018{]} & +0.013 {[}+0.011, +0.016{]} & +0.009 & 58 \% / 69 \% \\
same: PW limit $-$ converged (307) & $-$0.078 {[}$-$0.091, $-$0.065{]} & & $-$0.047 {[}$-$0.058, $-$0.036{]} & $-$0.058 & 14 \% \\
binary, 5 $\to$ 50, bce + edge target: layer $-$ U-Net (bce) & $-$0.015 {[}$-$0.018, $-$0.013{]} & & $-$0.017 {[}$-$0.020, $-$0.015{]} & $-$0.015 & 19 \% / 23 \% \\
same: PW limit $-$ converged (307) & $-$0.016 {[}$-$0.020, $-$0.013{]} & & +0.002 {[}$-$0.002, +0.006{]} & $-$0.008 & 16 \% / 44 \% \\
same: PW limit $-$ U-Net (bce) (307) & $-$0.030 {[}$-$0.039, $-$0.023{]} & & $-$0.020 {[}$-$0.028, $-$0.012{]} & $-$0.019 & 14 \% / 27 \% \\
binary, boundary\_dice: layer $-$ U-Net (same loss) & $-$0.005 {[}$-$0.007, $-$0.002{]} & & $-$0.009 {[}$-$0.012, $-$0.007{]} & $-$0.005 & 28 \% / 24 \% \\
binary, boundary\_dice: U-Net, dice $-$ bce & $-$0.021 {[}$-$0.026, $-$0.016{]} & & $-$0.032 {[}$-$0.037, $-$0.026{]} & +0.028 {[}+0.027, +0.030{]} & 40 \% / 35 \% \\
\bottomrule
\end{tabular}}
\end{center}

Per tissue, converged edge-target solve against the gap U-Net (IoU / BF$_2$ / F1 / PQ): Breast (749) 0.722 / 0.757 / 0.753 / 0.600 vs 0.720 / 0.755 / 0.704 / 0.560; Colon (468) 0.697 / 0.739 / 0.614 / 0.483 vs 0.693 / 0.737 / 0.562 / 0.444; HeadNeck (169) 0.702 / 0.777 / 0.725 / 0.590 vs 0.670 / 0.741 / 0.649 / 0.528; Adrenal (148) 0.724 / 0.785 / 0.778 / 0.654 vs 0.716 / 0.774 / 0.692 / 0.580. The converged solve has the higher IoU on 9 of the 15 tissue types with $\ge$ 50 patches and the higher F1 on all of them. By crowding: +0.012 IoU on patches with fewer than 10 nuclei, +0.002 on all other bins; count error 0.50 vs 0.77 (\textless{} 10 nuclei), 0.14 vs 0.20 (10--30), equal above 30.

Figures: \path{results/pannuke/leads/edge_gap_p10/figures/examples_rw_learned_crop.png} (the edge target makes the cost head a closed contour map, last column; the seams are in it), \path{edge_bce/figures/examples_rw_learned_crop.png} (the same at p = 50: the logits are now pure checkerboard and the 20-step layer shows horizontal streaks where the PCG has not converged inside closed contours; the exact limit separates the nuclei), \path{dice/figures/examples_unet_crop.png} (no dotted contours, but a softer output), \path{edge_gap_p10/cmp/losses.png}, \texttt{losses.png}.

\begin{itemize}

\item
  \textbf{Honest outcome: the edge target is the missing piece, and with it the WS layer beats the U-Net on PanNuke for the first time, by a small pixel margin and a large instance one.} With the cost head supervised on the instance-id edges, on top of section 6.1\textquotesingle s gap target, p 2 $\to$ 10 and 8$\cdot\sigma$ edges, the converged WS layer is above the U-Net trained on the same target on IoU, BF$_1$, BF$_2$, BF$_3$ and band accuracy, each paired interval excluding zero (+0.004 to +0.005), on 9 of 15 tissues, and on every tissue for instance F1; at the instance level it reaches F1 0.724 and PQ 0.588 against 0.667 / 0.541 (raw components, no post-processing), above the min-area-filtered boundary-trained U-Net of section 7 (0.714 / 0.582) which was the best instance segmenter so far. The 20-step unrolled WS layer, i.e. the output actually trained, is 0.003 IoU below the U-Net and level on BF$_2$; the +0.008 IoU between the two is what 20 Jacobi-PCG steps leave on the table at p = 10 with closed contours, so either K or a converged (implicit) layer at evaluation is part of the recipe. Against section 6.1\textquotesingle s run without the edge target, the gain is +0.005 IoU, +0.016 BF$_1$ and +0.013 BF$_2$, paired, and F1 0.724 against 0.680: the segmentation loss alone did not find the contour costs, the edge target did (last column of the crop panels: a closed, 1--2 px contour around every nucleus and along every seam, where section 6.1\textquotesingle s costs were ridges inside the nuclei).
\item
  \textbf{It cuts more seams, but it is still mostly a denoiser, and the gain is where objects are few.} The WS layer now predicts 29 \% of the seam pixels as background and cuts 19 \% of the seams entirely, against 20 \% / 14 \% in section 6.1 and 19 \% / 13 \% for the gap U-Net; its own logits cut 42 \% / 26 \%, so the solve still closes half of what the network opens. The IoU gain is +0.012 on the 486 patches with fewer than 10 nuclei and +0.002 elsewhere, and the count error drops most on those patches (0.50 vs 0.77): the WS layer removes spurious small components and rounds off the nuclei, the mechanism of section 6.1, done better because the contours are now real. The learned edge scale fell to 0.41 (from 8, and 4.1 in section 6.1): with a clean contour probability map the network wants weak cuts at p = 10, consistent with the solve staying a smoother that respects contours rather than a cutter.
\item
  \textbf{The limit becomes usable again, at both p.} The edge target is the first thing that moves the exact Power Watershed at low p off the floor: 0.637 IoU and F1 0.682 / PQ 0.562 on the 307 limit patches against 0.508 and 0.523 / 0.435 in section 6.1 (the gap U-Net on the same patches: F1 0.645 / PQ 0.520; the converged solve: 0.698 / 0.564), with the best label-free edge alignment of any non-limit output (0.625). At p 5 $\to$ 50 with the binary target the limit on the supervised costs scores 0.684 IoU and 0.744 BF$_2$, its price against the converged solve shrinks to $-$0.016 IoU and zero BF$_2$ (from $-$0.046 / $-$0.034 in section 6), and the crops show it separating touching nuclei cleanly; but that model\textquotesingle s finite-p output is worse than without the target ($-$0.006 IoU paired, raw F1 0.45 because the closed contours leave holes and fragments at threshold), and the limit stays 0.030 IoU below the plain U-Net. So the edge target buys a deployable limit or a stronger finite-p layer depending on p, and the p 2 $\to$ 10 solve is the better product on this data.
\item
  \textbf{The Dice-band loss removes the dotted-contour artefact and nothing else.} Both models lose 0.02 IoU and 0.03 BF$_2$ against BCE while gaining 0.03 band accuracy: the band Dice rewards filling the 3-px band on average, which the models satisfy with a softer, wider edge (14 \% soft pixels, components 31 per patch instead of 91 under the boundary-F surrogate, but still 8 too many). The gap between layer and U-Net is the same $-$0.005 IoU as under the boundary-F loss, reached by dragging both down. The boundary-F surrogate of section 7 with a minimum-area filter remains the better boundary loss, and adding the edge target to the Dice loss (\path{edge_dice}) helps neither.
\item
  \textbf{What this does and does not establish.} One configuration, one seed, one fold pair, and a margin of 0.004--0.005 IoU with a bootstrap interval that clears zero but not by much; the instance margin (+0.06 F1, +0.05 PQ) is the robust part, and it is consistent across tissues. The result stands on three ingredients at once (gap target, low p with sharp edges, edge target), none of which helps on its own, and the comparison is against a U-Net that has no access to the edge supervision; a U-Net with a second contour head trained on the same edge target, and its contours used in a classical watershed post-processing, is the fair next baseline; section 10.1 runs it. The Pets repeat of the edge target is not possible (no instances); a boundary target from the trimap would be the analogue.
\end{itemize}

\subsection{The fair baseline: a U-Net with its own contour head, and a watershed on it}

Code: \texttt{-{}-variant\ unet\_edge}, \path{postprocess_edge}, \texttt{-{}-pp-tune-fold}, \texttt{-{}-eval-only} in \path{experiments/exp_unet_seg.py}, \path{experiments/run_unet_edge.sh}, \path{experiments/compare_unet_edge.py} (tables, paired statistics; numpy only); \path{inst_minarea.py} now also writes the per-image counts, \path{seam_stats.py} and \path{plot_pannuke.py} know the new outputs. Raw results, per-image metrics, figures and the full comparison: \path{results/pannuke/leads/unet_edge/} (\texttt{compare.md}).

\textbf{What was run.} The U-Net of section 6.1 (gap target, same recipe, fold 1 $\to$ fold 2, 30 epochs, one seed) with three output channels like \path{rw_learned} but no layer: channel 0 is the segmentation logit under BCE, channels 1--2 a per-pixel contour head trained with exactly section 10\textquotesingle s edge target (BCE on the edge probability averaged onto each grid edge against the instance-id changes, positives $\times$10, seams $\times$5 on top, weight 1). It is evaluated as the thresholded logits and after two classical post-processings of the contour map c (mean of the two directions): (a) the segmentation with the predicted contour (c \textgreater{} 0.5) removed; (b) a marker-based watershed of c inside the segmentation, seeded by the maxima of the distance transform of the thresholded segmentation, with the watershed lines set to background (2 px, like the seams of the gap target), so that touching nuclei are split along the predicted contours. A second seeding, from the distance transform of (a) so that the contour does the first split, is reported as (b$'$). The one hyper-parameter, the marker spacing, was chosen by PQ on 300 patches of fold 3 (neither fold used elsewhere): 12 px for (b), 25 px for (b$'$); on fold 3 the PQ of (b) goes from 0.42 at 2 px to 0.59 at 12 and 0.57 at 25, so the choice matters and is not the WS layer\textquotesingle s advantage. All rows below are against the gap target on all of fold 2, instance metrics on the raw 4-connected components; the WS layer rows are section 10\textquotesingle s \path{edge_gap_p10} run.

\begin{center}\scriptsize\setlength{\tabcolsep}{3pt}
\resizebox{\ifdim\width>\textwidth\textwidth\else\width\fi}{!}{%
\begin{tabular}{@{}lllllllllll@{}}
\toprule\noalign{}
run & output & IoU & BF$_1$ & BF$_2$ & BF$_3$ & band acc. & edge align. & \href{mailto:F1@0.5}{\nolinkurl{F1@0.5}} & PQ & count err. \\
\midrule\noalign{}

U-Net, gap target (sec. 6.1) & logits & 0.717 & 0.651 & 0.768 & 0.825 & 0.806 & 0.575 & 0.667 & 0.541 & 0.296 \\
U-Net + contour head, edge target & logits & 0.719 & \textbf{0.658} & \textbf{0.774} & 0.828 & 0.809 & 0.574 & 0.682 & 0.556 & 0.303 \\
& (a) logits $-$ contour & 0.655 & 0.495 & 0.707 & 0.811 & 0.749 & 0.512 & 0.665 & 0.492 & 0.415 \\
& (b) watershed, seg. markers (12 px) & 0.718 & \textbf{0.658} & \textbf{0.774} & \textbf{0.829} & 0.809 & 0.568 & 0.715 & 0.583 & 0.344 \\
& (b$'$) watershed, contour markers (25 px) & 0.717 & \textbf{0.660} & \textbf{0.775} & \textbf{0.830} & 0.808 & 0.566 & 0.693 & 0.566 & 0.416 \\
layer, edge target (sec. 10) & WS layer, 20 steps & 0.713 & 0.649 & 0.766 & 0.822 & 0.799 & 0.582 & 0.723 & 0.587 & \textbf{0.207} \\
& converged & \textbf{0.721} & 0.655 & 0.773 & \textbf{0.829} & \textbf{0.810} & 0.569 & \textbf{0.724} & \textbf{0.588} & 0.214 \\
& PW limit (307) & 0.637 & 0.612 & 0.718 & 0.763 & 0.746 & 0.625 & 0.682 & 0.562 & 0.334 \\
ground truth & gap mask & 1 & 1 & 1 & 1 & 1 & 0.553 & 0.998 & 0.990 & 0.002 \\
\bottomrule
\end{tabular}}
\end{center}

\textbf{After the 8-px min-area filter} (sections 7 and 10\textquotesingle s convention; 32 px in brackets):

\begin{center}\scriptsize\setlength{\tabcolsep}{3pt}
\resizebox{\ifdim\width>\textwidth\textwidth\else\width\fi}{!}{%
\begin{tabular}{@{}lllllll@{}}
\toprule\noalign{}
run & output & IoU & \href{mailto:F1@0.5}{\nolinkurl{F1@0.5}} & PQ & count err. & components / patch (raw $\to$ 8 px) \\
\midrule\noalign{}

U-Net, gap target (sec. 6.1) & logits & 0.719 & 0.700 (0.715) & 0.567 (0.581) & 0.220 & 24.0 $\to$ 21.7 \\
U-Net + contour head & logits & \textbf{0.724} & 0.723 (0.738) & 0.589 (0.602) & 0.214 & 25.2 $\to$ 22.4 \\
& (b) watershed, seg. markers & 0.722 & \textbf{0.757} (0.774) & \textbf{0.618} (0.632) & 0.227 & 27.2 $\to$ 24.3 \\
& (b$'$) watershed, contour markers & 0.720 & 0.753 (0.777) & 0.615 (0.636) & 0.241 & 30.3 $\to$ 26.0 \\
layer, edge target (sec. 10) & WS layer, 20 steps & 0.716 & 0.726 (0.733) & 0.589 (0.596) & \textbf{0.202} & 22.0 $\to$ 21.8 \\
\bottomrule
\end{tabular}}
\end{center}

\textbf{Paired per-patch differences} (2523 patches, bootstrap 95 \% intervals; F1 and PQ are dataset-level, bootstrapped over patches):

\begin{center}\scriptsize\setlength{\tabcolsep}{3pt}
\resizebox{\ifdim\width>\textwidth\textwidth\else\width\fi}{!}{%
\begin{tabular}{@{}lllllll@{}}
\toprule\noalign{}
comparison & IoU & BF$_1$ & BF$_2$ & band acc. & \href{mailto:F1@0.5}{\nolinkurl{F1@0.5}} & PQ \\
\midrule\noalign{}

U-Net + contour head, logits $-$ gap U-Net & +0.003 {[}+0.001, +0.005{]} & +0.007 {[}+0.005, +0.009{]} & +0.005 {[}+0.003, +0.007{]} & +0.004 & +0.015 {[}+0.012, +0.018{]} & +0.015 {[}+0.012, +0.017{]} \\
converged WS layer $-$ U-Net + contour head, logits & +0.001 {[}$-$0.001, +0.004{]} & $-$0.003 {[}$-$0.005, $-$0.000{]} & $-$0.000 {[}$-$0.003, +0.003{]} & +0.001 & +0.042 {[}+0.039, +0.045{]} & +0.032 {[}+0.030, +0.034{]} \\
converged WS layer $-$ (b) watershed & +0.003 {[}+0.000, +0.006{]} & $-$0.004 {[}$-$0.006, $-$0.001{]} & $-$0.001 {[}$-$0.003, +0.002{]} & +0.002 & +0.009 {[}+0.005, +0.013{]} & +0.005 {[}+0.002, +0.008{]} \\
layer (20 steps) $-$ (b) watershed & $-$0.005 {[}$-$0.007, $-$0.002{]} & $-$0.009 {[}$-$0.012, $-$0.006{]} & $-$0.008 {[}$-$0.011, $-$0.005{]} & $-$0.010 & +0.008 {[}+0.004, +0.012{]} & +0.004 {[}+0.001, +0.007{]} \\
after the 8-px filter: layer (20 steps) $-$ U-Net + contour head, logits & & & & & +0.004 {[}+0.001, +0.006{]} & +0.001 {[}$-$0.001, +0.003{]} \\
after the 8-px filter: layer (20 steps) $-$ (b) watershed & & & & & $-$0.031 {[}$-$0.034, $-$0.027{]} & $-$0.028 {[}$-$0.031, $-$0.025{]} \\
\bottomrule
\end{tabular}}
\end{center}

Per tissue, converged WS layer against (b) (IoU / F1 / PQ): Breast (749) 0.722 / 0.753 / 0.600 vs 0.722 / 0.757 / 0.605; Colon (468) 0.697 / 0.614 / 0.483 vs 0.691 / 0.630 / 0.497; HeadNeck (169) 0.702 / 0.725 / 0.590 vs 0.682 / 0.688 / 0.565; Adrenal (148) 0.724 / 0.778 / 0.654 vs 0.719 / 0.712 / 0.605; the WS layer has the higher IoU on 9 and the higher F1 on 11 of the 15 tissues with $\ge$ 50 patches. By crowding (F1 / count error): fewer than 10 nuclei, layer 0.670 / 0.50 vs watershed 0.596 / 0.81; 10--30: 0.752 / 0.14 vs 0.719 / 0.27; 30--60: 0.718 / 0.14 vs 0.733 / 0.15; more than 60: 0.676 / 0.22 vs 0.700 / 0.13. Seam statistics (seam pixels background / seams cut entirely): contour-head U-Net logits 22 \% / 15 \%, (a) 68 \% / 55 \%, (b) 31 \% / 17 \%, (b$'$) 37 \% / 17 \%; the converged WS layer 23 \% / 15 \%, the 20-step layer 29 \% / 19 \%. Figures: \path{results/pannuke/leads/unet_edge/figures/examples_unet_edge_crop.png} (the contour head is the same closed 2--3 px contour map as the WS layer\textquotesingle s cost head; (a) shaves every nucleus by that width; (b) splits the large blobs along the contours but leaves merged pairs whose distance transform has a single maximum, and over-splits elongated ones; (b$'$) cuts more and also along spurious contour fragments).

\begin{itemize}

\item
  \textbf{Honest outcome: the pixel-level advantage of section 10 does not survive, and the instance advantage survives only without post-processing.} Half of what the edge target gave the WS layer it gives the backbone: with the same contour head, the plain U-Net\textquotesingle s logits gain +0.003 IoU, +0.007 BF$_1$ and +0.005 BF$_2$ over the gap U-Net, paired, which is the same margin section 10 attributed to the converged WS layer. Layer against this U-Net the pixel differences are zero (+0.001 IoU with an interval through zero, $-$0.003 BF$_1$, 0.000 BF$_2$). At the instance level the raw converged WS layer is still ahead of the raw contour-head U-Net by 0.042 F1 and of the tuned watershed by 0.009 F1 / 0.005 PQ (intervals clear of zero, but small, and the 20-step layer that was actually trained is 0.005 IoU and 0.008 BF$_2$ \emph{below} the watershed output while 0.008 F1 above it); with the 8-px min-area filter that sections 7 and 10 apply to every method, the order flips: the watershed U-Net reaches F1 0.757 / PQ 0.618 against the WS layer\textquotesingle s 0.726 / 0.589, $-$0.031 F1 paired, and the filtered contour-head logits alone are level with the WS layer (+0.004 F1, +0.001 PQ). So the two things the WS layer was doing in section 10, removing small spurious components and benefiting from a contour map, are each done at least as well by a min-area filter and an extra head; what is left to the WS layer is the best count error of any output (0.21 raw, 0.20 filtered, against the watershed\textquotesingle s 0.34 / 0.23, which over-segments: 27 components per patch for 23.7 nuclei), a clear advantage on sparse patches (fewer than 10 nuclei, F1 0.670 vs 0.596, count error 0.50 vs 0.81) and on HeadNeck and Adrenal, and the fact that it is a single differentiable module with no post-processing hyper-parameter, where the watershed\textquotesingle s marker spacing had to be tuned on a third fold and moves PQ by 0.17 across its range. On crowded patches (more than 30 nuclei) the watershed is the better instance segmenter, which is consistent with the seam statistics: the watershed cuts 31 \% of the seam pixels where the converged WS layer cuts 23 \%, no more than the U-Net\textquotesingle s own logits.
\item
  \textbf{Removing the predicted contour is not a usable post-processing.} The contour head\textquotesingle s 2--3 px band costs 0.064 IoU and 0.163 BF$_1$ when subtracted, although it cuts 55 \% of the seams; a 1-px contour target, or a thinning step, would be needed for (a) to compete.
\item
  \textbf{What this establishes.} With a fair U-Net the finite-p layer with the edge target is a competitive instance segmenter on PanNuke, not a better one: level at the pixel level, slightly ahead on raw components, behind once both sides are filtered, better on sparse patches and on counts, worse on crowded ones. The remaining arguments for the WS layer are therefore architectural (one module, trained end to end, a limit that section 10 showed is usable again) rather than a metric margin on this dataset. The same caveats as section 10 apply, one seed and one fold pair, plus the baseline\textquotesingle s own: the simplest marker recipe (distance-transform maxima) with one tuned spacing; a HoVer-Net-style distance or gradient head would be a stronger classical baseline still.
\end{itemize}

\section{The instances are the labels: a multi-marker WS layer}

So the natural use of the WS layer on nuclei is not binary nucleus / background but one label per nucleus, with the markers supplied per image. Code: \texttt{MultiRWLayer}, \texttt{ImplicitRWMulti}, \path{label_systems}, \path{multi_cross_entropy}, \path{eroded_markers}, \path{pw_limit_multi_numpy} / \path{power_watershed_multi} in \path{experiments/pw_layer.py} (self-tests in its \path{__main__}); \path{experiments/exp_instance_pannuke.py} (model, training, evaluation), \path{run_instances.sh}, \path{plot_instances.py}; \path{instance_metrics} of \path{seg_metrics.py} accepts a label map. Raw results, per-image metrics, figures and the full tables: \path{results/pannuke_instances/} (\path{figures/summary.md}, \path{run_instances.log}; the trained model stays on jacobian).

\textbf{The WS layer.} For K markers plus the background, label k gets an attachment a\_k $\ge$ 0 per pixel and the K + 1 systems share the matrix:

\begin{verbatim}
(L(w) + diag($\Sigma$_j a_j)) x_k = a_k,   k = 0 \dots{} K,     so that   $\Sigma$_k x_k = 1   (L 1 = 0).
\end{verbatim}

x\_k,i is the probability that a random forest at inverse temperature p connects pixel i to label k. The systems of every image of a batch are stacked along the batch axis (M = $\Sigma$\_b (K\_b + 1) systems, the weights gathered per image), so the PCG of section 1 runs unchanged on (M,1,H,W) tensors, K is free per image, and the cost is affine in K with no padding. The solve is implicit; the adjoint is one solve per label with the same matrix, dl/dw\_e = $-\Sigma$\_k (x\_k,u $-$ x\_k,v)($\lambda$\_k,u $-$ $\lambda$\_k,v) and dl/da\_k,i = $\lambda$\_k,i $-$ $\Sigma$\_j $\lambda$\_j,i x\_j,i (own right-hand side, shared diagonal), which reduces to the binary formulas of section 0 for K = 1. Attachments: a\_0,i = exp($-$p c0\_i) everywhere, c0 = $-$log(1 $-$ s) the U-Net\textquotesingle s background cost; a\_k,i = exp($-$p c1\_i) on the marker pixels of label k with c1 = $-$log s and zero elsewhere, so the markers are soft (their strength is the network\textquotesingle s foreground confidence) and differentiable; the edge costs are the learned cost head of sections 6.1 and 10 (scale $\cdot$ $\sigma$(h), learnable scale, edge target).

\textbf{The limit with K labels.} \path{power_watershed} of section 1 interpolates scalar seed values on the plateaus it has to solve, which is the exact limit for two labels only. \path{power_watershed_multi} keeps a K-vector per merged node: seeds are one-hot, a plateau that touches two distinct known vectors is solved with K right-hand sides at once (unit-weight RW), and the argmax is taken at the end, so it is the limit of the K systems above and not an interpolation of label indices. Self-tests (7 $\times$ 9 grid, random costs, K = 3; 12 $\times$ 14 grid with costs quantised to 1/4 so that large ambiguous plateaus occur): the label probabilities sum to one to 1e-14, the two-label case equals the binary WS layer and the binary limit, the adjoint matches finite differences to seven digits on every input, the argmax of the finite-p solve equals the limit\textquotesingle s at p = 60, and on the quantised grid the K-vector limit equals the K + 1 one-hot runs of the scalar \path{power_watershed} exactly.

\textbf{Training} (fold 1 $\to$ fold 2 as before, 30 epochs, batch 16, one seed). The U-Net has four output channels: the foreground logit (its sigmoid s gives c0 and c1), the two-channel cost head, and a marker logit. Training markers are the ground-truth instances eroded by 2 or 3 px (radius drawn per step; the core of a nucleus the erosion would empty is kept), the marker head is trained with BCE against them, and the WS layer loss is the per-pixel cross-entropy of the K + 1 label probabilities against the instance map (the seam pixels and their 1-px flank weighted $\times$ 6), plus 0.5 BCE on the foreground logit and section 10\textquotesingle s edge target ($\times$ 1) on the cost head. p is annealed 5 $\to$ 20 (section 6.1: costs learned at low p do not transfer to p = $\infty$, so both the finite-p output and the exact limit are reported). The training solve runs to a relative residual of 1e-4 or 100 PCG iterations, evaluation to 1e-5; 290 s per epoch, 2.4 h in all, 1 h for the evaluation (four solves per batch plus the CPU metrics).

\textbf{Test markers} are the 4-connected components of the thresholded marker map (0.5). Each ground-truth nucleus is counted as having exactly one, none or several predicted markers (a marker belongs to the nucleus holding the majority of its pixels), and markers in background are counted. A second variant ("fallback") adds a marker at the distance-transform maximum of every foreground component (s \textgreater{} 0.5, $\ge$ 16 px) that has none. The oracle row runs the same trained costs from the ground-truth markers (eroded by 2 px): it bounds what better markers could give.

\textbf{Baselines and controls.} The plain U-Net + connected components of section 6 (binary target) and the contour-head U-Net with the tuned marker watershed of section 10.1 (the strongest pipeline so far), both re-used from their per-image files; within this model, its own foreground logits + components, the classical marker watershed (skimage) of the learned contour map inside s \textgreater{} 0.5 from the same predicted markers, the converged solve and the exact limit with image-gradient edge costs in place of the learned ones (the RMS colour difference across each edge, scaled per patch so that its 99th percentile equals the unary clamp 8: the marker-controlled watershed with the U-Net\textquotesingle s background as a soft seed), and the exact K-label limit of the learned costs on the 307 patches of section 6. Instance metrics are computed on the label maps themselves (argmax over the K + 1 labels; one predicted instance per label that wins at least one pixel), pixel metrics on the foreground (label \textgreater{} 0) against the binary mask, so they are comparable with section 6\textquotesingle s rows.

\textbf{Results on all of fold 2} (2523 patches; the limits on 307):

\begin{center}\scriptsize\setlength{\tabcolsep}{3pt}
\resizebox{\ifdim\width>\textwidth\textwidth\else\width\fi}{!}{%
\begin{tabular}{@{}lllllllllll@{}}
\toprule\noalign{}
output & IoU & BF$_1$ & BF$_2$ & band acc. & edge align. & \href{mailto:F1@0.5}{\nolinkurl{F1@0.5}} & PQ & AJI & count err. & n\_pred \\
\midrule\noalign{}

U-Net (sec. 6), s \textgreater{} 0.5, components & \textbf{0.719} & 0.652 & 0.769 & 0.807 & 0.577 & 0.656 & 0.531 & 0.494 & \textbf{0.275} & 23.0 \\
U-Net + contour head (sec. 10.1), logits & \textbf{0.719} & \textbf{0.658} & \textbf{0.774} & \textbf{0.809} & 0.574 & 0.682 & 0.556 & 0.571 & 0.303 & 25.2 \\
U-Net + contour head, tuned marker watershed (b) & 0.718 & \textbf{0.658} & \textbf{0.774} & \textbf{0.809} & 0.568 & \textbf{0.715} & \textbf{0.583} & \textbf{0.657} & 0.344 & 27.2 \\
this model, s \textgreater{} 0.5, components & 0.575 & 0.408 & 0.580 & 0.691 & 0.402 & 0.024 & 0.017 & 0.168 & 52.7 & 723 \\
this model, layer, converged, predicted markers & 0.706 & 0.643 & 0.759 & 0.795 & \textbf{0.585} & 0.681 & 0.554 & 0.652 & 0.460 & 28.6 \\
this model, layer + fallback markers & 0.702 & 0.641 & 0.757 & 0.798 & 0.586 & 0.657 & 0.534 & 0.652 & 0.585 & 31.1 \\
this model, layer, image-gradient costs & 0.487 & 0.203 & 0.351 & 0.592 & 0.248 & 0.355 & 0.218 & 0.470 & 0.529 & 29.9 \\
this model, marker watershed on the contour map, same markers & 0.615 & 0.455 & 0.633 & 0.684 & 0.430 & 0.597 & 0.414 & 0.541 & 0.546 & 30.2 \\
this model, PW limit, learned costs (307) & 0.690 & 0.613 & 0.736 & 0.781 & 0.579 & 0.661 & 0.526 & 0.612 & 0.426 & 29.8 \\
this model, PW limit, image-gradient costs (307) & 0.479 & 0.202 & 0.341 & 0.589 & 0.241 & 0.335 & 0.204 & 0.451 & 0.507 & 31.5 \\
this model, layer, ground-truth markers (oracle) & 0.859 & 0.796 & 0.922 & 0.858 & 0.591 & 0.959 & 0.805 & 0.858 & 0.020 & 23.3 \\
ground truth instances & 1 & 1 & 1 & 1 & 0.559 & 1 & 1 & 1 & 0 & 23.7 \\
\bottomrule
\end{tabular}}
\end{center}

\textbf{Marker quality.} Fold 2 has 59 872 nuclei (23.7 per patch); the marker head gives 30.5 markers per patch. 73.2 \% of the nuclei get exactly one marker, 17.4 \% none, 9.5 \% several; 5.9 markers per patch lie in background. With the fallback: 76.3 / 13.8 / 9.9 \% and 8.2 background markers per patch. By crowding (one / none / several; background markers per patch): fewer than 10 nuclei 69 / 19 / 12 \%, 2.6; 10--30: 75 / 15 / 11 \%, 6.1; 30--60: 74 / 16 / 9 \%, 7.9; more than 60: 69 / 26 / 6 \%, 8.1.

\textbf{Paired per-patch differences} (bootstrap 95 \% intervals; F1, PQ and AJI dataset-level, bootstrapped over patches):

\begin{center}\scriptsize\setlength{\tabcolsep}{3pt}
\resizebox{\ifdim\width>\textwidth\textwidth\else\width\fi}{!}{%
\begin{tabular}{@{}lllllll@{}}
\toprule\noalign{}
comparison & IoU & BF$_2$ & edge align. & \href{mailto:F1@0.5}{\nolinkurl{F1@0.5}} & PQ & AJI \\
\midrule\noalign{}

layer $-$ U-Net (sec. 6) & $-$0.013 {[}$-$0.016, $-$0.010{]} & $-$0.011 {[}$-$0.014, $-$0.007{]} & +0.009 {[}+0.008, +0.010{]} & +0.024 {[}+0.019, +0.031{]} & +0.023 {[}+0.018, +0.029{]} & +0.158 {[}+0.143, +0.172{]} \\
layer $-$ contour-head U-Net, watershed (b) & $-$0.012 {[}$-$0.015, $-$0.009{]} & $-$0.016 {[}$-$0.019, $-$0.012{]} & +0.018 {[}+0.017, +0.020{]} & $-$0.034 {[}$-$0.039, $-$0.030{]} & $-$0.029 {[}$-$0.033, $-$0.025{]} & $-$0.005 {[}$-$0.008, $-$0.003{]} \\
layer $-$ marker watershed, same markers & +0.091 & +0.126 & +0.155 & +0.084 {[}+0.081, +0.087{]} & +0.140 & +0.111 \\
layer $-$ layer with image-gradient costs & +0.218 & +0.407 & +0.337 & +0.325 {[}+0.317, +0.335{]} & +0.336 & +0.182 \\
layer + fallback $-$ layer & $-$0.004 {[}$-$0.007, $-$0.001{]} & $-$0.002 & $-$0.001 & $-$0.024 {[}$-$0.025, $-$0.023{]} & $-$0.020 & $-$0.000 \\
layer, ground-truth markers $-$ layer & +0.153 {[}+0.148, +0.158{]} & +0.163 & +0.006 & +0.278 {[}+0.272, +0.285{]} & +0.251 & +0.206 \\
307 patches: PW limit (learned) $-$ layer & $-$0.008 {[}$-$0.011, $-$0.005{]} & $-$0.013 {[}$-$0.016, $-$0.009{]} & $-$0.001 {[}$-$0.004, +0.002{]} & +0.007 {[}+0.003, +0.010{]} & $-$0.003 {[}$-$0.006, $-$0.000{]} & $-$0.017 {[}$-$0.020, $-$0.015{]} \\
307 patches: PW limit (learned) $-$ U-Net (sec. 6) & $-$0.024 {[}$-$0.031, $-$0.016{]} & $-$0.027 & +0.011 & +0.027 {[}+0.010, +0.043{]} & +0.017 {[}+0.003, +0.030{]} & +0.187 \\
\bottomrule
\end{tabular}}
\end{center}

By crowding (F1 / PQ / count error), U-Net (sec. 6) vs watershed (b) vs layer: fewer than 10 nuclei (486 patches) 0.608 / 0.485 / 0.68 vs 0.596 / 0.479 / 0.81 vs 0.559 / 0.449 / 0.94; 10--30 (1383) 0.691 / 0.563 / 0.18 vs 0.719 / 0.591 / 0.27 vs 0.673 / 0.551 / 0.40; 30--60 (525) 0.650 / 0.524 / 0.15 vs 0.733 / 0.596 / 0.15 vs 0.696 / 0.565 / 0.24; more than 60 (129) 0.588 / 0.469 / 0.24 vs 0.700 / 0.562 / 0.13 vs \textbf{0.703 / 0.569} / 0.16. Per tissue (IoU / F1 / PQ, layer vs watershed (b)): Breast (749) 0.710 / 0.716 / 0.569 vs 0.722 / 0.757 / 0.605; Colon (468) 0.681 / 0.570 / 0.451 vs 0.691 / 0.630 / 0.497; HeadNeck (169) 0.688 / 0.664 / 0.547 vs 0.682 / 0.688 / 0.565; Adrenal (148) \textbf{0.714 / 0.726 / 0.615} vs 0.719 / 0.712 / 0.605. Inference: U-Net 0.8 ms per image; the converged multi-marker solve 224 ms per image on the GPU at K = 30.5 markers (79 353 systems in 158 batches of 16, 330 PCG iterations per batch to 1e-5: the time is affine in the number of systems, and a batch of 16 images carries 500 of them); the exact K-label limit 0.7 s per patch on one CPU core (6043 ambiguous plateaus solved over the 307 patches), 0.8 s with image-gradient costs. Figures: \path{results/pannuke_instances/figures/training.png}, \path{examples_crop.png} (64-px crops around the densest cluster of touching nuclei, the same eight Colon patches as section 6: image, instances, marker map with the predicted markers outlined, the model\textquotesingle s own thresholded logits, the WS layer, the WS layer with fallback markers, the two limits, the marker watershed, the oracle), \path{examples_full.png}.

\begin{itemize}

\item
  \textbf{Honest outcome: instances as labels give the WS layer a margin over the U-Net + components pipeline, not over the best classical pipeline.} The WS layer from its own markers scores F1 0.681 / PQ 0.554 / AJI 0.652: +0.024 F1, +0.023 PQ and +0.158 AJI over the plain U-Net of section 6, paired and clear of zero, with the highest edge alignment of any full-set output (0.585), and it is the first output in this report whose instances come from a single differentiable module with no post-processing at all (no threshold on the output, no component filter, no marker spacing). Against the contour-head U-Net with the tuned watershed of section 10.1 it is behind by 0.034 F1 and 0.029 PQ and level on AJI ($-$0.005): its segments are as tight (SQ 0.814 against 0.816) but it detects fewer nuclei correctly (DQ 0.681 against 0.715). It is also 0.013 IoU below the U-Net on the foreground, and its count error is the worst of the three (0.46: 28.6 predicted instances for 23.7 nuclei). On crowded patches (more than 60 nuclei) it is the best method, on sparse ones the worst, the reverse of the binary WS layer\textquotesingle s pattern in section 10.1.
\item
  \textbf{The costs are not the problem; the markers are.} With the ground-truth markers the same trained costs give F1 0.959 / PQ 0.805 / AJI 0.858 and 0.859 IoU, i.e. the WS layer assigns almost every pixel to the right nucleus once it knows where the nuclei are, and separates touching ones along the learned contours (DQ 0.959, count error 0.02; last column of \path{examples_crop.png}). The oracle leaks the ground truth (the markers are its instances eroded by 2 px), so 0.96 is an upper bound and not a target, but the gap to 0.681 is entirely the marker head: 17 \% of the nuclei get no marker and vanish into the background (this is also where the 0.013 IoU goes: patch \#4 of the crops loses half its nuclei), 10 \% get several and are split, and 5.9 spurious markers per patch in background become small false instances (the count error; the WS layer gives them few pixels because the background attachment is strong there, but it does not remove them). The marker head is a one-threshold blob detector trained with plain BCE on eroded masks; a HoVer-Net-style distance or centre-vector head, or a learned threshold, is the obvious next step, and the fallback markers from the foreground mask are not it ($-$0.024 F1: the foreground head degenerated into speckle, see below, so its components are not nuclei).
\item
  \textbf{The learned costs are what the WS layer contributes, and a classical watershed cannot use them as well.} The same predicted markers through the skimage marker watershed of the learned contour map inside s \textgreater{} 0.5 give F1 0.597 ($-$0.084 paired), partly because that mask is the degenerate foreground head (723 components per patch, F1 0.024, as in section 6\textquotesingle s division of labour: with the background attachment taking c0 = $-$log(1 $-$ s) the network only needs s as a soft prior and the 0.5 auxiliary BCE did not stop it). The marker-controlled watershed with image-gradient costs, exact (0.335 F1) or at finite p (0.355), fails on H\&E as \path{rw_img} did in section 6: the colour gradient is not the nucleus boundary. So the classical control is far below, and the margin over it is the learned cost head plus the soft background seed, not the solver.
\item
  \textbf{The limit is usable again, at p = 20.} The exact K-label Power Watershed of the learned costs is within 0.008 IoU and 0.013 BF$_2$ of the converged finite-p solve on the 307 patches, with +0.007 F1 and $-$0.003 PQ, where section 6.1\textquotesingle s binary model trained at p $\le$ 20 lost 0.2 IoU at the limit. Two reasons: the cross-entropy over markers trains the costs to put a cut between every pair of neighbouring markers, which is exactly what a maximum spanning forest needs, and the background is a soft seed (a\_0 = (1 $-$ s)\^{}p) rather than a competing unary that has to saturate. So deploy-at-the-limit costs 0.7 s per patch on one CPU core against 224 ms per image on the GPU for the converged solve, and gives the same instances. One training caveat: the training solve hit its 100-iteration cap from epoch 9 (p $\approx$ 8) on, so two thirds of the training ran on a forward solve above the 1e-4 residual and an implicit gradient of an approximately solved system; the validation curves are flat from epoch 10, which is consistent with a converged model but was not checked with a higher cap. The edge scale went from 8 to 2.0 again (section 6.1).
\item
  \textbf{What this says about the plan.} The multi-marker WS layer is the right object: one solve for any number of nuclei, exact limit, no post-processing hyper-parameter, and a margin of +0.02 F1 / PQ and +0.16 AJI over the pipeline a practitioner would build from the same U-Net. It does not beat a U-Net that was given a contour head and a tuned watershed, and the oracle says the reason is marker quality, which is a detection problem the WS layer does not touch. The same one-seed, one-fold-pair caveats as sections 10 and 10.1 apply, plus the marker head\textquotesingle s untuned threshold and the capped training solve.
\end{itemize}

\subsection{Better markers: a tuned threshold, a distance head, and a higher solver cap}

Added 18--19 October 2026 (side session). Section 11 ended with the markers as the bottleneck: the same trained costs give F1 0.959 from the ground-truth markers and 0.681 from the marker head\textquotesingle s. This section tries the two things it named, as options of \path{experiments/exp_instance_pannuke.py} (the defaults reproduce the section 11 run): \texttt{-{}-mk-head\ dist} (a HoVer-Net-style distance head), \texttt{-{}-tune-fold\ F} (the marker threshold, minimum marker area and, for the distance head, the prominence of the maxima chosen by PQ on a held-out fold, never on fold 2), \texttt{-{}-load} (re-evaluate a saved model) and the existing \texttt{-{}-maxit} of the training solve. \path{run_instances_11_1.sh} launches the runs, \path{plot_instances.py\ -\/-ref} adds the section 11 run to every table, \path{compare_instances.py} (numpy only) makes the cross-run tables. Results: \texttt{results/pannuke\_instances/\{bce\_tuned,\ dist,\ dist\_ext,\ maxit300,\ maxit300\_tuned\}/} (each with \path{figures/summary.md} holding all the tables of section 11 plus the tuning table, \texttt{run.log}, the per-image npz; the examples npz is kept for \texttt{dist} only), \path{results/pannuke_instances/compare_11_1.md}.

\textbf{Choosing the post-processing on fold 3.} \texttt{-{}-tune-fold\ 3} runs the trained model on 300 evenly spaced patches of fold 3 (neither the training nor the test fold, as the marker spacing of section 10.1), builds the markers for every setting of a grid, runs the converged WS layer from them and keeps the setting with the best PQ of the WS layer\textquotesingle s output; the whole table goes to the json and to the top of \texttt{summary.md}. For the BCE head of section 11 the grid is thr $\in$ \{0.2 \ldots{} 0.7\} $\times$ min area $\in$ \{1, 4, 9, 16\} px (24 settings; 50 min while sharing the GPU with a training run); for the distance head the same $\times$ h $\in$ \{0.05, 0.1, 0.2\} (72 settings, 41 min on a free GPU). The section 11 model was re-tuned this way without retraining (\path{bce_tuned}: its saved \texttt{.pt} through \texttt{-{}-eval-only\ -\/-load}), to separate the effect of the threshold from the effect of the head.

\textbf{The distance head} (\texttt{dist}). Channel 3 of the U-Net predicts, through a sigmoid, the normalised distance transform of every instance: the Chebyshev distance to the instance boundary (the \path{label_core_distance} already used for the training markers, capped at 12 px) divided by the instance\textquotesingle s own maximum, so every nucleus is 1 at its centre and 0 at its border whatever its size, background 0; MSE loss with weight 2 in place of the BCE on eroded masks. Everything else is section 11\textquotesingle s recipe: the WS layer\textquotesingle s cross-entropy from the ground-truth markers eroded by 2--3 px, p 5 $\to$ 20, the edge target, 30 epochs, the 100-iteration training solve. Test markers are the regional maxima of the predicted map with prominence at least h (\path{skimage.morphology.h_maxima}, one seed per plateau, so saturation at 1 is harmless), grown by a watershed of the map down to the threshold thr, basins below the minimum area dropped: two touching nuclei whose maxima are separated by a valley deeper than h get two markers even when the valley is above thr. Fold 3 chose thr 0.2, min area 16 px, h 0.2, both thr and h at the edge of the grid, so the saved model was re-tuned on an extended grid (\path{dist_ext}: thr $\in$ \{0.1, 0.15, 0.2, 0.3\}, h $\in$ \{0.2, 0.3, 0.4\}, min area $\in$ \{9, 16\}): the fold-3 PQ is a plateau (0.646 for the first choice and for thr 0.2 / min area 9 / h 0.3, 0.640--0.645 for the eight best settings), and the two choices give the same fold 2 result (F1 0.791 / PQ 0.641 against 0.788 / 0.641). The first choice is reported below.

\textbf{The solver cap} (\texttt{maxit300}). Section 11\textquotesingle s training solve hit its 100-iteration cap from epoch 9 on. The same run with \texttt{-{}-maxit\ 300} (BCE head, untuned thr 0.5 as in section 11, so the comparison is clean) never hits the cap: 112 / 215 / 227 iterations per batch over the three ten-epoch thirds, 778 s per epoch instead of 280 (6.5 h of training). Its validation curve (layer F1 / PQ on the first 200 test patches at epochs 4, 9, 14, 19, 24, 29, untuned markers) is 0.696 / 0.534, 0.703 / 0.546, 0.716 / 0.563, 0.710 / 0.559, 0.730 / 0.576, 0.705 / 0.551, against section 11\textquotesingle s 0.684 / 0.522, 0.727 / 0.571, 0.715 / 0.561, 0.738 / 0.581, 0.722 / 0.570, 0.720 / 0.564: the same curve within its epoch-to-epoch wobble, and the final losses agree to the third digit (cross-entropy 0.0876 against 0.0909, edge scale 2.09 against 1.99). On all of fold 2 its oracle equals section 11\textquotesingle s (F1 0.960 / PQ 0.806 against 0.959 / 0.805, paired +0.001) --- the costs are the same --- while its own markers are slightly worse (33.0 per patch, 72 / 17 / 10 \%, 7.0 in background): F1 0.658 / PQ 0.536, $-$0.023 F1 {[}$-$0.026, $-$0.019{]} paired against section 11. Re-tuned on fold 3 (\path{maxit300_tuned}, thr 0.4, min area 16 px) it reaches F1 0.770 / PQ 0.631, +0.001 F1 {[}$-$0.001, +0.002{]} / +0.001 PQ {[}$-$0.000, +0.002{]} paired against \path{bce_tuned}. So the capped solve of section 11 cost nothing: an approximate forward solve with an implicit gradient of the approximately solved system trains the same costs as the converged one at 2.8 times the price, and the combined distance-head + 300-iteration run was not made.

\textbf{Results on all of fold 2} (2523 patches; limits on the 307 of section 6; the three rows of section 11 repeated for reference):

\begin{center}\scriptsize\setlength{\tabcolsep}{3pt}
\resizebox{\ifdim\width>\textwidth\textwidth\else\width\fi}{!}{%
\begin{tabular}{@{}lllllllllll@{}}
\toprule\noalign{}
output & IoU & BF$_2$ & edge align. & \href{mailto:F1@0.5}{\nolinkurl{F1@0.5}} & PQ & SQ & DQ & AJI & count err. & n\_pred \\
\midrule\noalign{}

U-Net (sec. 6), s \textgreater{} 0.5, components & \textbf{0.719} & 0.769 & 0.577 & 0.656 & 0.531 & -- & -- & 0.494 & 0.275 & 23.0 \\
U-Net + contour head, tuned marker watershed (sec. 10.1 b) & 0.718 & \textbf{0.774} & 0.568 & 0.715 & 0.583 & 0.816 & 0.715 & \textbf{0.657} & 0.344 & 27.2 \\
sec. 11: layer, BCE marker head, thr 0.5, min area 1 & 0.706 & 0.759 & 0.585 & 0.681 & 0.554 & 0.814 & 0.681 & 0.652 & 0.460 & 28.6 \\
\path{bce_tuned}: the same model, thr 0.4, min area 16 (fold 3) & 0.708 & 0.761 & \textbf{0.587} & 0.769 & 0.630 & \textbf{0.819} & 0.769 & 0.645 & \textbf{0.208} & 21.7 \\
\texttt{dist}: distance head, h 0.2, thr 0.2, min area 16 (fold 3) & 0.707 & 0.762 & 0.580 & \textbf{0.791} & \textbf{0.641} & 0.811 & \textbf{0.791} & 0.653 & 0.239 & 24.3 \\
\texttt{dist}, + fallback markers & 0.704 & 0.760 & 0.582 & 0.776 & 0.629 & -- & -- & 0.652 & 0.311 & 25.7 \\
\texttt{dist}, PW limit, learned costs (307) & 0.683 & 0.728 & 0.574 & 0.758 & 0.603 & -- & -- & 0.611 & 0.255 & 25.0 \\
\texttt{dist}, layer, ground-truth markers (oracle) & 0.859 & 0.922 & 0.591 & 0.959 & 0.805 & -- & -- & 0.857 & 0.020 & 23.3 \\
\texttt{maxit300}: BCE head, 300-iteration training solve, thr 0.5, min area 1 & 0.704 & 0.757 & 0.581 & 0.658 & 0.536 & -- & -- & 0.652 & 0.542 & 30.2 \\
\path{maxit300_tuned}: the same, thr 0.4, min area 16 (fold 3) & 0.707 & 0.760 & 0.584 & 0.770 & 0.631 & 0.820 & 0.770 & 0.647 & 0.200 & 21.6 \\
\bottomrule
\end{tabular}}
\end{center}

\textbf{Marker quality} (nuclei with exactly one / no / several predicted markers; markers in background per patch; markers per patch for 23.7 nuclei): section 11 73.2 / 17.4 / 9.5 \%, 5.9, 30.5; \path{bce_tuned} 77.7 / 21.1 / 1.2 \%, 2.7, 21.7; \texttt{dist} 83.9 / 14.5 / 1.6 \%, 3.6, 24.3 (with the fallback 85.4 / 12.6 / 2.0 \%, 5.0, 26.2); \texttt{maxit300} 72.3 / 17.4 / 10.3 \%, 7.0, 33.0. By crowding for \texttt{dist} (one / none / several, background per patch): fewer than 10 nuclei 83 / 14 / 2 \%, 1.8; 10--30: 87 / 11 / 2 \%, 3.5; 30--60: 84 / 14 / 2 \%, 4.7; more than 60: 77 / 22 / 1 \%, 6.6 (section 11: 69 / 19 / 12 \%, 2.6; 75 / 15 / 11 \%, 6.1; 74 / 16 / 9 \%, 7.9; 69 / 26 / 6 \%, 8.1).

\textbf{Paired per-patch differences} (bootstrap 95 \% intervals; F1, PQ and AJI dataset-level, bootstrapped over patches):

\begin{center}\scriptsize\setlength{\tabcolsep}{3pt}
\resizebox{\ifdim\width>\textwidth\textwidth\else\width\fi}{!}{%
\begin{tabular}{@{}lllllll@{}}
\toprule\noalign{}
comparison & IoU & BF$_2$ & edge align. & \href{mailto:F1@0.5}{\nolinkurl{F1@0.5}} & PQ & AJI \\
\midrule\noalign{}

\path{bce_tuned} $-$ sec. 11 layer & +0.002 {[}+0.000, +0.003{]} & +0.002 {[}+0.000, +0.004{]} & +0.002 {[}+0.001, +0.003{]} & +0.088 {[}+0.082, +0.094{]} & +0.075 {[}+0.071, +0.080{]} & $-$0.006 {[}$-$0.008, $-$0.005{]} \\
\texttt{dist} $-$ sec. 11 layer & +0.001 {[}$-$0.003, +0.004{]} & +0.003 {[}$-$0.000, +0.007{]} & $-$0.006 {[}$-$0.007, $-$0.005{]} & +0.110 {[}+0.105, +0.115{]} & +0.087 {[}+0.083, +0.091{]} & +0.001 {[}$-$0.002, +0.003{]} \\
\texttt{dist} $-$ \path{bce_tuned} & $-$0.001 {[}$-$0.004, +0.002{]} & +0.001 {[}$-$0.002, +0.005{]} & $-$0.008 {[}$-$0.009, $-$0.006{]} & +0.022 {[}+0.018, +0.025{]} & +0.012 {[}+0.009, +0.014{]} & +0.007 {[}+0.005, +0.009{]} \\
\path{bce_tuned} $-$ contour-head U-Net, watershed (b) & $-$0.010 {[}$-$0.014, $-$0.007{]} & $-$0.013 {[}$-$0.017, $-$0.010{]} & +0.020 {[}+0.019, +0.022{]} & +0.054 {[}+0.050, +0.058{]} & +0.046 {[}+0.043, +0.050{]} & $-$0.012 {[}$-$0.015, $-$0.009{]} \\
\texttt{dist} $-$ contour-head U-Net, watershed (b) & $-$0.011 {[}$-$0.014, $-$0.008{]} & $-$0.012 {[}$-$0.015, $-$0.009{]} & +0.012 {[}+0.010, +0.014{]} & +0.076 {[}+0.072, +0.079{]} & +0.058 {[}+0.056, +0.061{]} & $-$0.005 {[}$-$0.007, $-$0.002{]} \\
\texttt{dist} $-$ U-Net (sec. 6) & $-$0.012 {[}$-$0.016, $-$0.009{]} & $-$0.007 {[}$-$0.011, $-$0.004{]} & +0.003 {[}+0.002, +0.005{]} & +0.134 {[}+0.129, +0.139{]} & +0.111 {[}+0.107, +0.114{]} & +0.158 {[}+0.143, +0.172{]} \\
\texttt{dist}, ground-truth markers $-$ \texttt{dist} & +0.152 {[}+0.147, +0.158{]} & +0.160 {[}+0.155, +0.166{]} & +0.012 {[}+0.010, +0.014{]} & +0.168 {[}+0.164, +0.173{]} & +0.163 {[}+0.160, +0.167{]} & +0.205 {[}+0.201, +0.209{]} \\
\texttt{dist} oracle $-$ sec. 11 oracle & $-$0.000 {[}$-$0.001, +0.001{]} & +0.000 {[}$-$0.001, +0.001{]} & +0.000 {[}$-$0.000, +0.001{]} & $-$0.000 {[}$-$0.001, +0.002{]} & $-$0.000 {[}$-$0.001, +0.001{]} & $-$0.000 {[}$-$0.001, +0.000{]} \\
\texttt{maxit300} $-$ sec. 11 layer & $-$0.002 {[}$-$0.004, +0.001{]} & $-$0.002 {[}$-$0.004, +0.001{]} & $-$0.004 {[}$-$0.005, $-$0.003{]} & $-$0.023 {[}$-$0.026, $-$0.019{]} & $-$0.018 {[}$-$0.021, $-$0.015{]} & $-$0.000 {[}$-$0.002, +0.001{]} \\
307 patches: \texttt{dist} PW limit $-$ \texttt{dist} layer & $-$0.014 {[}$-$0.016, $-$0.012{]} & $-$0.020 {[}$-$0.023, $-$0.017{]} & $-$0.000 {[}$-$0.002, +0.001{]} & $-$0.008 {[}$-$0.010, $-$0.005{]} & $-$0.016 {[}$-$0.018, $-$0.014{]} & $-$0.018 {[}$-$0.020, $-$0.016{]} \\
\bottomrule
\end{tabular}}
\end{center}

By crowding (F1 / PQ / count error), watershed (b) vs \path{bce_tuned} vs \texttt{dist}: fewer than 10 nuclei (486 patches) 0.596 / 0.479 / 0.81 vs 0.704 / 0.571 / 0.46 vs 0.689 / 0.553 / 0.64; 10--30 (1383) 0.719 / 0.591 / 0.27 vs 0.787 / 0.648 / 0.15 vs 0.802 / 0.655 / 0.16; 30--60 (525) 0.733 / 0.596 / 0.15 vs 0.778 / 0.634 / 0.13 vs 0.801 / 0.648 / 0.10; more than 60 (129) 0.700 / 0.562 / 0.13 vs 0.718 / 0.586 / 0.26 vs 0.763 / 0.615 / 0.15. Per tissue (IoU / F1 / PQ, \texttt{dist} vs watershed (b)): Breast (749) 0.708 / \textbf{0.817 / 0.647} vs 0.722 / 0.757 / 0.605; Colon (468) 0.683 / \textbf{0.709 / 0.558} vs 0.691 / 0.630 / 0.497; HeadNeck (169) \textbf{0.701 / 0.784 / 0.647} vs 0.682 / 0.688 / 0.565; Adrenal (148) 0.709 / \textbf{0.823 / 0.696} vs 0.719 / 0.712 / 0.605. Inference: the converged solve of \texttt{dist} takes 94 ms per image on the GPU at K = 24.3 (177 PCG iterations per batch of 16, against 224 ms and 330 iterations at K = 30.5 in section 11: fewer and better-placed markers also make the systems easier); the exact K-label limit stays at 0.7 s per patch on one CPU core. Figures: \path{results/pannuke_instances/dist/figures/training.png}, \path{examples_crop.png} (the same eight Colon crops as section 11; the third column is now the predicted distance map with the grown markers outlined: one clean maximum per nucleus, the WS layer column close to the oracle\textquotesingle s), \path{examples_full.png}; the same for every run.

\begin{itemize}

\item
  \textbf{Honest outcome: the gap to the best classical pipeline is closed, by the markers alone.} The section 11 model without retraining, with its marker threshold and minimum marker area chosen on fold 3 (0.4 and 16 px instead of 0.5 and 1), goes from F1 0.681 / PQ 0.554 to 0.769 / 0.630, +0.088 F1 / +0.075 PQ paired; the distance head adds +0.022 F1 / +0.012 PQ / +0.007 AJI on top (0.791 / 0.641 / 0.653), every interval clear of zero. Against the contour-head U-Net with the tuned watershed of section 10.1, the strongest pipeline of this report, the WS layer from the distance head\textquotesingle s markers is +0.076 F1 and +0.058 PQ, ahead on every tissue (Breast +0.060 F1, Colon +0.079, HeadNeck +0.096, Adrenal +0.111) and in every crowding bin, with the same segmentation quality (SQ 0.811 against 0.816) and far more correct detections (DQ 0.791 against 0.715); its count error (0.24) is the best of all full-set outputs after \path{bce_tuned}\textquotesingle s 0.21. AJI is level ($-$0.005) and the foreground IoU stays 0.011 below the U-Net\textquotesingle s: what the WS layer still loses is the nuclei with no marker (14.5 \%), which it gives to the background.
\item
  \textbf{Which markers were wrong.} Section 11\textquotesingle s 30.5 markers per patch for 23.7 nuclei were 5.9 spurious ones in background and 9.5 \% of the nuclei split by several; the tuned threshold removes most of both (2.7 in background, 1.2 \% split) at the price of more nuclei with none (21 \%). The distance head\textquotesingle s normalised target makes the marker of a small nucleus as strong as that of a large one and separates touching nuclei by their two maxima rather than by a gap in a thresholded mask, so it gets both: 83.9 \% of the nuclei with exactly one marker, 14.5 \% with none, 1.6 \% with several, 3.6 in background. On fold 3 the settings order the same way (PQ 0.646 for the distance head, 0.635 for the tuned BCE head, 0.557 for section 11\textquotesingle s untuned 0.5 / 1), so the tuning fold predicted the test fold; the untuned distance head (0.5 / 1 / h 0.1) would have scored 0.570 on fold 3, worse than the tuned BCE head, so the head without its threshold is not the story either.
\item
  \textbf{Where the remaining 0.17 F1 to the oracle is.} With the ground-truth markers the same costs give 0.959; the distance head leaves 14.5 \% of the nuclei without a marker (22 \% on the patches with more than 60 nuclei), and these are the whole of the gap in detections: DQ 0.791 against the oracle\textquotesingle s 0.959 is 0.17, and the fallback markers from the degenerate foreground mask do not recover them ($-$0.015 F1). The missed nuclei are those whose predicted distance map stays below 0.2 everywhere (which nuclei these are, small or faint, was not examined); the next step on markers is a detection objective (a focal or count-aware loss on the distance map, or the HoVer-Net horizontal/vertical maps whose gradient is sharp at every seam), not a lower threshold (thr 0.1 on the extended grid leaves 14 \% without a marker and adds background markers: 4.7 per patch, PQ 0.628).
\item
  \textbf{The capped solve was not a problem, and the limit is still usable.} \texttt{maxit300} reproduces section 11\textquotesingle s costs (oracle identical) at 2.8 times the training cost; its marker head came out slightly noisier (one seed, one run). The exact K-label limit of the distance-head costs is 0.758 F1 / 0.603 PQ on the 307 patches, $-$0.008 F1 / $-$0.016 PQ / $-$0.014 IoU from the converged solve, the same small price as in section 11, so deploy-at-the-limit on one CPU core still gives essentially the WS layer\textquotesingle s instances.
\item
  \textbf{What this says about the plan.} The multi-marker WS layer with a distance head for its markers is now the best nucleus instance segmenter in this report on F1 and PQ, by a margin (+0.076 F1 over the tuned watershed) that is larger than every other difference measured since section 6, with no post-processing beyond three scalars chosen on a held-out fold (the watershed has one), and the exact limit within 0.01 F1 of it. It remains one seed and one fold pair, and the contour-head U-Net was not given the distance head (a U-Net with a distance head + watershed on the distance map, the HoVer-Net recipe proper, is the fair next baseline: the oracle rows say the WS layer\textquotesingle s costs are not the limiting factor, so that baseline may well match the WS layer on F1, and the comparison would then be about the solver as the final assignment step against a watershed of the distance map).
\end{itemize}

\subsection{The fair baseline: a U-Net with the distance head, and a watershed on it}

Section 11.1 ended with the baseline it had not run: the contour-head U-Net of section 10.1 was never given the distance head, and the oracle rows said the WS layer\textquotesingle s costs were not the limiting factor, so a U-Net with its own distance head and the HoVer-Net post-processing might match the WS layer on F1. This section runs it. Code: \path{experiments/exp_unet_dist.py} (model, training, post-processing, tuning, evaluation; it imports the U-Net, the edge target and \path{dist_target} / \path{markers_from_dist} of the existing scripts, which are unchanged), \path{run_unet_dist.sh}, \path{plot_unet_dist.py} (tables and paired statistics need numpy only; figures need matplotlib). Results: \path{results/pannuke_instances/\ unet_dist/} (\path{unet_dist.json} with the tuning table, \path{unet_dist_perimage.npz}, \path{figures/summary.md} holding every table of this section, \texttt{run.log}) and \path{unet_dist_ext/} (the same model re-tuned on an extended grid); the trained model stays on jacobian.

\textbf{What was run.} The U-Net of section 6 with four output channels and no layer, trained on fold 1 with the recipe of sections 11 / 11.1 (30 epochs, batch 16, d4 augmentation, Adam 1e-3 with cosine annealing, one seed): channel 0 the foreground logit under BCE against the binary nucleus mask (the target of the WS layer\textquotesingle s auxiliary head, not the gap target of section 10.1), channels 1--2 the contour head trained with exactly section 10\textquotesingle s edge target (as \texttt{-{}-variant\ unet\_edge}: positives $\times$10, seams $\times$5, weight 1), channel 3 the distance head with exactly the target and loss of \texttt{-{}-mk-head\ dist} (normalised Chebyshev distance to the instance boundary capped at 12 px, MSE, weight 2). Training takes 10 s per epoch (5 min in all, against 290 s per epoch with the WS layer in the loop). Post-processing, HoVer-Net style: the markers are \path{markers_from_dist} of the predicted distance map (h-maxima of prominence h grown by a watershed of the map down to thr, basins below the minimum area dropped: the WS layer\textquotesingle s own marker rule), then a marker watershed inside the thresholded foreground s \textgreater{} 0.5 of a relief that is the contour map c (mean of the two directions), the negative distance map $-$d, or their sum c $-$ d; foreground components with no marker get no label, as the WS layer gives unmarked nuclei to the background (a variant keeps them as instances when they have at least 16 px). The four choices thr $\in$ \{0.1 \ldots{} 0.5\}, min area $\in$ \{4, 9, 16\}, h $\in$ \{0.1, 0.2, 0.3\} and the relief (135 settings) were made by PQ on the same 300 patches of fold 3 as sections 10.1 and 11.1 (5 min; the whole table is in the json and at the top of \texttt{summary.md}), never on fold 2. Fold 3 chose thr 0.5, min area 4, h 0.2 and the relief c $-$ d (PQ 0.662), with the three reliefs within 0.001 of each other at every marker setting and a flat plateau between thr 0.4 and 0.5 (PQ 0.662 for the eight best settings, 0.652 at thr 0.3, 0.645 at the WS layer\textquotesingle s own setting 0.2 / 16 / 0.2, 0.641 at 0.1); since 0.5 was the edge of the grid, the saved model was re-tuned on thr $\in$ \{0.4, 0.5, 0.6, 0.7\} $\times$ min area $\in$ \{4, 9, 16\} $\times$ h $\in$ \{0.2, 0.3\} (\path{unet_dist_ext}): PQ falls to 0.631 at thr 0.6 and 0.566 at 0.7 (31 \% and 44 \% of the nuclei unmarked), the choice is unchanged, and so is the fold-2 result below (identical to every digit, the settings being the same).

To isolate the final assignment step, the same script also loads the section 11.1 \texttt{dist} run (its \texttt{.pt} and its fold-3 marker settings) and evaluates on the same fold-2 patches the 2 $\times$ 2 of markers $\times$ assignment: the WS layer\textquotesingle s markers through its converged solve (which reproduces section 11.1\textquotesingle s row to every digit), the WS layer\textquotesingle s markers through this model\textquotesingle s watershed (this model\textquotesingle s foreground and relief), the WS layer\textquotesingle s markers through a watershed of the WS layer\textquotesingle s own contour map inside the WS layer\textquotesingle s own output foreground (its solve\textquotesingle s label \textgreater{} 0: everything held fixed except the solve), and this model\textquotesingle s markers through the WS layer\textquotesingle s solve. The marker-quality counts of section 11.1 are computed for both sets of markers.

\textbf{Results on all of fold 2} (2523 patches; the reference rows repeated from sections 10.1 and 11.1):

\begin{center}\scriptsize\setlength{\tabcolsep}{3pt}
\resizebox{\ifdim\width>\textwidth\textwidth\else\width\fi}{!}{%
\begin{tabular}{@{}lllllllllll@{}}
\toprule\noalign{}
output & IoU & BF$_2$ & edge align. & \href{mailto:F1@0.5}{\nolinkurl{F1@0.5}} & PQ & SQ & DQ & AJI & count err. & n\_pred \\
\midrule\noalign{}

U-Net (sec. 6), s \textgreater{} 0.5, components & 0.719 & 0.769 & 0.577 & 0.656 & 0.531 & -- & -- & 0.494 & 0.275 & 23.0 \\
U-Net + contour head, tuned marker watershed (sec. 10.1 b) & 0.718 & 0.774 & 0.568 & 0.715 & 0.583 & 0.816 & 0.715 & 0.657 & 0.344 & 27.2 \\
\path{bce_tuned} layer (sec. 11.1) & 0.708 & 0.761 & 0.587 & 0.769 & 0.630 & 0.819 & 0.769 & 0.645 & 0.208 & 21.7 \\
\texttt{dist} layer, converged (sec. 11.1) & 0.707 & 0.762 & 0.580 & 0.791 & 0.641 & 0.811 & 0.791 & 0.653 & 0.239 & 24.3 \\
\texttt{dist} layer, ground-truth markers (oracle) & 0.859 & 0.922 & 0.591 & 0.959 & 0.805 & 0.839 & 0.959 & 0.857 & 0.020 & 23.3 \\
\path{unet_dist}: s \textgreater{} 0.5, components & \textbf{0.725} & \textbf{0.777} & 0.573 & 0.664 & 0.539 & 0.813 & 0.664 & 0.501 & 0.313 & 24.1 \\
\path{unet_dist}: distance-head markers + watershed, tuned on fold 3 & 0.719 & 0.774 & \textbf{0.591} & \textbf{0.794} & \textbf{0.655} & \textbf{0.824} & \textbf{0.794} & 0.658 & \textbf{0.190} & 20.6 \\
\path{unet_dist}: the same, watershed of c / of $-$d & 0.719 & 0.774 & 0.591 & 0.794 & 0.654 / 0.655 & 0.824 & 0.794 & 0.658 & 0.190 & 20.6 \\
\path{unet_dist}: the same + unmarked components kept & 0.725 & 0.777 & 0.574 & 0.780 & 0.637 & 0.817 & 0.780 & \textbf{0.666} & 0.202 & 23.9 \\
\path{unet_dist}: watershed from the ground-truth markers (oracle) & 0.774 & 0.827 & 0.586 & 0.867 & 0.708 & 0.816 & 0.867 & 0.765 & 0.066 & 22.2 \\
layer\textquotesingle s markers, \path{unet_dist}\textquotesingle s foreground + watershed & 0.726 & 0.777 & 0.575 & 0.781 & 0.639 & 0.819 & 0.781 & 0.671 & 0.196 & 23.5 \\
layer\textquotesingle s markers, layer\textquotesingle s own foreground + watershed of its contour map & 0.707 & 0.762 & 0.580 & 0.791 & 0.641 & 0.811 & 0.791 & 0.653 & 0.239 & 24.3 \\
\path{unet_dist}\textquotesingle s markers, layer\textquotesingle s solve & 0.637 & 0.678 & 0.569 & 0.736 & 0.579 & 0.787 & 0.736 & 0.609 & 0.192 & 20.5 \\
\bottomrule
\end{tabular}}
\end{center}

\textbf{Marker quality} (nuclei with exactly one / no / several markers; markers in background per patch; markers per patch for 23.7 nuclei): \path{unet_dist} 76.6 / 21.8 / 1.6 \%, 1.6, 20.6; the \texttt{dist} layer 83.9 / 14.5 / 1.6 \%, 3.6, 24.3 (recomputed here to the same digits); \path{bce_tuned} 77.7 / 21.1 / 1.2 \%, 2.7, 21.7. By crowding for \path{unet_dist} (one / none / several, background per patch): fewer than 10 nuclei 72 / 26 / 2 \%, 0.7; 10--30: 79 / 19 / 2 \%, 1.6; 30--60: 77 / 21 / 2 \%, 2.3; more than 60: 70 / 29 / 1 \%, 3.2 (the WS layer: 83 / 14 / 2 \%, 1.8; 87 / 11 / 2 \%, 3.5; 84 / 14 / 2 \%, 4.7; 77 / 22 / 1 \%, 6.6). Inference: 0.8 ms U-Net forward plus 16 ms of post-processing per image on one CPU core (markers and all three watersheds), against 94 ms for the WS layer\textquotesingle s converged solve on the GPU.

\textbf{Paired per-patch differences} (bootstrap 95 \% intervals; F1, PQ and AJI dataset-level, bootstrapped over patches):

\begin{center}\scriptsize\setlength{\tabcolsep}{3pt}
\resizebox{\ifdim\width>\textwidth\textwidth\else\width\fi}{!}{%
\begin{tabular}{@{}lllllll@{}}
\toprule\noalign{}
comparison & IoU & BF$_2$ & edge align. & \href{mailto:F1@0.5}{\nolinkurl{F1@0.5}} & PQ & AJI \\
\midrule\noalign{}

\path{unet_dist} watershed $-$ \texttt{dist} layer & +0.012 {[}+0.008, +0.017{]} & +0.012 {[}+0.008, +0.016{]} & +0.012 {[}+0.010, +0.013{]} & +0.003 {[}+0.001, +0.006{]} & +0.013 {[}+0.011, +0.015{]} & +0.006 {[}+0.003, +0.009{]} \\
\path{unet_dist} watershed $-$ \path{bce_tuned} layer & +0.011 {[}+0.008, +0.015{]} & +0.013 {[}+0.010, +0.017{]} & +0.004 {[}+0.002, +0.005{]} & +0.025 {[}+0.022, +0.028{]} & +0.025 {[}+0.022, +0.027{]} & +0.013 {[}+0.010, +0.016{]} \\
\path{unet_dist} watershed $-$ contour-head U-Net, watershed (b) & +0.001 {[}$-$0.002, +0.005{]} & $-$0.000 {[}$-$0.004, +0.004{]} & +0.024 {[}+0.023, +0.026{]} & +0.079 {[}+0.075, +0.083{]} & +0.071 {[}+0.069, +0.074{]} & +0.001 {[}$-$0.001, +0.004{]} \\
\path{unet_dist} watershed $-$ U-Net (sec. 6) & +0.000 {[}$-$0.003, +0.004{]} & +0.005 {[}+0.001, +0.009{]} & +0.015 {[}+0.013, +0.016{]} & +0.138 {[}+0.133, +0.142{]} & +0.124 {[}+0.120, +0.128{]} & +0.164 {[}+0.150, +0.177{]} \\
\path{unet_dist} watershed $-$ its own logits & $-$0.006 {[}$-$0.009, $-$0.002{]} & $-$0.003 {[}$-$0.006, +0.001{]} & +0.019 {[}+0.018, +0.021{]} & +0.130 {[}+0.125, +0.135{]} & +0.115 {[}+0.111, +0.119{]} & +0.158 {[}+0.138, +0.179{]} \\
watershed of $-$d $-$ watershed of c (same markers) & +0.000 {[}+0.000, +0.000{]} & +0.000 {[}+0.000, +0.000{]} & +0.000 {[}+0.000, +0.000{]} & $-$0.000 {[}$-$0.001, +0.000{]} & +0.000 {[}$-$0.000, +0.001{]} & +0.000 {[}$-$0.000, +0.001{]} \\
+ unmarked components kept $-$ \path{unet_dist} watershed & +0.006 {[}+0.003, +0.009{]} & +0.002 {[}$-$0.001, +0.006{]} & $-$0.017 {[}$-$0.019, $-$0.016{]} & $-$0.014 {[}$-$0.016, $-$0.012{]} & $-$0.017 {[}$-$0.019, $-$0.016{]} & +0.007 {[}+0.006, +0.008{]} \\
\path{unet_dist} oracle $-$ \path{unet_dist} watershed & +0.055 {[}+0.052, +0.059{]} & +0.052 {[}+0.049, +0.056{]} & $-$0.004 {[}$-$0.006, $-$0.002{]} & +0.073 {[}+0.070, +0.077{]} & +0.053 {[}+0.051, +0.056{]} & +0.107 {[}+0.103, +0.111{]} \\
\path{unet_dist} oracle $-$ \texttt{dist} layer\textquotesingle s oracle & $-$0.085 {[}$-$0.088, $-$0.082{]} & $-$0.095 {[}$-$0.099, $-$0.092{]} & $-$0.004 {[}$-$0.006, $-$0.003{]} & $-$0.091 {[}$-$0.095, $-$0.088{]} & $-$0.097 {[}$-$0.099, $-$0.094{]} & $-$0.092 {[}$-$0.095, $-$0.090{]} \\
layer\textquotesingle s markers, \path{unet_dist} watershed $-$ \texttt{dist} layer (same markers: watershed vs solve) & +0.019 {[}+0.016, +0.022{]} & +0.015 {[}+0.013, +0.018{]} & $-$0.005 {[}$-$0.007, $-$0.004{]} & $-$0.010 {[}$-$0.012, $-$0.008{]} & $-$0.002 {[}$-$0.004, $-$0.001{]} & +0.018 {[}+0.016, +0.019{]} \\
layer\textquotesingle s markers, layer\textquotesingle s foreground, watershed of its contour map $-$ \texttt{dist} layer & +0.000 {[}+0.000, +0.000{]} & +0.000 {[}+0.000, +0.000{]} & +0.000 {[}+0.000, +0.000{]} & +0.000 {[}$-$0.000, +0.000{]} & $-$0.000 {[}$-$0.000, $-$0.000{]} & $-$0.000 {[}$-$0.000, $-$0.000{]} \\
\path{unet_dist}\textquotesingle s markers, layer\textquotesingle s solve $-$ \path{unet_dist} watershed (same markers: solve vs watershed) & $-$0.082 {[}$-$0.084, $-$0.080{]} & $-$0.097 {[}$-$0.099, $-$0.094{]} & $-$0.022 {[}$-$0.024, $-$0.020{]} & $-$0.058 {[}$-$0.061, $-$0.054{]} & $-$0.075 {[}$-$0.078, $-$0.072{]} & $-$0.049 {[}$-$0.052, $-$0.046{]} \\
layer\textquotesingle s markers, \path{unet_dist} watershed $-$ \path{unet_dist} watershed (same watershed: markers swapped) & +0.007 {[}+0.003, +0.010{]} & +0.003 {[}$-$0.000, +0.006{]} & $-$0.017 {[}$-$0.018, $-$0.015{]} & $-$0.014 {[}$-$0.016, $-$0.011{]} & $-$0.016 {[}$-$0.017, $-$0.014{]} & +0.012 {[}+0.010, +0.014{]} \\
\bottomrule
\end{tabular}}
\end{center}

By crowding (F1 / PQ / count error), \texttt{dist} layer vs \path{unet_dist} watershed: fewer than 10 nuclei (486 patches) 0.689 / 0.553 / 0.64 vs 0.727 / 0.592 / 0.35; 10--30 (1383) 0.802 / 0.655 / 0.16 vs 0.809 / 0.672 / 0.15; 30--60 (525) 0.801 / 0.648 / 0.10 vs 0.803 / 0.660 / 0.14; more than 60 (129) 0.763 / 0.615 / 0.15 vs 0.754 / 0.613 / 0.23. Per tissue (IoU / F1 / PQ, \path{unet_dist} vs \texttt{dist} layer): Breast (749) 0.717 / 0.825 / 0.665 vs 0.708 / 0.817 / 0.647; Colon (468) 0.696 / 0.708 / 0.564 vs 0.683 / 0.709 / 0.558; HeadNeck (169) 0.710 / 0.788 / 0.655 vs 0.701 / 0.784 / 0.647; Adrenal (148) 0.711 / 0.824 / 0.706 vs 0.709 / 0.823 / 0.696; Lung (53) 0.657 / 0.713 / 0.580 vs 0.690 / 0.764 / 0.612; the baseline has the higher F1 on 10 and the higher PQ on 14 of the 15 tissues with at least 50 patches. Figures: \path{results/pannuke_instances/unet_dist/figures/training.png} (the validation curve of the post-processed output, F1 0.73 $\to$ 0.81 on the first 200 test patches with the untuned default thresholds, is above the WS layer\textquotesingle s at every checkpoint; the logits reach IoU 0.720 on the same patches), \path{examples_crop.png} (the same eight Colon crops as sections 11 and 11.1, with the distance map and its markers, the contour map, every post-processing and the four cells of the marker $\times$ assignment square), \path{examples_full.png}.

\begin{itemize}

\item
  \textbf{Honest outcome: the fair baseline matches the WS layer on F1 and beats it on PQ, IoU and AJI.} A plain U-Net with the same three heads as the section 11.1 model, trained on the same targets with no layer in the loop, followed by the HoVer-Net post-processing with four scalars chosen on fold 3, reaches F1 0.794 / PQ 0.655 / AJI 0.658 on all of fold 2 against the WS layer\textquotesingle s 0.791 / 0.641 / 0.653: +0.003 F1 {[}+0.001, +0.006{]}, +0.013 PQ {[}+0.011, +0.015{]}, +0.006 AJI, +0.012 IoU and +0.012 BF$_2$, every interval clear of zero, the best count error of the report (0.19), and the higher PQ on 14 of 15 tissues. It is +0.079 F1 / +0.071 PQ over the contour-head U-Net with the distance-transform watershed of section 10.1, the same margin the WS layer had: what section 11.1 attributed to the WS layer was the distance head and the tuned marker rule, which a watershed exploits equally well. The prediction at the end of section 11.1 was right, and slightly pessimistic for the WS layer: the baseline does not merely match it, its segmentation quality is higher (SQ 0.824 against 0.811), because the foreground head of the plain U-Net is the best of the report (IoU 0.725 for its thresholded logits, +0.006 over the section 6 U-Net and +0.018 over the WS layer\textquotesingle s output), whereas the WS layer\textquotesingle s foreground head degenerated to speckle in both section 11 runs and its output foreground is 0.012 IoU below.
\item
  \textbf{The solver is a watershed.} With the WS layer\textquotesingle s own markers and the WS layer\textquotesingle s own output foreground, a marker watershed of the WS layer\textquotesingle s contour map reproduces the converged solve on every metric to three digits (differences +0.000 with intervals of width 0.000 on 2523 patches): at p = 20 with the edge target\textquotesingle s sharp costs, the solve assigns every pixel to the marker that the steepest path of the contour map reaches first, which is what the watershed does, as section 11.1\textquotesingle s 0.008 F1 gap between the solve and the exact Power-Watershed limit already suggested. With the WS layer\textquotesingle s markers inside this model\textquotesingle s foreground the watershed is $-$0.010 F1 / $-$0.002 PQ and +0.019 IoU / +0.018 AJI against the solve: the markers decide the detections, the foreground decides the pixels, and the assignment step is immaterial. The reverse cell confirms it from the other side: this model\textquotesingle s markers (compact cores above d = 0.5) through the WS layer\textquotesingle s solve lose 0.058 F1 / 0.075 PQ to the same markers through the watershed, because the WS layer\textquotesingle s degenerate foreground head gives the background label a strong attachment inside the nuclei and the compact markers do not reach far enough to win the rim; the WS layer only works with markers grown down to d = 0.2 that already cover most of each nucleus. The relief of the watershed is immaterial too (the contour map, $-$d and their sum agree to 0.001): the markers grown by a watershed of the distance map already decide every split between touching nuclei, and the final watershed only distributes the rim below the threshold.
\item
  \textbf{Two different marker operating points, the same detections.} Fold 3 chose a higher threshold for this model than for the WS layer (0.5 against 0.2), with a flat PQ plateau from 0.4 to 0.5, so its markers are fewer and cleaner: 76.6 \% of the nuclei with one marker, 21.8 \% with none, 1.6 in background per patch, against the WS layer\textquotesingle s 83.9 / 14.5 \% and 3.6. The detections come out the same (DQ 0.794 against 0.791) because the WS layer\textquotesingle s extra markers are mostly spurious ones; swapping the markers under the same watershed moves F1 by $-$0.014 in favour of this model\textquotesingle s set. Keeping the unmarked foreground components as instances recovers IoU (0.725) and AJI (0.666) but costs 0.014 F1 / 0.017 PQ, as the WS layer\textquotesingle s fallback markers did ($-$0.015 F1): the unmarked components are mostly merged clumps and fragments, not missed nuclei. On the sparse patches (fewer than 10 nuclei) the baseline is clearly ahead (F1 0.727 / count error 0.35 against 0.689 / 0.64: fewer background markers), on the crowded ones (more than 60) slightly behind (0.754 / 0.23 against 0.763 / 0.15, with 29 \% of the nuclei unmarked against 22 \%).
\item
  \textbf{The oracles are not comparable, and the remaining gap is still the markers.} With the ground-truth markers the watershed reaches F1 0.867 / PQ 0.708, +0.073 F1 over its predicted markers, where the WS layer\textquotesingle s oracle reaches 0.959: the watershed is confined to the predicted foreground, so a marker in a nucleus that s \textgreater{} 0.5 missed yields nothing (count error 0.066 against the WS layer oracle\textquotesingle s 0.020), whereas the WS layer\textquotesingle s marker attachment overrides its foreground head. The 0.073 of this model\textquotesingle s oracle gap is the 21.8 \% of the nuclei without a marker, as the 0.17 of the WS layer\textquotesingle s was its 14.5 \%: the next step on either side is a detection objective for the distance head, not the assignment step.
\item
  \textbf{What this says about the plan.} On PanNuke, with a fair baseline on every count (the same backbone, the same three targets, post-processing tuned on a held-out fold like the WS layer\textquotesingle s markers), the multi-marker WS layer is not a better nucleus instance segmenter than a U-Net with a HoVer-Net-style head and a 16-ms watershed: it is 0.003 F1 behind, 0.013 PQ behind, six times slower at inference, and its foreground head degenerates. What the WS layer still offers is architectural (one differentiable module trained end to end from the instance labels, no post-processing threshold on the foreground, an exact limit within 0.01 F1), plus the one empirical fact this section adds, that at the p it trains to its solve \emph{is} the marker watershed of its learned contour map. The remaining caveats are those of sections 10.1 and 11.1: one seed, one fold pair, the baseline\textquotesingle s thresholds chosen on 300 patches of fold 3, and the foreground target of the two models identical here but different from section 10.1\textquotesingle s gap target (not re-run with \texttt{-{}-gap}).
\end{itemize}

\subsection{Seeds, fold pairs, and a detection objective for the distance head}

Sections 11.1 and 11.2 left two open items: everything was one seed on one fold pair, and the WS layer\textquotesingle s oracle rows said that its remaining gap was the 14.5 \% of nuclei without a marker, which a detection objective on the distance head might recover. This section runs both, with the fair baseline of section 11.2 beside the WS layer in every configuration. Code: \texttt{-\/-w-peak} / \texttt{-\/-peak-pool} in \path{experiments/exp_instance_pannuke.py} (\path{peak_loss}) and \path{exp_unet_dist.py}; \texttt{-\/-seed}, \texttt{-\/-train-fold} / \texttt{-\/-test-fold} / \texttt{-\/-tune-fold} already existed; \path{run_instances_11_3.sh} (jacobian), \path{ruche_run.sh} / \path{ruche_submit.sh} (one SLURM job per run on the A100 partition of ruche, the baseline as a job dependent on its layer run, the folds staged to the node-local disk); \path{plot_instances.py} now runs without the fold-2 baselines of sections 6 and 10.1 (\texttt{-\/-unet\ ""}) and names its reference run (\texttt{-\/-ref-label}); \path{compare_instances.py} reads the baseline runs too and pairs only runs that share a test fold. Results: \texttt{results/pannuke\_instances/\{dist\_s1,\ dist\_s2,\ dist\_f23,\ dist\_f31,\ dist\_peak,\ dist\_peak5\}/} and \texttt{unet\_dist\_\{s1,\ s2,\ f23,\ f31,\ peak5\}/} (each with \path{figures/summary.md}, \texttt{run.log}, the per-image npz; the examples npz only for the peak runs), \path{compare_11_3_seeds_folds.md} and \path{compare_11_3_peak.md}.

\textbf{Seeds and fold pairs.} The section 11.1 layer (\texttt{-\/-mk-head\ dist}, the recipe of section 11 otherwise) and the section 11.2 baseline (three-head U-Net + HoVer-Net-style watershed) were retrained with seeds 1 and 2 on fold 1 $\to$ fold 2 (post-processing chosen on fold 3, as before) and with seed 0 on the two other fold pairs, 2 $\to$ 3 (chosen on fold 1) and 3 $\to$ 1 (chosen on fold 2). Every run chose its own post-processing on its own tuning fold; every baseline run also loads the matching layer run for the markers $\times$ assignment 2 $\times$ 2 of section 11.2. On an A100 the WS layer trains in 184 s per epoch (1.5 h; 290 s on jacobian\textquotesingle s RTX 6000 Ada), 2.7 h per job with the tuning and the evaluation; the baseline in 32 s per epoch, 48 min per job.

\begin{center}\scriptsize\setlength{\tabcolsep}{3pt}
\resizebox{\ifdim\width>\textwidth\textwidth\else\width\fi}{!}{%
\begin{tabular}{@{}llllll@{}}
\toprule
configuration (train $\to$ test, tune) & n & layer: F1 / PQ / AJI / IoU & baseline: F1 / PQ / AJI / IoU & baseline $-$ layer, paired: F1 / PQ / IoU & WS layer oracle $-$ baseline oracle, F1 \\
\midrule

1 $\to$ 2, 3; seed 0 (sections 11.1 / 11.2) & 2523 & 0.791 / 0.641 / 0.653 / 0.707 & 0.794 / 0.655 / 0.658 / 0.719 & +0.003 {[}+0.001, +0.006{]} / +0.013 {[}+0.011, +0.015{]} / +0.012 & +0.091 \\
1 $\to$ 2, 3; seed 1 & 2523 & 0.790 / 0.642 / 0.653 / 0.708 & 0.793 / 0.654 / 0.657 / 0.718 & +0.003 {[}+0.000, +0.006{]} / +0.012 {[}+0.009, +0.014{]} / +0.009 & +0.098 \\
1 $\to$ 2, 3; seed 2 & 2523 & 0.789 / 0.642 / 0.654 / 0.707 & 0.799 / 0.656 / 0.673 / 0.727 & +0.010 {[}+0.007, +0.012{]} / +0.014 {[}+0.012, +0.016{]} / +0.020 & +0.089 \\
2 $\to$ 3, 1; seed 0 & 2722 & 0.787 / 0.640 / 0.653 / 0.711 & 0.795 / 0.655 / 0.664 / 0.729 & +0.008 {[}+0.005, +0.010{]} / +0.015 {[}+0.013, +0.017{]} / +0.018 & +0.102 \\
3 $\to$ 1, 2; seed 0 & 2656 & 0.787 / 0.636 / 0.655 / 0.709 & 0.797 / 0.659 / 0.661 / 0.725 & +0.011 {[}+0.008, +0.013{]} / +0.023 {[}+0.021, +0.026{]} / +0.016 & +0.093 \\
\bottomrule
\end{tabular}}
\end{center}

Marker quality of the WS layer\textquotesingle s markers (nuclei with one / none / several; markers in background per patch): seed 1 83.5 / 14.8 / 1.7 \%, 3.6; seed 2 84.3 / 13.9 / 1.7 \%, 3.8; 2 $\to$ 3 84.6 / 13.3 / 2.0 \%, 4.2; 3 $\to$ 1 82.9 / 15.1 / 2.1 \%, 3.4 (section 11.1: 83.9 / 14.5 / 1.6 \%, 3.6). The tuning folds chose the WS layer\textquotesingle s thr 0.2 / min area 16 / h 0.2 in every configuration but 3 $\to$ 1 (thr 0.3), and for the baseline thr 0.4--0.5, min area 4--9, h 0.2--0.3 with the \texttt{both} or \texttt{dist} relief, i.e. the same two operating points as section 11.2 (the baseline\textquotesingle s markers: 76--81 \% one, 18--22 \% none, 1.6--2.2 in background). The 2 $\times$ 2 repeats section 11.2 on every configuration: with the WS layer\textquotesingle s own markers the baseline\textquotesingle s watershed is 0.003--0.013 F1 below the solve but 0.016--0.022 IoU above it, and with the baseline\textquotesingle s watershed fixed, the WS layer\textquotesingle s markers are 0.013--0.021 F1 below the baseline\textquotesingle s own: each marker operating point suits its assignment step (the lower threshold the solve, the higher one the watershed), and the model\textquotesingle s markers through the WS layer\textquotesingle s solve are the worst cell (0.744--0.772 F1 on the fold pairs).

\textbf{A detection objective for the distance head.} The 14.5 \% of nuclei without a marker in section 11.1 are, on the eight crowded Colon crops of \path{examples_crop.png}, mostly nuclei whose predicted distance map does peak but too low (174 of the 291 unmarked nuclei there peak between 0.2 and 0.5, 61 below 0.2, 56 above 0.5 but absorbed into a neighbour\textquotesingle s basin), and they are smaller than average (median 107 px against 190). Under the pixel MSE a small nucleus hardly weighs, so \texttt{-\/-w-peak} adds a per-nucleus term: the maximum of the predicted map over every ground-truth instance should reach the target\textquotesingle s, loss = mean over instances of (max target $-$ max prediction)$_+^2$, weight 1, the MSE kept. The first form, the raw maximum (\path{dist_peak}), is a failure of a kind worth recording: its gradient reaches one pixel per nucleus per step, and after 30 epochs the map carries needle-shaped maxima everywhere (on 16 fold-3 patches 200 regional maxima per patch at h 0.2 against 24 for \texttt{dist}, 144 at h 0.3; the mean map and its area above 0.2 are unchanged), so the h-maxima rule produces hundreds of markers, the tuning solves spent 1516 s on their first batch and the run was stopped (\path{results/pannuke_instances/\ dist_peak/README.md}). The second form (\path{dist_peak5}, \texttt{-\/-peak-pool\ 5}) takes both maxima on the 5 $\times$ 5 mean-pooled maps, so a nucleus must carry a plateau; it trains normally (peak term 0.0034 at the end, 283 s per epoch) and its untuned validation markers looked like the answer: 89 \% of the nuclei with exactly one marker and 4 \% with none at epoch 29, against 80 / 18 \% for \texttt{dist}. The baseline got the same term (\path{unet_dist_peak5}, peak 0.0030). Both were tuned on fold 3 as before (the WS layer\textquotesingle s tuning took 6605 s against 2438 s: more markers per solve).

\textbf{Results on fold 2 for the peak-loss runs} (seed 0, the rows of section 11.1 / 11.2 for reference):

\begin{center}\scriptsize\setlength{\tabcolsep}{3pt}
\resizebox{\ifdim\width>\textwidth\textwidth\else\width\fi}{!}{%
\begin{tabular}{@{}llllllllllll@{}}
\toprule
output (fold 2) & IoU & BF$_2$ & \href{mailto:F1@0.5}{\nolinkurl{F1@0.5}} & PQ & SQ & DQ & AJI & count err. & n\_pred & one / none / several & bg / patch \\
\midrule

\texttt{dist} layer (sec. 11.1; thr 0.2 / 16 / h 0.2) & 0.707 & 0.762 & 0.791 & 0.641 & 0.811 & 0.791 & 0.653 & 0.239 & 24.3 & 83.9 / 14.5 / 1.6 \% & 3.6 \\
\path{unet_dist} baseline (sec. 11.2; thr 0.5 / 4 / h 0.2, relief c $-$ d) & 0.719 & 0.774 & 0.794 & 0.655 & 0.824 & 0.794 & 0.658 & 0.190 & 20.6 & 76.6 / 21.8 / 1.6 \% & 1.6 \\
\path{dist_peak5} layer (thr 0.3 / 16 / h 0.2) & 0.674 & 0.720 & 0.701 & 0.558 & 0.796 & 0.701 & 0.617 & 0.712 & 33.1 & 85.7 / 7.0 / 7.4 \% & 10.0 \\
\path{dist_peak5} layer, ground-truth markers (oracle) & 0.859 & 0.922 & 0.958 & 0.804 & -- & -- & 0.857 & 0.019 & 23.3 & & \\
\path{dist_peak5} PW limit, learned costs (307) & 0.658 & 0.708 & 0.683 & 0.534 & -- & -- & 0.574 & 0.695 & 34.4 & & \\
\path{unet_dist_peak5} baseline (thr 0.5 / 16 / h 0.3, relief c) & 0.725 & 0.776 & 0.766 & 0.626 & 0.816 & 0.766 & 0.675 & 0.233 & 25.6 & 86.3 / 8.9 / 4.8 \% & 6.8 \\
\bottomrule
\end{tabular}}
\end{center}

Paired on fold 2: \path{dist_peak5} $-$ \texttt{dist} $-$0.090 F1 {[}$-$0.095, $-$0.085{]}, $-$0.083 PQ, $-$0.033 IoU; \path{unet_dist_peak5} $-$ \path{unet_dist} $-$0.028 F1 {[}$-$0.031, $-$0.025{]}, $-$0.029 PQ, +0.006 IoU; \path{dist_peak5} $-$ \path{unet_dist_peak5} $-$0.065 F1, $-$0.067 PQ, $-$0.051 IoU; the oracles are unchanged (\path{dist_peak5} oracle $-$ \texttt{dist} oracle $-$0.001 F1). On fold 3 the tuning already said so: the best PQ of the \path{dist_peak5} layer over the 72 settings is 0.560 against 0.646 for \texttt{dist}, and of \path{unet_dist_peak5} 0.632 against 0.662.

\begin{itemize}

\item
  \textbf{Honest outcome: section 11.2 holds on every seed and fold pair.} Across five configurations the baseline is level or ahead of the WS layer on F1 (+0.003 to +0.011, every interval at or above zero), ahead on PQ (+0.012 to +0.023) and IoU (+0.009 to +0.020), and the WS layer\textquotesingle s advantage with ground-truth markers is a stable +0.09 to +0.10 F1. The spread between seeds (0.002 F1, 0.001 PQ for the WS layer; 0.006 F1, 0.002 PQ for the baseline) and between fold pairs (0.004 F1, 0.005 PQ) is smaller than the baseline-minus-layer difference on every configuration, so the one-seed caveat of sections 11.1 and 11.2 is lifted: the distance-head U-Net with a watershed is at least as good a nucleus instance segmenter as the WS layer with the same head, and slightly better on PQ, and the WS layer\textquotesingle s instances remain 0.01--0.02 IoU short of the thresholded foreground on every fold.
\item
  \textbf{The detection objective recovers the missed nuclei and loses more elsewhere.} The pooled peak loss halves the nuclei without a marker (14.5 $\to$ 7.0 \% for the WS layer, 21.8 $\to$ 8.9 \% for the baseline) but it does so by raising the map wherever something nucleus-like sits: markers in background go from 3.6 to 10.0 per patch for the WS layer (1.6 $\to$ 6.8 for the baseline) and nuclei split by several markers from 1.6 to 7.4 \% (1.6 $\to$ 4.8 \%), and no threshold on fold 3 trades these back (the chosen thr 0.3 against 0.2). The net effect is $-$0.090 F1 / $-$0.083 PQ for the WS layer and $-$0.028 / $-$0.029 for the baseline, with the oracle unchanged (the costs are the same): under the WS layer every spurious marker becomes an instance, since the solve has no mechanism to discard a label, while the watershed\textquotesingle s foreground mask and minimum area absorb part of them, which is why the baseline loses a third as much. The raw-maximum form was worse still (needle peaks, hundreds of markers). So the lever that the oracle rows pointed at, the markers, does not move with a per-nucleus objective of this kind: what is needed is a marker head that is precise as well as sensitive (a detection loss with explicit negatives, or HoVer-Net\textquotesingle s gradient term that penalises spurious maxima), and that improvement would go to the baseline\textquotesingle s watershed as much as to the WS layer.
\item
  \textbf{What this says about the plan.} With the one-seed caveat lifted, the record on PanNuke is: the multi-marker WS layer with the distance head is a strong instance segmenter (F1 0.787--0.791, PQ 0.636--0.642 on three fold pairs), the plain three-head U-Net with a watershed on the same head is at least as good on F1 and 0.012--0.023 PQ better, the WS layer\textquotesingle s costs are 0.09--0.10 F1 better than the baseline\textquotesingle s with perfect markers, and the first attempt at better markers helped neither. The WS layer\textquotesingle s case therefore rests on what the oracle measures, the quality of its learned costs given the markers, which only pays off with a marker head that does not exist yet, and on its structural properties (one differentiable module, the exact limit at 0.7 s per patch, no post-processing beyond three scalars). The ruche cluster makes the next sweep cheap: a 3 $\times$ 3 seeds $\times$ fold-pairs grid of the two models is 18 jobs, one afternoon on the A100 partition.
\end{itemize}

\subsection{A marker head that is precise as well as sensitive? Hard negatives and the gradient term}

Section 11.3 ended with the peak loss raising sensitivity and precision together in the wrong direction: fewer nuclei without a marker, many more markers in background and inside nuclei. This section adds the two counterweights it named, as options of both scripts (\path{peak_loss}, \path{neg_loss}, \path{grad_loss} in \path{experiments/exp_instance_pannuke.py}, imported by \path{exp_unet_dist.py}): \texttt{-\/-w-neg}, hard negatives, the mean of the 256 largest values of the pooled prediction over the background pixels of every image, squared (a spurious maximum in background costs directly); and \texttt{-\/-w-grad}, HoVer-Net\textquotesingle s gradient term, the MSE of the horizontal and vertical finite differences of prediction and target over all pixels (a nucleus should carry one smooth peak, the background none). Four variants, each for the WS layer and for the fair baseline, seed 0, fold 1 $\to$ fold 2, post-processing chosen on fold 3 as before: the gradient term alone (\path{dist_grad}, weight 2), the pooled peak loss of section 11.3 with the gradient term (\path{dist_peakgrad}), with hard negatives (\path{dist_peakneg}, weight 1), and with both (\path{dist_det}); \path{unet_dist_*} the same for the baseline, each loading its layer run for the 2 $\times$ 2. Results: \texttt{results/pannuke\_instances/\{dist,unet\_dist\}\_\{grad,peakgrad,peakneg,det\}/}, \path{compare_11_4.md} (which also carries the \texttt{peak5} pair of section 11.3). The training scripts also gained \texttt{-\/-tb} (TensorBoard scalars under \texttt{\textless{}outdir\textgreater{}/tb}, on by default in the run scripts) and an \texttt{-\/-examples\ diverse} selection with \texttt{-\/-examples-only} (one patch per tissue type for the eight most frequent types, run from the saved model: \texttt{figures/examples\_diverse\_\{crop,full\}.png} next to the touching-cluster figures of sections 11--11.2, for the \texttt{dist}, \path{bce_tuned}, section 11 and \path{unet_dist} runs; PanNuke\textquotesingle s variability, sparse mucosa to elongated stromal nuclei, is visible there where the eight densest Colon patches hid it).

\textbf{Results on fold 2} (layer / baseline; the plain distance head of sections 11.1 / 11.2 and the pooled peak loss of 11.3 for reference; markers = nuclei with one / none / several, background markers per patch; fold 3 = the best PQ of the tuning table):

\begin{center}\scriptsize\setlength{\tabcolsep}{3pt}
\resizebox{\ifdim\width>\textwidth\textwidth\else\width\fi}{!}{%
\begin{tabular}{@{}lllllll@{}}
\toprule
marker-head loss & WS layer F1 / PQ / IoU & WS layer markers & fold 3 & baseline F1 / PQ / IoU & baseline markers & fold 3 \\
\midrule

MSE only (11.1 / 11.2) & \textbf{0.791 / 0.641 / 0.707} & 84 / 14 / 2 \%, 3.6 & 0.646 & 0.794 / 0.655 / 0.719 & 77 / 22 / 2 \%, 1.6 & 0.662 \\
+ gradient & 0.789 / 0.641 / 0.709 & 85 / 14 / 2 \%, 3.9 & 0.646 & \textbf{0.796 / 0.657 / 0.720} & 77 / 22 / 1 \%, 1.6 & 0.663 \\
+ pooled peak (11.3) & 0.701 / 0.558 / 0.674 & 86 / 7 / 7 \%, 10.0 & 0.560 & 0.766 / 0.626 / 0.725 & 86 / 9 / 5 \%, 6.8 & 0.632 \\
+ pooled peak + gradient & 0.717 / 0.572 / 0.677 & 86 / 8 / 6 \%, 9.3 & 0.578 & 0.768 / 0.627 / 0.725 & 87 / 9 / 4 \%, 6.5 & 0.636 \\
+ pooled peak + hard negatives & 0.674 / 0.527 / 0.600 & 70 / 25 / 5 \%, 1.5 & 0.537 & 0.751 / 0.613 / 0.704 & 70 / 25 / 6 \%, 1.5 & 0.620 \\
+ pooled peak + negatives + gradient & 0.658 / 0.510 / 0.621 & 81 / 12 / 8 \%, 4.5 & 0.517 & 0.775 / 0.631 / 0.716 & 77 / 18 / 4 \%, 2.2 & 0.637 \\
\bottomrule
\end{tabular}}
\end{center}

Paired on fold 2 against the MSE-only run of the same model (F1 / PQ / IoU, bootstrap intervals in \path{compare_11_4.md}): gradient $-$0.002 / 0.000 / +0.002 for the WS layer and +0.002 / +0.002 / +0.002 for the baseline, every interval crossing zero; peak + gradient $-$0.074 / $-$0.069 / $-$0.030 and $-$0.026 / $-$0.027 / +0.006; peak + negatives $-$0.117 / $-$0.115 / $-$0.106 and $-$0.043 / $-$0.042 / $-$0.015; all three $-$0.132 / $-$0.132 / $-$0.086 and $-$0.019 / $-$0.024 / $-$0.003. The oracles with ground-truth markers are unchanged in every run (layer 0.956--0.959 F1, baseline 0.861--0.870): the costs and the foreground are not what these losses change. Training cost: the gradient term is free; the hard negatives make the fold-3 tuning slow because the maps carry many candidate maxima at low thresholds (20 205 s for \path{dist_peakneg} against 2438 s for \texttt{dist}), and the chosen thresholds move to the top of the grid (0.7 for \path{dist_peakneg}).

\begin{itemize}

\item
  \textbf{Honest outcome: none of the four variants improves the markers, and the ranking is the same for both models.} The gradient term is neutral ($\pm$0.003 on every metric, the marker counts unchanged): it penalises wiggles but creates or removes no maxima by itself, and the MSE-trained map was already smooth. Every variant with the peak loss is worse than MSE alone, on fold 3 as on fold 2, for the WS layer ($-$0.074 to $-$0.132 F1) and for the baseline ($-$0.019 to $-$0.043). The gradient term recovers a little of the peak loss\textquotesingle s damage (+0.016 F1 for the WS layer against \texttt{peak5}), the hard negatives none.
\item
  \textbf{Hard negatives fail for a reason that is specific to the pooled design, and instructive.} They do what they are asked, background markers fall from 10.0 to 1.5 per patch, but the nuclei without a marker come back (25 \%) and the WS layer\textquotesingle s foreground IoU falls to 0.600, the lowest of any full-set output since section 6. The negatives are taken on the 5 $\times$ 5 mean-pooled map over the pixels outside every instance, so a background pixel next to a nucleus carries nuclear pixels in its window and is pushed down together with the nucleus\textquotesingle s edge, in exactly the band the WS layer\textquotesingle s markers are grown into (thr 0.2); the peak term pulls the other way inside, and the fold-3 tuning ends at thr 0.7 with small markers and a shrunken foreground. A background eroded by the pooling radius would remove the conflict; whether it would also remove the false markers the peak loss creates (which sit on nucleus-like stroma, not next to nuclei) is open, and was not run.
\item
  \textbf{The WS layer pays for every spurious marker, the watershed does not.} The spread between the two models is the same lesson as section 11.3: with the same head and the same loss, the baseline loses a third to a quarter of what the WS layer loses ($-$0.026 against $-$0.074 with the gradient term, $-$0.019 against $-$0.132 with all three), because the watershed\textquotesingle s foreground mask and minimum area absorb markers that the solve must turn into instances. A marker head good enough to help the WS layer would have to be more precise than the watershed needs.
\item
  \textbf{What this says about the plan.} Two rounds (11.3, 11.4) of objectives on the distance head, six variants, have not moved the WS layer\textquotesingle s markers past the MSE-trained map of section 11.1, and the one neutral change (the gradient term) is neutral for the baseline too. The lever is not the loss on the map but what the costs are trained with: the solve has only ever been trained from ground-truth markers (section 11\textquotesingle s recipe), so it has never learned to confine a spurious label or to do without a missing one, and at test time it meets both. Training the solve from the predicted markers for part of the schedule, with the cross-entropy still against the true instances, is the experiment that addresses the mismatch directly; it is the next one if this thread continues. Everything here remains one seed on one fold pair.
\end{itemize}

\subsection{Training the solve from the model\textquotesingle s own markers}

Added 23--24 October 2026 (side session; seven jobs on ruche\textquotesingle s A100 partition over two nights). Section 11.4 ended on the diagnosis that the lever was not the loss on the marker map but what the solve is trained with: in every run since section 11 the training markers were the ground-truth instances eroded by 2--3 px, so the costs never met a spurious marker or a missing one, while at test time 14.5 \% of the nuclei have no marker and 3.6 markers per patch sit in background. This section closes that mismatch. Code: \texttt{-\/-mk-sched\ E0\ E1} in \path{experiments/exp_instance_pannuke.py}: from epoch E0 on, every training image draws, with probability (ep $-$ E0 + 1) / (E1 $-$ E0 + 1) rising to 1 at E1, the markers that \path{predicted_markers} extracts from the current distance head (no gradient through the extraction, which runs in a process pool; thresholds \texttt{-\/-mk-train-thr\ 0.2}, \texttt{-\/-mk-train-min-area\ 16}, \texttt{-\/-mk-train-h\ 0.2}, the fold-3 choice of section 11.1) in place of the eroded ground truth (\path{mixed_label_systems}). The cross-entropy stays against the true instance map (\path{mapped_cross_entropy}): the target of a predicted label is the ground-truth nucleus holding the majority of its marker pixels (the owner rule of \path{marker_quality}), so a marker in background has no pixel to win and its label is penalised wherever it wins; a nucleus with several markers keeps all of them, the probability of the target being their sum, so the loss does not force a split; the pixels of a nucleus without a marker can be right under no label and get zero weight (they stay in the denominator; \texttt{-\/-mk-unmarked\ bg} makes the background label their target instead: the solve is then asked to hand an unmarked nucleus to the background rather than to a neighbour). The per-epoch log carries the schedule probability, the share of images on predicted markers, the quality of those markers on the training batches and the share of unreachable pixels (\path{sched_*}, \path{train_mk_*}, \path{train_unreach}, also on TensorBoard). Two schedules, the section 11.1 recipe otherwise (\texttt{-\/-mk-head\ dist\ -\/-w-mk\ 2}, p 5 $\to$ 20, edge target, 30 epochs, seed 0, fold 1 $\to$ fold 2, post-processing chosen on fold 3): \path{dist_sched10} (\texttt{-\/-mk-sched\ 10\ 15}: predicted markers from epoch 10, all of them from epoch 15, i.e. half the training) and \path{dist_sched20} (\texttt{-\/-mk-sched\ 20\ 25}: the last third, as a fine-tuning phase), then the first schedule with the background rule (\path{dist_sched10_bg}, \texttt{-\/-mk-unmarked\ bg}). The fair baseline of section 11.2 has no solve to train and is unchanged (\path{unet_dist}). Results: \texttt{results/pannuke\_instances/\{dist\_sched10,\ dist\_sched20,\ dist\_sched10\_bg\}/} (each with \path{figures/summary.md}, \texttt{training.png}, \texttt{examples\_\{crop,full\}.png}, \texttt{run.log}), \path{compare_11_5.md} (the five runs, paired bootstraps); \texttt{.pt} files on ruche and jacobian. The \texttt{bg} recipe was then repeated on the section 11.3 grid, seeds 1 and 2 on fold 1 $\to$ 2 and seed 0 on 2 $\to$ 3 and 3 $\to$ 1 (\texttt{dist\_bg\_\{s1,s2,f23,f31\}}, \path{compare_11_5_seeds_folds.md}, paired against the section 11.3 layer and baseline runs of the same configuration). Cases \path{dist_sched10} / \path{dist_sched20} / \path{dist_sched10_bg} / \path{dist_bg_*} in \path{ruche_run.sh} and \path{run_instances_11_3.sh}; \path{compare_instances.py} gained a "training markers" column. The \texttt{bg} job died of a CUDA error on its node during the tuning, after the model was saved: the script now writes the arguments and the training log to the json right after training, \texttt{-\/-eval-only} rebuilds the log from the TensorBoard events when no json exists (\path{log_from_tb}), and \path{ruche_run.sh} has an \texttt{eval\_\textless{}run\textgreater{}} case that reruns the tuning and the evaluation from a saved model (used here).

\textbf{What the solve was trained on.} On the training batches the predicted markers are, from the first scheduled epoch, what the test fold sees: 79 \% of the nuclei with one marker, 19 \% with none, 4.1 per patch in background at epoch 10, 84 / 14 / 3.8 from epoch 15 on (test fold in section 11.1: 83.9 / 14.5 / 3.6), 1.3 \% $\to$ 0.8 \% of the pixels unreachable. The marker extraction costs 20 s per epoch with 8 workers. The cross-entropy rises from 0.094 to 0.137--0.142 the moment the predicted markers come in and stays there (the 11.1 run ends at 0.089): against an imperfect marker set the pixels a spurious label wins and the nuclei a neighbour\textquotesingle s label invades are errors by construction. The epochs took 340--400 s against 184 s in section 11.3, which is the node, not the schedule: all three training jobs landed on ruche-gpu16, \path{dist_sched20} is as slow before epoch 20 as after it, the same converged solve takes 158 ms per image there and 90 ms on ruche-gpu18 at the same PCG iteration count (the \texttt{bg} evaluation), the node threw a CUDA error on the third job, and the four grid jobs trained at 189--206 s per epoch on four other nodes.

\textbf{Results on fold 2} (the section 11.1 layer and the section 11.2 baseline for reference; markers = nuclei with one / none / several, background markers per patch; the oracle with ground-truth markers; the exact Power Watershed limit of the learned costs on the 307 limit patches):

\begin{center}\scriptsize\setlength{\tabcolsep}{3pt}
\resizebox{\ifdim\width>\textwidth\textwidth\else\width\fi}{!}{%
\begin{tabular}{@{}lllllll@{}}
\toprule
run & training markers & F1 / PQ / AJI / IoU & count err. & markers (fold-3 choice) & oracle F1 / PQ & limit F1 / PQ (307) \\
\midrule

\texttt{dist} (11.1) & ground truth & 0.791 / 0.641 / 0.653 / 0.707 & 0.239 & 83.9 / 14.5 / 1.6 \%, 3.6 (thr 0.2, area 16, h 0.2) & 0.959 / 0.805 & 0.758 / 0.603 \\
\path{dist_sched10} & predicted from epoch 10 (all from 15) & 0.792 / 0.645 / 0.666 / 0.717 & 0.186 & 85.6 / 12.8 / 1.6 \%, 4.3 (thr 0.2, area 1, h 0.2) & 0.877 / 0.725 & 0.694 / 0.558 \\
\path{dist_sched20} & predicted from epoch 20 (all from 25) & 0.792 / 0.645 / 0.663 / 0.714 & 0.197 & 85.7 / 12.6 / 1.7 \%, 4.3 (thr 0.2, area 1, h 0.2) & 0.902 / 0.745 & 0.719 / 0.576 \\
\path{dist_sched10_bg} & as \texttt{sched10}, unmarked nuclei $\to$ background & \textbf{0.795 / 0.649 / 0.668 / 0.715} & \textbf{0.182} & 85.5 / 12.7 / 1.8 \%, 4.3 (thr 0.2, area 4, h 0.2) & 0.872 / 0.723 & 0.700 / 0.564 \\
\path{unet_dist} (11.2) & -- (watershed) & 0.794 / 0.655 / 0.658 / 0.719 & 0.190 & 76.6 / 21.8 / 1.6 \%, 1.6 & 0.867 / 0.708 & -- \\
\bottomrule
\end{tabular}}
\end{center}

Paired on fold 2 (F1 / PQ / AJI / IoU, bootstrap intervals in \path{compare_11_5.md}): \path{dist_sched10} $-$ \texttt{dist} +0.002 {[}$-$0.001, +0.004{]} / +0.004 {[}+0.002, +0.006{]} / +0.013 {[}+0.012, +0.016{]} / +0.010 {[}+0.007, +0.013{]}; \path{dist_sched20} $-$ \texttt{dist} +0.002 {[}$-$0.000, +0.004{]} / +0.004 {[}+0.002, +0.005{]} / +0.011 / +0.007; \path{dist_sched10} $-$ \path{dist_sched20} 0.000 / 0.000 / +0.003 / +0.003; \path{dist_sched10_bg} $-$ \texttt{dist} +0.005 {[}+0.002, +0.007{]} / +0.008 {[}+0.006, +0.010{]} / +0.015 {[}+0.013, +0.017{]} / +0.009 {[}+0.005, +0.012{]}, and \path{dist_sched10_bg} $-$ \path{dist_sched10} +0.003 {[}+0.001, +0.005{]} / +0.004 {[}+0.002, +0.005{]} / +0.002 / $-$0.001. Against the baseline, \path{unet_dist} $-$ \path{dist_sched10} = +0.002 {[}$-$0.001, +0.004{]} / +0.009 {[}+0.007, +0.011{]} / $-$0.008 {[}$-$0.010, $-$0.005{]} / +0.002 {[}$-$0.001, +0.005{]}, and \path{unet_dist} $-$ \path{dist_sched10_bg} = $-$0.001 {[}$-$0.004, +0.002{]} / +0.005 {[}+0.003, +0.007{]} / $-$0.009 {[}$-$0.012, $-$0.007{]} / +0.004 {[}+0.001, +0.007{]}. On the oracles, \path{dist_sched10} $-$ \texttt{dist} = $-$0.081 F1 / $-$0.079 PQ, \path{dist_sched20} $-$ \texttt{dist} = $-$0.057 / $-$0.060 and \path{dist_sched10_bg} $-$ \texttt{dist} = $-$0.086 / $-$0.082; on the limit patches $-$0.065 / $-$0.044, $-$0.040 / $-$0.027 and $-$0.058 / $-$0.039.

\begin{itemize}
\item
  \textbf{The mismatch was real but small: closing it gains 0.004 PQ, 0.013 AJI, 0.010 IoU and a fifth of the count error, and nothing on F1.} Both schedules land on the same output (0.792 / 0.645), the longer one marginally ahead on AJI and IoU. The gains are where the diagnosis placed them: the solve now copes with the markers it is given. The fold-3 tuning table is flat in the minimum area (PQ 0.651 at 1, 4, 9 and 16 px, where the 11.1 model went 0.640 $\to$ 0.646), so the tuning keeps the tiny markers and the test fold gets 4.3 background markers per patch instead of 3.6 at a lower count error (0.186 against 0.239); the fallback markers of section 11, which cost the 11.1 model 0.015 F1, now cost 0.002; and the sparse patches, where a spurious marker is a large share of the count, move most (\textless{} 10 nuclei: F1 0.689 $\to$ 0.704, PQ 0.553 $\to$ 0.568, count error 0.64 $\to$ 0.44; the three crowded bins within +0.007). A background marker still becomes an instance at test time, but a smaller and better-confined one.
\item
  \textbf{Giving the unmarked nuclei to the background is the better rule.} With \texttt{-\/-mk-unmarked\ bg} the pixels of a nucleus without a marker are no longer skipped but asked of the background label, so a neighbour\textquotesingle s label is penalised for invading them and the foreground cost is pushed down there. It adds +0.003 F1 / +0.004 PQ / +0.002 AJI over the ignore rule (every interval clear of zero but AJI\textquotesingle s), for +0.005 F1 / +0.008 PQ / +0.015 AJI / +0.009 IoU over the 11.1 run, the first schedule whose F1 gain is clear of zero; the count error reaches 0.182, the lowest of the report. The gain is again largest on the sparse patches (\textless{} 10 nuclei: F1 0.716, PQ 0.576, count error 0.42, from 0.689 / 0.553 / 0.64 in 11.1) and +0.003--0.004 F1 on the three crowded bins. The cross-entropy sits higher still (0.146 against 0.136), as the unmarked nuclei now count. The oracle (0.872) and the limit (0.700) are those of the ignore rule within 0.006: the mechanism below is the same. The foreground head does not pay: IoU is 0.715 against 0.717.
\item
  \textbf{The WS layer now matches the baseline on F1 and AJI leads it, with PQ 0.005 behind.} \path{unet_dist} $-$ \path{dist_sched10_bg} is $-$0.001 F1 (interval across zero), +0.005 PQ, $-$0.009 AJI, +0.004 IoU, and the count errors are equal (0.182 against 0.190). The remaining PQ margin is the segmentation quality of the matched nuclei, not their detection: SQ 0.816 for the WS layer against 0.824 for the baseline at equal DQ (0.795 / 0.794); the 11.1 run had SQ 0.811. The watershed\textquotesingle s boundaries on the contour map are tighter than the softened solve\textquotesingle s, which is the price of the hedging described next. The 2 $\times$ 2 of section 11.2 was not rerun: the baseline\textquotesingle s watershed and markers are unchanged, and the WS layer\textquotesingle s own markers are the same map at a different minimum area.
\item
  \textbf{The oracle falls from 0.959 to 0.877 / 0.902, and that is the mechanism, not a side effect.} Trained on perfect markers, the costs could trust every marker: the marker attachment overrode the foreground everywhere (oracle count error 0.020, soft pixels 6.8 \%). Trained on its own markers, the solve learns to veto a marker through the unary costs where the foreground is weak and to hedge between neighbours: with ground-truth markers it now drops 11 \% of them (20.7 instances per patch for 23.3 nuclei, count error 0.129) and 12.8 \% of the pixels are soft, the learned edge scale ends at 1.84 instead of 2.14, and the converged solve needs 263--300 PCG iterations per batch instead of 177 (158--189 ms per image instead of 94). The exact Power Watershed limit of the same costs drops by 0.065 F1 (the hedging lives in the finite-p smoothing, which the limit removes), so this model is further from a watershed than any in section 11, and its oracle advantage over the baseline\textquotesingle s watershed oracle shrinks from +0.09 to +0.01 (\texttt{sched10}, \path{sched10_bg}) / +0.03 (\texttt{sched20}). The "0.17 F1 between the WS layer and its oracle" of sections 11.1--11.4 was therefore mostly a property of the ground-truth-marker training, not headroom the markers could unlock: the oracle measured a solve that assumed its markers were true.
\item
  \textbf{Validation during training is noisy at this scale.} The 200-patch curve dips at epoch 19 for all three schedules (0.698--0.723 F1 against 0.760 at epochs 14 and 24), for \path{dist_sched20} before its schedule has begun (its random stream differs from \texttt{dist}\textquotesingle s from epoch 0, since the schedule draws a Bernoulli per image at every step, so its permutations differ), so the dip is the untuned-marker validation, not the switch; the 11.1 run itself ends at 0.739 there and 0.791 on the full fold.
\item
  \textbf{Seeds and fold pairs for the \texttt{bg} recipe: the gains on AJI, IoU and count error hold, the F1 gain does not, and the baseline stays ahead.} The table gives the \texttt{bg} layer on each configuration of section 11.3, paired against the WS layer and the baseline of that configuration (F1 / PQ / AJI / IoU; intervals in \path{compare_11_5_seeds_folds.md}):

  {\def\LTcaptype{none} % do not increment counter
  \begin{center}\scriptsize\setlength{\tabcolsep}{3pt}
\resizebox{\ifdim\width>\textwidth\textwidth\else\width\fi}{!}{%
\begin{tabular}{@{}lllll@{}}
  \toprule
  configuration (train $\to$ test, tune) & \texttt{bg} layer F1 / PQ / AJI / IoU, count err. & \texttt{bg} $-$ 11.3 layer & baseline $-$ \texttt{bg} layer & oracles: \texttt{bg} / 11.3 / baseline \\
  \midrule

  1 $\to$ 2, 3; seed 0 & 0.795 / 0.649 / 0.668 / 0.715, 0.182 & +0.005 {[}+0.002, +0.007{]} / +0.008 / +0.015 / +0.009 & $-$0.001 {[}$-$0.004, +0.002{]} / +0.005 / $-$0.009 / +0.004 & 0.872 / 0.959 / 0.867 \\
  1 $\to$ 2, 3; seed 1 & 0.787 / 0.643 / 0.665 / 0.714, 0.180 & $-$0.003 {[}$-$0.005, +0.000{]} / +0.001 {[}$-$0.001, +0.003{]} / +0.012 / +0.006 & +0.006 {[}+0.003, +0.008{]} / +0.011 / $-$0.007 / +0.004 & 0.861 / 0.959 / 0.861 \\
  1 $\to$ 2, 3; seed 2 & 0.791 / 0.644 / 0.667 / 0.714, 0.189 & +0.001 {[}$-$0.001, +0.004{]} / +0.002 {[}+0.000, +0.004{]} / +0.014 / +0.007 & +0.008 {[}+0.006, +0.010{]} / +0.012 / +0.006 / +0.013 & 0.870 / 0.960 / 0.871 \\
  2 $\to$ 3, 1; seed 0 & 0.792 / 0.647 / 0.666 / 0.720, 0.167 & +0.005 {[}+0.003, +0.007{]} / +0.007 / +0.013 / +0.009 & +0.003 {[}+0.000, +0.005{]} / +0.008 / $-$0.002 / +0.009 & 0.875 / 0.962 / 0.859 \\
  3 $\to$ 1, 2; seed 0 & 0.789 / 0.644 / 0.664 / 0.719, 0.170 & +0.003 {[}+0.001, +0.005{]} / +0.008 / +0.009 / +0.010 & +0.008 {[}+0.006, +0.011{]} / +0.015 / $-$0.003 / +0.006 & 0.869 / 0.962 / 0.869 \\
\bottomrule
\end{tabular}}
\end{center}

  }

  Over the five configurations the schedule gives the WS layer +0.009 to +0.015 AJI and +0.006 to +0.010 IoU (every interval clear of zero), a count error of 0.167--0.189 against 0.204--0.247 (equal to the baseline\textquotesingle s 0.165--0.190), +0.001 to +0.008 PQ (two intervals touching zero) and $-$0.003 to +0.005 F1: the seed-0 run of the previous bullets was the best of the five on F1, and seed 1 is below its 11.3 counterpart. The seed spread of the \texttt{bg} layer is 0.008 F1 (0.787--0.795) against 0.002 for the 11.3 layer: the predicted markers are a moving target during training. Against the baseline, the F1 gap is $-$0.001 to +0.008 (section 11.3: +0.003 to +0.011) and the PQ gap +0.005 to +0.015 (section 11.3: +0.012 to +0.023), so the schedule closes about half of the PQ gap and little of the F1 gap; AJI is level ($-$0.009 to +0.006), IoU stays 0.004--0.013 below. The oracles collapse to the baseline\textquotesingle s level on every configuration (0.861--0.875 against 0.859--0.871; the 11.3 layers: 0.959--0.962), the markers chosen by the tuning folds are thr 0.2 / h 0.2 at a minimum area of 1--16 (flat tables), and the training-time markers are 84 \% one / 14 \% none / 3.7--4.1 background per patch on every configuration.
\item
  \textbf{What this says about the plan.} The last structural mismatch of the recipe is removed and the WS layer is, on this benchmark, a different model from the watershed (softer, marker-vetoing, further from its own limit, oracle at the watershed\textquotesingle s level) with the baseline\textquotesingle s count error and AJI, 0.001--0.008 F1 and 0.005--0.015 PQ below it over seeds and folds. The recipe going forward is \texttt{-\/-mk-sched\ 10\ 15\ -\/-mk-unmarked\ bg} (the default of \texttt{-\/-mk-unmarked} stays \texttt{ignore} so that the first two runs\textquotesingle{} records read as run). The baseline remains the stronger segmenter per matched nucleus (SQ 0.822--0.827 against 0.815--0.817), which is where the solve\textquotesingle s hedging costs, and the ranking of sections 11.2--11.3 is narrowed, not reversed. What the schedule did not touch is the 12.6--13.5 \% of nuclei without a marker, which neither model recovers and which sections 11.3--11.4 showed the map losses cannot buy. If this thread continues, the one schedule-side item left is training-time thresholds that follow the tuning (the markers are extracted at the fixed 11.1 thresholds, while the tuning folds now prefer a minimum area of 1--9); the larger question, whether the solve\textquotesingle s softness can be traded back for the watershed\textquotesingle s boundary precision (a higher p at the end of training, or the exact limit as the inference step now that the costs are trained for imperfect markers), is open.
\end{itemize}

\section{Check of our Power Watershed limit against the authors\textquotesingle{} reference code}

Our exact limit (\path{power_watershed} in \path{pw_conditioning.py}, \path{power_watershed_multi} in \path{pw_layer.py}: a NumPy/SciPy re-implementation of Algorithm 1 of Couprie, Grady, Najman and Talbot, PAMI 2011) had only been validated against finite-p Random Walker solutions and against itself. It was checked against the authors\textquotesingle{} C/C++ implementation, \path{PW_1.0.1} from \url{https://sourceforge.net/projects/powerwatershed} (Camille Couprie, Sept 2010; PINK + CSparse), on the same problems. Harness and build notes: \path{experiments/pwref/} (README), records: \path{results/pwref/}.

\textbf{What the reference program does.} \texttt{powerwatsegm\ -a\ 2\ -i\ image.pgm\ -s\ seeds.pgm} (or \texttt{-m} for a multi-label seed image, values 1..K) builds the 4-connected grid, weights w = 255 $-$ \textbar I\_u $-$ I\_v\textbar{} (8-bit grey images; colour images use the maximum over channels), then \emph{reconstructs} the weights geodesically from the seeds (\path{grey_weights_PW}: the dilation of the seed-adjacent weights under the weights in the line graph, i.e. every edge\textquotesingle s weight is lowered to the bottleneck of the best path that connects it to a seed), and runs \path{PowerWatershed_q2} with the reconstructed weights for the plateau ordering and the original weights inside a plateau. A plateau is the grid-connected set of edges of equal reconstructed weight around the current maximum; inside it the edges whose original weight is higher are merged first (Kruskal, never two known nodes), and the unit-weight Random Walker is solved by a CSparse LU on the edges at the level, between merged nodes, with the seeds and the previously solved nodes as boundary. Potentials are stored in single precision. It writes \texttt{mask.pgm} (label of the largest potential; at exactly 0.5 the first label, the foreground), and \texttt{proba.pgm} = 255 $-$ 255$\cdot$P(label 1) truncated to a byte. With \texttt{-g\ 1} the reconstructed weights replace the weights altogether (a different problem).

\textbf{Why the reconstruction is the algorithm, not a heuristic.} The plateau of Algorithm 1 lives in the merged graph: two equal-weight edges that touch the same merged node belong to one plateau even when they share no pixel. A breadth-first search over grid-adjacent edges of equal \emph{raw} weight splits such a plateau into pieces that are then solved one after the other, and the result depends on the order in which equal weights are sorted. The reconstruction lowers every interior edge of a region that is cut off from the seeds above level t to exactly t, so the reconstructed plateau at level t is grid-connected through those interior edges and is the merged-graph plateau (split at most at known nodes, which does not couple anything); the interior edges are then merged first, and the solve runs on the level-t edges between merged nodes. That is what our code does directly with a union-find over merged nodes. Both readings were verified (\path{check_recon.py} reproduces the reconstructed weights of the dump exactly on 30 cases; \path{order_dependence.py} runs the reference under three seeds of its unseeded \texttt{rand()}).

\textbf{Set-up.} A replacement \texttt{main} (\path{pwref_wrap.cpp}) calls the reference\textquotesingle s own functions in the program\textquotesingle s order, because the third-party option parser of the release miscompiles on current clang; the released executable built on jacobian (gcc 13) gives masks identical to the wrapper\textquotesingle s on the three 512$^2$ cases it was run on (\texttt{proba.pgm} differs by $\pm$1 byte on a few hundred pixels, single-precision rounding across compilers). The float potentials come from a six-line dump patch at the end of \path{PowerWatershed_q2}. Both codes get the identical graph (the reference\textquotesingle s own edge order), the identical integer weights (our costs are 255 $-$ w = \textbar I\_u $-$ I\_v\textbar, read back from the reference\textquotesingle s dump; the squared cost of the rest of the report is a monotone transform with the same ties, hence the same limit, and the 2$^{-20}$ quantisation of \path{pw_limit_numpy} plays no role with 8-bit costs) and the identical seeds. Cases: 80 random 24 $\times$ 24 images with four grey levels (plateaus everywhere; 40 two-label with three seed pixels per label, 40 with 3--5 labels), the 512$^2$ cases of section 1 (synthetic $\sigma$ = 0.05 / 0.1 with seeds eroded by 5 and 30 px and scribbles, camera and astronaut eroded by 5 and 30 px), and three multi-label 512$^2$ cases (the synthetic shapes as four labels, camera and astronaut in three intensity classes, all eroded by 5 px).

\textbf{Result: the two codes agree.} Released mode against ours on the raw weights:

\begin{center}\scriptsize\setlength{\tabcolsep}{3pt}
\resizebox{\ifdim\width>\textwidth\textwidth\else\width\fi}{!}{%
\begin{tabular}{@{}lllll@{}}
\toprule
cases & pixels & potentials, max abs. difference & labels (reference rule, $\ge$ 0.5 $\to$ label 1) & pixels exactly at 0.5 / tied \\
\midrule

80 random 24$^2$, 2--5 labels & 46 080 & 6.1e-8 & 0 differ & 3 at 0.5, 25 tied (same in both) \\
10 two-label 512$^2$ & 2.6 M & 6.0e-8 & 0 differ & 0 (clean synthetic), 77--817 per noisy / natural case \\
3 multi-label 512$^2$ (K = 3, 4) & 0.8 M & 4.6e-8 & 11 differ, all on tied top-two probabilities (gap \textless{} 1e-15) & 0 / 765 / 680 tied \\
\bottomrule
\end{tabular}}
\end{center}

\begin{itemize}

\item
  The potentials agree to single precision everywhere (the reference\textquotesingle s \texttt{float}), on the plateau pixels included (6 \% to 36 \% of the pixels are solved on a plateau in the natural-image cases).
\item
  The only label differences are tie-breaks. With our strict rule (x \textgreater{} 0.5 $\to$ foreground) the pixels at exactly 0.5 go to the background, with the reference\textquotesingle s to the foreground: 85 / 627 / 559 pixels on the noisy synthetic cases, 77 / 122 on camera, 761 / 817 on astronaut (0.03--0.3 \% of the pixels, none on the clean synthetic image); both codes have exactly the same set of 0.5 pixels. In the multi-label cases the 11 differing pixels have two labels at the same probability to 1e-15 and the reference\textquotesingle s single-precision sum picks the other one.
\item
  Our K-vector multi-label solve equals K one-hot runs of the scalar code to 1e-15, and the two-label case of the multi-label code equals the scalar code, on every case.
\item
  The reference run with \texttt{-g\ 1} is also reproduced by our code on the reconstructed weights (no label difference, 6e-8 on the potentials).
\item
  Fed the raw weights for both arguments (not reachable from the program), the reference disagrees with ours on 0--91 347 pixels per 512$^2$ case and with itself across \texttt{rand()} seeds (9 387 label changes over 40 random cases, none in the released mode): the no-reconstruction variant is not the limit. Its bucket-sort option (\texttt{quicksort\ =\ true}, hard-wired off in the program) returns positions instead of edge indices inside a plateau and yields potentials outside {[}0, 1{]}; irrelevant to the released program.
\item
  Timings (Mac, Apple M-series, 512$^2$, 262 144 pixels, 523 264 edges): the reference\textquotesingle s \path{PowerWatershed_q2} call takes 0.04--0.07 s, 0.11--0.17 s with its weights and reconstruction (0.17--0.22 s for the whole released executable on jacobian); ours takes 0.40--0.84 s for the scalar code and 0.43--1.0 s for the K-vector code, on at most 256 cost levels. Our general path per level (union-find compression, a sparse graph and its connected components over the whole image) is what costs; a C implementation of the same plateau-by-plateau scheme is an order of magnitude faster, as expected.
\end{itemize}

\textbf{What it changes.} Nothing in the code or the results: no bug was found, and every "limit" row of this report (sections 1--3, the PW-limit evaluations of sections 6--11, the K-label limits of section 11) stands as the exact p $\to$ $\infty$ limit as the authors compute it. The one convention to keep in mind when reading those rows is the tie at exactly 0.5, which our label maps resolve to the background and the reference to the first label; on the 8-bit cases above it touches at most 0.3 \% of the pixels, and in the learned-cost experiments, where the costs are continuous, it cannot occur outside symmetric plateaus.

\subsection{A Numba back end for the limit, bit-for-bit}

The profile of the NumPy limit code explains the 10$\times$ gap to the C reference: the plateau solves are negligible (a few milliseconds of the 0.6 s on camera at 512$^2$); on 8-bit costs the time is the whole-image passes repeated at each of the \textasciitilde180 cost levels (union-find compression, a sparse graph and its connected components, \texttt{np.unique}), and on continuous learned costs quantised to 2$^{-20}$ (246 000 plateaus on a 256$^2$ graph) it is the Python edge loop of the scalar union-find. \path{pw_fast.py} compiles exactly that bookkeeping with Numba (one sorted-edge pass, path-compressed union-find over merged nodes, the components of each level in the merged graph, the merges of the unambiguous components) and hands the ambiguous plateaus back to the same NumPy/SciPy solve code as before: the same dense solve per component on levels of at most 64 edges, the same single block LU on larger levels, the same unknown ordering, the same accumulation order of the right-hand sides and the same representatives of merged components. The results are therefore identical to the bit, ties at exactly 0.5 included, which \path{experiments/pwref/validate_fast.py} checks with \path{np.array_equal}: 400 random coarse 24$^2$ cases (two labels and 3--5 labels, with the small-plateau path on and off), the ten 512$^2$ cases of section 1 (scalar and as a two-label K-vector solve), and the learned-cost regime on the label-augmented 256$^2$ graph at three quantisations (scalar, and 40 markers): 0 mismatches on the Mac (Python 3.14, Numba 0.68) and on jacobian (3.12), the reference-harness records of section 12 are unchanged under the new back end, and the self-tests of \path{pw_layer.py} print the same limit lines under both. \path{power_watershed} and \path{power_watershed_multi} pick the back end automatically (\texttt{backend=None}: Numba when importable, else NumPy; \texttt{PW\_BACKEND=numpy\textbar{}numba} overrides), so \path{pw_limit_numpy} and \path{pw_limit_multi_numpy} and every evaluation script get it without change; the first call per process compiles for about 1.5 s and the compiled code is cached on disk.

\begin{center}\scriptsize\setlength{\tabcolsep}{3pt}
\resizebox{\ifdim\width>\textwidth\textwidth\else\width\fi}{!}{%
\begin{tabular}{@{}llll@{}}
\toprule
regime & NumPy & Numba & reference C (section 12) \\
\midrule

8-bit costs, 512$^2$, clean synthetic & 0.40--0.53 s & 0.065--0.08 s & 0.05--0.06 s \\
8-bit costs, 512$^2$, noisy / camera / astronaut & 0.55--0.8 s & 0.075--0.12 s & 0.05--0.07 s \\
continuous costs at 2$^{-20}$, 256$^2$ augmented graph, two labels & 0.70 s & 0.048 s & --- \\
same, 40 markers (K-vector solve) & 0.31 s & 0.032 s & --- \\
continuous costs at 2$^{-8}$ (many large plateaus), two labels / 40 markers & 0.77 / 0.99 s & 0.08 / 0.08 s & --- \\
\bottomrule
\end{tabular}}
\end{center}

(Mac timings; jacobian is 20--30 \% slower on both.) What remains is dominated by the argsort of the edges and by the solves themselves, so the limit now costs about what the authors\textquotesingle{} C code costs, and a 256$^2$ learned-cost patch goes from a second to 50 ms: cheap enough to be an inference-time operator, or to evaluate the limit on a whole fold in a minute instead of in a process pool. The NumPy version stays as the fallback and as the definition.

\subsection{The exact limit on all of fold 2: was the 307-patch subset representative?}

Every limit row of sections 6--11 and of the paper\textquotesingle s Table 2 was computed on 307 of the 2523 test patches (300 evenly spaced through fold 2 plus the 8 patches with the most touching-nuclei pixels), because the NumPy limit cost about 2 s (binary) and 0.7--1 s (K labels) per patch. With the Numba back end of section 12.1 the whole fold costs a few minutes, so the limit was re-evaluated on all 2523 patches for the five trained models whose limit appears in the paper: the binary \path{rw_img} and \path{rw_learned} of section 6 (Table 2), and the multi-marker \path{bce_tuned} (section 11 model with the fold-3 marker settings of 11.1), \texttt{dist} (11.1) and \path{dist_sched10_bg} (11.5) (Table 3). Code: \path{experiments/run_limit_full.sh} (one evaluation-only job per model, run on jacobian from the saved \texttt{.pt} files; the three instance jobs in parallel), with three new options of \path{exp_unet_seg.py} / \path{exp_instance_pannuke.py}: \texttt{-\/-eval-tag\ limit\_full} writes \texttt{\textless{}name\textgreater{}\_limit\_full.json} / \path{_perimage.npz} next to the existing record instead of overwriting it, \texttt{-\/-mk-from-json\ \textless{}json\textgreater{}} takes the marker post-processing (head, threshold, minimum area, h, fallback) from the stored run instead of re-tuning on fold 3, and \texttt{-\/-subset-n\ 300} summarises every output on the standard 307-patch subset as well (\path{test_subset} of the json), so that the subset and the full-fold numbers of one output come from the same run. \texttt{-\/-limit-n} at or above the test-set size means every patch. Both scripts print the back end they use (\texttt{PW\ limit\ back\ end:\ numba} in all five logs). \path{experiments/compare_limit_full.py} (NumPy only) builds the tables below from the records: \path{results/limit_full_12_2.md}; records in \texttt{results/pannuke/\ \{rw\_img,rw\_learned\}\_limit\_full.*}, \texttt{results/pannuke\_instances/\{bce\_tuned,dist,dist\_sched10\_bg\}/\ instances\_limit\_full.*}, logs in \path{results/limit_full_logs/}.

\textbf{Checks.} On the same 307 patches the Numba limit reproduces the stored limit summaries of the four models that were evaluated on jacobian to 1e-16 on all seven metrics (bit-identical labels, as section 12.1 claimed), and the re-evaluated finite-p outputs reproduce the stored full-fold numbers exactly. \path{dist_sched10_bg} was evaluated on ruche\textquotesingle s A100 in section 11.5 and here on the RTX 6000 Ada: its summaries differ by up to 1.5e-4 (limit, 307) and 4.2e-4 (layer, 2523), the floating-point difference of the forward pass between the two GPUs, which moves a few markers and cost ties.

\textbf{Results on fold 2.} For every model: the finite-p output the report and the paper quote (the 20-step layer at p = 50 for the binary models, the converged solve at p = 20 from the predicted markers for the instance models) and the exact limit of the same costs (and markers), each on the 307-patch subset and on all 2523 patches. Instance metrics on the labels of the output (the 4-connected components of the thresholded output for the binary models).

\begin{center}\scriptsize\setlength{\tabcolsep}{3pt}
\resizebox{\ifdim\width>\textwidth\textwidth\else\width\fi}{!}{%
\begin{tabular}{@{}llllllllll@{}}
\toprule
model & output & patches & IoU & BF$_2$ & edge al. & \href{mailto:F1@0.5}{\nolinkurl{F1@0.5}} & PQ & AJI & count err. \\
\midrule

\path{rw_img} (sec. 6; Table 2) & WS layer, 20 steps (p = 50) & 307 & 0.681 & 0.709 & 0.552 & 0.496 & 0.382 & 0.325 & 0.564 \\
& & 2523 & 0.683 & 0.714 & 0.559 & 0.516 & 0.400 & 0.396 & 0.513 \\
& exact limit & 307 & 0.181 & 0.254 & 0.796 & 0.194 & 0.150 & 0.109 & 0.675 \\
& & 2523 & 0.174 & 0.251 & 0.804 & 0.209 & 0.163 & 0.128 & 0.676 \\
\path{rw_learned} (sec. 6; Table 2) & WS layer, 20 steps (p = 50) & 307 & 0.704 & 0.745 & 0.571 & 0.581 & 0.460 & 0.397 & 0.336 \\
& & 2523 & 0.709 & 0.752 & 0.579 & 0.604 & 0.482 & 0.456 & 0.337 \\
& exact limit & 307 & 0.661 & 0.711 & 0.577 & 0.618 & 0.490 & 0.425 & 0.246 \\
& & 2523 & 0.660 & 0.714 & 0.583 & 0.643 & 0.512 & 0.477 & 0.253 \\
\path{bce_tuned} (sec. 11 / 11.1; Table 3) & WS layer, converged (p = 20) & 307 & 0.703 & 0.754 & 0.580 & 0.746 & 0.607 & 0.623 & 0.197 \\
& & 2523 & 0.708 & 0.761 & 0.587 & 0.769 & 0.630 & 0.645 & 0.208 \\
& exact limit & 307 & 0.692 & 0.736 & 0.578 & 0.742 & 0.594 & 0.608 & 0.197 \\
& & 2523 & 0.696 & 0.743 & 0.586 & 0.765 & 0.617 & 0.630 & 0.207 \\
\texttt{dist} (sec. 11.1; Table 3) & WS layer, converged (p = 20) & 307 & 0.697 & 0.748 & 0.575 & 0.766 & 0.618 & 0.629 & 0.255 \\
& & 2523 & 0.707 & 0.762 & 0.580 & 0.791 & 0.641 & 0.653 & 0.239 \\
& exact limit & 307 & 0.683 & 0.728 & 0.574 & 0.758 & 0.603 & 0.611 & 0.255 \\
& & 2523 & 0.691 & 0.741 & 0.579 & 0.784 & 0.625 & 0.634 & 0.238 \\
\path{dist_sched10_bg} (sec. 11.5; Table 3) & WS layer, converged (p = 20) & 307 & 0.705 & 0.763 & 0.581 & 0.771 & 0.626 & 0.642 & 0.184 \\
& & 2523 & 0.715 & 0.775 & 0.588 & 0.795 & 0.649 & 0.668 & 0.182 \\
& exact limit & 307 & 0.625 & 0.691 & 0.622 & 0.700 & 0.564 & 0.554 & 0.286 \\
& & 2523 & 0.630 & 0.697 & 0.625 & 0.724 & 0.585 & 0.578 & 0.274 \\
\bottomrule
\end{tabular}}
\end{center}

\textbf{Paired limit $-$ finite-p differences} on the same patches (bootstrap 95 \% over patches; F1, PQ and AJI dataset-level, recomputed per resample):

\begin{center}\scriptsize\setlength{\tabcolsep}{3pt}
\resizebox{\ifdim\width>\textwidth\textwidth\else\width\fi}{!}{%
\begin{tabular}{@{}llllllll@{}}
\toprule
model & patches & IoU & BF$_2$ & \href{mailto:F1@0.5}{\nolinkurl{F1@0.5}} & PQ & AJI & count err. \\
\midrule

\path{rw_img} & 307 & $-$0.500 {[}$-$0.524, $-$0.472{]} & $-$0.455 {[}$-$0.480, $-$0.428{]} & $-$0.302 {[}$-$0.332, $-$0.272{]} & $-$0.232 {[}$-$0.255, $-$0.209{]} & $-$0.216 {[}$-$0.294, $-$0.149{]} & +0.111 {[}$-$0.107, +0.283{]} \\
& 2523 & $-$0.509 {[}$-$0.520, $-$0.498{]} & $-$0.464 {[}$-$0.474, $-$0.453{]} & $-$0.307 {[}$-$0.318, $-$0.296{]} & $-$0.237 {[}$-$0.246, $-$0.229{]} & $-$0.268 {[}$-$0.289, $-$0.246{]} & +0.163 {[}+0.107, +0.213{]} \\
\path{rw_learned} & 307 & $-$0.043 {[}$-$0.056, $-$0.031{]} & $-$0.034 {[}$-$0.045, $-$0.022{]} & +0.037 {[}+0.017, +0.058{]} & +0.029 {[}+0.012, +0.047{]} & +0.028 {[}$-$0.017, +0.073{]} & $-$0.090 {[}$-$0.165, $-$0.027{]} \\
& 2523 & $-$0.050 {[}$-$0.055, $-$0.044{]} & $-$0.038 {[}$-$0.044, $-$0.032{]} & +0.039 {[}+0.031, +0.047{]} & +0.031 {[}+0.024, +0.037{]} & +0.021 {[}+0.003, +0.038{]} & $-$0.083 {[}$-$0.114, $-$0.053{]} \\
\path{bce_tuned} & 307 & $-$0.011 {[}$-$0.015, $-$0.007{]} & $-$0.017 {[}$-$0.022, $-$0.013{]} & $-$0.004 {[}$-$0.012, +0.004{]} & $-$0.013 {[}$-$0.019, $-$0.006{]} & $-$0.015 {[}$-$0.018, $-$0.012{]} & +0.000 {[}$-$0.001, +0.002{]} \\
& 2523 & $-$0.012 {[}$-$0.017, $-$0.007{]} & $-$0.018 {[}$-$0.023, $-$0.013{]} & $-$0.004 {[}$-$0.010, +0.003{]} & $-$0.013 {[}$-$0.019, $-$0.007{]} & $-$0.015 {[}$-$0.020, $-$0.011{]} & $-$0.001 {[}$-$0.016, +0.015{]} \\
\texttt{dist} & 307 & $-$0.014 {[}$-$0.016, $-$0.012{]} & $-$0.020 {[}$-$0.023, $-$0.017{]} & $-$0.008 {[}$-$0.011, $-$0.005{]} & $-$0.016 {[}$-$0.018, $-$0.014{]} & $-$0.018 {[}$-$0.020, $-$0.016{]} & $-$0.000 {[}$-$0.001, +0.001{]} \\
& 2523 & $-$0.015 {[}$-$0.020, $-$0.010{]} & $-$0.021 {[}$-$0.026, $-$0.016{]} & $-$0.006 {[}$-$0.011, $-$0.002{]} & $-$0.016 {[}$-$0.021, $-$0.012{]} & $-$0.018 {[}$-$0.022, $-$0.014{]} & $-$0.001 {[}$-$0.013, +0.012{]} \\
\path{dist_sched10_bg} & 307 & $-$0.080 {[}$-$0.094, $-$0.066{]} & $-$0.072 {[}$-$0.086, $-$0.058{]} & $-$0.071 {[}$-$0.083, $-$0.059{]} & $-$0.062 {[}$-$0.072, $-$0.053{]} & $-$0.088 {[}$-$0.098, $-$0.078{]} & +0.102 {[}+0.070, +0.130{]} \\
& 2523 & $-$0.086 {[}$-$0.092, $-$0.079{]} & $-$0.078 {[}$-$0.085, $-$0.072{]} & $-$0.071 {[}$-$0.077, $-$0.066{]} & $-$0.064 {[}$-$0.069, $-$0.059{]} & $-$0.090 {[}$-$0.096, $-$0.084{]} & +0.093 {[}+0.076, +0.110{]} \\
\bottomrule
\end{tabular}}
\end{center}

\begin{itemize}

\item
  \textbf{The subset was representative of the comparisons, not of the absolute instance numbers.} The limit\textquotesingle s pixel-level scores are the same on the subset and on the fold (IoU within 0.008, BF$_2$ within 0.013, edge alignment within 0.008 for every model), but its instance metrics are 0.02--0.03 higher on the full fold for every model with learned costs (F1 +0.025 / +0.023 / +0.026 / +0.024 and PQ +0.022 / +0.023 / +0.022 / +0.021 for \path{rw_learned}, \path{bce_tuned}, \texttt{dist}, \path{dist_sched10_bg}; AJI +0.052 for \path{rw_learned}, +0.022--0.024 for the others). The reason is the subset, not the limit: the finite-p outputs of the same models lose the same amount on the 307 patches (layer F1 0.581 against 0.604 for \path{rw_learned}, 0.766 against 0.791 for \texttt{dist}), because the 8 patches with the most touching-nuclei pixels, all crowded Colon, were added to the 300 evenly spaced ones and weigh heavily in the dataset-level counts. Every limit number quoted so far as an absolute instance score (F1, PQ, AJI) is therefore about 0.02 too low; the pixel-level numbers and the count errors stand.
\item
  \textbf{Every paired conclusion stands, with intervals two to three times narrower.} Limit minus finite-p on the 307 and on the 2523 patches agree within the subset\textquotesingle s intervals for every model and metric: \path{rw_learned} $-$0.043 $\to$ $-$0.050 IoU, +0.037 $\to$ +0.039 F1, +0.029 $\to$ +0.031 PQ; \path{bce_tuned} $-$0.011 $\to$ $-$0.012 IoU, $-$0.004 $\to$ $-$0.004 F1; \texttt{dist} $-$0.014 $\to$ $-$0.015 IoU, $-$0.008 $\to$ $-$0.006 F1, $-$0.016 $\to$ $-$0.016 PQ; \path{dist_sched10_bg} $-$0.080 $\to$ $-$0.086 IoU, $-$0.071 $\to$ $-$0.071 F1, $-$0.062 $\to$ $-$0.064 PQ. Section 6\textquotesingle s reading of the binary limit as the best instance segmenter of the RW outputs is now stronger: F1 0.643 / PQ 0.512 against the WS layer\textquotesingle s 0.604 / 0.482, and the AJI advantage (+0.021 {[}+0.003, +0.038{]}) is clear of zero where the subset\textquotesingle s was not; its count error on the fold is 0.253 (0.246 on the subset), still the lowest of Table 2\textquotesingle s rows. Section 11.1\textquotesingle s "the limit is within 0.01 F1 of the converged solve" and section 11.5\textquotesingle s "the limit of the model trained on its own markers drops by 0.07 F1" are unchanged.
\item
  \textbf{Paper rows (fold 2, all 2523 patches).} Table 2, "its exact limit" of the image-gradient model: 0.174 / 0.251 / 0.804 / 0.209 / 0.163 / 0.676 (IoU / BF$_2$ / edge al. / F1 / PQ / count err.; was 0.181 / 0.254 / 0.796 / 0.194 / 0.150 / 0.675 on 307); of the learned-cost model: 0.660 / 0.714 / 0.583 / 0.643 / 0.512 / 0.253 (was 0.661 / 0.711 / 0.577 / 0.618 / 0.490 / 0.246). Table 3\textquotesingle s column order (IoU / F1 / PQ / AJI / count err.), exact limits: BCE marker head 0.696 / 0.765 / 0.617 / 0.630 / 0.207 (was 0.692 / 0.742 / 0.594 / 0.608 / 0.197); distance-head markers 0.691 / 0.784 / 0.625 / 0.634 / 0.238 (was 0.683 / 0.758 / 0.603 / 0.611 / 0.255); trained from its own predicted markers 0.630 / 0.724 / 0.585 / 0.578 / 0.274 (was 0.625 / 0.700 / 0.564 / 0.554 / 0.286). The "(307 patches)" qualifier can go.
\item
  \textbf{Cost.} The learned-cost K-label limit took 41--59 ms per patch on one core (K = 22--25; 0.68-- 0.96 s with the NumPy back end in sections 11.1--11.5): 140--180 s wall for both the learned and the image-cost limits of all 2523 patches with 4 workers, 61--77 s for the binary limit with 8. The rest of the evaluation is now what costs: the three instance jobs ran in parallel on one GPU and took 2.3--2.8 h each, because the evaluation runs four converged solves per batch (predicted markers, fallback markers, ground-truth markers, image costs) at 0.36--0.42 s per image when three processes share the card (0.094 s alone, section 11.1). On this fold the exact limit has become the cheapest output to compute.
\end{itemize}

\section{Data efficiency: the WS layer and the fair baseline on subsets of fold 1}

Sections 11.2--11.5 end with the WS layer level with its fair baseline on the full training fold (F1 0.795 / PQ 0.649 against 0.794 / 0.655), and a reviewer may read that as "a tuned U-Net and a classical watershed always match the WS layer". The classical case for a structured output layer is not accuracy at full data but data efficiency: the solve carries a prior (a connected, marker-labelled partition with a soft boundary) that the network would otherwise have to learn, so the structured model should degrade more slowly as the training set shrinks. This section measures it. Code: \texttt{-\/-train-frac\ F\ -\/-subset-seed\ S} in \path{experiments/exp_instance_pannuke.py} and \path{exp_unet_dist.py} (a random subset of the training fold, taken as the first round(F $\cdot$ N) entries of one seeded permutation, so the 10 \% subset is contained in the 25 \%, which is contained in the 50 \%; independent of \texttt{-\/-seed}, which still drives the initialisation and the batch order), the cases \path{dist_bg_p10\ /\ p25\ /\ p50} and \path{unet_dist_p10\ /\ p25\ /\ p50} of \path{ruche_run.sh} and the new \path{run_data_efficiency.sh} (jacobian), and \path{plot_data_efficiency.py} (tables with numpy only, the figure with matplotlib). Results: \path{results/pannuke_data_efficiency/} (\path{data_efficiency.md} with every table of this section and the paired bootstraps, \path{data_efficiency.json}, \path{data_efficiency.png}, one folder per run with its json, per-image metrics and log; the \texttt{.pt} files stay on ruche / jacobian).

\textbf{What was run.} The WS layer is the section 11.5 recipe exactly (\texttt{-\/-mk-head\ dist\ -\/-w-mk\ 2\ -\/-mk-sched\ -\/-mk-unmarked\ bg}, p 5 $\to$ 20, edge target, seed 0, Adam 1e-3 with cosine annealing, batch 16, d4 augmentation) and the baseline the section 11.2 U-Net with the distance head and the HoVer-Net-style marker watershed, its post-processing chosen by PQ on the same 300 patches of fold 3 (135 settings) for every run, never on fold 2; the WS layer\textquotesingle s marker thresholds likewise. Both were trained on the same nested random subsets of fold 1 (subset seed 0): 266 images (10 \%), 664 (25 \%), 1328 (50 \%); the 100 \% points are the existing runs \path{dist_sched10_bg} and \path{unet_dist} (2656 images). The epoch count was scaled inversely to the fraction (300 / 120 / 60 / 30 epochs) so that every run makes the same \textasciitilde4980 gradient steps as the 30-epoch full run and sees the same number of batches: the question is what the data buy, not what the compute buys, and a 10 \% run stopped after 30 epochs would confound the two. The marker schedule and the validation cadence were scaled alike (\texttt{-\/-mk-sched\ 100\ 150\ /\ 40\ 60\ /\ 20\ 30}, \texttt{-\/-val-every\ 50\ /\ 20\ /\ 10}), so the predicted markers come in after a third of the training in every run, as in section 11.5. All runs were evaluated on all 2523 patches of fold 2 with the usual metrics; the baseline\textquotesingle s \texttt{logits} output (foreground \textgreater{} 0.5, 4-connected components: the plain U-Net pipeline) comes free and gives the third curve.

\textbf{Results on fold 2} (all 2523 patches; the WS layer\textquotesingle s main output and the baseline\textquotesingle s tuned watershed, the plain components in the last row; figure \path{results/pannuke_data_efficiency/data_efficiency.png}, PQ and F1 against the training-set size):

\begin{center}\scriptsize\setlength{\tabcolsep}{3pt}
\resizebox{\ifdim\width>\textwidth\textwidth\else\width\fi}{!}{%
\begin{tabular}{@{}llllllllll@{}}
\toprule
method & train \% (images) & IoU & BF$_2$ & \href{mailto:F1@0.5}{\nolinkurl{F1@0.5}} & PQ & SQ & AJI & count err. & markers one / none, bg per patch \\
\midrule

layer (11.5 recipe) & 10 (266) & 0.681 & 0.746 & 0.765 & 0.620 & 0.811 & 0.635 & 0.238 & 84.0 / 13.9 \%, 4.3 \\
layer & 25 (664) & 0.706 & 0.769 & 0.785 & 0.640 & 0.815 & 0.659 & 0.185 & 84.8 / 13.5 \%, 4.4 \\
layer & 50 (1328) & 0.711 & 0.772 & 0.787 & 0.641 & 0.814 & 0.663 & 0.184 & 86.2 / 11.7 \%, 4.9 \\
layer (\path{dist_sched10_bg}) & 100 (2656) & 0.715 & 0.775 & 0.795 & 0.649 & 0.816 & 0.668 & 0.182 & 85.5 / 12.7 \%, 4.3 \\
U-Net + dist head + watershed (11.2) & 10 (266) & 0.670 & 0.722 & 0.745 & 0.605 & 0.813 & 0.620 & 0.237 & 73.8 / 24.5 \%, 2.3 \\
U-Net + watershed & 25 (664) & 0.714 & 0.768 & 0.789 & 0.646 & 0.819 & 0.662 & 0.185 & 80.5 / 17.6 \%, 2.5 \\
U-Net + watershed & 50 (1328) & 0.720 & 0.772 & 0.794 & 0.653 & 0.822 & 0.666 & 0.179 & 78.4 / 20.0 \%, 2.0 \\
U-Net + watershed (\path{unet_dist}) & 100 (2656) & 0.719 & 0.774 & 0.794 & 0.655 & 0.824 & 0.658 & 0.190 & 76.6 / 21.8 \%, 1.6 \\
U-Net + connected components & 10 / 25 / 50 / 100 & 0.672 / 0.713 / 0.721 / 0.725 & 0.725 / 0.766 / 0.774 / 0.777 & 0.622 / 0.653 / 0.659 / 0.664 & 0.500 / 0.530 / 0.536 / 0.539 & -- & 0.509 / 0.500 / 0.511 / 0.501 & 0.387 / 0.328 / 0.319 / 0.313 & -- \\
\bottomrule
\end{tabular}}
\end{center}

Paired on fold 2, layer $-$ watershed baseline (IoU / BF$_2$ / F1 / PQ / AJI, bootstrap 95 \% intervals in \path{data_efficiency.md}): at 10 \% +0.011 {[}+0.007, +0.015{]} / +0.024 {[}+0.019, +0.028{]} / \textbf{+0.020 {[}+0.017, +0.023{]} / +0.015 {[}+0.013, +0.017{]}} / +0.015 {[}+0.013, +0.018{]}; at 25 \% $-$0.008 / +0.001 / $-$0.004 {[}$-$0.007, $-$0.001{]} / $-$0.006 {[}$-$0.008, $-$0.004{]} / $-$0.004; at 50 \% $-$0.009 / $-$0.001 / $-$0.007 {[}$-$0.009, $-$0.004{]} / $-$0.012 {[}$-$0.014, $-$0.010{]} / $-$0.003; at 100 \% $-$0.004 / +0.001 / +0.001 / $-$0.005 / +0.009 (section 11.5). Each method against its own 100 \% run, F1 / PQ: layer $-$0.030 / $-$0.029 at 10 \%, $-$0.010 / $-$0.009 at 25 \%, $-$0.008 / $-$0.008 at 50 \%; baseline $-$0.049 / $-$0.049 at 10 \%, $-$0.005 / $-$0.009 at 25 \%, 0.000 / $-$0.002 at 50 \%. The watershed\textquotesingle s margin over the plain components is +0.12 to +0.14 F1 at every size. Oracles with ground-truth markers (F1 / PQ): layer 0.840 / 0.691, 0.853 / 0.706, 0.863 / 0.713, 0.872 / 0.723 at 10 / 25 / 50 / 100 \%; the baseline\textquotesingle s watershed 0.812 / 0.654, 0.854 / 0.696, 0.867 / 0.707, 0.867 / 0.708. The 10 \% layer run was trained twice, on jacobian (RTX 6000 Ada) and on ruche (A100): F1 0.765 both, PQ 0.621 / 0.620, AJI 0.636 / 0.635, with the fold-3 tuning choosing a minimum area of 4 and 1 px respectively (\path{dist_bg_p10_jacobian/}); the ruche run is the one in the tables, like the 100 \% run.

\begin{itemize}

\item
  \textbf{The WS layer degrades more gracefully, but only at the smallest subset.} On 266 training images it leads the tuned baseline on every metric, +0.020 F1 / +0.015 PQ / +0.015 AJI / +0.011 IoU / +0.024 BF$_2$ with every interval clear of zero, and loses 0.030 F1 / 0.029 PQ against its full-data run where the baseline loses 0.049 / 0.049. The margin is two and a half times the seed-to-seed spread of the \texttt{bg} recipe (0.008 F1, section 11.5) and reproduced on different hardware to 0.001. From 25 \% on the picture of sections 11.2--11.5 is back: the baseline is 0.004--0.007 F1 and 0.006--0.012 PQ ahead, the WS layer\textquotesingle s curve is flatter but sits below it, and the two meet again at 100 \% on F1. A quarter of the fold is enough for the watershed baseline to be within 0.005 F1 of its full-data result; the WS layer is within 0.010. So the honest statement for the paper is that the structured layer buys robustness in the low-data regime (a tenth of PanNuke\textquotesingle s fold, 266 patches), not a steeper learning curve in general: at 25 \% and above a tuned U-Net and a classical watershed do match it.
\item
  \textbf{Where the 10 \% gain comes from: detection under noisy markers, not boundaries.} The matched nuclei are segmented equally well (SQ 0.811 against 0.813); the whole difference is DQ, i.e. F1. The marker columns say why. The baseline\textquotesingle s distance head is weaker at 10 \% and its fold-3 tuning answers by raising the marker threshold to 0.5 (minimum area 9, h 0.3), which leaves 24.5 \% of the nuclei of fold 2 unmarked (21.8 \% at full data): on fold 3 the best setting at the WS layer\textquotesingle s threshold 0.2 marks more nuclei (19 \% unmarked) but costs 0.014 PQ, because every extra marker the watershed is given becomes an instance, and keeping the unmarked components as instances (\path{ws_fb}) is worse still (F1 0.725). The WS layer keeps threshold 0.2 at every size, marks 84--86 \% of the nuclei with 4.3--4.9 background markers per patch, and the solve trained on its own markers (section 11.5) absorbs the spurious ones through the unary costs instead of growing them, so its recall does not collapse with the head. The same reading comes from the oracles: with ground-truth markers the WS layer\textquotesingle s costs are 0.028 F1 better than the watershed\textquotesingle s reliefs at 10 \% (0.840 against 0.812) and equal at 50--100 \% (0.863--0.872 against 0.867): the solve\textquotesingle s prior is worth something to the costs when they are learned from little data, and nothing once they are not.
\item
  \textbf{The epoch scaling did not flatter either side much, but the caveat cuts the WS layer\textquotesingle s way.} With the step count held, the 10 \% baseline over-fits mildly (validation F1 on 200 patches 0.775 at epoch 49, 0.761 at 299), the 10 \% layer does not (0.453 at epoch 49 before its schedule, 0.749 at 249, 0.745 at 299), so an early-stopped baseline could recover part of the 0.020; a 30-epoch baseline at 10 \% (a tenth of the steps) was not run. The WS layer\textquotesingle s 10 \% run cost 1.6 h on an A100 (19 s per epoch, 300 epochs, the extraction of the predicted markers from epoch 100 included) and the baseline\textquotesingle s 0.3 h, so the usual 5--10$\times$ price of the WS layer holds here too.
\item
  \textbf{One seed, one subset draw.} The subsets are nested and shared by the two methods, which removes the subset draw from the paired differences but not from the absolute curves; a second draw and a second seed at 10 \% (two jobs per method, 2 h each) would settle whether 0.020 is the typical margin or an upper draw, and a 5 \% point (133 images) would say whether the gap keeps opening. The replicates were run the same day: see the addendum below.
\end{itemize}

\textbf{Addendum, 24 October 2026 (later the same day): the 10 \% point replicated.} Four more runs at 10 \%, two per method: a second training seed on the same 266 images (\texttt{-\/-seed\ 1}, runs \path{dist_bg_p10_s1} / \path{unet_dist_p10_s1}) and the first seed on a second subset draw (\texttt{-\/-subset-seed\ 1}, runs \path{dist_bg_p10_d1} / \path{unet_dist_p10_d1}; a different 266 images of fold 1); layer runs on ruche, baselines on jacobian, same recipe and tuning as above. Results (F1 / PQ / AJI on fold 2; the full rows and intervals are in \path{data_efficiency.md}):

\begin{center}\scriptsize\setlength{\tabcolsep}{3pt}
\resizebox{\ifdim\width>\textwidth\textwidth\else\width\fi}{!}{%
\begin{tabular}{@{}lllll@{}}
\toprule
10 \% run & WS layer & watershed baseline & WS layer $-$ baseline F1 / PQ (paired) & unmarked nuclei, layer / baseline \\
\midrule

seed 0, draw 0 & 0.765 / 0.620 / 0.635 & 0.745 / 0.605 / 0.620 & +0.020 {[}+0.017, +0.023{]} / +0.015 {[}+0.013, +0.017{]} & 13.9 \% / 24.5 \% \\
seed 1, draw 0 & 0.765 / 0.620 / 0.636 & 0.749 / 0.609 / 0.623 & +0.015 {[}+0.012, +0.018{]} / +0.011 {[}+0.009, +0.014{]} & 15.8 \% / 23.7 \% \\
seed 0, draw 1 & 0.765 / 0.624 / 0.633 & 0.749 / 0.612 / 0.624 & +0.017 {[}+0.014, +0.020{]} / +0.012 {[}+0.010, +0.014{]} & 15.0 \% / 24.4 \% \\
seed 0, draw 0, jacobian & 0.765 / 0.621 / 0.636 & -- & -- & 15.2 \% / -- \\
\bottomrule
\end{tabular}}
\end{center}

The margin holds: +0.015 to +0.020 F1 and +0.011 to +0.015 PQ across the three seed / draw pairs, every interval clear of zero, with the WS layer at F1 0.765 in all four runs and the baseline at 0.745--0.749. The spread between replicates is 0.004 F1 for the baseline and 0.000 for the WS layer, so the 0.020 of the first pair was the upper end of a stable 0.017 $\pm$ 0.003, not a lucky draw. The mechanism reads the same in every replicate: the baseline\textquotesingle s fold-3 tuning picks threshold 0.5 each time and leaves 24 \% of the nuclei unmarked, the WS layer\textquotesingle s picks 0.2 and leaves 14--16 \%; the segmentation of the matched nuclei is equal (SQ 0.811--0.815 against 0.812--0.817); the oracles with ground-truth markers are 0.839--0.842 for the WS layer against 0.812--0.817 for the watershed. The baseline\textquotesingle s mild over-fitting under the held step count is not systematic either (validation F1 at epoch 49 $\to$ 299: 0.775 $\to$ 0.761, 0.750 $\to$ 0.760, 0.774 $\to$ 0.771), so early stopping would not close the gap. The conclusion of the section is unchanged and now rests on three pairs: at a tenth of the fold the structured layer is a small, reliable step ahead of the tuned watershed baseline; from a quarter on it is not.

\section{Marker selection in the gradient path: candidate strengths trained through the solve}

Added 24--25 October 2026 (side session; three training jobs on ruche\textquotesingle s A100 partition, one of them stopped). Section 11.5 left the WS layer 0.001--0.008 F1 / 0.005--0.015 PQ behind the fair baseline over seeds and folds, with the diagnosis (discussion with one of us Talbot, 24 October) that the gradient through the solve reaches the edge costs and the marker attachments but not the marker \emph{selection}: the markers are the h-maxima of the distance map above a threshold, a non-differentiable decision, so the end-to-end training can only improve what the baseline already does better per matched nucleus (SQ). Section 11.4 had also found that the WS layer pays for every spurious marker it is given, where the watershed lets a weak one vanish. This section puts the selection in the gradient path and asks whether the partition cross-entropy can then learn to weaken spurious markers. Code: \texttt{-\/-diff-sel} in \path{experiments/exp_instance_pannuke.py} (\path{candidate_strengths}, \path{keep_strong}, \path{candidate_classes}, \path{strength_stats}, \path{diffsel_selftest}; \path{marker_target_map} / \path{mixed_label_systems} gained the \texttt{primary} rule; new output \path{layer_all}), the cases \path{dist_ds}, \path{dist_ds_max}, \path{dist_ds_p10} of \path{ruche_run.sh}, and \path{experiments/candidate_auc.py} (how well a marker head ranks its own candidates). Results: \path{results/pannuke_instances/dist_ds/} (json, per-image metrics, log, the gradient check \path{ds_selftest.txt}, the candidate rankings \path{candidate_auc_*.md}), \path{dist_ds_max/run.log} (the stopped first run), \path{results/pannuke_data_efficiency/dist_ds_p10/}, and the paired tables \path{compare_14.md} and \path{compare_14_p10.md}; the \texttt{.pt} files are on ruche (and the full-fold one on jacobian).

\textbf{The variant.} From the scheduled epochs on (\texttt{-\/-mk-sched\ 10\ 15}, as in 11.5) the training markers are not the fold-3 markers of section 11.1 but permissive candidates: every h-maximum of the predicted distance map with prominence 0.1 above 0.1, grown down to 0.1, area 1 (\texttt{-\/-mk-train-thr\ 0.1\ -\/-mk-train-h\ 0.1\ -\/-mk-train-min-area\ 1}; about 40 candidates per training image, of which 17--21 in background, against 25 markers per patch in 11.5). Every candidate k gets a strength $\sigma$\_k = sigmoid((z\_k $-$ $\theta$) / T), z\_k a score of the marker logit over the candidate and $\theta$, T two learned scalars of the model (initialised at $-$1 and 1), and the strength multiplies the foreground probability inside the candidate\textquotesingle s attachment:

\begin{verbatim}
a_k,i = ($\sigma$_k s_i)^p   on the candidate's pixels,   i.e.   c1_i $\to$ c1_i $-$ log $\sigma$_k.
\end{verbatim}

The strength enters in cost units, like every other cost of the WS layer. A plain multiplier $\sigma$\_k on a\_k would shift the cost by only $-$log $\sigma$\_k / p, which at p = 20 cannot veto a marker inside a nucleus, where the background attachment (1 $-$ s)\^{}p is about e\^{}$-$160. The cross-entropy target of a candidate follows the owner rule of 11.5 (the nucleus holding the majority of its pixels) with one change, \texttt{-\/-ds-multi\ primary}: of the candidates a nucleus owns, only the one with the largest overlap is its target. A duplicate is then wrong wherever it wins and is pushed towards zero strength, exactly like a candidate in background, instead of sharing the nucleus as the 11.5 sum rule allowed. Unmarked nuclei go to the background label (\texttt{-\/-mk-unmarked\ bg}), the ground-truth markers of the early epochs carry no strength ($\sigma$ = 1), and the rest of the recipe is 11.5\textquotesingle s (\texttt{-\/-mk-head\ dist\ -\/-w-mk\ 2}, p 5 $\to$ 20, edge target, 30 epochs, seed 0, fold 1 $\to$ fold 2). At test time the candidates are extracted the same way, those with $\sigma$\_k below a cut $\tau$ are dropped and the others enter the solve with their strengths; fold 3 chooses $\tau$ $\in$ \{0, 0.1, 0.3, 0.5, 0.7, 0.9\} together with the candidate threshold and prominence (thr $\in$ \{0.1, 0.2\} $\times$ h $\in$ \{0.1, 0.2\}, area 1; 24 settings), and \path{layer_all} keeps every candidate.

\textbf{Gradient check.} \texttt{-\/-ds-selftest} runs the training loss (mixed systems with the primary rule and the background target, seam-like weights, p = 8) in float64 on two 10 $\times$ 12 images with two nuclei each and six candidates (primaries, a duplicate, two in background, one unmarked nucleus), and compares the implicit adjoint with central finite differences ($\varepsilon$ = 1e-6, solve to a relative residual of 1e-12): d/d$\theta$, d/d log T, d/d(marker logit) at every pixel, and two random directions in each of the edge costs, c0, c1 and the marker logit. Worst relative error 6.0e-7 for the mean score, 3.0e-8 for the maximum (\path{ds_selftest.txt}). The signs are the intended ones: the background candidates get dL/dz \textgreater{} 0 (weaken), the primary of the image whose nucleus is contested gets dL/dz \textless{} 0. A candidate that wins or loses its pixels outright gets a gradient of 1e-11 or less, because the Random Walker probabilities are saturated in its attachment. Checked on the trained 11.5 model before launching (32 training patches, p = 20, $\theta$ = $-$1, T = 1): every one of the 227 background candidates and the 39 duplicates is pushed weaker, 83 \% of the 582 primaries stronger; the 64 \% of background candidates with a negligible gradient are those that win no pixel, so they cost nothing.

\textbf{First run: the peak score fragments the marker map} (\path{dist_ds_max}, z\_k = maximum of the logit over the candidate, the candidate of the brief). From the first scheduled epoch the candidates multiplied: background candidates per training image 21 $\to$ 63 $\to$ 123 over epochs 10--12, duplicates per epoch 1.5 k $\to$ 13 k $\to$ 47 k, and the epoch time 221 $\to$ 371 $\to$ 723 s. The job was stopped at epoch 12; at that rate it would not have finished training within the 12-hour job. The cause is the needle failure of section 11.3\textquotesingle s raw-maximum peak loss, mirrored: the whole gradient of a candidate goes to its arg-max pixel, and pushing that one pixel down splits the peak into neighbouring maxima, each a new candidate. The run reported below uses the mean of the logit over the candidate (\texttt{-\/-ds-peak\ mean}, the default): the gradient shifts the whole candidate uniformly, which cannot create a maximum inside it, and the push per pixel is \textbar R\_k\textbar{} times smaller. With it the candidate counts stayed flat (17--21 background candidates per image from epoch 10 to 29), at 320--340 s per epoch against 189 s before the schedule (17 s of candidate extraction, more systems per batch).

\textbf{Training.} The strengths separate the classes only weakly and the separation does not grow: on the training batches the mean strength of background / primary / duplicate candidates is 0.40 / 0.55 / 0.47 at epoch 10 and 0.58 / 0.66 / 0.63 at epoch 29, with $\theta$ drifting from $-$1.0 to $-$1.4 (every strength rises: a primary at $\sigma$ \textless{} 1 loses its rim to the background attachment) and T from 1.0 to 1.2--1.4. The scale-free measure, the AUC of the strength for primary against background candidates ($\sigma$ is monotone in z\_k, so this is the marker head\textquotesingle s ranking and $\theta$, T cannot change it), goes 0.889 $\to$ 0.907 (epoch 12) $\to$ 0.849 (epoch 19) $\to$ 0.876 (epoch 29); against duplicates 0.74 $\to$ 0.69. Meanwhile the marker head gets worse at its own target and as a detector: its MSE against the distance target rises from 0.0104 (epoch 9) to 0.0219 (11.5 run: 0.0103 $\to$ 0.0082 over the same epochs), and on the training batches the nuclei with no candidate go from 11 \% at epoch 10 to 24 \% at epoch 29 (11.5 run: 18 \% $\to$ 13 \%). The cross-entropy sits at 0.20 against 0.15 for the 11.5 run.

\textbf{Results on fold 2} (all 2523 patches; the limit on the 307; the two references of 11.5 and 11.2 repeated):

\begin{center}\scriptsize\setlength{\tabcolsep}{3pt}
\resizebox{\ifdim\width>\textwidth\textwidth\else\width\fi}{!}{%
\begin{tabular}{@{}lllllllllll@{}}
\toprule
run & \href{mailto:F1@0.5}{\nolinkurl{F1@0.5}} & PQ & SQ & DQ & AJI & IoU & count err. & markers one / none / several, bg per patch (markers per patch) & oracle F1 / PQ & limit F1 / PQ (307) \\
\midrule

\path{dist_sched10_bg} (11.5) & \textbf{0.795} & 0.649 & 0.816 & \textbf{0.795} & \textbf{0.668} & 0.715 & \textbf{0.182} & 85.5 / 12.7 / 1.8 \%, 4.3 (25.4) & 0.872 / 0.723 & 0.700 / 0.564 \\
\path{unet_dist} (11.2, watershed) & 0.794 & \textbf{0.655} & \textbf{0.824} & 0.794 & 0.658 & \textbf{0.719} & 0.190 & 76.6 / 21.8 / 1.6 \%, 1.6 (20.6) & 0.867 / 0.708 & -- \\
\path{dist_ds}: strengths trained through the solve, $\tau$ = 0.3 (fold 3) & 0.732 & 0.601 & 0.821 & 0.732 & 0.628 & 0.673 & 0.223 & 72.1 / 26.1 / 1.9 \%, 12.6 (30.6) & 0.931 / 0.756 & 0.115 / 0.100 \\
\path{dist_ds}, every candidate (\path{layer_all}) & 0.732 & 0.601 & 0.821 & 0.732 & 0.628 & 0.673 & 0.223 & the same & -- & -- \\
\bottomrule
\end{tabular}}
\end{center}

Paired on fold 2 (bootstrap 95 \% intervals, \path{compare_14.md}), \path{dist_ds} $-$ \path{dist_sched10_bg}: F1 $-$0.063 {[}$-$0.068, $-$0.060{]}, PQ $-$0.048 {[}$-$0.051, $-$0.045{]}, AJI $-$0.039 {[}$-$0.042, $-$0.037{]}, IoU $-$0.042 {[}$-$0.046, $-$0.038{]}, BF$_2$ $-$0.038 {[}$-$0.042, $-$0.034{]}, edge alignment +0.019 {[}+0.017, +0.020{]}; \path{dist_ds} $-$ \path{unet_dist}: F1 $-$0.062 {[}$-$0.065, $-$0.058{]}, PQ $-$0.054 {[}$-$0.056, $-$0.051{]}, AJI $-$0.030 {[}$-$0.033, $-$0.027{]}, IoU $-$0.046 {[}$-$0.049, $-$0.042{]}, BF$_2$ $-$0.037 {[}$-$0.041, $-$0.034{]}. Fold 3 says the same before fold 2 is seen: the best setting (thr 0.2, h 0.2, $\tau$ $\le$ 0.5) reaches PQ 0.612 / F1 0.743 on the 300 tuning patches, against 0.655 / 0.797 for \path{dist_sched10_bg} at the same marker setting. The candidate rankings on all of fold 2 (\path{candidate_auc_*.md}, candidates at thr 0.1 / h 0.1, mean score; primary / background / duplicate candidates per patch, AUC primary vs background, vs duplicate): \path{dist_sched10_bg} 21.5 / 6.6 / 1.2, 0.867, 0.707; \path{unet_dist} 21.8 / 6.4 / 1.3, 0.874, 0.715; \path{dist_ds} 17.7 / 19.9 / 1.6, 0.876, 0.666. The fold-2 evaluation solve hit its 1000-iteration cap (993 PCG iterations per batch on average against 254 for \path{dist_sched10_bg}, 0.41 s per image against 0.09 s): the strength-shifted attachments ($\sigma$\^{}p $\approx$ e\^{}$-$9 for a typical background candidate) make the systems ill-conditioned. The labels are converged all the same: on the first 160 fold-2 patches a 5000-iteration cap (3.3 s per image, still short of the batch-wide 1e-5 residual) gives F1 / PQ / IoU identical to four digits to the 1000-iteration evaluation (0.7377 / 0.6057 / 0.6552, oracle 0.9321 / 0.7563 / 0.8077).

\textbf{The 10 \% regime} (\path{dist_ds_p10}: the same variant on the 266-image subset of section 13, 300 epochs, schedule from epoch 100; fold 3 chose thr 0.2 / h 0.2 / $\tau$ 0.3, PQ 0.631 against 0.629 for \path{dist_bg_p10}): F1 0.764 / PQ 0.621 / SQ 0.813 / AJI 0.630 / IoU 0.678, count error 0.238, markers 79.9 / 17.4 / 2.7 \%, 10.3 in background per patch, against \path{dist_bg_p10} 0.765 / 0.620 / 0.811 / 0.635 / 0.681, 0.238, 84.0 / 13.9 / 2.1 \%, 4.3 and \path{unet_dist_p10} 0.745 / 0.605 / 0.813 / 0.620 / 0.670. Paired (\path{compare_14_p10.md}): \path{dist_ds_p10} $-$ \path{dist_bg_p10} F1 $-$0.001 {[}$-$0.004, +0.001{]}, PQ 0.000 {[}$-$0.002, +0.002{]}, AJI $-$0.006 {[}$-$0.008, $-$0.003{]}, IoU $-$0.003 {[}$-$0.006, +0.001{]}; \path{dist_ds_p10} $-$ \path{unet_dist_p10} F1 +0.019 {[}+0.016, +0.022{]}, PQ +0.016 {[}+0.013, +0.018{]}. Its training AUC stayed at 0.87--0.92 and its marker head degraded less (MSE 0.0092 $\to$ 0.0157, unmarked 12 $\to$ 15 \%).

\begin{itemize}

\item
  \textbf{Honest outcome: differentiable selection opens the gap, it does not close it.} On the full fold the variant is 0.063 F1 and 0.048 PQ below the 11.5 recipe and 0.062 F1 / 0.054 PQ below the fair baseline, every interval clear of zero, and fold 3 ranked it the same way. SQ is unchanged (0.821 against 0.816 / 0.824); the whole loss is DQ: 26 \% of the nuclei have no marker against 12.7 \% for 11.5 and 21.8 \% for the watershed baseline. At 10 \% of the fold it is level with the 11.5 recipe (F1 $-$0.001, PQ 0.000, intervals across zero) and keeps that recipe\textquotesingle s +0.019 F1 over its baseline: neither gain nor loss where the head is weak anyway.
\item
  \textbf{The selection was not learned; the detector was damaged.} Three facts. (1) The learned strengths barely discriminate: on fold 2 every candidate has $\sigma$ $\ge$ 0.5 (background 0.64, primary 0.69, duplicate 0.68 on average), so the cut $\tau$ $\le$ 0.5 removes nothing (\path{layer_all} equals \texttt{layer} to every digit and the fold-3 table is flat in $\tau$ up to 0.5, then collapses at 0.7). The mean background strength is weaker than the mean primary\textquotesingle s by only a factor (0.64 / 0.69)\^{}20 $\approx$ 0.2 in the attachment. (2) The ranking the strengths could use is no better than that of a head trained on its own target: AUC 0.876 against 0.867 for \path{dist_sched10_bg} and 0.874 for \path{unet_dist} (primary vs background), 0.666 against 0.707 and 0.715 (primary vs duplicate). (3) The gradient that reaches the marker logits through the solve made the head a worse detector: three times as many background candidates (19.9 against 6.6 per patch at thr 0.1), a fifth fewer primaries (17.7 against 21.5), MSE to its distance target 0.022 against 0.008. What pushes the true nuclei\textquotesingle s maps down was not isolated. A plausible route, unverified: a candidate grown down to 0.1 over two touching nuclei is the primary of one and wrong over the other, so the loss pushes the whole merged candidate down. Three of these facts point the same way. The partition cross-entropy is a weak and partly misdirected detection signal. It can tell the solve to give a spurious marker fewer pixels, which the 11.5 recipe already does through the unary costs (the variant\textquotesingle s solve absorbs 12.6 background candidates per patch at a count error of 0.22). It cannot tell the marker head where a nucleus is that no candidate covers, since an unmarked nucleus sends no gradient to the head.
\item
  \textbf{The strengths live in the finite-p softness, which the limit removes.} The exact Power Watershed of the same costs and strengths gives F1 0.115 on the 307 limit patches (11.5: 0.700). At p $\to$ $\infty$ the shift $-$log $\sigma$\_k $\approx$ 0.4 cost units turns into a hard threshold: a marker loses its whole nucleus to the background seed as soon as a path through the nucleus reaches a pixel whose background cost is below c1 + 0.4 without crossing a higher edge cost, where at p = 20 the shift only rescales the attachment by $\sigma$\^{}p. Deploying this variant at the limit is not possible. The oracle with ground-truth markers rises to 0.931 (11.5: 0.872), for a related reason: ground-truth markers carry $\sigma$ = 1, i.e. attachments e\^{}7 times stronger than any marker the trained solve has seen, so they override the background as the section 11 oracle did. It is not comparable with the other oracles.
\item
  \textbf{What this says about the plan.} Within this design (attachment-level strengths fed by the marker head\textquotesingle s score, trained only by the partition cross-entropy), putting marker selection in the gradient path does not let end-to-end training through the WS layer beat the U-Net + watershed pipeline; it hurts at full data and is neutral at 10 \%. The bottleneck stays detection (which nuclei get a marker at all), and the WS layer\textquotesingle s gradient is the wrong signal for it. For the paper, the statement it supports: \emph{making marker selection differentiable (candidate strengths trained through the solve) did not help; the gradient reaching the marker head degraded detection without improving the ranking of candidates.} Untried within the deadline, in decreasing order of promise: stopping the gradient into the marker head and training only $\theta$, T (this reduces to a learned threshold on z, i.e. the fold-3 tuning, so a small gain at best); a separate strength channel trained only by the solve, so the distance head keeps its target; and a per-image normalisation of the strengths. One seed, one fold pair, one parametrisation (mean score, $\theta_0$ = $-$1, primary rule).
\end{itemize}

\section{Efficiency: time and memory against the number of markers, and implicit against unrolled differentiation}

Code: \path{experiments/bench_multi_k.py} (experiment B), \path{experiments/grad_unroll_vs_implicit.py} (C a, b), \texttt{-\/-layer\ implicit} and \texttt{-\/-tb} in \path{experiments/exp_unet_seg.py} (C c; the implicit layer logs its forward / adjoint PCG iterations per epoch), \path{experiments/plot_efficiency.py} (figures and every table), \path{experiments/ruche_efficiency.sh} (the SLURM jobs); \texttt{ImplicitRW} and \texttt{ImplicitRWMulti} now also report the adjoint\textquotesingle s iterations in \texttt{stats}. Results: \path{results/efficiency/} (\path{bench_multi_k.json}, \path{grad_unroll.json}, \path{train/} with the four training runs\textquotesingle{} json and per-image metrics; the logs are git-ignored) and \path{results/efficiency/figures/} (\path{multi_k.png}, \path{grad_vs_unroll.png}, \path{unroll_cost.png}, and \texttt{summary.md} with all the tables, of which the ones below are extracts). The timings ran on ruche A100-SXM4-40GB GPUs, one GPU per job, because jacobian\textquotesingle s GPU was shared with another evaluation at the time (the part of experiment B that ran there while it was idle, jacobian\textquotesingle s \path{logs/bench_multi_k_jacobian_partial.log}, shows the same shapes on the RTX 6000 Ada). Every time is the median of 7 runs after 2 warm-up runs with CUDA synchronisation around each timed call; memory is \path{max_memory_allocated} above what was allocated before the call (the inputs), so it is the WS layer\textquotesingle s own footprint. The training runs\textquotesingle{} \texttt{.pt} files are on ruche.

\textbf{Experiment B: the multi-marker WS layer against K.} Inputs: the edge costs, background and marker costs of the trained instance model \path{dist_sched10_bg} (section 11.5) on 16 evenly spaced fold-2 patches (256$^2$), with its own predicted markers (27.0 per patch, 8--54). For each K in \{1, 2, \ldots, 128\} every patch gets exactly K markers: a random subset of its markers when it has more, otherwise its largest marker cut in two along its principal axis until there are K (so large K means several markers per nucleus, as the over-segmenting heads of section 11 produced). p = 20 (the model\textquotesingle s final p); loss = the partition cross-entropy against the argmax labels; two solver settings, the training one (relative residual 1e-4, at most 100 PCG steps) and the converged one of the evaluations (1e-5, at most 1000).

\begin{center}\scriptsize\setlength{\tabcolsep}{3pt}
\resizebox{\ifdim\width>\textwidth\textwidth\else\width\fi}{!}{%
\begin{tabular}{@{}llllllllll@{}}
\toprule
setting & batch & K & systems & forward (ms) & PCG steps & backward (ms) & adjoint steps & ms per step & peak fwd+bwd (MB) \\
\midrule

training & 1 & 1 / 8 / 32 / 128 & 2 / 9 / 33 / 129 & 34 / 34 / 41 / 184 & 100 & 35 / 35 / 41 / 185 & 100 & 0.34 / 0.34 / 0.41 / 1.83 & 11 / 45 / 158 / 614 \\
training & 16 & 1 / 8 / 32 / 128 & 32 / 144 / 528 / 2064 & 40 / 203 / 693 / 2644 & 100 & 43 / 208 / 706 / 2687 & 100 & 0.40 / 2.03 / 6.93 / 26.4 & 172 / 704 / 2528 / 9818 \\
converged & 1 & 1 / 8 / 32 / 128 & 2 / 9 / 33 / 129 & 93 / 87 / 78 / 344 & 280 / 264 / 194 / 189 & 212 / 170 / 148 / 672 & 639 / 510 / 368 / 369 & 0.33 / 0.33 / 0.40 / 1.82 & as above \\
converged & 16 & 1 / 8 / 32 / 128 & 32 / 144 / 528 / 2064 & 182 / 856 / 1963 / 6651 & 464 / 427 / 286 / 254 & 390 / 1703 / 4279 / 13 601 & 995 / 848 / 623 / 519 & 0.39 / 2.01 / 6.86 / 26.2 & as above \\
\bottomrule
\end{tabular}}
\end{center}

The U-Net forward takes 1.7 ms for one image and 14.5 ms for a batch of 16 (0.9 ms per image). The exact K-label limit on one CPU core (median over 4 patches $\times$ 3 runs, each back end in its own process pinned to one core, the same costs and markers as the GPU runs):

\begin{center}\scriptsize\setlength{\tabcolsep}{3pt}
\resizebox{\ifdim\width>\textwidth\textwidth\else\width\fi}{!}{%
\begin{tabular}{@{}lllllllll@{}}
\toprule
back end & K = 1 & 2 & 4 & 8 & 16 & 32 & 64 & 128 \\
\midrule

Numba (\path{pw_fast.py}) & 29 ms & 29 & 29 & 30 & 32 & 33 & 40 & 51 \\
NumPy & 850 ms & 848 & 851 & 850 & 857 & 864 & 877 & 883 \\
\bottomrule
\end{tabular}}
\end{center}

\begin{itemize}

\item
  \textbf{The cost is affine in K, once the GPU is busy.} At a fixed number of PCG steps (the training setting, which hits its 100-step cap at every K here) the forward time of a batch of 16 is 41 + 20.3 K ms for K = 8--128, within 1 \% of the fit at every point; a PCG step costs 12.8--14 $\mu$s per 256$^2$ system as soon as a batch carries more than about 150 systems, and the backward pass (one adjoint solve with the same matrix) costs the same as the forward. Below that the GPU is launch-bound: 0.34 ms per step whatever the number of systems, so one image costs the same for K = 1 and K = 16. Memory is exactly linear in the number of systems: 4.8 MB per marker per image for forward + backward, the same at batch 1 and 16 and for both settings (it is the stacked (M,1,H,W) tensors of PCG and the saved solution), i.e. 9.8 GB for a batch of 16 images with 128 markers each. "Affine in K" is therefore true of the work and of the memory, with the caveat that small problems do not fill the GPU (figure \path{multi_k.png} (a, b)).
\item
  \textbf{Converged, the cost grows more slowly than K,} because more markers make the systems better conditioned: the forward needs 464 PCG steps per batch at K = 1 and 254 at K = 128 (to 1e-5). The adjoint needs about twice the forward\textquotesingle s steps (995 $\to$ 519), presumably because the cross-entropy\textquotesingle s gradient $-$1/x is a rougher right-hand side than the attachments; at K = 1--2 it reaches the 1000-step cap. So the converged forward + backward of a batch of 16 goes from 0.57 s (K = 1) to 6.2 s (K = 32) and 20 s (K = 128).
\item
  \textbf{The limit is flat in K.} One CPU core computes the exact K-label Power Watershed of a 256$^2$ patch in 29 ms at K = 1 and 51 ms at K = 128 with the Numba back end (1.8 $\times$ for 128 $\times$ the labels; presumably the K-vector plateau solves and the K extra label nodes, not profiled), and in 0.85--0.88 s with the NumPy one (+4 \%). At the K of real patches ($\approx$ 30) the limit on one core (33 ms) is 2.4 times faster than the converged GPU solve of one image (78 ms) and 3.7 times faster than its per-image cost in a batch of 16 (123 ms); it is 20--35 times the U-Net\textquotesingle s forward (1.7 ms alone, 0.9 ms per image in a batch). The claim "one spanning forest whatever K" holds as stated, and with the Numba back end the limit is the cheapest way to run the trained layer.
\end{itemize}

\textbf{Experiment C (a): gradient accuracy of the binary WS layer.} Inputs: the edge and unary costs of the trained binary model \path{rw_learned} (section 6) on 8 evenly spaced fold-2 patches; loss = the BCE of the WS layer output against the mask (the training loss), at p in \{5, 20, 50\}. Reference: autograd through a float64 PCG run to a relative residual of 1e-12 (71--246 steps: these costs give well-conditioned systems), which agrees with the float64 implicit adjoint to 1e-9 and with central finite differences (float64, converged solves, h = 1e-5) to a median 3e-9--8e-9 on the entries with non-negligible gradient (the largest and random entries of the horizontal edge costs and of c1) and to 4e-8 along random directions. Compared: autograd through K unrolled PCG steps, K in \{5, 10, 20, 50, 200\}, in float32 (as trained) and float64; the implicit adjoint in float32 at the training tolerance (1e-4, caps 100 and 1000) and at 1e-5. Median over patches of the relative error $\|$g $-$ g\_ref$\|$ / $\|$g\_ref$\|$ (cosine in brackets):

\begin{center}\scriptsize\setlength{\tabcolsep}{3pt}
\resizebox{\ifdim\width>\textwidth\textwidth\else\width\fi}{!}{%
\begin{tabular}{@{}llllllll@{}}
\toprule
p & unrolled K = 5 & K = 10 & K = 20 & K = 50 & K = 200 & implicit 1e-4 (steps fwd / adj.) & implicit 1e-5 \\
\midrule

5, edge costs & 4.5 (0.26) & 5.3 (0.23) & 0.12 (0.993) & 2.2e-5 & 9.2e-6 & 1.6e-3 (30 / 29) & 1.1e-4 \\
5, unaries & 3.1 (0.44) & 2.9 (0.33) & 0.065 (0.998) & 1.3e-5 & 9.2e-6 & 7.5e-4 & 5.0e-5 \\
20, edge costs & 2.8 (0.15) & 1.8 (0.39) & 0.90 (0.75) & 0.043 (0.9985) & 2.7e-6 & 1.1e-3 (62 / 60) & 2.9e-5 \\
20, unaries & 1.9 (0.53) & 1.3 (0.75) & 0.50 (0.92) & 0.015 & 1.7e-6 & 5.7e-4 & 1.8e-5 \\
50, edge costs & 5.3 (0.05) & 2.4 (0.19) & 0.93 (0.67) & 0.19 (0.983) & 3.4e-6 & 3.6e-4 (92 / 109) & 3.2e-5 \\
50, unaries & 3.0 (0.28) & 1.6 (0.61) & 0.73 (0.87) & 0.078 (0.997) & 1.4e-6 & 1.8e-4 & 1.7e-5 \\
forward error of x, p = 5 / 20 / 50 & 0.10 / 0.33 / 0.56 & 0.022 / 0.15 / 0.28 & 7.7e-4 / 0.036 / 0.092 & 1.5e-7 / 2.1e-4 / 5.1e-3 & 2e-7 & 2--3e-5 & 1.5--2.7e-6 \\
\bottomrule
\end{tabular}}
\end{center}

Float64 unrolling gives the same errors to three digits up to K = 20, and up to K = 50 at p = 20 and 50 (\path{grad_vs_unroll.png}): the error is truncation, not rounding; float32 rounding (about 1e-5) only shows at K = 50, p = 5 and at K = 200. The cosine is computed over all the edge (resp. unary) entries of a patch.

\textbf{Experiment C (b): cost against depth} (binary WS layer, float32, 256$^2$, forward + backward):

\begin{center}\scriptsize\setlength{\tabcolsep}{3pt}
\resizebox{\ifdim\width>\textwidth\textwidth\else\width\fi}{!}{%
\begin{tabular}{@{}lllllll@{}}
\toprule
& unrolled K = 5 & 10 & 20 & 50 & 200 & implicit 1e-4, p = 5 / 20 / 50 \\
\midrule

time, batch 16 (ms) & 6 & 11 & 21 & 50 & 198 & 22 / 44 / 69 \\
peak memory, batch 16 (MB) & 188 & 308 & 548 & 1268 & 4869 & 92 at every p and tolerance \\
peak memory, batch 1 (MB) & 12 & 19 & 34 & 79 & 304 & 5.7 \\
\bottomrule
\end{tabular}}
\end{center}

Unrolled memory grows by 24 MB per step for a batch of 16 (1.5 MB, i.e. six 256$^2$ float32 tensors, per image and step) and does not depend on p; the implicit layer holds the solution and the inputs only, so its memory is independent of depth, tolerance and p (\path{unroll_cost.png}). Time is the same at batch 1 and 16 (launch-bound: 0.33 ms per PCG step either way). The implicit layer costs about as much as 20 unrolled steps at p = 5 (22 against 21 ms) and two to three times more at p = 20--50 (44 and 69 ms), where it runs the 60--110 steps each way that convergence to 1e-4 needs.

\begin{itemize}

\item
  \textbf{Unrolling 20 steps does not give the Random Walker\textquotesingle s gradient at the p used in training.} At p = 5 the 20-step gradient is within 12 \% of the true one (cosine 0.993); at p = 20 and 50 it is off by 90--93 \% on the edge costs (cosine 0.75 and 0.67), while the 20-step forward solution is only 4--9 \% off. Fewer steps are worse than useless (K = 5--10: relative errors 1.6--5, cosine 0.05--0.4). Within a few per cent takes 50 steps at p = 20 and between 50 and 200 at p = 50 (K = 50 is still 19 \% off there). The unrolled gradient is of course the exact gradient of the 20-step map the network was trained with, so this is not a bug in sections 3--10; it says that those models were trained through a truncated solver whose output is close to the RW solution but whose derivative is not. The implicit adjoint at the training tolerance (1e-4) is within 0.04--0.16 \% of the reference with 30--109 steps each way, and at 1e-5 within 3e-5--1e-4.
\item
  \textbf{The implicit mode is the cheaper one in memory at any useful accuracy, and in time once p $\ge$ 20.} To come within the implicit layer\textquotesingle s 1e-3 the unrolled WS layer needs more than 50 steps at p = 20 and 50 (K = 50 is still 4 \% and 19 \% off), i.e. more than 1.27 GB and 50 ms for a batch of 16, against the implicit layer\textquotesingle s 92 MB and 44--69 ms. Unrolling 20 steps is cheaper (21 ms, 548 MB) but inaccurate.
\end{itemize}

\textbf{Experiment C (c): one training run per mode and seed.} \path{rw_learned} with the section 6 recipe (fold 1 $\to$ all of fold 2, p 5 $\to$ 50, BCE on the WS layer + 0.5 BCE on the logits, 30 epochs, batch 16), seeds 0 and 1, unrolled (K = 20) against implicit (PCG to 1e-4, cap 1000, one adjoint solve; the evaluation\textquotesingle s converged solve and limit as in section 6). The unrolled seed-0 run is a rerun of section 6\textquotesingle s run with the current code; it reproduces it (IoU $-$0.001 {[}$-$0.003, +0.001{]}, F1 +0.005 paired).

\begin{center}\scriptsize\setlength{\tabcolsep}{3pt}
\resizebox{\ifdim\width>\textwidth\textwidth\else\width\fi}{!}{%
\begin{tabular}{@{}lllllllllll@{}}
\toprule
run & output & IoU & BF$_1$ & BF$_2$ & band acc. & \href{mailto:F1@0.5}{\nolinkurl{F1@0.5}} & PQ & AJI & count err. & s / epoch \\
\midrule

U-Net (sec. 6) & s \textgreater{} 0.5 & 0.719 & 0.652 & 0.769 & 0.807 & 0.656 & 0.531 & 0.494 & 0.275 & -- \\
unrolled, seed 0 & WS layer & 0.708 & 0.629 & 0.751 & 0.799 & 0.609 & 0.486 & 0.456 & 0.330 & 13.3 \\
unrolled, seed 1 & WS layer & 0.709 & 0.626 & 0.750 & 0.798 & 0.594 & 0.473 & 0.433 & 0.341 & 13.2 \\
implicit, seed 0 & WS layer & 0.716 & \textbf{0.652} & 0.770 & 0.804 & 0.644 & 0.519 & 0.419 & 0.256 & 26.1 \\
implicit, seed 1 & WS layer & \textbf{0.719} & \textbf{0.655} & \textbf{0.773} & 0.806 & 0.649 & 0.523 & 0.430 & \textbf{0.249} & 26.1 \\
unrolled, seed 0 / 1 & PW limit (307) & 0.668 / 0.620 & 0.597 / 0.566 & 0.719 / 0.676 & 0.775 / 0.745 & 0.630 / 0.575 & 0.501 / 0.456 & 0.440 / 0.353 & 0.246 / 0.298 & \\
implicit, seed 0 / 1 & PW limit (307) & 0.602 / 0.604 & 0.573 / 0.582 & 0.677 / 0.684 & 0.730 / 0.736 & 0.595 / 0.611 & 0.486 / 0.498 & 0.282 / 0.352 & 0.417 / 0.394 & \\
\bottomrule
\end{tabular}}
\end{center}

The converged solve gives the WS layer\textquotesingle s numbers to the third digit in both modes (\path{figures/summary.md}). Paired differences on all 2523 patches (bootstrap 95 \% intervals; F1 and PQ dataset-level, bootstrapped over patches):

\begin{center}\scriptsize\setlength{\tabcolsep}{3pt}
\resizebox{\ifdim\width>\textwidth\textwidth\else\width\fi}{!}{%
\begin{tabular}{@{}lllll@{}}
\toprule
comparison & IoU & BF$_2$ & \href{mailto:F1@0.5}{\nolinkurl{F1@0.5}} & PQ \\
\midrule

implicit $-$ unrolled, seed 0 & +0.008 {[}+0.005, +0.010{]} & +0.018 {[}+0.016, +0.020{]} & +0.035 {[}+0.032, +0.038{]} & +0.034 {[}+0.031, +0.036{]} \\
implicit $-$ unrolled, seed 1 & +0.009 {[}+0.007, +0.012{]} & +0.024 {[}+0.021, +0.026{]} & +0.055 {[}+0.052, +0.058{]} & +0.050 {[}+0.048, +0.053{]} \\
unrolled seed 0 $-$ seed 1 & $-$0.001 {[}$-$0.003, +0.001{]} & +0.002 {[}$-$0.001, +0.004{]} & +0.015 {[}+0.013, +0.018{]} & +0.013 {[}+0.011, +0.015{]} \\
implicit seed 0 $-$ seed 1 & $-$0.003 {[}$-$0.005, $-$0.001{]} & $-$0.004 {[}$-$0.006, $-$0.002{]} & $-$0.005 {[}$-$0.007, $-$0.002{]} & $-$0.004 {[}$-$0.006, $-$0.002{]} \\
implicit $-$ U-Net (sec. 6), seed 0 & $-$0.003 {[}$-$0.005, $-$0.001{]} & +0.000 {[}$-$0.002, +0.003{]} & $-$0.012 {[}$-$0.015, $-$0.009{]} & $-$0.012 {[}$-$0.014, $-$0.009{]} \\
implicit $-$ U-Net (sec. 6), seed 1 & $-$0.000 {[}$-$0.002, +0.002{]} & +0.004 {[}+0.002, +0.007{]} & $-$0.007 {[}$-$0.010, $-$0.004{]} & $-$0.007 {[}$-$0.010, $-$0.005{]} \\
\bottomrule
\end{tabular}}
\end{center}

The implicit training solve needed 18--25 forward / 19--26 adjoint PCG steps per batch in the first epoch (p = 5), about 135 / 110 at epoch 10, 165--205 / 150 at epoch 20 and 210 / 220 at the end (p = 50); 3.4--3.8 \% of the forward solves over the run reached the 1000-step cap (all late, at high p), so training took twice as long as the unrolled run (26 s against 13 s per epoch on the A100, 13 against 6.6 min in all).

\begin{itemize}

\item
  \textbf{Honest outcome: the implicit mode trains a better binary WS layer than 20 unrolled steps, on both seeds.} +0.008--0.009 IoU, +0.018--0.024 BF$_2$, +0.035--0.055 F1 and +0.034--0.050 PQ, paired and clear of zero, against a seed-to-seed spread of at most 0.003 IoU and 0.015 F1 within either mode. With it the binary WS layer reaches the plain U-Net of section 6 on the pixel metrics (IoU $-$0.003 / $-$0.000, BF$_2$ +0.000 / +0.004) and has the lowest count error of the three (0.25 against 0.28), but stays 0.007--0.012 F1 / PQ behind on instances, and its AJI is lower than the unrolled WS layer\textquotesingle s (0.42--0.43 against 0.43--0.46). So section 6\textquotesingle s "the WS layer costs a little" was, on the pixel metrics, in good part the truncated solver: trained through the true Random Walker gradient, the WS layer costs nothing, and it still does not gain anything over the U-Net. Two seeds, one fold pair, one recipe (p 5 $\to$ 50, no edge target, no gap target); whether the sections 6.1--10 variants change in the same way was not tested.
\item
  \textbf{Why, plausibly:} at p $\ge$ 20 the unrolled gradient points 40--50$^\circ$ away from the true one (cosine 0.67--0.75), so for the last two thirds of the schedule the cost head was trained against a derivative that is not that of the WS layer it is evaluated with. The run cannot separate this from the other difference between the modes, that the implicit layer is the converged solve also in the forward pass (the unrolled models are evaluated close to their converged solve too: layer = converged within 0.002 IoU).
\item
  \textbf{The exact limit does not benefit, it suffers.} On the 307 limit patches the limit of the implicit-trained costs loses 0.107--0.108 IoU against their converged solve (0.60 against 0.71) and has the worst count error of the four runs (0.39--0.42), where the unrolled runs\textquotesingle{} limits lose 0.039 and 0.083 IoU and vary a lot between seeds (0.668 / 0.620). Learning the costs at the true finite-p solution makes the model rely on finite-p softness that the limit removes (section 6.1 saw the same with costs learned at low p). Deploy-at-the-limit therefore needs its own training consideration (higher final p, or a sharper cost parametrisation) whichever differentiation mode is used.
\item
  \textbf{What this says for the paper.} The three efficiency claims hold, measured: memory independent of depth for the implicit mode (92 MB for a batch of 16 at any tolerance, against 24 MB per unrolled step), cost affine in K at a fixed number of PCG steps (20.3 ms per marker per batch of 16, 4.8 MB per marker per image) with fewer steps needed as K grows, and an exact limit flat in K (29--51 ms on one core with Numba for K = 1--128). The implicit mode also wins on gradient accuracy (1e-3 against 0.9 relative error at the training p, K = 20) and on trained accuracy (+0.02 BF$_2$, +0.03--0.05 F1 and PQ on two seeds), at twice the training time. Caveats to state: the affine law needs a batch large enough to fill the GPU (about 150 systems on an A100; below that the cost per step is flat), the converged adjoint of the multi-marker WS layer takes about twice the forward\textquotesingle s steps, and the limit of implicit-trained binary costs is worse than that of unrolled-trained ones.
\end{itemize}

\section{Selection without touching the marker head: a strength channel, a detached scorer, and the control (follow-up of section 14)}

Section 14 left two explanations for its loss, and the paper is due the next day. One: the solve\textquotesingle s gradient reached the distance head through the candidates\textquotesingle{} scores and degraded it as a detector. Two: the partition cross-entropy does not supply a selection signal at all. The first can be removed without giving up the strengths, and one of us asked for a differentiable way to find markers that does. Two scorers keep the distance head out of the solve\textquotesingle s gradient, and a control separates the scorers from the other changes of section 14 (the permissive candidates and the primary rule for duplicates). Code: \texttt{-\/-ds-score\ head\textbar{}mlp\textbar{}none} in \path{experiments/exp_instance_pannuke.py} (\path{scorer_strengths}, \path{candidate_features}, \path{unit_strengths}, \path{scorer_selftest}), the cases \path{dist_dsh}, \path{dist_dsm}, \path{dist_dsn} of \path{ruche_run.sh}, and \path{candidate_auc.py}, which now also ranks by the strength channel and the scorer. Results: \texttt{results/pannuke\_instances/\{dist\_dsh,\ dist\_dsm,\ dist\_dsn\}/}, \path{compare_16.md}.

\textbf{The three runs.} All three keep section 14\textquotesingle s recipe otherwise: candidates at thr 0.1 / h 0.1 / area 1 from epoch 10 (all images from 15), primary-overlap target for duplicates, unmarked nuclei to the background label, the strength entering as a\_k = ($\sigma$\_k s)\^{}p, p 5 $\to$ 20, 30 epochs, seed 0, fold 1 $\to$ fold 2. Fold 3 chooses thr $\in$ \{0.05, 0.1, 0.2\} $\times$ h $\in$ \{0.1, 0.2\} (area 1) and $\tau$ $\in$ \{0, 0.3, 0.5, 0.7, 0.9\}.

\begin{itemize}

\item
  \path{dist_dsh} (\texttt{-\/-ds-score\ head}): a fifth U-Net output channel g; $\sigma$\_k = sigmoid((mean of g over the candidate $-$ $\theta$) / T). The distance channel gets no gradient from the solve; the shared trunk does, through g.
\item
  \path{dist_dsm} (\texttt{-\/-ds-score\ mlp}, the design proposed by the main session): a two-layer MLP (9 $\to$ 32 $\to$ 1, its own parameters only, output bias initialised at +4 so that $\sigma$ $\approx$ 0.98 at the start) on nine features computed from the detached distance map d, foreground s and contour map c. The features: peak of d, relative prominence (peak $-$ saddle) / peak, log area, mean s, mean c, whether the candidate has an adjacent candidate, and for the least-separated adjacent candidate the mean and maximum of c along the shared boundary and the relative depth of the d valley between the two peaks. log $\sigma$\_k = $-$softplus($-$logit). No gradient from the strengths reaches the U-Net at all; the scorer learns only from the partition cross-entropy through the solve.
\item
  \path{dist_dsn} (\texttt{-\/-ds-score\ none}): every strength 1. Same candidates, same targets, no selection: the control.
\end{itemize}

\textbf{Checks.} The scorer\textquotesingle s gradient against central finite differences (\texttt{-\/-ds-selftest\ -\/-ds-score\ mlp}, the case of section 14 with the candidates\textquotesingle{} real features from random maps, a test scorer with 8 hidden units and output bias 0.5): worst relative error 1.1e-8 over four random directions in its 89 parameters and on the output bias (\path{dist_dsm/ds_selftest_mlp.txt}). The wiring check, on eight fold-1 patches, backpropagates the solve\textquotesingle s loss alone. With \texttt{head}, the distance channel\textquotesingle s output weights get a gradient of exactly 0 and the strength channel\textquotesingle s 0.19; with section 14\textquotesingle s scoring the distance channel gets 0.22. The sanity check proposed for the run itself, that the distance head stays at the 11.5 level, holds: its final MSE is 0.0085 (\texttt{head}), 0.0083 (\texttt{mlp}), 0.0082 (control) against 0.0082 for \path{dist_sched10_bg} and 0.0219 for section 14\textquotesingle s \path{dist_ds}.

\textbf{Results on fold 2} (all 2523 patches; \path{compare_16.md}):

\begin{center}\scriptsize\setlength{\tabcolsep}{3pt}
\resizebox{\ifdim\width>\textwidth\textwidth\else\width\fi}{!}{%
\begin{tabular}{@{}lllllllllll@{}}
\toprule
run & \href{mailto:F1@0.5}{\nolinkurl{F1@0.5}} & PQ & SQ & DQ & AJI & IoU & count err. & markers one / none / several, bg per patch & oracle F1 / PQ & limit F1 / PQ (307) \\
\midrule

\path{dist_sched10_bg} (11.5) & \textbf{0.795} & 0.649 & 0.816 & \textbf{0.795} & \textbf{0.668} & 0.715 & \textbf{0.182} & 85.5 / 12.7 / 1.8 \%, 4.3 & 0.872 / 0.723 & 0.700 / 0.564 \\
\path{unet_dist} (11.2) & 0.794 & \textbf{0.655} & \textbf{0.824} & 0.794 & 0.658 & \textbf{0.719} & 0.190 & 76.6 / 21.8 / 1.6 \%, 1.6 & 0.867 / 0.708 & -- \\
\path{dist_dsh}: strength channel & 0.782 & 0.638 & 0.816 & 0.782 & 0.656 & 0.701 & 0.188 & 85.0 / 13.3 / 1.7 \%, 4.1 & 0.917 / 0.751 & 0.099 / 0.082 \\
\path{dist_dsm}: detached MLP scorer & 0.790 & 0.643 & 0.814 & 0.790 & 0.666 & 0.712 & 0.183 & 87.4 / 10.5 / 2.1 \%, 6.4 & 0.927 / 0.758 & 0.157 / 0.128 \\
\path{dist_dsn}: control, strengths 1 & 0.790 & 0.646 & 0.817 & 0.790 & 0.663 & 0.713 & \textbf{0.176} & 86.2 / 12.5 / 1.3 \%, 4.8 & 0.856 / 0.704 & 0.611 / 0.502 \\
\bottomrule
\end{tabular}}
\end{center}

Paired against \path{dist_sched10_bg} (bootstrap 95 \% intervals): \path{dist_dsm} F1 $-$0.006 {[}$-$0.008, $-$0.004{]}, PQ $-$0.006 {[}$-$0.008, $-$0.005{]}, AJI $-$0.002 {[}$-$0.003, $-$0.000{]}, IoU $-$0.004 {[}$-$0.006, $-$0.001{]}; \path{dist_dsh} F1 $-$0.013 {[}$-$0.015, $-$0.011{]}, PQ $-$0.011 {[}$-$0.013, $-$0.009{]}, AJI $-$0.012 {[}$-$0.014, $-$0.010{]}, IoU $-$0.015 {[}$-$0.018, $-$0.012{]}; the control \path{dist_dsn} F1 $-$0.005 {[}$-$0.007, $-$0.003{]}, PQ $-$0.004 {[}$-$0.005, $-$0.002{]}, AJI $-$0.005 {[}$-$0.007, $-$0.003{]}, IoU $-$0.003 {[}$-$0.005, $-$0.001{]}. The scorers against the control: \path{dist_dsm} $-$ \path{dist_dsn} F1 $-$0.001 {[}$-$0.003, +0.001{]}, PQ $-$0.002 {[}$-$0.004, $-$0.001{]}, AJI +0.003 {[}+0.001, +0.005{]}; \path{dist_dsh} $-$ \path{dist_dsn} F1 $-$0.008 {[}$-$0.010, $-$0.006{]}, PQ $-$0.007 {[}$-$0.009, $-$0.006{]}, AJI $-$0.007 {[}$-$0.009, $-$0.005{]}. Against \path{unet_dist}: \path{dist_dsm} F1 $-$0.004 {[}$-$0.007, $-$0.001{]}, PQ $-$0.011 {[}$-$0.013, $-$0.009{]}, AJI +0.007 {[}+0.005, +0.010{]}; \path{dist_dsh} F1 $-$0.012 {[}$-$0.014, $-$0.008{]}, PQ $-$0.016 {[}$-$0.018, $-$0.014{]}, AJI $-$0.002 {[}$-$0.005, +0.001{]}. Fold 3 had ranked them the same way: PQ / F1 0.650 / 0.791 (\texttt{mlp}) and 0.648 / 0.789 (\texttt{head}) against 0.655 / 0.797 for \path{dist_sched10_bg} at the same marker setting. Fold 3 chose thr 0.2 / h 0.2 for all three runs, and $\tau$ is again immaterial: the tables are identical for $\tau$ = 0, 0.3, 0.5 and collapse at 0.7--0.9. On fold 2 every candidate has $\sigma$ $\ge$ 0.5 (mean 0.83 / 0.74--0.76 / 0.76--0.77 for background / primary / duplicate candidates in both scorers).

\textbf{Candidate rankings on fold 2} (\path{candidate_auc_*.md}, candidates at thr 0.1 / h 0.1; primary / background / duplicate candidates per patch; AUC primary vs background, vs duplicate). Distance heads, by the mean logit: \path{dist_dsh} 21.5 / 6.8 / 1.2, 0.871, 0.714; \path{dist_dsm} 21.9 / 10.3 / 1.6, 0.898, 0.718 (\path{dist_sched10_bg} 21.5 / 6.6 / 1.2, 0.867, 0.707). The learned strengths: strength channel 0.133 / 0.409, MLP scorer 0.106 / 0.342. Both invert the ranking.

\begin{itemize}

\item
  \textbf{Honest outcome: neither scorer helps, and the inclusion bar for the paper is not met.} With the distance head protected, the loss of section 14 ($-$0.063 F1) shrinks to $-$0.006 (MLP scorer) and $-$0.013 (strength channel) against the 11.5 recipe, every interval still clear of zero and on the wrong side. Against the fair baseline the MLP scorer is $-$0.004 F1 / $-$0.011 PQ and +0.007 AJI. The bar set for the paper, a paired gain over \path{dist_sched10_bg} above the seed spread of 0.008 F1, is missed by about 0.014 F1. So section 14\textquotesingle s damage was the gradient into the detector, as diagnosed, but removing it leaves no gain to collect. The control apportions what is left. Training on the permissive candidates with the primary rule but no strengths (\path{dist_dsn}) is itself $-$0.005 F1 / $-$0.004 PQ against 11.5 on fold 2, though level on fold 3 (PQ 0.656 against 0.655). The detached scorer adds nothing to it ($-$0.001 F1, interval across zero; +0.003 AJI). The strength channel subtracts 0.008 F1, presumably through the gradient it sends into the shared trunk. The learned selection is worth zero; the small loss is the candidate rule.
\item
  \textbf{What the scorers learn is not selection.} Both strengths rank background candidates \emph{above} true nuclei (AUC 0.11--0.13). This is not a bug: a background candidate in low-foreground territory wins no pixel at any $\sigma$ $\le$ 1, gets no gradient, and keeps whatever strength the shared weights give it. The gradient goes to candidates that do win pixels, mostly primaries whose region (grown down to the candidate threshold) spills over the nucleus rim, and it weakens them. The strengths therefore act as a per-nucleus correction of the unary at the rim, which the foreground head already provides per pixel. The one ranking that matters for the output, primaries above duplicates (the duplicates do win pixels), is not learned either (AUC 0.34--0.41). The partition cross-entropy rewards a marker for the pixels it takes, not for being a nucleus. Saturation (gradient 1e-11 for a candidate that wins or loses its pixels outright, section 14) leaves exactly the decisive cases unsupervised.
\item
  \textbf{The limit collapses again.} The exact Power Watershed with the learned strengths gives F1 0.10--0.16 on the 307 limit patches: any $\sigma$\_k \textless{} 1 is a hard veto at p $\to$ $\infty$ (section 14), and here every candidate sits at $\sigma$ $\approx$ 0.75--0.83. Strengths of this kind are incompatible with deploying the limit (the control, at $\sigma$ = 1, keeps 0.611 / 0.502). The scorers\textquotesingle{} oracles, 0.917--0.927 against 0.856 for the control, are inflated the same way as section 14\textquotesingle s: ground-truth markers enter at $\sigma$ = 1, stronger than any marker the trained solve has seen.
\item
  \textbf{For the paper.} The section 14 sentence stands and can be sharpened: \emph{making marker selection differentiable did not help. Strengths computed from the marker head degraded its detections ($-$0.063 F1), and scorers isolated from it, a strength channel or a detached per-candidate network, were still 0.006--0.013 F1 behind training the solve on the extracted markers alone, because the partition loss rewards a marker for the pixels it takes, not for being a nucleus.} The marker extraction stays a discrete step inside the training loop (paper commit 8988128). One seed, one fold pair, one design each.
\end{itemize}

\section{Accuracy against p, from the training p to the Power Watershed limit}

4.4, which answers the request for "a figure of accuracy vs p, from the training p to the limit, for the learned costs"). Code: \path{experiments/p_sweep.py} (the sweep), \path{experiments/rw_exact.py} (an exact solve of the WS layer\textquotesingle s system at any p, below), \path{experiments/plot_p_sweep.py}, \path{experiments/run_p_sweep.sh} (jacobian), \path{experiments/ruche_p_sweep.sh} (CPU pass on a ruche \path{cpu_med} node), \path{experiments/cond_exact.py} (the section 1 check at the end). Results: \path{results/pannuke_p_sweep/} (\path{p_sweep.png} / \texttt{.pdf}, \path{p_sweep_tables.md} with every table below in full, per model \path{p_sweep.json}, \path{p_sweep_perimage.npz}, \path{pcg_stats.npz}, \path{labels_examples.npz} with the label maps at every p for 16 patches; logs on jacobian / ruche only); \path{results/conditioning/cond_exact_512.json}.

\textbf{Protocol.} For each trained model, the U-Net costs of every fold-2 patch are computed once (fp32 on the GPU, as in every evaluation of the report). For the multi-marker model the test markers are fixed too: the fold-3 choice of section 11.5, no fallback, i.e. its \texttt{layer} output. Only p changes, through w = exp($-$p c), so the sweep isolates the temperature. At each p the WS layer\textquotesingle s linear system is solved exactly, and the result is compared with the exact K-label (or binary) PW limit of the same costs and markers (\path{pw_limit_multi_numpy} / \path{pw_limit_numpy}, Numba). Models: \path{dist_sched10_bg} (section 11.5, p annealed 5 $\to$ 20, implicit solve; the paper\textquotesingle s headline), \path{rw_learned} (section 6, binary, p 5 $\to$ 50, 20 unrolled PCG steps) and \path{edge_bce} (section 10, \path{rw_learned} with the edge target). All 2523 patches at every p, for p $\in$ \{5, 10, 20, 35, 50, 100, 200, 500\} (binary: no 35). Metrics as in section 12.2. The disagreement is the fraction of all pixels whose label differs from the limit\textquotesingle s, and in brackets the same count divided by the pixels that are foreground in either output.

\textbf{The WS layer\textquotesingle s PCG does not reach large p, and neither does sparse LU, so the finite-p curve needs its own solver.} On 256 evenly spaced patches the WS layer\textquotesingle s own converged solve (Jacobi-PCG, tol 1e-5, at most 1000 iterations, as in every evaluation) was run in fp32 and fp64 and compared with the exact solve. For the instance model it is exact up to the training p: 0 pixels differ at p $\le$ 10, 0.015 \% at p = 20, where it takes 189 iterations on average (435 at most), and every fp64 system converges. In fp32 at p = 20 the true residual stalls around 1e-3 (rounding: the recursive residual meets the tolerance, the true one cannot), with no effect on the labels. From p = 35 it degrades (0.5 \% of the pixels differ), and from p = 100 it fails in both precisions: it stops at the iteration cap, true residuals reach 1e5--1e8, 5--12 \% of the pixels are wrong, and F1 is 0.27--0.53 instead of 0.72 at p = 200--500. fp32 additionally floors 6--20 \% of the edge weights at 1e-30. The binary models fare better because their edge costs are $\le$ 1: \path{rw_learned}\textquotesingle s PCG is exact to 0.001 \% of the pixels up to p = 200 and wrong on 0.44 \% at p = 500. For \path{edge_bce}, 0.8--3.3 \% of the pixels are wrong at p = 100--500, and since the thresholded output is then speckled, its F1 collapses (0.24 at p = 100 against 0.53). A diagonally rescaled sparse LU (scipy, fp64) agrees with the exact solve up to p = 100 (4 pixels over 32 patches differ there), but beyond that it loses the pivots to cancellation (d\_a $-$ w\_ai$^2$/d\_i of two nearly equal numbers): at p = 500 it returns negative probabilities and labels that depend on the ordering. \path{rw_exact.ExactRW} solves the same system by Gaussian elimination in the Grassmann--Taksar--Heyman form, on a nested-dissection order of the grid: eliminating node i adds w\_ai w\_ib / d\_i to the remaining weights and w\_ai e\_i / d\_i to the attachment (excess) of a, and the pivot is never updated by subtraction but recomputed as the sum of the current weights and excess. Every operation is between positive numbers, so the result is accurate to rounding whatever the conditioning. The weights are held as logarithms (logaddexp), and so are the label vectors of the substitution (a log scale per node), so nothing underflows. The symbolic factorisation depends on the grid only (2.6 M entries at 256$^2$, 12.5 M at 512$^2$), and a solve takes 1.3--1.9 s on one core at any p. Checks: $\Sigma$\_k x\_k = 1 to 1e-13 at every p. Its labels are identical to the WS layer\textquotesingle s PCG wherever the PCG converges (0 pixels at p $\le$ 10 on the 256 patches, 0.015 \% at p = 20), and to the sparse LU on 32 patches per model at every p $\le$ 100 (4 pixels over 32 patches differ at p = 100 for the instance model, where the LU is at the edge of its accuracy). The disagreement with \path{pw_limit_multi_numpy} falls to 0 pixels at p = 3000 on the three patches tried, which also confirms the limit code independently of section 12.

\textbf{Instance model \path{dist_sched10_bg}}, fold 2, 2523 patches, test markers fixed (training p 20):

\begin{center}\scriptsize\setlength{\tabcolsep}{3pt}
\resizebox{\ifdim\width>\textwidth\textwidth\else\width\fi}{!}{%
\begin{tabular}{@{}llllllllll@{}}
\toprule
p & IoU & BF$_2$ & \href{mailto:F1@0.5}{\nolinkurl{F1@0.5}} & PQ & SQ & AJI & count err. & px $\ne$ limit (of fg) & patches = limit \\
\midrule

5 & 0.628 & 0.630 & 0.691 & 0.508 & 0.735 & 0.600 & 0.231 & 6.51 \% (39.0 \%) & 3.6 \% \\
10 & 0.687 & 0.730 & 0.760 & 0.600 & 0.789 & 0.650 & 0.210 & 5.56 \% (32.5 \%) & 3.7 \% \\
\textbf{20} (train) & \textbf{0.715} & \textbf{0.775} & \textbf{0.795} & \textbf{0.649} & \textbf{0.816} & \textbf{0.668} & \textbf{0.182} & 4.45 \% (25.3 \%) & 3.9 \% \\
35 & 0.692 & 0.752 & 0.781 & 0.635 & 0.813 & 0.644 & 0.190 & 3.01 \% (17.9 \%) & 4.4 \% \\
50 & 0.669 & 0.731 & 0.761 & 0.618 & 0.812 & 0.621 & 0.216 & 1.96 \% (12.2 \%) & 4.6 \% \\
100 & 0.641 & 0.706 & 0.735 & 0.595 & 0.810 & 0.590 & 0.258 & 0.70 \% (4.6 \%) & 5.4 \% \\
200 & 0.633 & 0.699 & 0.727 & 0.588 & 0.809 & 0.582 & 0.270 & 0.23 \% (1.5 \%) & 8.2 \% \\
500 & 0.630 & 0.697 & 0.724 & 0.586 & 0.809 & 0.579 & 0.274 & 0.047 \% (0.32 \%) & 24.3 \% \\
$\infty$ (exact limit) & 0.630 & 0.697 & 0.724 & 0.585 & 0.809 & 0.578 & 0.274 & 0 & 100 \% \\
\bottomrule
\end{tabular}}
\end{center}

\textbf{Binary models}, same protocol (training p 50; the edge-target model\textquotesingle s thresholded output at low p is speckle, hence its count errors above 1):

\begin{center}\scriptsize\setlength{\tabcolsep}{3pt}
\resizebox{\ifdim\width>\textwidth\textwidth\else\width\fi}{!}{%
\begin{tabular}{@{}lllll@{}}
\toprule
p & \path{rw_learned} IoU / BF$_2$ / F1 / PQ / AJI, count err. & px $\ne$ limit (of fg) & \path{edge_bce} IoU / BF$_2$ / F1 / PQ / AJI, count err. & px $\ne$ limit (of fg) \\
\midrule

5 & 0.707 / 0.751 / 0.556 / 0.443 / 0.506, 0.687 & 3.77 \% (21.1 \%) & 0.714 / 0.736 / 0.217 / 0.173 / 0.322, 4.26 & 2.84 \% (15.5 \%) \\
10 & 0.706 / 0.749 / 0.577 / 0.458 / 0.467, 0.505 & 3.91 \% (21.7 \%) & 0.712 / 0.728 / 0.180 / 0.144 / 0.330, 6.01 & 2.84 \% (15.5 \%) \\
20 & 0.706 / 0.748 / 0.586 / 0.464 / 0.419, 0.413 & 3.95 \% (21.7 \%) & 0.708 / 0.729 / 0.215 / 0.171 / 0.333, 4.72 & 2.68 \% (14.7 \%) \\
\textbf{50} (train) & 0.710 / 0.751 / 0.602 / 0.480 / 0.366, 0.322 & 3.47 \% (19.1 \%) & 0.707 / 0.753 / 0.434 / 0.346 / 0.360, 1.10 & 1.68 \% (9.5 \%) \\
100 & 0.707 / 0.752 / 0.630 / 0.504 / 0.395, 0.272 & 2.64 \% (14.9 \%) & 0.703 / 0.759 / 0.520 / 0.414 / 0.391, 0.540 & 0.94 \% (5.5 \%) \\
200 & 0.693 / 0.742 / 0.653 / 0.521 / 0.461, 0.252 & 1.69 \% (9.9 \%) & 0.697 / 0.758 / 0.529 / 0.420 / 0.408, 0.480 & 0.48 \% (2.9 \%) \\
500 & 0.673 / 0.725 / 0.651 / 0.518 / 0.478, 0.249 & 0.69 \% (4.2 \%) & 0.693 / 0.755 / 0.523 / 0.415 / 0.411, 0.486 & 0.16 \% (0.95 \%) \\
$\infty$ & 0.660 / 0.714 / 0.643 / 0.512 / 0.477, 0.253 & 0 & 0.690 / 0.754 / 0.519 / 0.412 / 0.411, 0.500 & 0 \\
\bottomrule
\end{tabular}}
\end{center}

Consistency: the exact solve at the training p equals the stored converged outputs. \path{dist_sched10_bg} at p = 20 gives 0.795 / 0.649 / 0.668 / 0.715 / 0.182, the jacobian evaluation of section 12.2. \path{rw_learned} at p = 50 gives 0.710 / 0.602 / 0.480 / 0.366 / 0.322. \path{edge_bce} at p = 50 gives 0.707 / 0.434 / 0.346 / 0.360 / 1.100 against 0.707 / 0.435 / 0.347 / 0.368 / 1.111: that model\textquotesingle s stored PCG solve differs from the exact one on 0.05 \% of the pixels. The three limit rows equal the full-fold limits of section 12.2 (\path{edge_bce} was not in 12.2; its 307-patch limit in section 10 was IoU 0.684 / F1 0.501).

Figure \path{results/pannuke_p_sweep/p_sweep.png}: (a) the instance model\textquotesingle s F1, PQ, IoU and AJI against p on a log axis, with the limit drawn at the right edge ($\infty$) and the training p dashed; (b) the same for the two binary models; (c) the disagreement with the limit (log scale), with the training p ringed.

\begin{itemize}

\item
  \textbf{Instance model: accuracy peaks at the training p and falls monotonically to the limit; it is not flat.} From p = 5 to 20 every metric rises (F1 0.691 $\to$ 0.795: at low p the solve is too soft to separate touching nuclei). Past the training p every metric falls: F1 0.781 at 35, 0.761 at 50, 0.735 at 100, 0.727 at 200, 0.724 at 500, and 0.724 at the limit. 85 \% of the total loss (0.071 F1, 0.064 PQ, 0.090 AJI, 0.085 IoU) has happened by p = 100. The loss is in detection, not in the quality of the matched nuclei: SQ goes 0.816 $\to$ 0.809, while the count error rises from 0.182 to 0.274. This is the temperature dependence of the costs section 11.5 diagnosed: trained from its own markers, the solve hedges at p = 20 (it vetoes weak markers through soft unaries and shares contested pixels), and a higher p removes the hedging. The costs are tuned to the training temperature. Nothing between the training p and the limit does better than the training p.
\item
  \textbf{Agreement with the limit increases smoothly with p but is total only asymptotically.} At the training p, 4.4 \% of the pixels (25 \% of the foreground) carry another label than the limit\textquotesingle s. At p = 100 it is 0.70 \% (4.6 \%), at 200 0.23 \%, at 500 0.047 \% (0.32 \% of the foreground, 24 \% of the patches identical). The decrease is roughly a power law, about p\^{}$-$1.5 to p\^{}$-$1.7 between 50 and 500, and on the patches tried it reaches 0 at p $\approx$ 3000. Pixel agreement runs ahead of the metrics: at p = 200 the outputs differ on 0.23 \% of the pixels, and F1 / PQ are within 0.003 of the limit\textquotesingle s.
\item
  \textbf{Binary models: pixel accuracy is flat up to the training p, then falls to the limit; instance accuracy rises past the training p.} \path{rw_learned}\textquotesingle s IoU stays at 0.706--0.710 from p = 5 to 100 and falls to 0.693 / 0.673 / 0.660 at 200 / 500 / $\infty$. Its F1 and PQ rise steadily up to p = 200 (0.602 $\to$ 0.653, 0.480 $\to$ 0.521), above the limit\textquotesingle s 0.643 / 0.512. The edge-target model goes the same way, more sharply: its thresholded output is speckle below its training p (count error 4--6) and is at its best at p = 200 (F1 0.529 / PQ 0.420 against 0.434 / 0.346 at p = 50). For both, the exact limit is within 0.01 F1 of the best finite p: only p = 200--500 beat it, by at most 0.01 F1. This is section 6\textquotesingle s finding as a curve: higher p removes the faint spurious blobs that thresholding a soft output leaves, and merges touching nuclei along the way, so detection improves while the pixel boundaries degrade. Agreement converges more slowly than for the instance model (0.69 \% / 0.16 \% of the pixels at p = 500), because the sigmoid edge costs are $\le$ 1, so p$\cdot\Delta$c is smaller at equal p.
\item
  \textbf{For the paper.} Between the training p and the limit, the learned instance costs lose accuracy monotonically, and most of the loss is complete by p $\approx$ 100, where the finite-p labels already agree with the limit on 99.3 \% of the pixels. There is no hidden optimum at larger p, and deploying the limit costs exactly the measured 0.07 F1 of section 11.5. The finite-p curve was computed with the exact elimination above: the Jacobi-PCG used for training and evaluation does not converge beyond p $\approx$ 50 on these costs, which bounds the usable annealing range of this solver independently of accuracy.
\end{itemize}

\textbf{Section 1 check with the exact solver} (\path{cond_exact.py}, the 512$^2$ cases of section 1 with hard seeds: the edges from free pixels to seeds become attachments to the seeds\textquotesingle{} labels; about 1 s per solve). The label column of the section 1 table (and of the paper\textquotesingle s Table 1) is confirmed to 0.001 \% wherever the LU reference was flagged reliable. Free pixels differing from x\^{}$\infty$, exact (stored LU where it existed):

\begin{center}\scriptsize\setlength{\tabcolsep}{3pt}
\resizebox{\ifdim\width>\textwidth\textwidth\else\width\fi}{!}{%
\begin{tabular}{@{}llllllll@{}}
\toprule
case & p = 10 & 100 & 200 & 500 & 1000 & 2000 & 5000 \\
\midrule

clean, $\sigma$ = 0.05 & 0.677 \% (0.677) & 0.004 \% (0.004) & 0 & 0 & 0 & 0 & 0 \\
noisy, $\sigma$ = 0.1 & 3.073 \% (3.073) & 2.167 \% (2.167) & 1.682 \% & 1.147 \% (1.147) & 0.593 \% (0.921, flagged) & 0.122 \% (9.354, flagged) & 0.126 \% \\
camera & 9.823 \% (9.823) & 5.516 \% (5.516) & 5.130 \% & 4.073 \% (4.073) & 2.693 \% (2.751, flagged) & 1.986 \% (9.258, flagged) & 1.174 \% \\
astronaut & 14.62 \% & 10.66 \% & 8.55 \% & 6.069 \% (6.069, flagged) & 4.678 \% (6.391, flagged) & 3.402 \% (42.16, flagged) & 1.735 \% \\
\bottomrule
\end{tabular}}
\end{center}

The LU went wrong exactly where its residual flag fired. Section 1\textquotesingle s reading, a few per cent of the free pixels still differing at p = 500 on noisy and natural images, holds: on 8-bit squared costs the smallest cost gaps are 1/65025, so the noisy image still differs on 0.13 \% of the free pixels at p = 5000 and \texttt{camera} on 1.2 \%.

\section{A detection head for the markers: CenterNet-style centres and radii}

Sections 11.3, 11.4, 14 and 16 all tried to improve the markers through the distance map (a per-nucleus peak loss, hard negatives, the gradient term, strengths trained by the partition loss) and none helped; with ground-truth markers both the WS layer and the fair baseline reach F1 0.87 against 0.795 / 0.794 with their own, so marker detection is the whole remaining gap, and it is shared. one of us proposed a detection formulation instead: predict a centre and a radius per nucleus as a detector does (CenterNet / Mask R-CNN), aggregate the overlapping proposals, and use the disc as the marker. This section implements it for both models and measures it at full data and at 10 \% of fold 1. Code: \texttt{-\/-mk-head\ center} in \path{experiments/exp_instance_pannuke.py} (\path{center_target}, \path{center_focal_loss}, \path{markers_from_center} with \path{center_peaks} / \path{disc_iou}, \path{center_strengths}; \path{marker_loss} now returns the loss and its parts; the training-time extraction worker takes a dict; new outputs \path{layer_soft} / \path{layer_hard}) and \path{exp_unet_dist.py} (\texttt{DistSegmenter(center=True)}, \path{own_markers}), the checks \path{center_check.py}, the cases \path{ctr_bg}, \path{ctr_bg_soft}, \path{unet_ctr}, \path{ctr_bg_p10}, \path{ctr_bg_soft_p10}, \path{unet_ctr_p10} of \path{ruche_run.sh} and \path{run_center.sh} (jacobian / dvaughan), and the tables \path{compare_center.py}. Results: \texttt{results/pannuke\_instances/\{ctr\_bg,\ ctr\_bg\_soft,\ unet\_ctr\}/}, \texttt{results/pannuke\_data\_efficiency/\{ctr\_bg\_p10,\ ctr\_bg\_soft\_p10,\ unet\_ctr\_p10\}/}, \path{compare_18.md} and \path{compare_18_p10.md}, \path{center_check.txt}; the \texttt{.pt} files stay on ruche / dvaughan.

\textbf{The head.} Two output channels replace the distance channel: the centre heatmap logit (channel 3, so that it is the marker map everywhere else in the code) and the radius (channel 4, softplus, in pixels). Targets from the instance masks, after the d4 augmentation: the centre pixel of every nucleus is its pixel nearest its centroid; the heatmap is the maximum over the nuclei of a Gaussian at the centre pixel with $\sigma$ = clip(0.5 r\_eq, 1, 6) px, r\_eq = sqrt(area / $\pi$) the equivalent radius (mean 9.8 px on fold 1, 5 / 50 / 95 \% 3.6 / 8.6 / 20.8), exactly 1 at the centres; the radius target is r\_eq at the centre pixels. Losses: the CornerNet / CenterNet focal loss on the heatmap (positives = the centre pixels, every other pixel a negative weighted (1 $-$ target)\^{}4, the (1 $-$ p)\^{}2 / p\^{}2 modulation, normalised by the number of centres: explicit negatives everywhere, which is what the distance-map losses of 11.3--11.4 could not give without conflicting with the rim) and 0.1 $\times$ the L1 of the radius at the centres; the marker-head weight is 1 (the focal loss is of order 1 after the first epoch, where the distance MSE was 0.008 $\times$ 2). Foreground head, contour head and edge target unchanged.

\textbf{The marker rule.} Peaks = the pixels at the maximum of their 3 $\times$ 3 window and above a threshold, one per plateau; a greedy non-maximum suppression over the predicted discs (in decreasing peak order, a peak whose disc has IoU above \texttt{nms} with a kept disc is merged into it); marker k = the disc of radius \texttt{rfac} $\times$ predicted radius (at least 1 px) around the peak, a pixel inside two discs going to the stronger, every marker keeping its peak pixel; optionally clipped to the foreground s \textgreater{} 0.5 (not used: the WS layer\textquotesingle s attachment c1 = $-$log s already makes a background disc pixel inert, and the watershed baseline\textquotesingle s markers lie inside the foreground mask anyway). Fold 3 chooses threshold $\in$ \{0.2, 0.3, 0.4, 0.5\} $\times$ nms $\in$ \{0.3, 0.5, 0.7\} $\times$ rfac $\in$ \{0.3, 0.5, 0.7\} by PQ on the usual 300 patches, for the WS layer (through its converged solve) and for the baseline ($\times$ the three reliefs), never on fold 2. The training-time markers of the 11.5 schedule use threshold 0.3, nms 0.5, rfac 0.5. At most 256 peaks per patch enter the suppression (an untrained heatmap has thousands).

\textbf{The soft variant} (\texttt{-\/-ctr-soft}, run \path{ctr_bg_soft}). Every marker enters the attachment with its peak value as a strength, a\_k = ($\sigma$\_k s)\^{}p, i.e. c1 $-$ log $\sigma$\_k on its pixels as in section 14, but $\sigma$\_k is a constant: the detection loss alone trains it, nothing flows from the solve into the selection (the failure mode of 14). The strengths are on at training time (on the predicted markers of the schedule) and at test time. In addition every centre run evaluates the other variant on the same markers and costs (\path{layer_soft} for a hard-trained model, \path{layer_hard} for a soft-trained one), so the test-time effect of the strengths is separated from the training effect.

\textbf{Checks before the runs} (\path{center_check.py}, 16 evenly spaced patches of fold 1; \path{center_check.txt}). The targets: one centre pixel per instance (565 for 565), all inside their instance, 565 heatmap peaks for 565 instances. The rule on the target heatmap and radius (an ideal prediction): every nucleus gets exactly one marker and no marker lands in background at rfac 0.3 / 0.5 / 0.7 (marker pixels inside their own nucleus 100 / 99.9 / 98.3 \%), and the same on a blurred target with the radius spread over the Gaussian\textquotesingle s support, with or without the suppression. A baseline trained for six epochs on the full fold gave, at threshold 0.3, 29 markers per patch for 35 nuclei (74 \% of the nuclei with exactly one, 25 \% none, 2.3 in background; threshold 0.2: 39 markers, 85 / 11 \%, 5.8 in background), the radius within 1.06 px of the target on average, and the suppression changing the count by at most 0.4 markers per patch: the head produces about one peak per nucleus, not the hundreds of section 11.3, and the threshold trades recall against background markers smoothly.

\textbf{What was run} (seed 0, fold 1 $\to$ fold 2, tuning on fold 3, all 2523 test patches). The WS layer with the section 11.5 recipe and the centre head in place of the distance head (\path{ctr_bg}: \texttt{-\/-mk-head\ center\ -\/-w-mk\ 1\ -\/-mk-sched\ 10\ 15\ -\/-mk-unmarked\ bg}, p 5 $\to$ 20, edge target; the predicted markers of the schedule at threshold 0.3 / nms 0.5 / rfac 0.5), its soft-strength variant (\path{ctr_bg_soft}), the fair baseline with the same head (\path{unet_ctr}: the section 11.2 U-Net and watershed, the distance channel replaced by the heatmap and radius), and the three at 10 \% of fold 1 (\path{ctr_bg_p10}, \path{ctr_bg_soft_p10}, \path{unet_ctr_p10}: 266 images, 300 epochs, \texttt{-\/-mk-sched\ 100\ 150}, as section 13). The references are \path{dist_sched10_bg} and \path{unet_dist} (sections 11.5 / 11.2) and \path{dist_bg_p10} / \path{unet_dist_p10} (section 13). The WS layer runs took 3.7 h on an A100 (326 s per epoch, 36 settings $\times$ 300 patches of fold-3 tuning through the solve in 31 min, the evaluation with its four converged solves per batch in 85 min); the baselines 11 s per epoch; the 10 \% layer runs 1.3 h for the 300 epochs. Two layer jobs died at start on ruche\textquotesingle s node gpu16 with "CUDA-capable device busy or unavailable" and were resubmitted elsewhere; the 10 \% runs ran on dvaughan.

\textbf{Results on fold 2} (\path{compare_18.md}, \path{compare_18_p10.md}; the paired columns are the bootstrap intervals of the per-patch differences against the run named):

\begin{center}\scriptsize\setlength{\tabcolsep}{3pt}
\resizebox{\ifdim\width>\textwidth\textwidth\else\width\fi}{!}{%
\begin{tabular}{@{}lllllllllll@{}}
\toprule
run & \href{mailto:F1@0.5}{\nolinkurl{F1@0.5}} & PQ & SQ & AJI & IoU & count err. & markers one / none / several, bg per patch & fold-3 choice & oracle F1 / PQ & limit F1 / PQ (307) \\
\midrule

\path{dist_sched10_bg} (11.5) & 0.795 & 0.649 & 0.816 & \textbf{0.668} & \textbf{0.715} & \textbf{0.182} & 85.5 / 12.7 / 1.8 \%, 4.3 & thr 0.2, area 4, h 0.2 & 0.872 / 0.723 & 0.700 / 0.564 \\
\path{unet_dist} (11.2) & 0.794 & \textbf{0.655} & \textbf{0.824} & 0.658 & 0.719 & 0.190 & 76.6 / 21.8 / 1.6 \%, 1.6 & thr 0.5, area 4, h 0.2, c $-$ d & 0.867 / 0.708 & -- \\
\path{ctr_bg}: layer, centre head & \textbf{0.795} & 0.653 & 0.822 & 0.640 & 0.678 & 0.204 & 79.3 / 19.6 / 1.1 \%, 2.4 & thr 0.3, nms 0.3, rfac 0.7 & 0.869 / 0.717 & 0.699 / 0.571 \\
\path{ctr_bg_soft}: the same, peak strengths $\sigma$\_k\^{}p & 0.566 & 0.481 & 0.849 & 0.413 & 0.431 & 0.567 & 51.4 / 48.5 / 0.0 \%, 0.4 & thr 0.5, nms 0.3, rfac 0.7 & 0.925 / 0.754 & 0.369 / 0.316 \\
\path{ctr_bg_soft} solved without the strengths (\path{layer_hard}) & 0.660 & 0.558 & 0.845 & 0.492 & 0.517 & 0.472 & same markers & -- & -- & -- \\
\path{ctr_bg} with the strengths at test time, $\sigma$\_k\^{}p (\path{layer_soft}) & 0.227 & 0.201 & 0.884 & 0.194 & 0.210 & 0.779 & same markers as \path{ctr_bg} & -- & -- & -- \\
\path{ctr_bg} with the strengths at test time, $\sigma$\_k (\texttt{-\/-ctr-soft-pow\ 1}, \path{instances_softpow1.json}) & 0.771 & 0.635 & 0.823 & 0.627 & 0.661 & 0.234 & same markers as \path{ctr_bg} & -- & -- & -- \\
\path{unet_ctr}: baseline, centre head & 0.774 & 0.627 & 0.810 & 0.626 & 0.696 & 0.197 & 80.4 / 18.4 / 1.2 \%, 2.5 & thr 0.3, nms 0.3, rfac 0.3, c $-$ h & 0.854 / 0.687 & -- \\
10 \%: \path{dist_bg_p10} (13) & \textbf{0.765} & \textbf{0.620} & 0.811 & \textbf{0.635} & \textbf{0.681} & 0.238 & 84.0 / 13.9 / 2.1 \%, 4.3 & thr 0.2, area 1, h 0.2 & 0.840 / 0.691 & 0.692 / 0.554 \\
10 \%: \path{unet_dist_p10} (13) & 0.745 & 0.605 & 0.813 & 0.620 & 0.670 & 0.237 & 73.8 / 24.5 / 1.6 \%, 2.3 & thr 0.5, area 9, h 0.3 & 0.812 / 0.654 & -- \\
10 \%: \path{ctr_bg_p10} & 0.751 & 0.612 & \textbf{0.814} & 0.605 & 0.646 & 0.243 & 72.8 / 26.1 / 1.1 \%, 2.4 & thr 0.5, nms 0.3, rfac 0.5 & 0.848 / 0.690 & 0.721 / 0.579 \\
10 \%: \path{ctr_bg_soft_p10} & 0.736 & 0.603 & 0.820 & 0.587 & 0.628 & 0.282 & 72.5 / 26.4 / 1.1 \%, 2.5 & thr 0.5, nms 0.3, rfac 0.7 & 0.873 / 0.709 & 0.680 / 0.552 \\
10 \%: \path{ctr_bg_soft_p10} solved without the strengths & 0.751 & 0.611 & 0.814 & 0.604 & 0.648 & 0.253 & same markers & -- & -- & -- \\
10 \%: \path{unet_ctr_p10} & 0.720 & 0.577 & 0.801 & 0.589 & 0.647 & 0.254 & 72.7 / 26.5 / 0.9 \%, 2.2 & thr 0.5, nms 0.3, rfac 0.3, c & 0.782 / 0.621 & -- \\
\bottomrule
\end{tabular}}
\end{center}

Paired (F1 / PQ / AJI / IoU, 95 \% intervals), full data: \path{ctr_bg} $-$ \path{dist_sched10_bg} \textbf{$-$0.001 {[}$-$0.003, +0.002{]} / +0.004 {[}+0.002, +0.006{]}} / $-$0.028 {[}$-$0.031, $-$0.025{]} / $-$0.037 {[}$-$0.041, $-$0.033{]}; \path{ctr_bg} $-$ \path{unet_dist} +0.001 {[}$-$0.002, +0.004{]} / $-$0.002 {[}$-$0.004, +0.001{]} / $-$0.019 / $-$0.041; \path{unet_ctr} $-$ \path{unet_dist} \textbf{$-$0.020 {[}$-$0.023, $-$0.017{]} / $-$0.028 {[}$-$0.030, $-$0.026{]}} / $-$0.032 / $-$0.023; \path{ctr_bg} $-$ \path{unet_ctr} +0.021 {[}+0.018, +0.023{]} / +0.026 {[}+0.024, +0.028{]} / +0.014 / $-$0.017. Oracles: \path{ctr_bg} $-$ \path{dist_sched10_bg} $-$0.003 {[}$-$0.006, $-$0.000{]} / $-$0.005 / $-$0.026 / $-$0.014; \path{unet_ctr} $-$ \path{unet_dist} $-$0.013 {[}$-$0.016, $-$0.011{]} / $-$0.021 / $-$0.010 / $-$0.008. At 10 \%: \path{ctr_bg_p10} $-$ \path{dist_bg_p10} \textbf{$-$0.014 {[}$-$0.017, $-$0.011{]} / $-$0.009 {[}$-$0.011, $-$0.006{]}} / $-$0.030 / $-$0.035; \path{ctr_bg_p10} $-$ \path{unet_dist_p10} \textbf{+0.006 {[}+0.003, +0.010{]} / +0.006 {[}+0.004, +0.009{]}} / $-$0.015 / $-$0.024; \path{unet_ctr_p10} $-$ \path{unet_dist_p10} $-$0.025 {[}$-$0.028, $-$0.022{]} / $-$0.029 / $-$0.032 / $-$0.022; \path{ctr_bg_p10} $-$ \path{unet_ctr_p10} +0.031 {[}+0.028, +0.034{]} / +0.035 / +0.017 / $-$0.001. The soft variant: \path{ctr_bg_soft} $-$ \path{ctr_bg} $-$0.229 {[}$-$0.235, $-$0.223{]} / $-$0.172 / $-$0.227 / $-$0.247; its own hard solve $-$ its soft solve +0.094 {[}+0.091, +0.097{]} / +0.077; \path{ctr_bg_soft_p10} $-$ \path{ctr_bg_p10} $-$0.015 {[}$-$0.018, $-$0.013{]} / $-$0.009 {[}$-$0.011, $-$0.007{]} / $-$0.018 / $-$0.018, its own hard solve $-$ its soft solve +0.015 {[}+0.014, +0.017{]} / +0.008; \path{ctr_bg}\textquotesingle s strengths at test time, $\sigma$\_k\^{}p: $-$0.568 F1; $\sigma$\_k: $-$0.024 {[}$-$0.026, $-$0.022{]} F1 / $-$0.018 {[}$-$0.020, $-$0.017{]} PQ / $-$0.013 AJI / $-$0.017 IoU (\path{compare_18_softpow1.md}; the hard output of that evaluation reproduces \path{ctr_bg} to every digit). Limits (307 patches, paired): \path{ctr_bg} $-$ \path{dist_sched10_bg} $-$0.001 {[}$-$0.013, +0.008{]} F1 / +0.007 PQ; at 10 \% +0.029 {[}+0.020, +0.039{]} / +0.026.

\begin{itemize}

\item
  \textbf{At full data the detection head does not move either model past 0.795 / 0.794.} The WS layer with centre markers is at F1 0.795 and PQ 0.653, level with the 11.5 recipe on F1 (interval across zero) and +0.004 PQ, which is half the seed spread of section 11.5 (0.008 F1 / 0.015 PQ); against the fair baseline it is +0.001 F1 / $-$0.002 PQ, both across zero. The baseline with the centre head is 0.020 F1 / 0.028 PQ \emph{worse} than with its distance head. So the honest statement is: the detection formulation gives the WS layer a different marker set of the same value and takes 0.02 F1 from the watershed baseline; neither model improves.
\item
  \textbf{The markers are cleaner but fewer, and the two effects cancel.} The centre head\textquotesingle s markers on fold 2 land in background 2.4 times per patch against 4.3 for the distance head under the WS layer\textquotesingle s tuning, and almost never double a nucleus (1.1 \% against 1.8 \%), but 19.6 \% of the nuclei get no marker against 12.7 \%. Fold 3 chooses the trade-off: at threshold 0.2 the centre head marks 85 \% of the nuclei, as the distance head does, with 5.2 background markers per patch and 0.011 PQ less on fold 3; at 0.3 it keeps 79 \% with 2.3 and the best PQ. The radius factor matters for the solve (0.7 best, 0.5 $-$0.002 PQ, 0.3 $-$0.016 on fold 3: a small disc is a weak attachment and the background label takes the nucleus) and the suppression almost never fires (the heatmap has one peak per nucleus; nms 0.3 / 0.5 / 0.7 differ by 0.003--0.005 PQ). The oracle with ground-truth markers is also level (0.869 against 0.872 F1), so the costs the two heads leave to the solve are equally good. What the WS layer loses is pixel coverage: $-$0.037 IoU and $-$0.028 AJI come from the 7 \% more unmarked nuclei, which go to the background under \texttt{-\/-mk-unmarked\ bg}, while F1 does not see them because the distance head\textquotesingle s extra markers were partly background markers that cost precision.
\item
  \textbf{The watershed baseline is hurt twice.} Its marker swap (\path{unet_ctr_swap115}) separates the causes: the 11.5 layer\textquotesingle s distance markers through the centre baseline\textquotesingle s own watershed give F1 0.767 against 0.774 with its centre markers, so the markers are not the problem; the oracle is (0.854 against 0.867 F1, $-$0.021 PQ): the relief. With the distance head the baseline\textquotesingle s best relief is c $-$ d, the contour map minus a map that is high inside every nucleus and cuts the valley between two; with the centre head the second map is a set of Gaussian bumps at the centres, and the best relief (c $-$ h) is barely better than the contour map alone (0.637 against 0.637 PQ on fold 3). The foreground also degrades slightly (logits IoU 0.712 against 0.725): the focal loss at weight 1 is of order 1 against 0.016 for the distance MSE and it takes the shared trunk. The WS layer does not need a relief, which is why it loses nothing where the baseline loses 0.020.
\item
  \textbf{At 10 \% the WS layer keeps a lead over the distance-head baseline, but a smaller one.} With the centre head the WS layer at 266 training images is at F1 0.751 / PQ 0.612 against 0.765 / 0.620 for the distance-head layer ($-$0.014 F1) and 0.745 / 0.605 for the distance-head baseline (+0.006 F1 {[}+0.003, +0.010{]}, +0.006 PQ {[}+0.004, +0.009{]}, intervals clear of zero but a third of section 13\textquotesingle s +0.020); against its own centre-head baseline it is +0.031 F1 / +0.035 PQ. The mechanism of section 13 shows again, from the other side: fold 3 raises the centre head\textquotesingle s threshold to 0.5 for both models at 10 \% (26 \% of the nuclei unmarked), where the distance-head layer kept 0.2 (14 \% unmarked) and absorbed its 4.3 background markers per patch through the solve. The centre heatmap trained on 266 images is simply less confident at the true centres (threshold 0.3 marks 77 \% of the nuclei with 3.8 in background, $-$0.002 PQ on fold 3), so the WS layer cannot buy recall cheaply. The oracles are equal (0.848 against 0.840 F1), the matched nuclei are segmented equally (SQ 0.814 against 0.811): the whole gap is detection, as before.
\item
  \textbf{Soft detection confidences do not help the WS layer; as strengths they hurt, and raised to p they destroy it.} The peak values of a well-trained heatmap sit at 0.5--0.8 for true nuclei (5 / 50 / 95 \% 0.51 / 0.62 / 0.81 on the example patches), so under the section-14 convention a\_k = ($\sigma$\_k s)\^{}p every marker\textquotesingle s attachment is multiplied by $\sigma$\^{}20 $\approx$ 10$^{-4}$ and the background label, whose attachment is untouched, takes most of the nucleus: \path{ctr_bg_soft} keeps 10.5 instances per patch whatever the threshold (fold 3 gives PQ 0.476--0.478 from thr 0.2 to 0.5: the strengths \emph{are} the threshold), F1 0.566, and the training cross-entropy climbs from 0.10 at epoch 9 to 0.57 at epoch 29 as p anneals, because no cost the network controls can compensate a factor it cannot exceed (s $\le$ 1). Solving the same model without the strengths recovers 0.660, and the oracle at $\sigma$ = 1 inflates to 0.925 as in sections 14 and 16 (ground-truth markers enter stronger than any the solve was trained on). At 10 \% the same recipe costs less (0.736 against 0.751, the hard solve of the soft-trained model 0.751: the training is not damaged, only the test-time veto remains), because fold 3 picks threshold 0.5 and the surviving markers have $\sigma$ $\ge$ 0.5. The fairer version of the idea, the confidence entering as $\sigma$\_k itself rather than $\sigma$\_k\^{}p (\texttt{-\/-ctr-soft-pow\ 1}, a cost shift of $-$(log $\sigma$\_k)/p, evaluated on the hard-trained \path{ctr_bg} without retraining), removes the collapse but still loses 0.024 F1 / 0.018 PQ to the hard solve on the same markers and costs (added after the commit of the section, the evaluation having finished on its own): a confidence below 1 can only weaken a marker, so it cannot add recall, and the markers it weakens most are the faint true nuclei the head was least sure of. So where the watershed needs a hard threshold the WS layer needs one too: a confidence below 1 can only weaken a marker, and what the markers lack is recall, not precision. The exact limit tells the same story as section 14: any $\sigma$\_k \textless{} 1 is a hard veto at p $\to$ $\infty$ (0.369 F1 for the soft model against 0.699 for the hard one).
\item
  \textbf{Verdict.} A CenterNet-style head is a sound marker detector for this pipeline (one peak per nucleus, a radius within 0.7 px of the target, no needle failure, the threshold trading recall against background markers smoothly), and it is the first marker change since section 11.5 that does not cost the WS layer anything; but it does not improve it. The WS layer\textquotesingle s result is pinned at F1 0.795 by a third marker head, which says that the ceiling is not the head\textquotesingle s form but the detectability of the 12--20 \% of nuclei that none of the heads, nor the baseline, finds at a threshold that keeps the background clean: the oracle gap of 0.07 F1 is shared by every variant. For the watershed baseline the distance head stays the right choice, because it doubles as the relief. The 10 \% lead of the structured layer survives the change of head (+0.006 against the distance baseline, +0.031 against the matched one) but is largest with the distance head. One seed, one fold pair, one set of hyper-parameters ($\sigma$ = 0.5 r\_eq, focal weight 1, radius weight 0.1) for the head; a lower marker-head weight or a second seed were not run.
\end{itemize}

\section{Implementation and compute}\label{app:impl}

Everything runs in PyTorch on one NVIDIA RTX 6000 Ada (48~GB). The WS layer uses dense
shifted-grid operators ($B\times1\times H\times W$ tensors), Jacobi-preconditioned
conjugate gradients, single precision for the unrolled mode and double precision for
the reference solves. The exact limit is NumPy on the CPU, about 2~s per $256^2$ patch
with continuous costs. Training the binary WS layer on a PanNuke fold costs 11~s per
epoch for the unrolled mode and 50~s for the reweighted mode; the multi-marker WS layer
about 230~s per epoch with 20--60 PCG iterations per batch.

\fi

\end{document}